\documentclass[runningheads]{llncs}
\usepackage{graphicx}
\usepackage{subfigure}
\usepackage{mathtools}
\usepackage{amsmath,amssymb}
\usepackage{color}
\usepackage[table]{xcolor}
\usepackage{multirow}
\usepackage{makecell}
\usepackage{placeins}
\usepackage{balance}
\usepackage{array}
\usepackage{wrapfig,lipsum,booktabs}
\newcolumntype{C}[1]{>{\centering\let\newline\\\arraybackslash\hspace{0pt}}m{#1}}
\usepackage[width=122mm,left=12mm,paperwidth=146mm,height=193mm,top=12mm,paperheight=217mm]{geometry}
\usepackage[pagebackref=true,breaklinks=true,letterpaper=true,colorlinks,bookmarks=false]{hyperref}

\hypersetup{linkcolor=[rgb]{0.8551,0.2333,0.2333}}
\hypersetup{citecolor=[rgb]{0.3333,0.2333,0.7551}}

\DeclareMathOperator*{\argmin}{arg\,min}
\DeclareMathOperator*{\argmax}{arg\,max}

\definecolor{sbd_color_1}{rgb}{0.5020,0.0000,0.0000}
\definecolor{sbd_color_2}{rgb}{0.0000,0.5020,0.0000}
\definecolor{sbd_color_3}{rgb}{0.5020,0.5020,0.0000}
\definecolor{sbd_color_4}{rgb}{0.0000,0.0000,0.5020}
\definecolor{sbd_color_5}{rgb}{0.5020,0.0000,0.5020}
\definecolor{sbd_color_6}{rgb}{0.0000,0.5020,0.5020}
\definecolor{sbd_color_7}{rgb}{0.5020,0.5020,0.5020}
\definecolor{sbd_color_8}{rgb}{0.2510,0.0000,0.0000}
\definecolor{sbd_color_9}{rgb}{0.7529,0.0000,0.0000}
\definecolor{sbd_color_10}{rgb}{0.2510,0.5020,0.0000}
\definecolor{sbd_color_11}{rgb}{0.7529,0.5020,0.0000}
\definecolor{sbd_color_12}{rgb}{0.2510,0.0000,0.5020}
\definecolor{sbd_color_13}{rgb}{0.7529,0.0000,0.5020}
\definecolor{sbd_color_14}{rgb}{0.2510,0.5020,0.5020}
\definecolor{sbd_color_15}{rgb}{0.7529,0.5020,0.5020}
\definecolor{sbd_color_16}{rgb}{0.0000,0.2510,0.0000}
\definecolor{sbd_color_17}{rgb}{0.5020,0.2510,0.0000}
\definecolor{sbd_color_18}{rgb}{0.0000,0.7529,0.0000}
\definecolor{sbd_color_19}{rgb}{0.5020,0.7529,0.0000}
\definecolor{sbd_color_20}{rgb}{0.0000,0.2510,0.5020}

\definecolor{city_color_1}{rgb}{0.5020,0.2510,0.5020}
\definecolor{city_color_2}{rgb}{0.9569,0.1373,0.9098}
\definecolor{city_color_3}{rgb}{0.2745,0.2745,0.2745}
\definecolor{city_color_4}{rgb}{0.4000,0.4000,0.6118}
\definecolor{city_color_5}{rgb}{0.7451,0.6000,0.6000}
\definecolor{city_color_6}{rgb}{0.6000,0.6000,0.6000}
\definecolor{city_color_7}{rgb}{0.9804,0.6667,0.1176}
\definecolor{city_color_8}{rgb}{0.8627,0.8627,0.0000}
\definecolor{city_color_9}{rgb}{0.4196,0.5569,0.1373}
\definecolor{city_color_10}{rgb}{0.5961,0.9843,0.5961}
\definecolor{city_color_11}{rgb}{0.2745,0.5098,0.7059}
\definecolor{city_color_12}{rgb}{0.8627,0.0784,0.2353}
\definecolor{city_color_13}{rgb}{1.0000,0.0000,0.0000}
\definecolor{city_color_14}{rgb}{0.0000,0.0000,0.5569}
\definecolor{city_color_15}{rgb}{0.0000,0.0000,0.2745}
\definecolor{city_color_16}{rgb}{0.0000,0.2353,0.3922}
\definecolor{city_color_17}{rgb}{0.0000,0.3137,0.3922}
\definecolor{city_color_18}{rgb}{0.0000,0.0000,0.9020}
\definecolor{city_color_19}{rgb}{0.4667,0.0431,0.1255}

\begin{document}

\title{Simultaneous Edge Alignment and Learning}

\titlerunning{Simultaneous Edge Alignment and Learning}

\author{
Zhiding Yu\inst{1}\and
Weiyang Liu\inst{3}\and
Yang Zou\inst{2}\and
Chen Feng\inst{4}\and
Srikumar Ramalingam\inst{5}\and
B. V. K. Vijaya Kumar\inst{2}\and
Jan Kautz\inst{1}
}

\authorrunning{Zhiding Yu et al.}

\institute{
NVIDIA\quad\email{\{zhidingy, jkautz\}@nvidia.com}\and
Carnegie Mellon University\quad\email{\{yzou2@andrew, kumar@ece\}.cmu.edu}\and
Georgia Institute of Technology\quad\email{wyliu@gatech.edu}\and
New York University\quad\email{cfeng@nyu.edu}\and
University of Utah\quad\email{srikumar@cs.utah.edu}\\
{[\href{https://github.com/Chrisding/seal}{GitHub}]~~
 [\href{https://chrisding.github.io/project/seal.htm}{Project Page}]~~
 [\href{https://www.youtube.com/watch?v=gpy20uGnlY4}{Demo}]~~
 [\href{https://chrisding.github.io/publications/ECCV18b_Slides.pdf}{Slides}]~~
 [\href{https://chrisding.github.io/publications/ECCV18b_Poster.pdf}{Poster}]}
}

\maketitle

\begin{abstract}
Edge detection is among the most fundamental vision problems for its role in perceptual grouping and its wide applications. Recent advances in representation learning have led to considerable improvements in this area. Many state of the art edge detection models are learned with fully convolutional networks (FCNs). However, FCN-based edge learning tends to be vulnerable to misaligned labels due to the delicate structure of edges. While such problem was considered in evaluation benchmarks, similar issue has not been explicitly addressed in general edge learning. In this paper, we show that label misalignment can cause considerably degraded edge learning quality, and address this issue by proposing a simultaneous edge alignment and learning framework. To this end, we formulate a probabilistic model where edge alignment is treated as latent variable optimization, and is learned end-to-end during network training. Experiments show several applications of this work, including improved edge detection with state of the art performance, and automatic refinement of noisy annotations.
\end{abstract}
\section{Introduction}
Over the past decades, edge detection played a significant role in computer vision. Early edge detection methods often formulate the task as a low-level or mid-level grouping problem where Gestalt laws and perceptual grouping play considerable roles in algorithm design~\cite{Kittler1983,Canny1986,mild-sugihara,hancock1990edge}. Latter works start to consider learning edges in a data-driven way, by looking into the statistics of features near boundaries~\cite{Konishi2003,Martin2004,Dollar2006,ren2008learning,Arbelaez2011,Arbelaez2014,Maire2014,dollar2015fast}. More recently, advances in deep representation learning~\cite{krizhevsky2012imagenet,Simonyan2015,He2016} have further led to significant improvements on edge detection, pushing the boundaries of state of the art performance~\cite{Xie2015,Hwang2015,Bertasius2015_de,kokkinos2016pushing,yang2016object} to new levels. The associated tasks also expended from the conventional binary edge detection problems to the recent more challenging category-aware edge detection problems~\cite{prasad2006learning,Hariharan2011,Bertasius2015_hfl,khoreva2016weakly,yu2017casenet}. As a result of such advancement, a wide variety of other vision problems have enjoyed the benefits of reliable edge detectors. Examples of these applications include, but are not limited to (semantic) segmentation~\cite{Arbelaez2011,yu2015generalized,Chen2016,Bertasius2015_hfl,bertasius2016semantic}, object proposal generation~\cite{zitnick2014edge,Bertasius2015_hfl,yang2016object}, object detection~\cite{Lim2013}, depth estimation~\cite{lineCurvedObjects,hoiem2005geometric}, and 3D vision~\cite{malik1989recovering,Karsch2013,Shan2014}, etc.

\begin{figure}[t]
\centering
\resizebox{1.0\textwidth}{!}{
\begin{tabular}{@{}cccccccccc@{}}
\cellcolor{sbd_color_1}\textcolor{white}{~~~~aero~~~~} &
\cellcolor{sbd_color_2}\textcolor{white}{~~~~bike~~~~} &
\cellcolor{sbd_color_3}\textcolor{white}{~~~~bird~~~~} &
\cellcolor{sbd_color_4}\textcolor{white}{~~~~boat~~~~} &
\cellcolor{sbd_color_5}\textcolor{white}{~~~~bottle~~~~} &
\cellcolor{sbd_color_6}\textcolor{white}{~~~~bus~~~~} &
\cellcolor{sbd_color_7}\textcolor{white}{~~~~car~~~~} &
\cellcolor{sbd_color_8}\textcolor{white}{~~~~cat~~~~} &
\cellcolor{sbd_color_9}\textcolor{white}{~~~~chair~~~~} &
\cellcolor{sbd_color_10}\textcolor{white}{~~~~cow~~~~} \\
\cellcolor{sbd_color_11}\textcolor{white}{~table~} &
\cellcolor{sbd_color_12}\textcolor{white}{~dog~} &
\cellcolor{sbd_color_13}\textcolor{white}{~horse~} &
\cellcolor{sbd_color_14}\textcolor{white}{~mbike~} &
\cellcolor{sbd_color_15}\textcolor{white}{~person~} &
\cellcolor{sbd_color_16}\textcolor{white}{~plant~} &
\cellcolor{sbd_color_17}\textcolor{white}{~sheep~} &
\cellcolor{sbd_color_18}\textcolor{white}{~sofa~} &
\cellcolor{sbd_color_19}\textcolor{white}{~train~} &
\cellcolor{sbd_color_20}\textcolor{white}{~tv~}
\end{tabular}
}
\centering
\subfigure[Original image]{\label{Fig1a}\includegraphics[width=.244\textwidth]{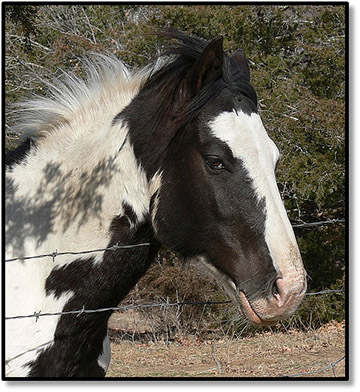}}
\subfigure[Ground truth]{\label{Fig1b}\includegraphics[width=.244\textwidth]{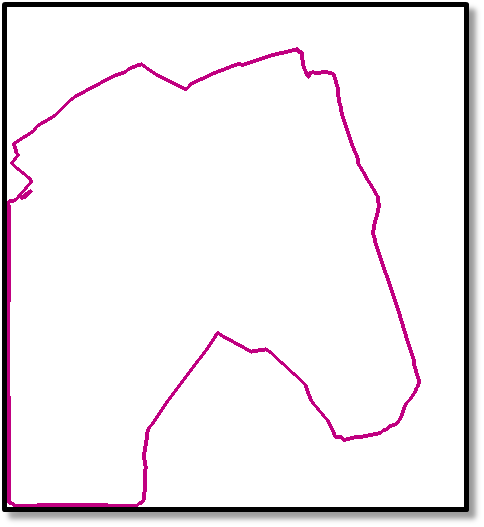}}
\subfigure[CASENet]{\label{Fig1c}\includegraphics[width=.244\textwidth]{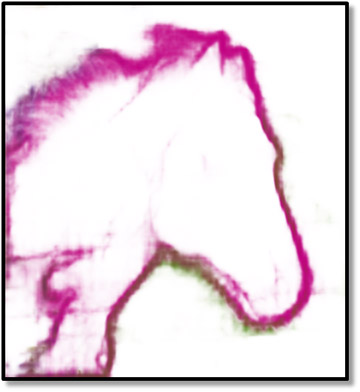}}
\subfigure[SEAL]{\label{Fig1d}\includegraphics[width=.244\textwidth]{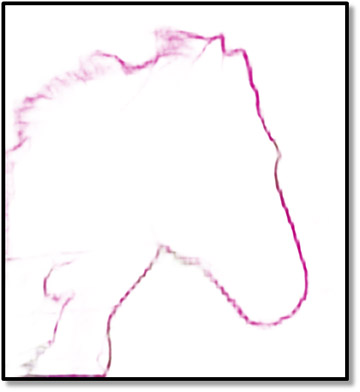}}
\centering
\resizebox{1.0\textwidth}{!}{
\begin{tabular}{@{}cccccccccc@{}}
\cellcolor{city_color_1}\textcolor{white}{~~road~~} &
\cellcolor{city_color_2}\textcolor{white}{~~sidewalk~~} &
\cellcolor{city_color_3}\textcolor{white}{~~building~~} &
\cellcolor{city_color_4}\textcolor{white}{~~wall~~} &
\cellcolor{city_color_5}\textcolor{white}{~~fence~~} &
\cellcolor{city_color_6}\textcolor{white}{~~pole~~} &
\cellcolor{city_color_7}~~traffic lgt~~ &
\cellcolor{city_color_8}~~traffic sgn~~ &
\cellcolor{city_color_9}\textcolor{white}{~~vegetation~} \\
\cellcolor{city_color_10}~~terrain~~ &
\cellcolor{city_color_11}\textcolor{white}{~~sky~~} &
\cellcolor{city_color_12}\textcolor{white}{~~person~~} &
\cellcolor{city_color_13}\textcolor{white}{~~rider~~} &
\cellcolor{city_color_14}\textcolor{white}{~~car~~} &
\cellcolor{city_color_15}\textcolor{white}{~~truck~~} &
\cellcolor{city_color_16}\textcolor{white}{~~bus~~} &
\cellcolor{city_color_17}\textcolor{white}{~~train~~} &
\cellcolor{city_color_18}\textcolor{white}{~~motorcycle~~} &
\cellcolor{city_color_19}\textcolor{white}{~~bike~~}
\end{tabular}
}
\centering
\subfigure[Original image]{\label{Fig1e}\includegraphics[width=.244\textwidth]{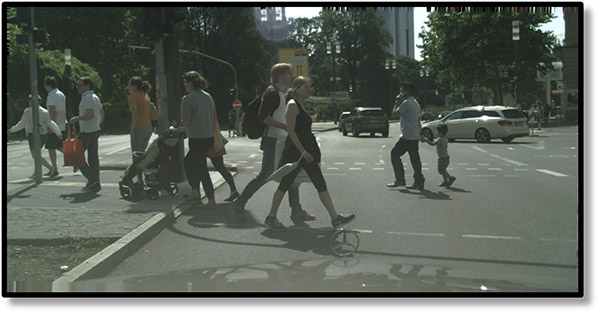}}
\subfigure[Ground truth]{\label{Fig1f}\includegraphics[width=.244\textwidth]{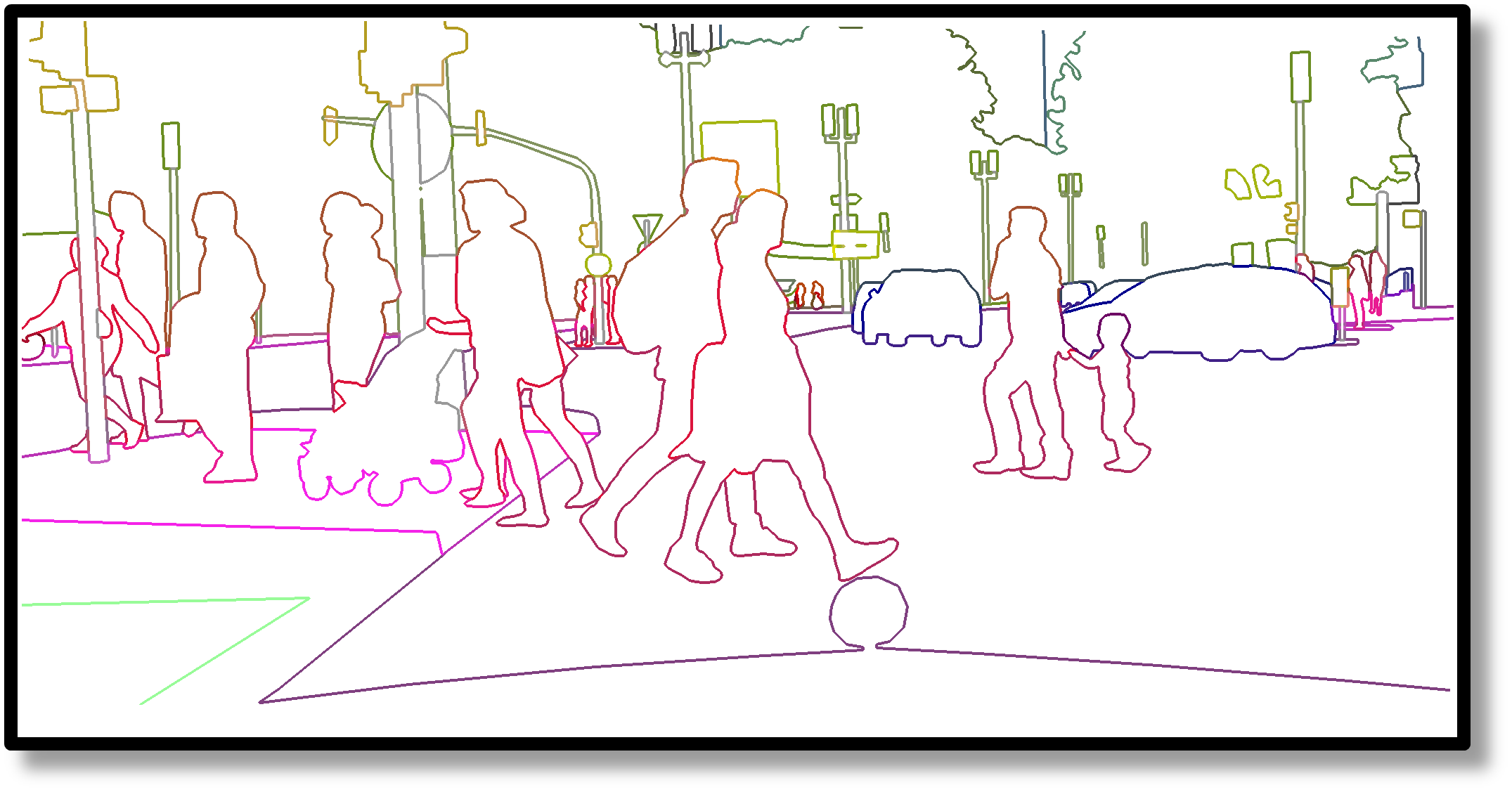}}
\subfigure[CASENet]{\label{Fig1g}\includegraphics[width=.244\textwidth]{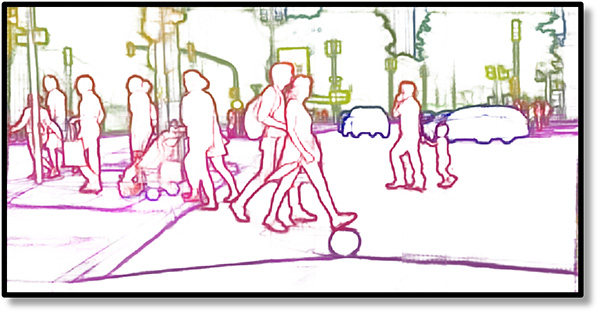}}
\subfigure[SEAL]{\label{Fig1h}\includegraphics[width=.244\textwidth]{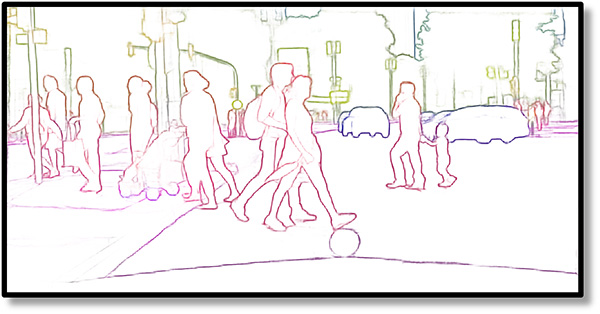}}
\caption{Examples of edges predicted by different methods on SBD (a-d) and Cityscapes (e-h). ``CASENet'' indicates the original CASENet from~\cite{yu2017casenet}. ``SEAL'' indicates the proposed framework trained with CASENet backbone. Best viewed in color.}
\centering
\label{Fig1}
\end{figure}

With the strong representation abilities of deep networks and the dense labeling nature of edge detection, many state of the art edge detectors are based on FCNs. Despite the underlying resemblance to other dense labeling tasks, edge learning problems face some typical challenges and issues. First, in light of the highly imbalanced amounts of positive samples (edge pixels) and negative samples (non-edge pixels), using reweighted losses where positive samples are weighted higher has become a predominant choice in recent deep edge learning frameworks~\cite{Xie2015,kokkinos2016pushing,khoreva2016weakly,liu2017richer,yu2017casenet}. While such a strategy to some extent renders better learning behaviors\footnote{E.g., more stabled training, and more balanced prediction towards smaller classes.}, it also induces thicker detected edges as well as more false positives. An example of this issue is illustrated in Fig.~\ref{Fig1c} and Fig.~\ref{Fig1g}, where the edge mapspredicted by CASENet~\cite{yu2017casenet} contains thick object boundaries. A direct consequence is that many local details are missing, which is not favored for other potential applications using edge detectors.

Another challenging issue for edge learning is the training label noise caused by inevitable misalignment during annotation. Unlike segmentation, edge learning is generally more vulnerable to such noise due to the fact that edge structures by nature are much more delicate than regions. Even slight misalignment can lead to significant proportion of mismatches between ground truth and prediction. In order to predict sharp edges, a model should learn to distinguish the few true edge pixels while suppressing edge responses near them. This already presents a considerable challenge to the model as non-edge pixels near edges are likely to be hard negatives with similar features, while the presence of misalignment further causes significant confusion by continuously sending false positives during training. The problem is further aggravated under reweighted losses, where predicting more false positives near the edge is be an effective way to decrease the loss due to the significant higher weights of positive samples.
\begin{figure}[t]
\centering
\includegraphics[height=.25\textwidth]{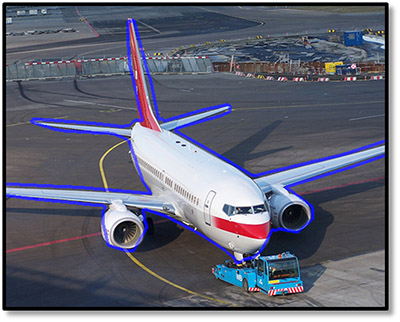}\label{Fig2a}
\includegraphics[height=.25\textwidth]{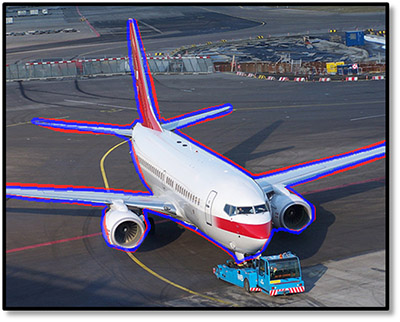}\label{Fig2b}
\includegraphics[height=.25\textwidth]{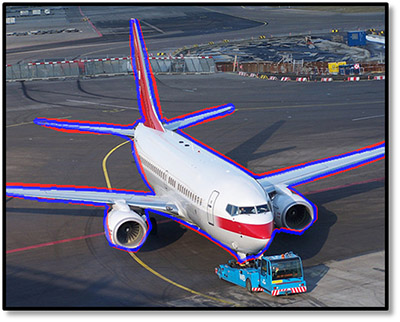}\label{Fig2c}
\caption{Evolution of edge alignment during training (progression from left to right). Blue color indicates the aligned edge labels learned by SEAL, while red color indicates the original human annotation. Overlapping pixels between the aligned edge labels and the original annotation are color-coded to be blue. Note how the aligned edge labels gradually tightens around the airplane as training progresses. Best viewed in color.}
\label{Fig2}
\end{figure}

Unfortunately, completely eliminating misalignment during annotation is almost impossible given the limit of human precision and the diminishing gain of annotation quality from additional efforts as a result. For datasets such as Cityscapes~\cite{cordts2016cityscapes} where high quality labels are generated by professional annotators, misalignment can still be frequently observed. For datasets with crowdsourcing annotations where quality control presents another challenge, the issue can become even more severe. Our proposed solution is an end-to-end framework towards Simultaneous Edge Alignment and Learning (SEAL). In particular, we formulate the problem with a probabilistic model, treating edge labels as latent variables to be jointly learned during training. We show that the optimization of latent edge labels can be transformed into a bipartite graph min-cost assignment problem, and present an end-to-end learning framework towards model training. Fig.~\ref{Fig2} shows some examples where the model gradually learns how to align noisy edge labels to more accurate positions along with edge learning.

Contrary to the widely believed intuition that reweighted loss benefits edge learning problems, an interesting and counter-intuitive observation made in this paper is that (regular) sigmoid cross-entropy loss works surprisingly well under the proposed framework despite the extremely imbalanced distribution. The underlying reason is that edge alignment significantly reduces the training confusion by increasing the purity of positive edge samples. Without edge alignment, on the other hand, the presence of label noise together with imbalanced distribution makes the model more difficult to correctly learn positive classes. As a result of the increased label quality and the benefit of better negative suppression using unweighted loss, our proposed framework produces state of the art detection performance with high quality sharp edges (see Fig.~\ref{Fig1d} and Fig.~\ref{Fig1h}).

\section{Related work}

\subsection{Boundary map correspondence}

Our work is partly motivated by the early work of boundary evaluation using precision-recall and F-measure~\cite{Martin2004}. To address misalignment between prediction and human ground truth, \cite{Martin2004} proposed to compute a one-to-one correspondence for the subset of matchable edge pixels from both domains by solving a min-cost assignment problem. However,~\cite{Martin2004} only considers the alignment between fixed boundary maps, while our work addresses a more complicated learning problem where edge alignment becomes part of the optimization with learnable inputs.

\subsection{Mask refinement via energy minimization}

Yang et al.~\cite{yang2016object} proposed to use dense-CRF to refine object mask and contour. Despite the similar goal, our method differs from~\cite{yang2016object} in that: 1. The refinement framework in~\cite{yang2016object} is a separate preprocessing step, while our work jointly learns refinement with the model in an end-to-end fashion. 2. The CRF model in~\cite{yang2016object} only utilizes low-level features, while our model considers both low-level and high-level information  via a deep network. 3. The refinement framework in~\cite{yang2016object} is segmentation-based, while our framework directly targets edge refinement.

\subsection{Object contour and mask learning}

A series of works~\cite{rupprecht2016deep,castrejon2017annotating,pinheiro2015learning} seek to learn object contours/masks in a supervised fashion. Deep active contour~\cite{rupprecht2016deep} uses learned CNN features to steer contour evolution given the input of an initialized contour. Polygon-RNN~\cite{castrejon2017annotating} introduced a semi-automatic approach for object mask annotation, by learning to extract polygons given input bounding boxes. DeepMask~\cite{pinheiro2015learning} proposed an object proposal generation method to output class-agnostic segmentation masks. These methods require accurate ground truth for contour/mask learning, while this work only assumes noisy ground truths and seek to refine them automatically.

\subsection{Noisy label learning}

Our work can be broadly viewed as a structured noisy label learning framework where we leverage abundant structural priors to correct label noise. Existing noisy label learning literatures have proposed directed graphical models~\cite{xiao2015learning}, conditional random fields (CRF)~\cite{vahdat2017toward}, neural networks~\cite{veit2017learning,wang2018iterative}, robust losses~\cite{patrini2017making} and knowledge graph~\cite{li2017learning} to model and correct image-level noisy labels. Alternatively, our work considers pixel-level labels instead of image-level ones.

\subsection{Virtual evidence in Bayesian networks}

Our work also shares similarity with virtual evidence~\cite{pearl1988probabilistic,bilmes2004virtual,liao2007training}, where the uncertainty of an observation is modeled by a distribution rather than a single value. In our problem, noisy labels can be regarded as uncertain observations which give conditional prior distributions over different configurations of aligned labels.

\section{A probabilistic view towards edge learning}

In many classification problems, training of the models can be formulated as maximizing the following likelihood function with respect to the parameters:
\begin{equation}\label{Eq:1}
\max_{\mathbf{W}}\mathcal{L}(\mathbf{W}) = P(\mathbf{y}|\mathbf{x}; \mathbf{W}),
\end{equation}
where $\mathbf{y}$, $\mathbf{x}$ and $\mathbf{W}$ indicate respectively training labels, observed inputs and model parameters. Depending on how the conditional probability is parameterized, the above likelihood function may correspond to
different types of models. For example, a generalized linear model function leads to the well known logistic regression. If the parameterization is formed as a layered representation, the model may turn into CNNs or multilayer perceptrons. One may observe that many traditional supervised edge learning models can also be regarded as special cases under the above probabilistic framework. Here, we are mostly concerned with edge detection using fully convolutional neural networks. In this case, the variable $\mathbf{y}$ indicates the set of edge prediction configurations at every pixel, while $\mathbf{x}$ and $\mathbf{W}$ denote the input image and the network parameters, respectively.

\section{Simultaneous edge alignment and learning}

To introduce the ability of correcting edge labels during training, we consider the following model. Instead of treating the observed annotation $\mathbf{y}$ as the fitting target, we assume there is an underlying ground truth $\hat{\mathbf{y}}$ that is more accurate than $\mathbf{y}$. Our goal is to treat $\hat{\mathbf{y}}$ as a latent variable to be jointly estimated during learning, which leads to the following likelihood maximization problem:
\begin{equation}\label{Eq:2}
\begin{split}
\max_{\hat{\mathbf{y}}, \mathbf{W}}\mathcal{L}(\hat{\mathbf{y}},\mathbf{W}) = P(\mathbf{y}, \hat{\mathbf{y}}|\mathbf{x}; \mathbf{W}) = P(\mathbf{y}|\hat{\mathbf{y}})P(\hat{\mathbf{y}}|\mathbf{x}; \mathbf{W}),\\
\end{split}
\end{equation}
where $\hat{\mathbf{y}}$ indicates the underlying true ground truth. The former part $P(\mathbf{y}|\hat{\mathbf{y}})$ can be regarded as an edge prior probabilistic model of an annotator generating labels given the observed ground truths, while the latter part $P(\hat{\mathbf{y}}|\mathbf{x}; \mathbf{W})$ is the standard likelihood of the prediction model.

\subsection{Multilabel edge learning}

Consider the multilabel edge learning setting where one assumes that $\mathbf{y}$ does not need to be mutually exclusive at each pixel. In other words, any pixel may correspond to the edges of multiple classes. The likelihood can be decomposed to a set of class-wise joint probabilities assuming the inter-class independence:
\begin{equation}\label{Eq:3}
\begin{split}
\mathcal{L}(\hat{\mathbf{y}},\mathbf{W}) = \prod_{k}P(\mathbf{y}^{k}|\hat{\mathbf{y}}^{k})P(\hat{\mathbf{y}}^{k}|\mathbf{x}; \mathbf{W}),\\
\end{split}
\end{equation}
where $\mathbf{y}^{k}\in\{0,1\}^N$ indicates the set of binary labels corresponding to the $k$-th class. A typical multilabel edge learning example which  alsoassumes inter-class independence is CASENet~\cite{yu2017casenet}. In addition, binary edge detection methods such as HED~\cite{Xie2015} can be viewed as special cases of multilabel edge learning.

\subsection{Edge prior model}

Solving Eq. (\ref{Eq:2}) is not easy given the additional huge search space of $\hat{\mathbf{y}}$. Fortunately, there is some prior knowledge one could leverage to effectively regularize $\hat{\mathbf{y}}$. One of the most important prior is that $\hat{\mathbf{y}}^{k}$ should not be too different from $\mathbf{y}^{k}$. In addition, we assume that edge pixels in $\mathbf{y}^{k}$ is generated from those in $\hat{\mathbf{y}}^{k}$ through an one-to-one assignment process, which indicates $|\mathbf{y}^{k}|=|\hat{\mathbf{y}}^{k}|$. In other words, let $y_{\mathbf{q}}^{k}$ denote the label of class $k$ at pixel $\mathbf{q}$, and similarly for $\hat{y}_{\mathbf{p}}^{k}$, there exists a set of one-to-one correspondences between edge pixels in $\hat{\mathbf{y}}^{k}$ and $\mathbf{y}^{k}$:
\begin{equation}\label{Eq:4}
\begin{split}
\mathcal{M}(\mathbf{y}^{k}, \hat{\mathbf{y}}^{k}) = &\{m(\cdot)|\forall \mathbf{u},\mathbf{v}\in\{\mathbf{q}|y_{\mathbf{q}}^{k}=1\}:\hat{y}_{m({\mathbf{u}})}^{k}=1,\\
&\hat{y}_{m({\mathbf{v}})}^{k}=1,\mathbf{u}\neq\mathbf{v} \Rightarrow m(\mathbf{u}) \neq m(\mathbf{v}) \}, \\
\end{split}
\end{equation}
where each $m(\cdot)$ is associated with a finite set of pairs:
\begin{equation}\label{Eq:5}
m(\cdot)\sim E_{m} = \{(\mathbf{p},\mathbf{q})|\hat{y}_{\mathbf{p}},y_{{\mathbf{q}}}=1, m(\mathbf{q})=\mathbf{p}\}.
\end{equation}
The edge prior therefore can be modeled as a product of Gaussian similarities maximized over all possible correspondences:
\begin{small}
\begin{equation}\label{Eq:6}
\begin{split}
P(\mathbf{y}^{k}|\hat{\mathbf{y}}^{k})& \propto \sup_{m\in\mathcal{M}(\mathbf{y}^{k},\hat{\mathbf{y}}^{k})}\prod_{(\mathbf{p},\mathbf{q})\in E_{m}}\exp\Big(-\frac{\|\mathbf{p}-\mathbf{q}\|^2}{2\sigma^2}\Big)\\
& =\exp\Big(-\inf_{m\in\mathcal{M}(\mathbf{y}^{k},\hat{\mathbf{y}}^{k})}\sum_{(\mathbf{p},\mathbf{q})\in E_{m}}\frac{\|\mathbf{p}-\mathbf{q}\|^2}{2\sigma^2}\Big), \\
\end{split}
\end{equation}
\end{small}
\hskip-1.2mm where $\sigma$ is the bandwidth that controls the sensitivity to misalignment. The misalignment is quantified by measuring the lowest possible sum of squared distances between pairwise pixels, which is determined by the tightest correspondence.

\subsection{Network likelihood model}

We now consider the likelihood of the prediction model, where we assume that the class-wise joint probability can be decomposed to a set of pixel-wise probabilities modeled by bernoulli distributions with binary configurations:
\begin{small}
\begin{equation}\label{Eq:7}
\begin{split}
P(\hat{\mathbf{y}}^{k}|\mathbf{x}; \mathbf{W}) = \prod_{\mathbf{p}}P(\hat{y_{\mathbf{p}}}^{k}|\mathbf{x};\mathbf{W})=\prod_{\mathbf{p}}h_{k}(\mathbf{p}|\mathbf{x};\mathbf{W})^{\hat{y}_{\mathbf{p}}^{k}}(1-h_{k}(\mathbf{p}|\mathbf{x};\mathbf{W}))^{(1-\hat{y}_{\mathbf{p}}^{k})}, \\
\end{split}
\end{equation}
\end{small}
\hskip-1.2mm where $\mathbf{p}$ is the pixel location index, and $h_k$ is the hypothesis function indicating the probability of the $k$-th class. We consider the prediction model as FCNs with $k$ sigmoid outputs. As a result, the hypothesis function in Eq. (\ref{Eq:7}) becomes the sigmoid function, which will be denoted as $\sigma(\cdot)$ in the rest part of this section.

\subsection{Learning}

Taking Eq. (\ref{Eq:6}) and (\ref{Eq:7}) into Eq. (\ref{Eq:3}), and taking log of the likelihood, we have:
\begin{small}
\begin{equation}\label{Eq:8}
\begin{split}
\log\mathcal{L}(\hat{\mathbf{y}},\mathbf{W})=&\sum_{k} \Big\{-\inf_{m\in\mathcal{M}(\mathbf{y}^{k},\hat{\mathbf{y}}^{k})}\sum_{(\mathbf{p},\mathbf{q})\in E_{m}}\frac{\|\mathbf{p}-\mathbf{q}\|^2}{2\sigma^2}\\
&+\sum_{\mathbf{p}}\Big[\hat{y}_{\mathbf{p}}^{k}\log\sigma_{k}(\mathbf{p}|\mathbf{x};\mathbf{W})+(1-\hat{y}_{\mathbf{p}}^{k})\log(1-\sigma_{k}(\mathbf{p}|\mathbf{x};\mathbf{W}))\Big]\Big\}, \\
\end{split}
\end{equation}
\end{small}
\hskip-1.2mm where the second part is the widely used sigmoid cross-entropy loss. Accordingly, learning the model requires solving the constrained optimization:
\begin{equation}\label{Eq:9}
\begin{split}
\min_{\hat{\mathbf{y}},\mathbf{W}}&~-\log\mathcal{L}(\hat{\mathbf{y}},\mathbf{W})\\
\mathrm{s.t.}&~~|\hat{\mathbf{y}}^{k}| = |\mathbf{y}^{k}|, \forall k
\end{split}
\end{equation}
Given a training set, we take an alternative optimization strategy where $\mathbf{W}$ is updated with $\hat{\mathbf{y}}$ fixed, and vice versa. When $\hat{\mathbf{y}}$ is fixed, the optimization becomes:
\begin{small}
\begin{equation}\label{Eq:10}
\begin{split}
\min_{\mathbf{W}}~\sum_{k}\sum_{\mathbf{p}}-\Big[\hat{y}_{\mathbf{p}}^{k}\log\sigma_{k}(\mathbf{p}|\mathbf{x};\mathbf{W})+(1-\hat{y}_{\mathbf{p}}^{k})\log(1-\sigma_{k}(\mathbf{p}|\mathbf{x};\mathbf{W}))\Big],\\
\end{split}
\end{equation}
\end{small}
\hskip-1.2mm which is the typical network training with the aligned edge labels and can be solved with standard gradient descent. When $\mathbf{W}$ is fixed, the optimization can be modeled as a constrained discrete optimization problem for each class:
\begin{small}
\begin{equation}\label{Eq:11}
\begin{split}
&\min_{\hat{\mathbf{y}}^{k}}~\inf_{m\in\mathcal{M}(\mathbf{y}^{k},\hat{\mathbf{y}}^{k})}\sum_{(\mathbf{p},\mathbf{q})\in E_{m}}\frac{\|\mathbf{p}-\mathbf{q}\|^2}{2\sigma^2} \\
&~~~~~~~~~-
\sum_{\mathbf{p}}\Big[\hat{y}_{\mathbf{p}}^{k}\log\sigma_{k}(\mathbf{p})+(1-\hat{y}_{\mathbf{p}}^{k})\log(1-\sigma_{k}(\mathbf{p}))\Big]\\
&~~\mathrm{s.t.}~~|\hat{\mathbf{y}}^{k}| = |\mathbf{y}^{k}|\\
\end{split}
\end{equation}
\end{small}
\hskip-1.2mm where $\sigma(\mathbf{p})$ denotes $\sigma(\mathbf{p}|\mathbf{x};\mathbf{W})$ for short.
Solving the above optimization is seemingly difficult, since one would need to enumerate all possible configurations of $\hat{\mathbf{y}}^k$ satisfying $|\hat{\mathbf{y}}^{k}| = |\mathbf{y}^{k}|$ and evaluate the associated cost. It turns out, however, that the above optimization can be elegantly transformed to a bipartite graph assignment problem with available solvers. We first have the following definition:
\begin{definition}\label{Def:7-1}
Let $\hat{\mathbf{Y}}=\{\hat{\mathbf{y}}||\hat{\mathbf{y}}|=|\mathbf{y}|\}$, a mapping space $\mathbf{M}$ is the space consisting all possible one-to-one mappings:
\begin{small}
\begin{displaymath}
\mathbf{M} = \{m|m\in\mathcal{M}(\mathbf{y},\hat{\mathbf{y}}), \hat{\mathbf{y}}\in \hat{\mathbf{Y}}\}
\end{displaymath}
\end{small}
\end{definition}
\begin{definition}\label{Def:7-2}
A label realization is a function which maps a correspondence to the corresponding label given :
\begin{displaymath}
\begin{split}
f_{L}:&\mathbf{Y}\times\mathbf{M}\mapsto \hat{\mathbf{Y}}\\
&f_{L}(\mathbf{y}, m) = \hat{\mathbf{y}}
\end{split}
\end{displaymath}
\end{definition}

\begin{lemma}\label{Lemma7-1}
The mapping $f_{L}(\cdot)$ is surjective.
\end{lemma}
\noindent\textbf{Remark:} Lemma \ref{Lemma7-1} shows that a certain label configuration $\hat{\mathbf{y}}$ may correspond to multiple underlying mappings. This is obviously true since there could be multiple ways in which pixels in $\mathbf{y}$ are assigned to the $\hat{\mathbf{y}}$.

\begin{lemma}\label{Lemma7-2}
Under the constraint $|\hat{\mathbf{y}}|=|\mathbf{y}|$, if:
\begin{small}
\begin{displaymath}
\begin{split}
&\hat{\mathbf{y}}^{*}=\argmin_{\hat{\mathbf{y}}}-\sum_{\mathbf{p}}\Big[\hat{y}_{\mathbf{p}}\log\sigma(\mathbf{p})+(1-\hat{y}_{\mathbf{p}})\log(1-\sigma(\mathbf{p})\Big]\\
&m^{*}=\argmin_{m\in\mathbf{M}}\sum_{(\mathbf{p},\mathbf{q})\in E_{m}}\Big[\log(1-\sigma(\mathbf{p}))-\log\sigma(\mathbf{p})\Big]\\
\end{split}
\end{displaymath}
\end{small}
then $f_{L}(\mathbf{y},m^{*})=\hat{\mathbf{y}}^{*}$.
\end{lemma}
\noindent\textbf{Proof:} Suppose in the beginning all pixels in $\hat{\mathbf{y}}$ are 0. The corresponding loss therefore is:
\begin{small}
\begin{displaymath}
\mathcal{C}_{N}(\mathbf{0})=-\sum_{\mathbf{p}}\log(1-\sigma(\mathbf{p}))
\end{displaymath}
\end{small}
\hskip-1.2mm Flipping $y_{\mathbf{p}}$ to 1 will accordingly introduce a cost $\log(1-\sigma(\mathbf{p}))-\log\sigma(\mathbf{p})$ at pixel $\mathbf{p}$. As a result, we have:
\begin{small}
\begin{displaymath}
\mathcal{C}_{N}(\hat{\mathbf{y}})= \mathcal{C}_{N}(\mathbf{0}) + \sum_{\mathbf{p}\in\{\mathbf{p}|\hat{y}_{\mathbf{p}}=1\}}\Big[\log(1-\sigma(\mathbf{p}))-\log\sigma(\mathbf{p})\Big]
\end{displaymath}
\end{small}
\hskip-1.2mm In addition, Lemma \ref{Lemma7-1} states that the mapping $f_{L}(\cdot)$ is surjective, which incites that the mapping search space $\mathbf{M}$ exactly covers $\hat{\mathbf{Y}}$. Thus the top optimization problem in Lemma \ref{Lemma7-2} can be transformed into the bottom problem.

Lemma \ref{Lemma7-2} motivates us to reformulate the optimization in Eq. (\ref{Eq:11}) by alternatively looking to the following problem:
\begin{small}
\begin{equation}\label{Eq:12}
\min_{m\in\mathbf{M}} \sum_{(\mathbf{p},\mathbf{q})\in E_{m}}\Big[\frac{\|\mathbf{p}-\mathbf{q}\|^2}{2\sigma^2}+\log(1-\sigma(\mathbf{p}))-\log\sigma(\mathbf{p})\Big]
\end{equation}
\end{small}

Eq. (\ref{Eq:12}) is a typical minimum cost bipartite assignment problem which can be solved by standard solvers, where the cost of each assignment pair $(\mathbf{p},\mathbf{q})$ is associated with the weight of a bipartite graphos edge. Following~\cite{Martin2004}, we formulate a sparse assignment problem and use the Goldbergos CSA package, which is the best known algorithms for min-cost sparse assignment~\cite{goldberg1995efficient,cherkassky1995implementing}. Upon obtaining the mapping, one can recover $\hat{\mathbf{y}}$ through label realization.

However, solving Eq. (\ref{Eq:12}) assumes an underlying relaxation where the search space contains $m$ which may not follow the infimum requirement in Eq. (\ref{Eq:11}). In other words, it may be possible that the minimization problem in Eq. (\ref{Eq:12}) is an approximation to Eq. (\ref{Eq:11}). The following theorem, however, proves the optimality of Eq. (\ref{Eq:12}):
\begin{theorem}\label{Theorem7-1}
Given a solver that minimizes Eq. (\ref{Eq:12}), the solution is also a minimizer of the problem in Eq. (\ref{Eq:11}).
\end{theorem}
\noindent\textbf{Proof:} We use contradiction to prove Theorem \ref{Theorem7-1}. Suppose there exists a solution of (\ref{Eq:12}) where:
\begin{small}
\begin{displaymath}
f_{L}(\mathbf{y},m^{*})=\hat{\mathbf{y}}, ~m^{*}\neq \argmin_{m\in\mathcal{M}(\mathbf{y}^{k},\hat{\mathbf{y}}^{k})}\sum_{(\mathbf{p},\mathbf{q})\in E_{m}}\frac{\|\mathbf{p}-\mathbf{q}\|^2}{2\sigma^2}
\end{displaymath}
\end{small}
\hskip-1.2mm There must exist another mapping $m'$ which satisfies:
\begin{small}
\begin{displaymath}
f_{L}(\mathbf{y},m')=\hat{\mathbf{y}}, \sum_{(\mathbf{p},\mathbf{q})\in E_{m'}}\frac{\|\mathbf{p}-\mathbf{q}\|^2}{2\sigma^2} < \sum_{(\mathbf{p},\mathbf{q})\in E_{m^{*}}}\frac{\|\mathbf{p}-\mathbf{q}\|^2}{2\sigma^2}
\end{displaymath}
\end{small}
\hskip-1.2mm Since $f_{L}(\mathbf{y},m')=f_{L}(\mathbf{y},m^{*})=\hat{\mathbf{y}}$, substituting $m'$ to (\ref{Eq:12}) leads to an even lower cost, which contradicts to the assumption that $m^{*}$ is the minimizer of (\ref{Eq:12}).

In practice, we follow the mini-batch SGD where $\hat{\mathbf{y}}$ of each image in the batch is updated via solving edge alignment, followed by updating $\mathbf{W}$ with gradient descent. $\hat{\mathbf{y}}$ is initialized as $\mathbf{y}$ in the first batch. The optimization can be written as a loss layer in a network, and is fully compatible with end-to-end training.

\subsection{Inference}
We now consider the inference problem given a trained model. Ideally, the inference problem of the model trained by Eq. (\ref{Eq:2}) would be the following:
\begin{equation}\label{Eq:13}
\hat{\mathbf{y}}^{*} = \argmax_{\hat{\mathbf{y}}}P(\mathbf{y}|\hat{\mathbf{y}})P(\hat{\mathbf{y}}|\mathbf{x}; \mathbf{W})
\end{equation}
However, in cases where $\mathbf{y}$ is not available during testing. we can alternatively look into the second part of (\ref{Eq:2}) which is the model learned under $\hat{\mathbf{y}}$:
\begin{equation}\label{Eq:14}
\hat{\mathbf{y}}^{*} = \argmax_{\hat{\mathbf{y}}}P(\hat{\mathbf{y}}|\mathbf{x}; \mathbf{W})
\end{equation}
Both cases can find real applications. In particular, (\ref{Eq:14}) corresponds to general edge prediction, whereas (\ref{Eq:13}) corresponds to refining noisy edge labels in a dataset. In the latter case, $\mathbf{y}$ is available and the inferred $\hat{\mathbf{y}}$ is used to output the refined label. In the experiment, we will show examples of both applications.

\section{Biased Gaussian kernel and Markov prior}
\begin{figure}[t]
\centering
\subfigure[Isotropic Gaussian]{\label{Fig3a}\includegraphics[height=.24\textwidth]{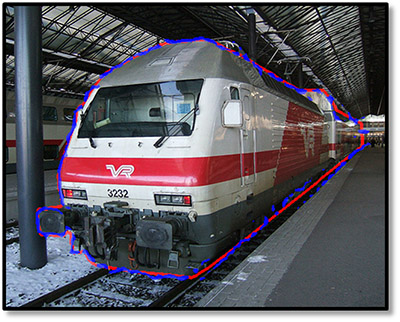}}
\subfigure[B.G.+MRF]{\label{Fig3c}\includegraphics[height=.24\textwidth]{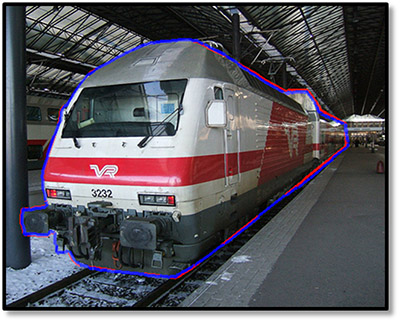}}
\subfigure[Illustration]{\label{Fig3d}\includegraphics[height=.24\textwidth]{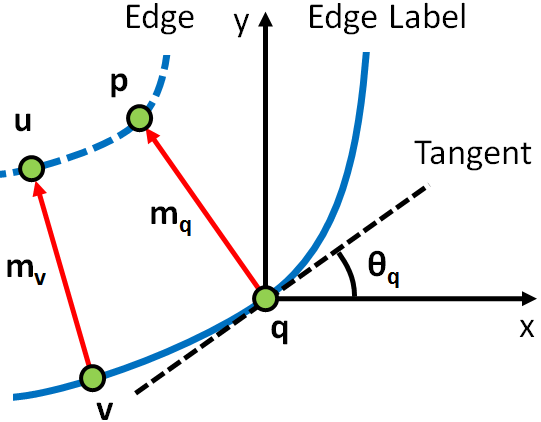}}
\caption{Examples of edge alignment using different priors and graphical illustration.}
\label{Fig3}
\end{figure}

The task of SEAL turns out not easy, as it tends to generate artifacts upon having cluttered background. A major reason causing this failure is the fragmented aligned labels, as shown in Fig. \ref{Fig3a}. This is not surprising since we assume an isotropic Gaussian kernel, where labels tend to break and shift along the edges towards easy locations. In light of this issue, we assume that the edge prior follows a biased Gaussian (B.G.), with the long axis of the kernel perpendicular to local boundary tangent. Accordingly, such model encourages alignment perpendicular to edge tangents while suppressing shifts along them.

Another direction is to consider the Markov properties of edges. Good edge labels should be relatively continuous, and nearby alignment vectors should be similar. Taking these into consideration, we can model the edge prior as:
\begin{small}
\begin{equation}\label{Eq:15}
\begin{split}
P(\mathbf{y}|\hat{\mathbf{y}}) \propto &\sup_{m\in\mathcal{M}(\mathbf{y},\hat{\mathbf{y}})}\prod_{(\mathbf{p},\mathbf{q})\in E_{m}}\exp(-\mathbf{m}_{\mathbf{q}}^{\top}\mathbf{\Sigma_{\mathbf{q}}}\mathbf{m}_{\mathbf{q}})\prod_{\substack{(\mathbf{u},\mathbf{v})\in E_{m},\\ \mathbf{v}\in \mathcal{N}(\mathbf{q})}}\exp(-\lambda \|\mathbf{m}_{\mathbf{q}}-\mathbf{m}_{\mathbf{v}}\|^{2})\\
\end{split}
\end{equation}
\end{small}
\hskip-1.2mm where $\lambda$ controls the strength of the smoothness. $\mathcal{N}(\mathbf{q})$ is the neighborhood of $\mathbf{q}$ defined by the geodesic distance along the edge. $\mathbf{m}_{\mathbf{q}} = \mathbf{p}-\mathbf{q}$, and $\mathbf{m}_{\mathbf{v}} = \mathbf{u}-\mathbf{v}$. An example of the improved alignment and a graphical illustration are shown in Fig. \ref{Fig3c} and Fig. \ref{Fig3d}. In addition, the precision matrix $\mathbf{\Sigma}_{\mathbf{q}}$ is defined as:
\[
\mathbf{\Sigma}_{\mathbf{q}} =
\begin{bmatrix}
    \frac{\cos(\theta_{\mathbf{q}})^2}{2\sigma_{x}^2} + \frac{\sin(\theta_{\mathbf{q}})^2}{2\sigma_{y}^2} & \frac{\sin(2\theta_{\mathbf{q}})}{4\sigma_{y}^2} - \frac{\sin(2\theta_{\mathbf{q}})}{4*\sigma_{x}^2}\\
    \frac{\sin(2\theta_{\mathbf{q}})}{4\sigma_{y}^2} - \frac{\sin(2\theta_{\mathbf{q}})}{4\sigma_{x}^2} & \frac{\sin(\theta_{\mathbf{q}})^2}{2\sigma_{x}^2} + \frac{\cos(\theta_{\mathbf{q}})^2}{2\sigma_{y}^2} \\
\end{bmatrix}
\]
where $\theta_{\mathbf{q}}$ is the angle between edge tangent and the positive x-axis, and $\sigma_y$ corresponds to the kernel bandwidth perpendicular to the edge tangent. With the new prior, the alignment optimization becomes the following problem:
\begin{small}
\begin{equation}\label{Eq:16}
\begin{split}
\min_{m\in\mathbf{M}} ~\mathcal{C}(m) = &~\mathcal{C}_{Unary}(m) + \mathcal{C}_{Pair}(m)\\
= &\sum_{(\mathbf{p},\mathbf{q})\in E_{m}}\Big[\mathbf{m}_{\mathbf{q}}^{\top}\mathbf{\Sigma_{\mathbf{q}}}\mathbf{m}_{\mathbf{q}}+\log((1-\sigma(\mathbf{p}))/\sigma(\mathbf{p}))\Big]\\
&~~~~~+\lambda\sum_{(\mathbf{p},\mathbf{q})\in E_{m}}\sum_{\substack{(\mathbf{u},\mathbf{v})\in E_{m},\\ \mathbf{v}\in \mathcal{N}(\mathbf{q})}}\|\mathbf{m}_{\mathbf{q}}-\mathbf{m}_{\mathbf{v}}\|^{2}\\
\end{split}
\end{equation}
\end{small}

Note that Theorem \ref{Theorem7-1} still holds for (\ref{Eq:16}). However, solving (\ref{Eq:16}) becomes more difficult as pairwise dependencies are included. As a result, standard assignment solvers can not be directly applied, and we alternatively decouple $\mathcal{C}_{Pair}$ as:
\begin{small}
\begin{equation}
\mathcal{C}_{Pair}(m, m') = \sum_{(\mathbf{p},\mathbf{q})\in E_{m}}\sum_{\substack{(\mathbf{u},\mathbf{v})\in E_{m'},\\ \mathbf{v}\in \mathcal{N}(\mathbf{q})}}\|\mathbf{m}_{\mathbf{q}}-\mathbf{m}_{\mathbf{v}}\|^{2}
\end{equation}
\end{small}
\hskip-1.2mm and take the iterated conditional mode like iterative approximation where the alignment of neighboring pixels are taken from the alignment in previous round:
\begin{small}
\begin{displaymath}
\begin{split}
\textbf{Initialize: }&m^{(0)} = \argmin_{m\in\mathbf{M}}~\mathcal{C}_{Unary}(m)\\
\textbf{Assign: }&m^{(t+1)} = \argmin_{m\in\mathbf{M}}~\mathcal{C}_{Unary}(m)+\mathcal{C}_{Pair}(m, m^{(t)})\\
\textbf{Update: }& \mathcal{C}_{Pair}(m, m^{(t)})\rightarrow\mathcal{C}_{Pair}(m, m^{(t+1)})\\
\end{split}
\end{displaymath}
\end{small}
\hskip-1.2mm where the \textbf{Assign} and \textbf{Update} steps are repeated multiple times. The algorithm converges very fast in practice. Usually two or even one \textbf{Assign} is sufficient.

\section{Experimental results}
In this section, we comprehensively test the performance of SEAL on category-ware semantic edge detection, where the detector not only needs to localize object edges, but also classify to a predefined set of semantic classes.

\subsection{Backbone network}
In order to guarantee fair comparison across different methods, a fixed backbone network is needed for controlled evaluation. We choose CASENet~\cite{yu2017casenet} since it is the current state of the art on our task. For additional implementation details such as choice of hyperparameters, please refer to the appendix.

\subsection{Evaluation benchmarks}
We follow~\cite{Hariharan2011} to evaluate edges with class-wise precision recall curves. However, the benchmarks of our work differ from~\cite{Hariharan2011} by imposing considerably stricter rules. In particular: 1. We consider non-suppressed edges inside an object as false positives, while \cite{Hariharan2011} ignores these pixels. 2. We accumulate false positives on any image, while the benchmark code from~\cite{Hariharan2011} only accumulates false positives of a certain class on images containing that class. Our benchmark can also be regarded as a multiclass extension of the BSDS benchmark~\cite{Martin2004}.

Both~\cite{Hariharan2011} and~\cite{Martin2004} by default thin the prediction before matching. We propose to match the raw predictions with unthinned ground truths whose width is kept the same as training labels. The benchmark therefore also considers the local quality of predictions. We refer to this mode as ``Raw'' and the previous conventional mode as ``Thin''. Similar to~\cite{Martin2004}, both settings use maximum F-Measure (MF) at optimal dataset scale (ODS) to evaluate the performance.

Another difference between the problem settings of our work and~\cite{Hariharan2011} is that we consider edges between any two instances as positive, even though the instances may belong to the same class. This differs from~\cite{Hariharan2011} where such edges are ignored. Our motivation on making such changes is two fold: 1. We believe instance-sensitive edges are important and it makes better sense to distinguish these locations. 2. The instance-sensitive setting may better benefit other potential applications where instances need to be distinguished.

\subsection{Experiment on the SBD dataset}
The Semantic Boundary Dataset (SBD)~\cite{Hariharan2011} contains 11355 images from the trainval set of PASCAL VOC2011~\cite{pascal-voc-2011}, with 8498 images divided as training set and 2857 images as test set. The dataset contains both category-level and instance-level semantic segmentation annotations, with semantic classes defined following the 20-class definitions in PASCAL VOC.

\subsubsection{Parameter analysis}
We set $\sigma_x=1$ and $\sigma_y > \sigma_x$ to favor alignment perpendicular to edge tangents. In addition, $\sigma_y$ and $\lambda$ are validated to determine their values. More details of the validation can be found in the appendix.

\begin{table*}[t!]
\centering
\caption{MF scores on the SBD test set. Results are measured by $\%$.\label{tb:sbd}}
\resizebox{\textwidth}{!}{\begin{tabular}{c|l|c|c|c|c|c|c|c|c|c|c|c|c|c|c|c|c|c|c|c|c|c}
Metric & Method	& aero & bike & bird & boat & bottle & bus & car & cat & chair & cow & table & dog & horse & mbike & person & plant & sheep & sofa & train & tv & mean \\
\hline \hline
\multirow{4}{0.1\linewidth}{\centering{MF\\(Thin)}}
& CASENet   & 83.6 & 75.3 & 82.3 & 63.1 & 70.5 & 83.5 & 76.5 & 82.6 & 56.8 & 76.3 & 47.5 & 80.8 & 80.9 & 75.6 & 80.7 & 54.1 & 77.7 & 52.3 & 77.9 & 68.0 & 72.3\\
& CASENet-S & \textbf{84.5} & \textbf{76.5} & \textbf{83.7} & \textbf{65.3} & 71.3 & \textbf{83.9} & \textbf{78.3} & 84.5 & \textbf{58.8} & 76.8 & 50.8 & 81.9 & \textbf{82.3} & \textbf{77.2} & 82.7 & \textbf{55.9} & 78.1 & 54.0 & \textbf{79.5} & 69.4 & \textbf{73.8}\\
& CASENet-C & 83.9 & 71.1 & 82.5 & 62.6 & 71.0 & 82.2 & 76.8 & 83.4 & 56.5 & \textbf{76.9} & 49.2 & 81.0 & 81.1 & 75.4 & 81.4 & 54.0 & \textbf{78.5} & 53.3 & 77.1 & 67.0 & 72.2\\
& SEAL      & \textbf{84.5} & \textbf{76.5} & \textbf{83.7} & 64.9 & \textbf{71.7} & 83.8 & 78.1 & \textbf{85.0} & \textbf{58.8} & 76.6 & \textbf{50.9} & \textbf{82.4} & 82.2 & 77.1 & \textbf{83.0} & 55.1 & 78.4 & \textbf{54.4} & 79.3 & \textbf{69.6} & \textbf{73.8}\\
\hline \hline
\multirow{4}{0.1\linewidth}{\centering{MF\\(Raw)}}
& CASENet   & 71.8 & 60.2 & 72.6 & 49.5 & 59.3 & 73.3 & 65.2 & 70.8 & 51.9 & 64.9 & 41.2 & 67.9 & 72.5 & 64.1 & 71.2 & 44.0 & 71.7 & 45.7 & 65.4 & 55.8 & 62.0\\
& CASENet-S & 75.8 & 65.0 & 78.4 & 56.2 & 64.7 & 76.4 & 71.8 & 75.2 & 55.2 & 68.7 & 45.8 & 72.8 & 77.0 & 68.1 & 76.5 & 47.1 & 75.5 & 49.0 & 70.2 & 60.6 & 66.5\\
& CASENet-C & 80.4 & 67.1 & 79.9 & 57.9 & 65.9 & 77.6 & 72.6 & 79.2 & 53.5 & 72.7 & 45.5 & 76.7 & 79.4 & 71.2 & 78.3 & \textbf{50.8} & 77.6 & 50.7 & 71.6 & 61.6 & 68.5\\
& SEAL      & \textbf{81.1} & \textbf{69.6} & \textbf{81.7} & \textbf{60.6} & \textbf{68.0} & \textbf{80.5} & \textbf{75.1} & \textbf{80.7} & \textbf{57.0} & \textbf{73.1} & \textbf{48.1} & \textbf{78.2} & \textbf{80.3} & \textbf{72.1} & \textbf{79.8} & 50.0 & \textbf{78.2} & \textbf{51.8} & \textbf{74.6} & \textbf{65.0} & \textbf{70.3}\\
\end{tabular}}
\end{table*}

\begin{table*}[t!]
\centering
\caption{MF scores on the re-annotated SBD test set. Results are measured by $\%$.\label{tb:sbd2}}
\resizebox{\textwidth}{!}{\begin{tabular}{c|l|c|c|c|c|c|c|c|c|c|c|c|c|c|c|c|c|c|c|c|c|c}
Metric & Method	& aero & bike & bird & boat & bottle & bus & car & cat & chair & cow & table & dog & horse & mbike & person & plant & sheep & sofa & train & tv & mean \\
\hline \hline
\multirow{4}{0.1\linewidth}{\centering{MF\\(Thin)}}
& CASENet   & 74.5 & 59.7 & 73.4 & 48.0 & 67.1 & 78.6 & 67.3 & 76.2 & 47.5 & 69.7 & 36.2 & 75.7 & 72.7 & 61.3 & 74.8 & 42.6 & 71.8 & 48.9 & 71.7 & 54.9 & 63.6\\
& CASENet-S & 75.9 & 62.4 & 75.5 & 52.0 & 66.7 & 79.7 & \textbf{71.0} & 79.0 & \textbf{50.1} & 70.0 & 39.8 & 77.2 & 74.5 & 65.0 & 77.0 & 47.3 & 72.7 & 51.5 & 72.9 & 57.3 & 65.9\\
& CASENet-C & \textbf{78.4} & 60.9 & 74.9 & 49.7 & 64.4 & 75.8 & 67.2 & 77.1 & 48.2 & 71.2 & 40.9 & 76.1 & 72.9 & 64.5 & 75.9 & \textbf{51.4} & 71.3 & 51.6 & 68.6 & 55.4 & 64.8\\
& SEAL      & 78.0 & \textbf{65.8} & \textbf{76.6} & \textbf{52.4} & \textbf{68.6} & \textbf{80.0} & 70.4 & \textbf{79.4} & 50.0 & \textbf{72.8} & \textbf{41.4} & \textbf{78.1} & \textbf{75.0} & \textbf{65.5} & \textbf{78.5} & 49.4 & \textbf{73.3} & \textbf{52.2} & \textbf{73.9} & \textbf{58.1} & \textbf{67.0}\\
\hline \hline
\multirow{4}{0.1\linewidth}{\centering{MF\\(Raw)}}
& CASENet   & 65.8 & 51.5 & 65.0 & 43.1 & 57.5 & 68.1 & 58.2 & 66.0 & 45.4 & 59.8 & 32.9 & 64.2 & 65.8 & 52.6 & 65.7 & 40.9 & 65.0 & 42.9 & 61.4 & 47.8 & 56.0\\
& CASENet-S & 68.9 & 55.8 & 70.9 & 47.4 & 62.0 & 71.5 & 64.7 & 71.2 & 48.0 & 64.8 & 37.3 & 69.1 & 68.9 & 58.2 & 70.2 & 44.3 & 68.7 & 46.1 & 65.8 & 52.5 & 60.3\\
& CASENet-C & \textbf{75.4} & 57.7 & 73.0 & 48.7 & 62.1 & 72.2 & 64.4 & 74.3 & 46.8 & 68.8 & 38.8 & 73.4 & 71.4 & \textbf{62.2} & 72.1 & \textbf{50.3} & 69.8 & 48.4 & 66.1 & 53.0 & 62.4\\
& SEAL      & 75.3 & \textbf{60.5} & \textbf{75.1} & \textbf{51.2} & \textbf{65.4} & \textbf{76.1} & \textbf{67.9} & \textbf{75.9} & \textbf{49.7} & \textbf{69.5} & \textbf{39.9} & \textbf{74.8} & \textbf{72.7} & 62.1 & \textbf{74.2} & 48.4 & \textbf{72.3} & \textbf{49.3} & \textbf{70.6} & \textbf{56.7} & \textbf{64.4}\\
\end{tabular}}
\end{table*}

\subsubsection{Results on the SBD test set}
We compare SEAL with CASENet, CASENet trained with regular sigmoid cross-entropy loss (CASENet-S), and CASENet-S trained on labels refined by dense-CRF following~\cite{yang2016object} (CASENet-C), with the results visualized in Fig. \ref{sbd} and quantified in Table~\ref{tb:sbd}. Results show that SEAL is on par with CASENet-S under the ``Thin'' setting, while significantly outperforms all other baselines when edge sharpness is taken into account.

\subsubsection{Results on the re-annotated SBD test set}
A closer analysis shows that SEAL actually outperforms CASENet-S considerably under the ``Thin'' setting. The original SBD labels turns out to be noisy, which can influence the validity of evaluation. We re-annotated more than 1000 images on SBD test set using LabelMe~\cite{russell2008labelme}, and report evaluation using these high-quality labels in Table~\ref{tb:sbd2}. Results indicates that SEAL outperforms CASENet-S in both settings.

\setlength{\intextsep}{0pt}
\begin{wrapfigure}{r}{0.3\textwidth}
\centering
\includegraphics[width=\linewidth]{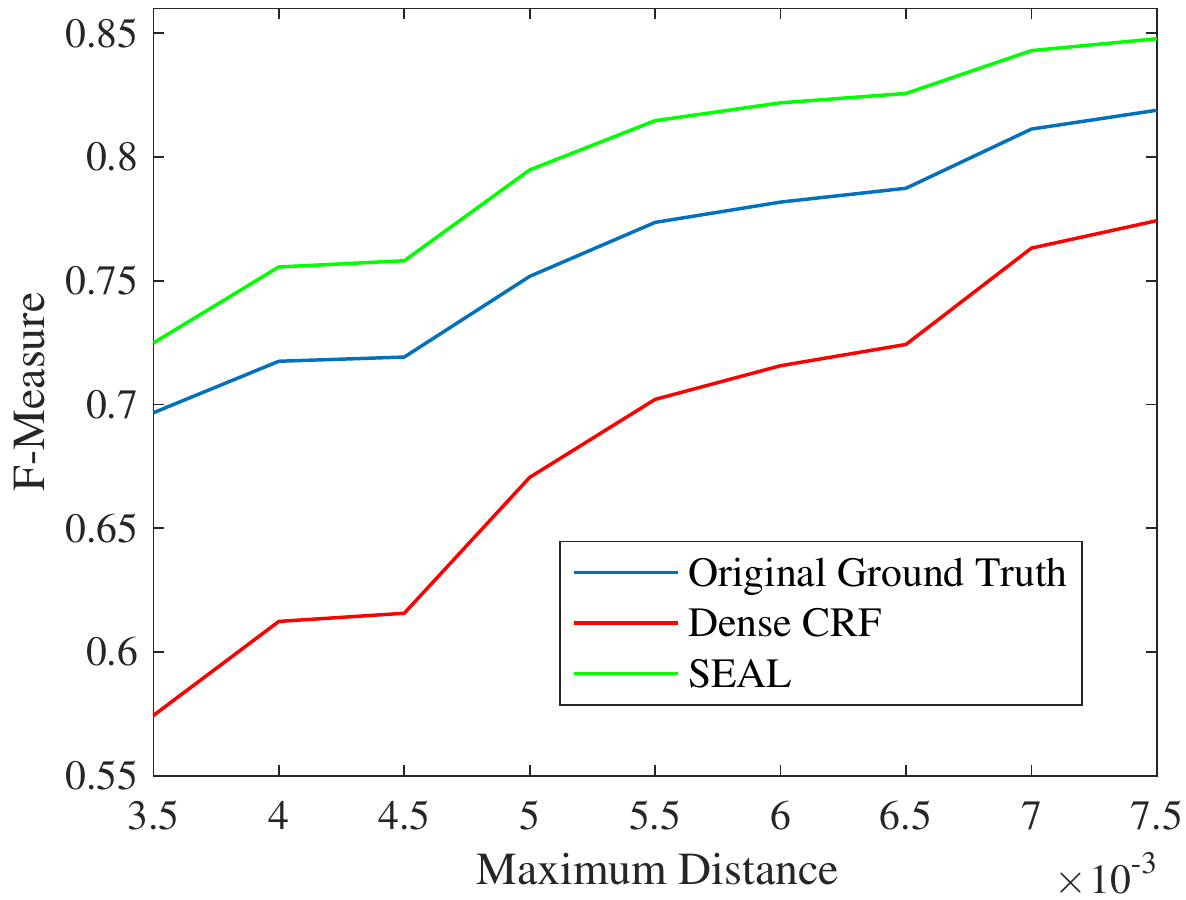}
\caption{MF vs. tolerance.}\label{align}
\end{wrapfigure}
\subsubsection{Results of SBD GT refinement}
We output the SEAL aligned labels and compare against both dense-CRF and original annotation. We match the aligned labels with re-annotated labels by varying the tolerance threshold and generating F-Measure scores. Fig.~\ref{align} shows that SEAL indeed can improve the label quality, while dense-CRF performs even worse than original labels. In fact, the result of CASENet-C also indicates the decreased model performance.

\setlength{\intextsep}{0pt}
\begin{wraptable}{r}{0.3\textwidth}
\centering
\caption{Non-IS results.\label{tb:non_is}}
\resizebox{\linewidth}{!}{\begin{tabular}{c|c|c|c|c}
Mode & CNet & CNet-S & CNet-C & SEAL\\
\hline \hline
\multirow{1}{0.3\linewidth}{\centering{Thin}}
& 63.6 & 66.4 & 64.7 & \textbf{66.9} \\
\hline
\multirow{1}{0.3\linewidth}{\centering{Raw}}
& 56.1 & 60.6 & 62.1 & \textbf{64.6} \\
\end{tabular}}
\end{wraptable}
\subsubsection{Non-Instance-insensitive (non-IS) mode}
We also train/evaluate under non-IS mode, with the evaluation using re-annotated SBD labels. Table \ref{tb:non_is} shows that the scores have high correlation with IS mode.

\subsubsection{Comparison with state of the art}
Although proposing different evaluation criteria, we still follow~\cite{yu2017casenet} by training SEAL with instance-insensitive labels and evaluating with the same benchmark and ground truths. Results in Table \ref{sota} show that this work outperforms previous state of the art by a significant margin.

\begin{table*}[t!]
\centering
\caption{Results on SBD test following the same benchmark and ground truths as~\cite{yu2017casenet}.\label{sota}}
\resizebox{\textwidth}{!}{\begin{tabular}{c|c|c|c|c|c|c|c|c|c|c|c|c|c|c|c|c|c|c|c|c|c}
Method & aero & bike & bird & boat & bottle & bus & car & cat & chair & cow & table & dog & horse & mbike & person & plant & sheep & sofa & train & tv & mean\\
\hline \hline
\cite{yu2017casenet} & 83.3 & 76.0 & 80.7 & 63.4 & 69.2 & 81.3 & 74.9 & 83.2 & 54.3 & 74.8 & 46.4 & 80.3 & 80.2 & 76.6 & 80.8 & 53.3 & 77.2 & 50.1 & 75.9 & 66.8 & 71.4\\
SEAL & \textbf{84.9} & \textbf{78.6} & \textbf{84.6} & \textbf{66.2} & \textbf{71.3} & \textbf{83.0} & \textbf{76.5} & \textbf{87.2} & \textbf{57.6} & \textbf{77.5} & \textbf{53.0} & \textbf{83.5} & \textbf{82.2} & \textbf{78.3} & \textbf{85.1} & \textbf{58.7} & \textbf{78.9} & \textbf{53.1} & \textbf{77.7} & \textbf{69.7} & \textbf{74.4}\\
\end{tabular}}
\end{table*}

\begin{table*}[t!]
\centering
\caption{MF scores on the Cityscapes validation set. Results are measured by $\%$.\label{tb:cityscapes}}
\resizebox{\textwidth}{!}{\begin{tabular}{c|l|c|c|c|c|c|c|c|c|c|c|c|c|c|c|c|c|c|c|c|c}
Metric & Method	& road & sidewalk & building & wall & fence & pole & t-light & t-sign & veg & terrain & sky & person & rider & car & truck & bus & train & motor & bike & mean \\
\hline \hline
\multirow{3}{0.1\linewidth}{\centering{MF\\(Thin)}}
& CASENet   & 86.2 & 74.9 & 74.5 & 47.6 & \textbf{46.5} & 72.8 & 70.0 & 73.3 & 79.3 & 57.0 & 86.5 & 80.4 & 66.8 & 88.3 & 49.3 & 64.6 & \textbf{47.8} & \textbf{55.8} & 71.9 & 68.1\\
& CASENet-S & \textbf{87.6} & 77.1 & \textbf{75.9} & \textbf{48.7} & 46.2 & \textbf{75.5} & \textbf{71.4} & 75.3 & 80.6 & 59.7 & 86.8 & 81.4 & 68.1 & \textbf{89.2} & \textbf{50.7} & \textbf{68.0} & 42.5 & 54.6 & 72.7 & \textbf{69.1}\\
& SEAL      & \textbf{87.6} & \textbf{77.5} & \textbf{75.9} & 47.6 & 46.3 & \textbf{75.5} & 71.2 & \textbf{75.4} & \textbf{80.9} & \textbf{60.1} & \textbf{87.4} & \textbf{81.5} & \textbf{68.9} & 88.9 & 50.2 & 67.8 & 44.1 & 52.7 & \textbf{73.0} & \textbf{69.1}\\
\hline \hline
\multirow{3}{0.1\linewidth}{\centering{MF\\(Raw)}}
& CASENet   & 66.8 & 64.6 & 66.8 & 39.4 & 40.6 & 71.7 & 64.2 & 65.1 & 71.1 & 50.2 & 80.3 & 73.1 & 58.6 & 77.0 & 42.0 & 53.2 & 39.1 & 46.1 & 62.2 & 59.6\\
& CASENet-S & 79.2 & 70.8 & 70.4 & 42.5 & 42.4 & 73.9 & 66.7 & 68.2 & 74.6 & 54.6 & 82.5 & 75.7 & 61.5 & 82.7 & 46.0 & 59.7 & 39.1 & 47.0 & 64.8 & 63.3\\
& SEAL      & \textbf{84.4} & \textbf{73.5} & \textbf{72.7} & \textbf{43.4} & \textbf{43.2} & \textbf{76.1} & \textbf{68.5} & \textbf{69.8} & \textbf{77.2} & \textbf{57.5} & \textbf{85.3} & \textbf{77.6} & \textbf{63.6} & \textbf{84.9} & \textbf{48.6} & \textbf{61.9} & \textbf{41.2} & \textbf{49.0} & \textbf{66.7} & \textbf{65.5}\\
\end{tabular}}
\end{table*}

\begin{figure*}[t!]
\resizebox{1.0\textwidth}{!}{
\begin{tabular}{@{}cccccccccc@{}}
\cellcolor{sbd_color_1}\textcolor{white}{~~~~aero~~~~} &
\cellcolor{sbd_color_2}\textcolor{white}{~~~~bike~~~~} &
\cellcolor{sbd_color_3}\textcolor{white}{~~~~bird~~~~} &
\cellcolor{sbd_color_4}\textcolor{white}{~~~~boat~~~~} &
\cellcolor{sbd_color_5}\textcolor{white}{~~~~bottle~~~~} &
\cellcolor{sbd_color_6}\textcolor{white}{~~~~bus~~~~} &
\cellcolor{sbd_color_7}\textcolor{white}{~~~~c~ar~~~~} &
\cellcolor{sbd_color_8}\textcolor{white}{~~~~cat~~~~} &
\cellcolor{sbd_color_9}\textcolor{white}{~~~~chair~~~~} &
\cellcolor{sbd_color_10}\textcolor{white}{~~~~cow~~~~} \\
\cellcolor{sbd_color_11}\textcolor{white}{~table~} &
\cellcolor{sbd_color_12}\textcolor{white}{~dog~} &
\cellcolor{sbd_color_13}\textcolor{white}{~horse~} &
\cellcolor{sbd_color_14}\textcolor{white}{~mbike~} &
\cellcolor{sbd_color_15}\textcolor{white}{~person~} &
\cellcolor{sbd_color_16}\textcolor{white}{~plant~} &
\cellcolor{sbd_color_17}\textcolor{white}{~sheep~} &
\cellcolor{sbd_color_18}\textcolor{white}{~sofa~} &
\cellcolor{sbd_color_19}\textcolor{white}{~train~} &
\cellcolor{sbd_color_20}\textcolor{white}{~tv~}
\end{tabular}
}
\centering\\
\quad\\
\includegraphics[width=.16\textwidth]{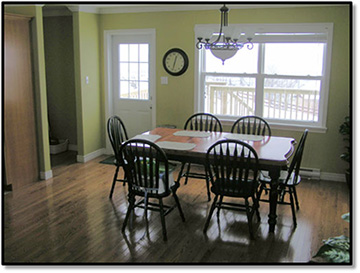}
\includegraphics[width=.16\textwidth]{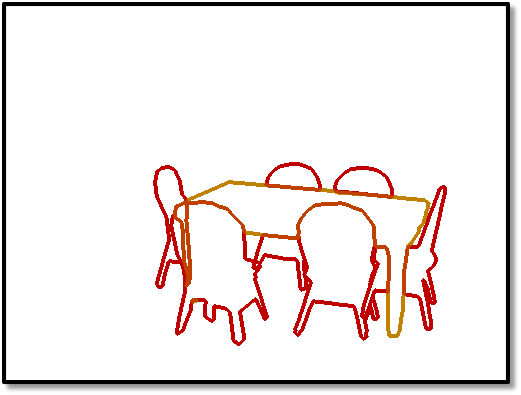}
\includegraphics[width=.16\textwidth]{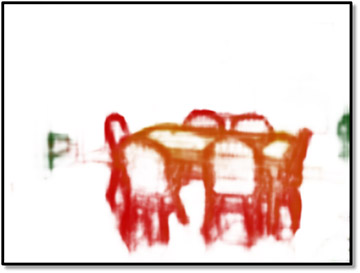}
\includegraphics[width=.16\textwidth]{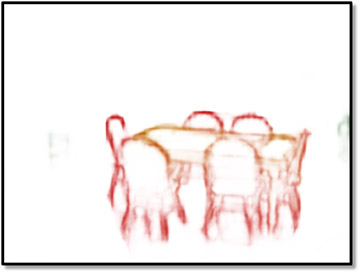}
\includegraphics[width=.16\textwidth]{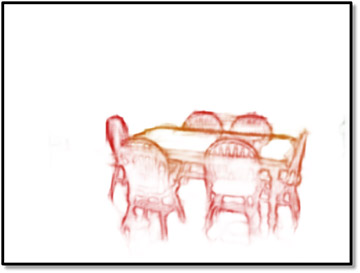}
\includegraphics[width=.16\textwidth]{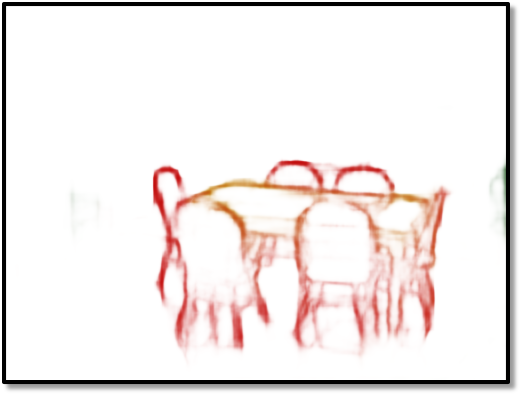}\\

\includegraphics[width=.16\textwidth]{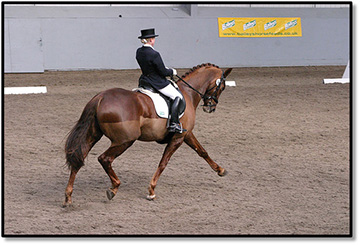}
\includegraphics[width=.16\textwidth]{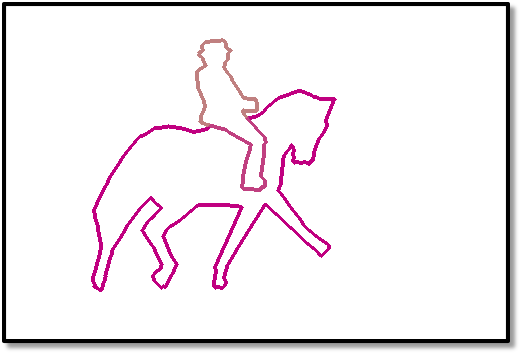}
\includegraphics[width=.16\textwidth]{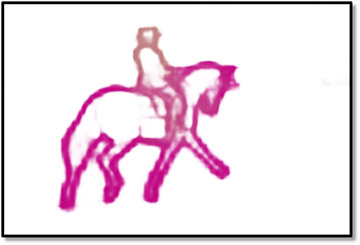}
\includegraphics[width=.16\textwidth]{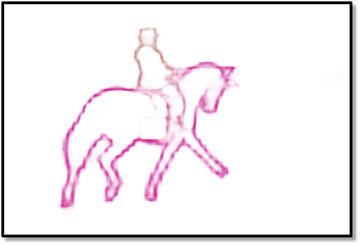}
\includegraphics[width=.16\textwidth]{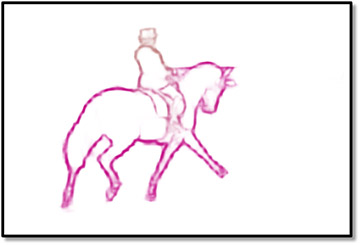}
\includegraphics[width=.16\textwidth]{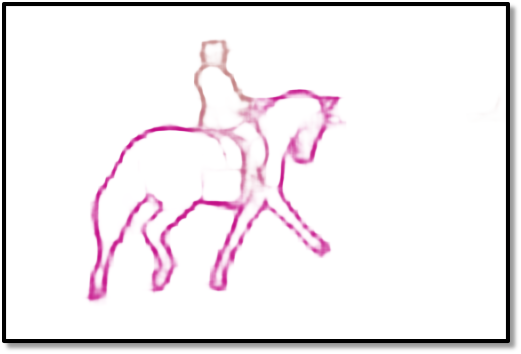}\\

\includegraphics[width=.16\textwidth]{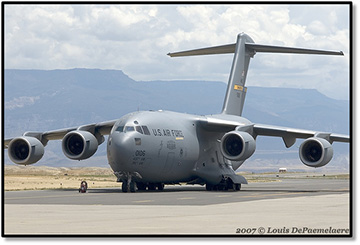}
\includegraphics[width=.16\textwidth]{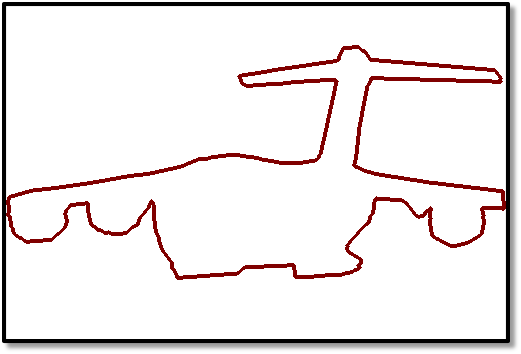}
\includegraphics[width=.16\textwidth]{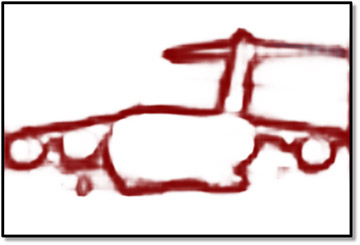}
\includegraphics[width=.16\textwidth]{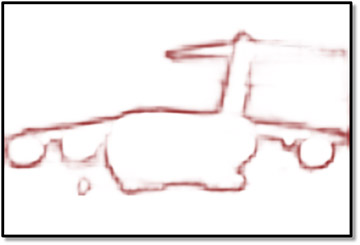}
\includegraphics[width=.16\textwidth]{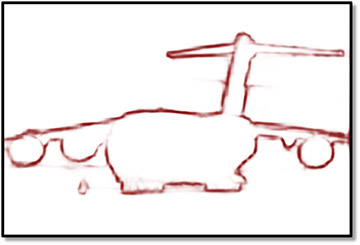}
\includegraphics[width=.16\textwidth]{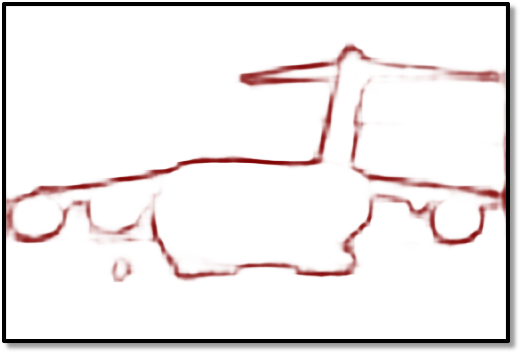}\\

\includegraphics[width=.16\textwidth]{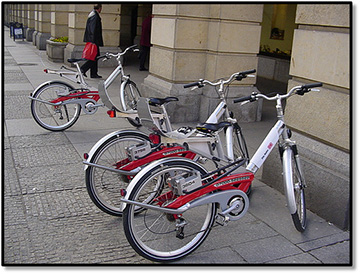}
\includegraphics[width=.16\textwidth]{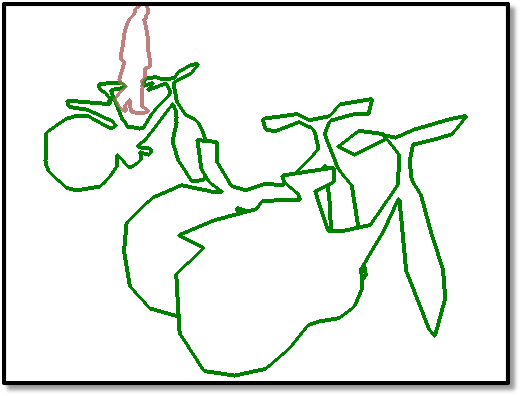}
\includegraphics[width=.16\textwidth]{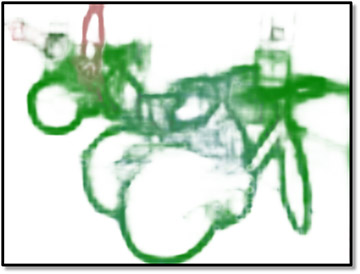}
\includegraphics[width=.16\textwidth]{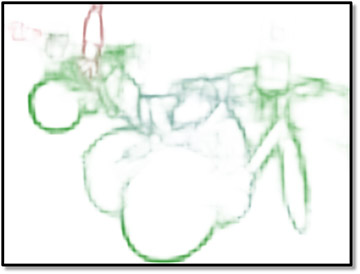}
\includegraphics[width=.16\textwidth]{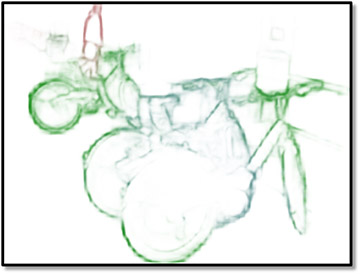}
\includegraphics[width=.16\textwidth]{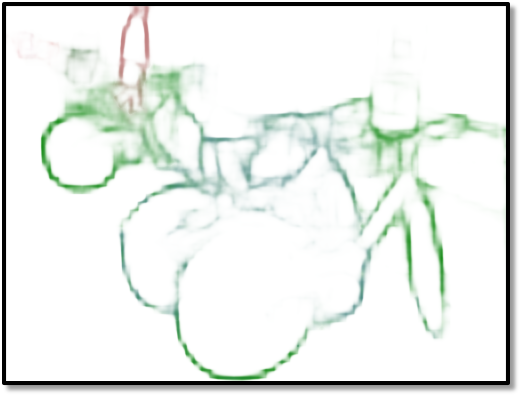}\\
\caption{Qualitative comparison among ground truth, CASENet, CASENet-S, CASENet-C, and SEAL (ordering from left to right). Best viewed in color.}
\label{sbd}
\end{figure*}

\subsection{Experiment on the Cityscapes dataset}
\subsubsection{Results on the validation set}
\begin{figure}[t]
\resizebox{1.0\textwidth}{!}{
\begin{tabular}{@{}cccccccccc@{}}
\cellcolor{city_color_1}\textcolor{white}{~~road~~} &
\cellcolor{city_color_2}\textcolor{white}{~~sidewalk~~} &
\cellcolor{city_color_3}\textcolor{white}{~~building~~} &
\cellcolor{city_color_4}\textcolor{white}{~~wall~~} &
\cellcolor{city_color_5}\textcolor{white}{~~fence~~} &
\cellcolor{city_color_6}\textcolor{white}{~~pole~~} &
\cellcolor{city_color_7}~~traffic lgt~~ &
\cellcolor{city_color_8}~~traffic sgn~~ &
\cellcolor{city_color_9}\textcolor{white}{~~vegetation~} \\
\cellcolor{city_color_10}~~terrain~~ &
\cellcolor{city_color_11}\textcolor{white}{~~sky~~} &
\cellcolor{city_color_12}\textcolor{white}{~~person~~} &
\cellcolor{city_color_13}\textcolor{white}{~~rider~~} &
\cellcolor{city_color_14}\textcolor{white}{~~car~~} &
\cellcolor{city_color_15}\textcolor{white}{~~truck~~} &
\cellcolor{city_color_16}\textcolor{white}{~~bus~~} &
\cellcolor{city_color_17}\textcolor{white}{~~train~~} &
\cellcolor{city_color_18}\textcolor{white}{~~motorcycle~~} &
\cellcolor{city_color_19}\textcolor{white}{~~bike~~}
\end{tabular}
}
\centering\\
\quad\\
\includegraphics[width=.193\textwidth]{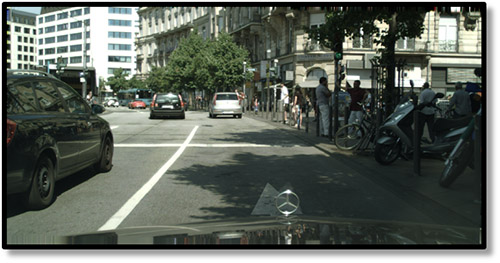}
\includegraphics[width=.193\textwidth]{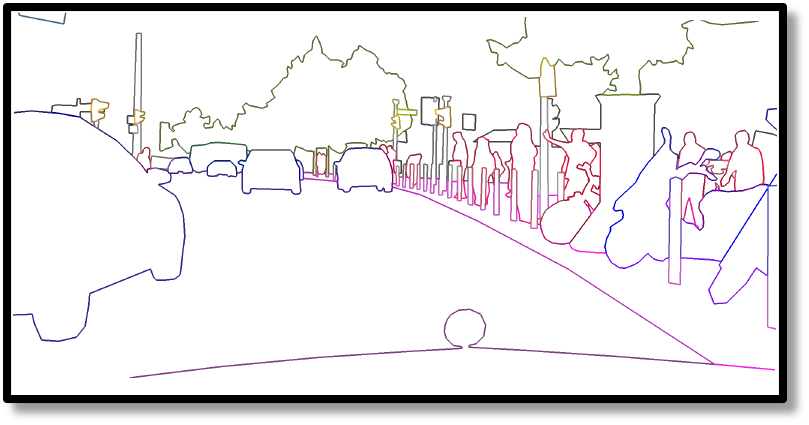}
\includegraphics[width=.193\textwidth]{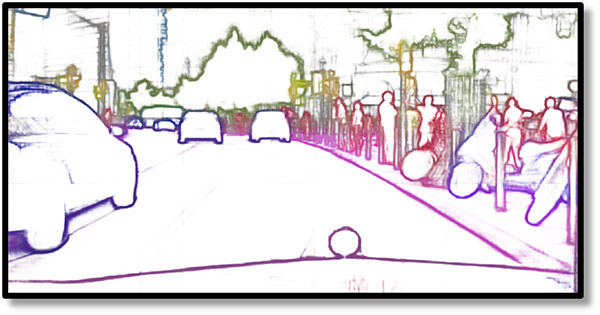}
\includegraphics[width=.193\textwidth]{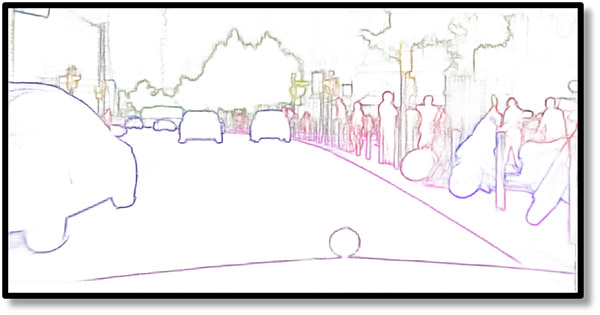}
\includegraphics[width=.193\textwidth]{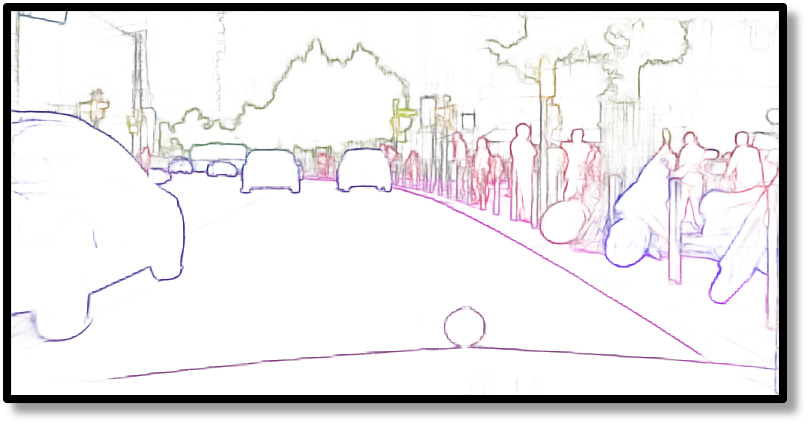}\\

\includegraphics[width=.193\textwidth]{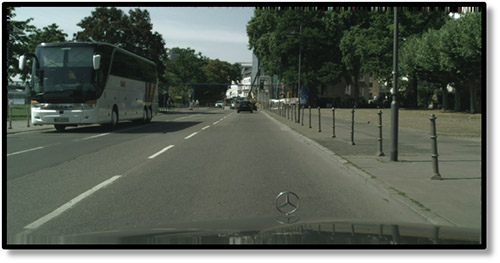}
\includegraphics[width=.193\textwidth]{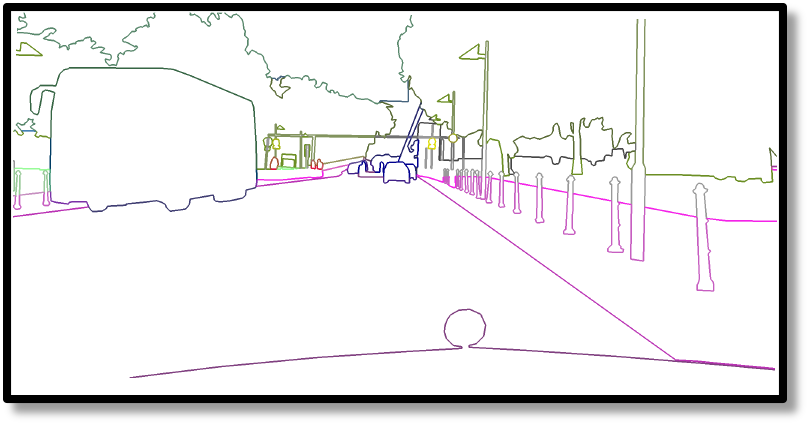}
\includegraphics[width=.193\textwidth]{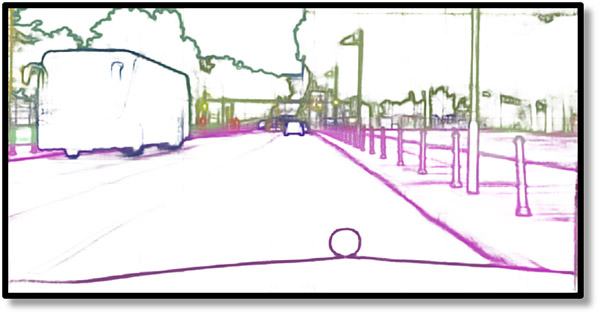}
\includegraphics[width=.193\textwidth]{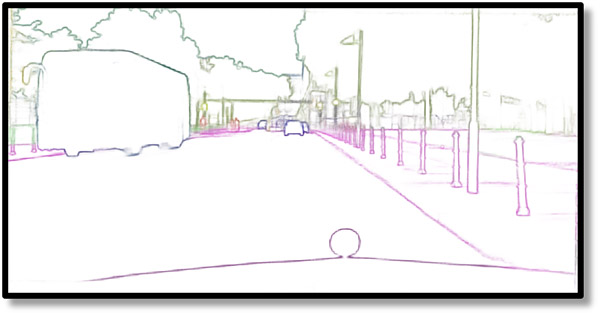}
\includegraphics[width=.193\textwidth]{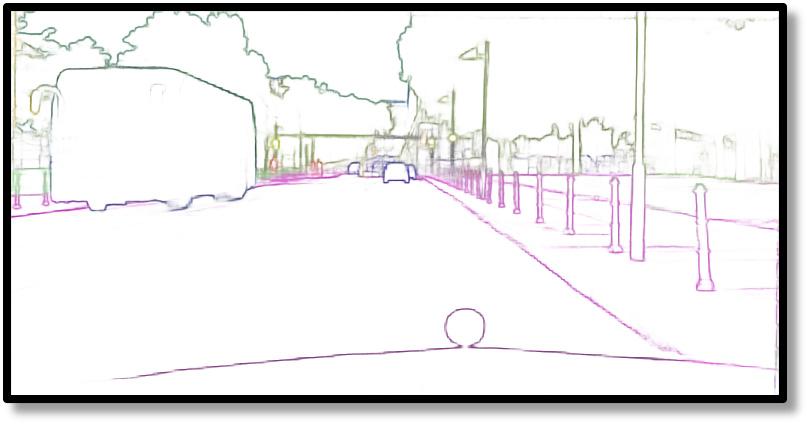}\\

\includegraphics[width=.193\textwidth]{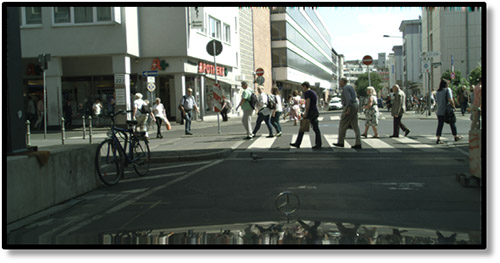}
\includegraphics[width=.193\textwidth]{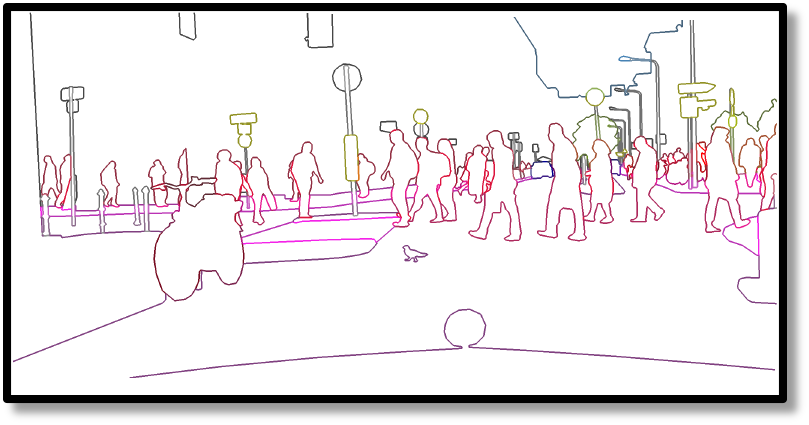}
\includegraphics[width=.193\textwidth]{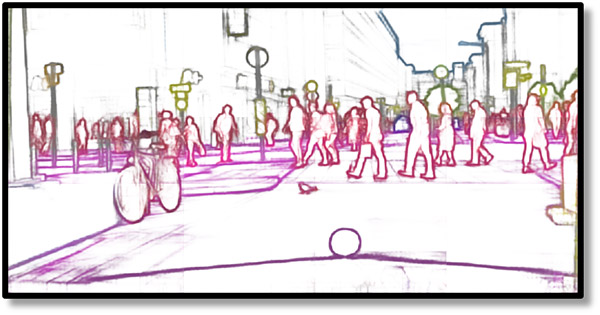}
\includegraphics[width=.193\textwidth]{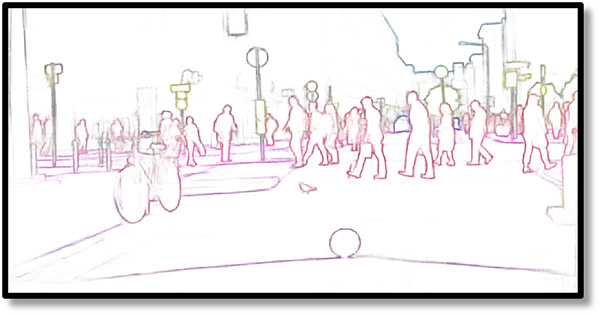}
\includegraphics[width=.193\textwidth]{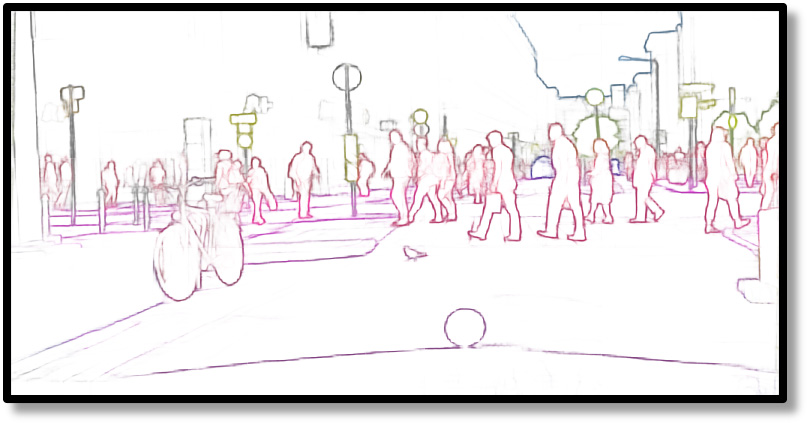}\\

\includegraphics[width=.193\textwidth]{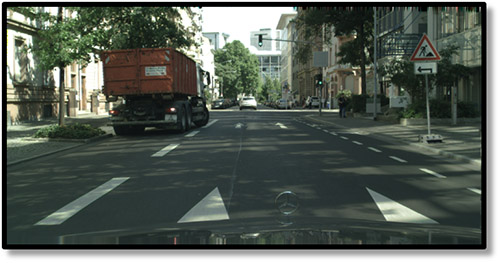}
\includegraphics[width=.193\textwidth]{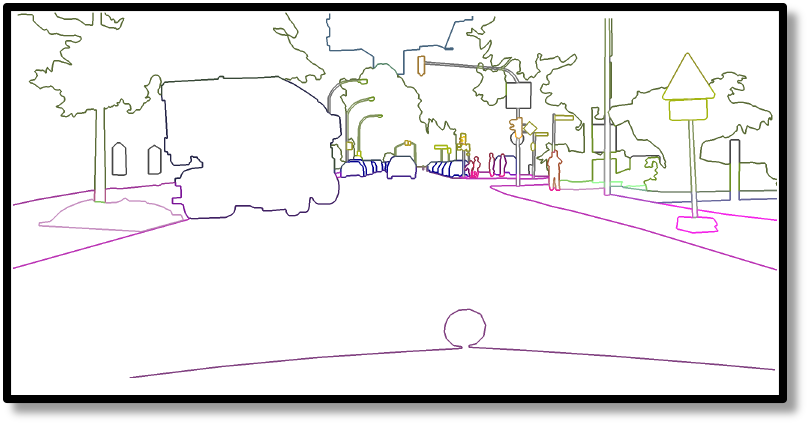}
\includegraphics[width=.193\textwidth]{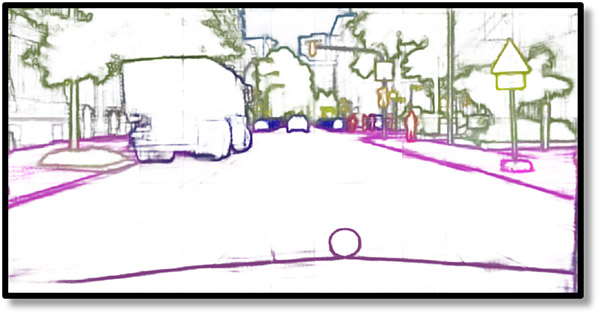}
\includegraphics[width=.193\textwidth]{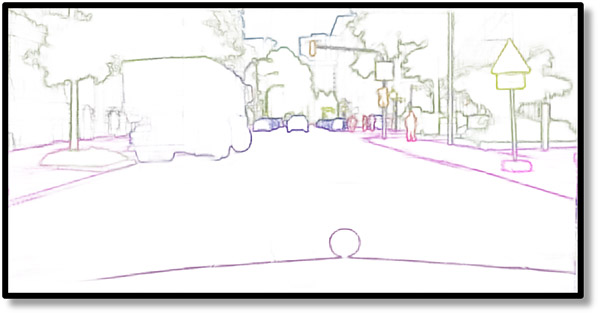}
\includegraphics[width=.193\textwidth]{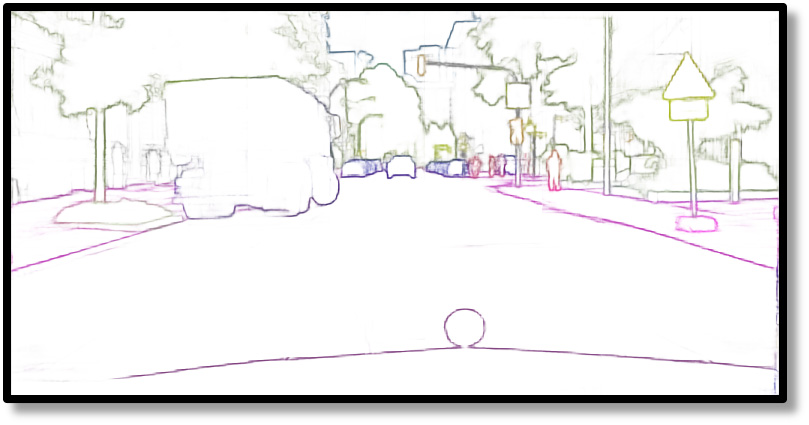}\\

\includegraphics[width=.193\textwidth]{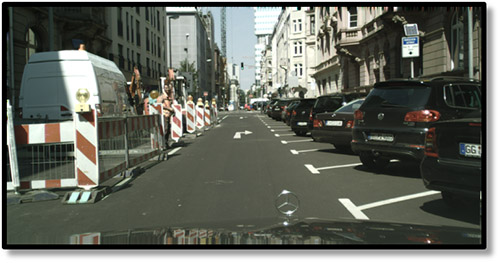}
\includegraphics[width=.193\textwidth]{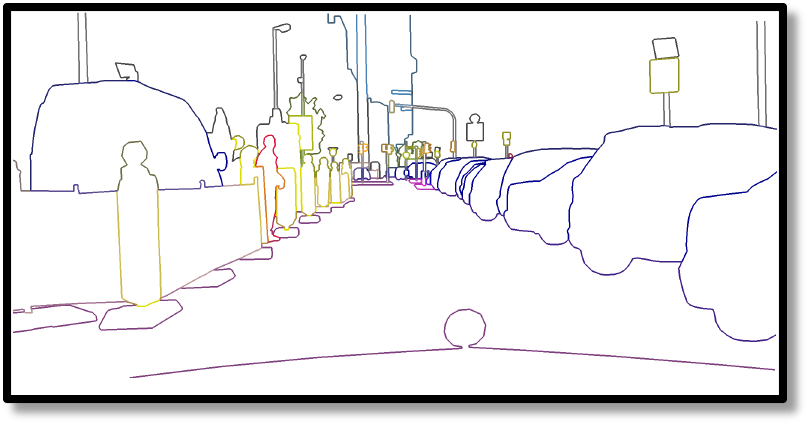}
\includegraphics[width=.193\textwidth]{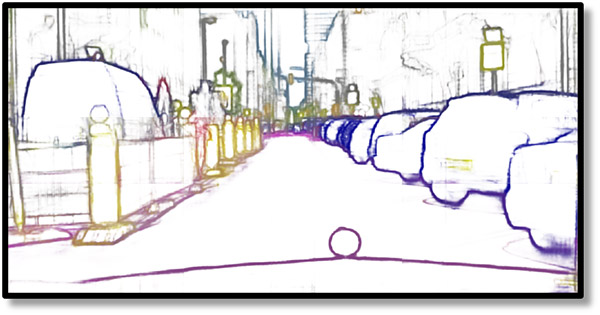}
\includegraphics[width=.193\textwidth]{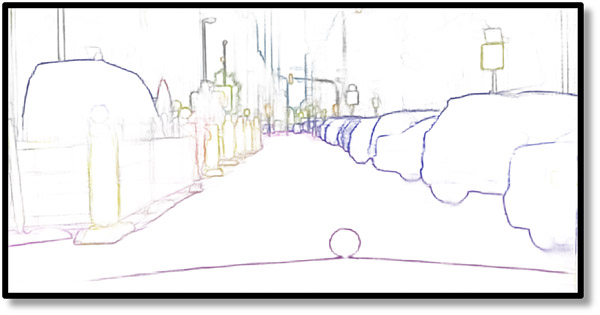}
\includegraphics[width=.193\textwidth]{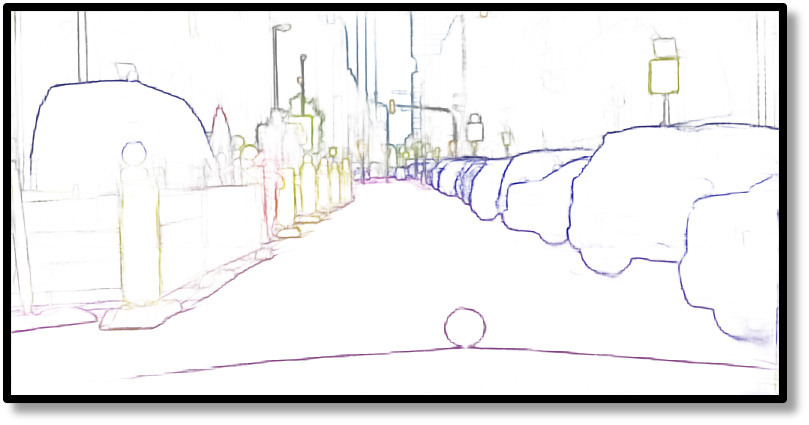}\\

\includegraphics[width=.193\textwidth]{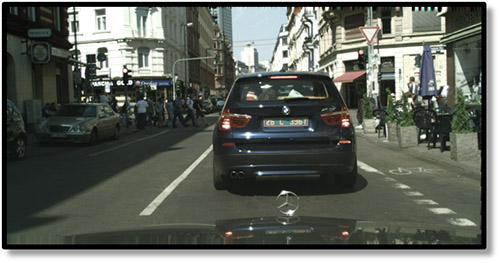}
\includegraphics[width=.193\textwidth]{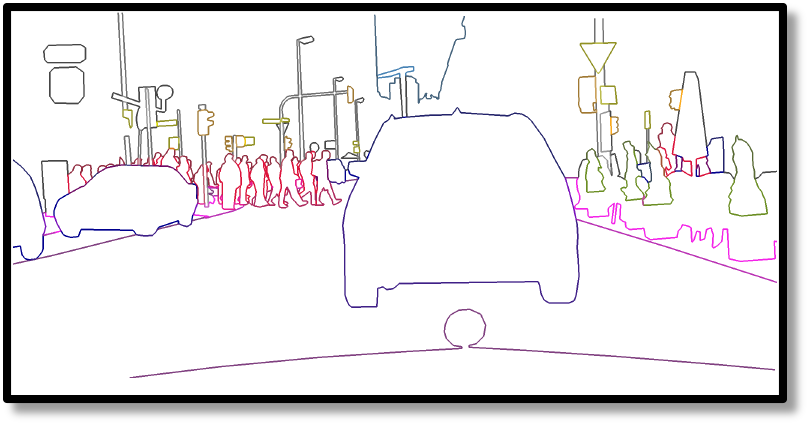}
\includegraphics[width=.193\textwidth]{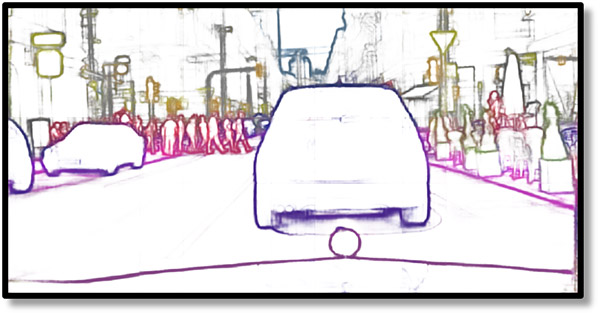}
\includegraphics[width=.193\textwidth]{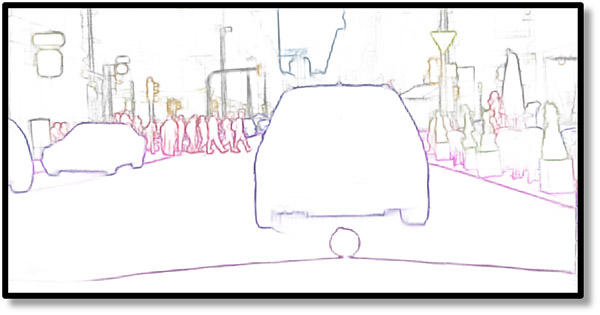}
\includegraphics[width=.193\textwidth]{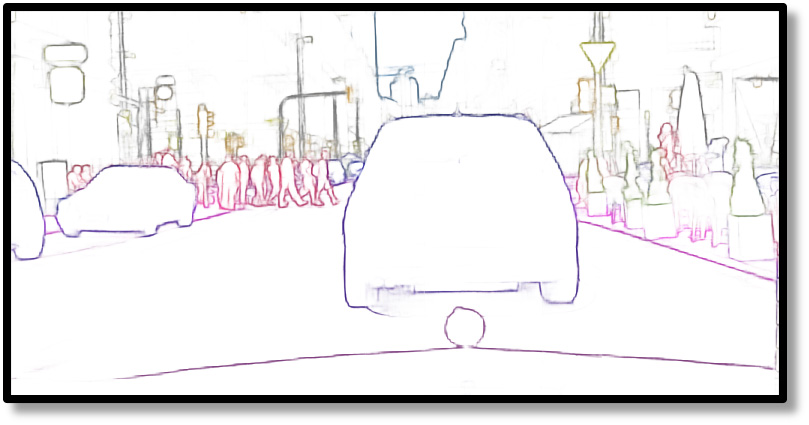}\\

\includegraphics[width=.193\textwidth]{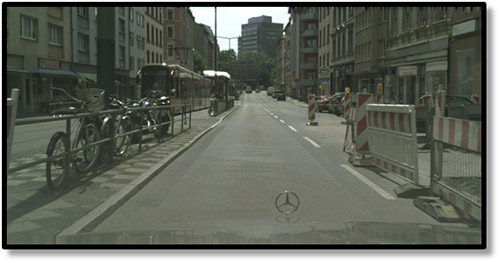}
\includegraphics[width=.193\textwidth]{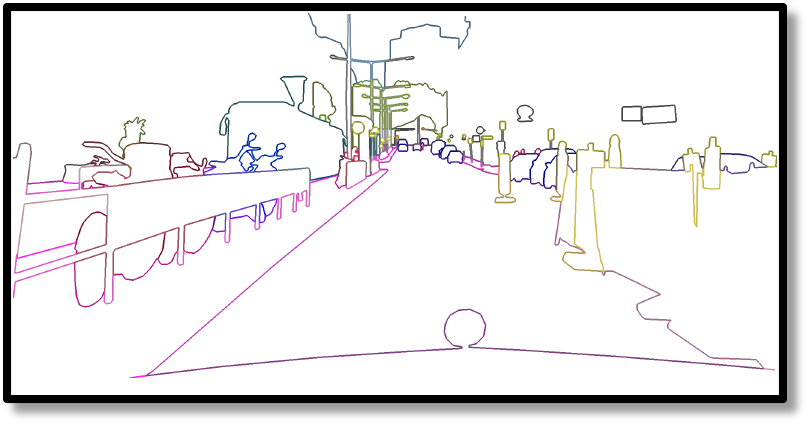}
\includegraphics[width=.193\textwidth]{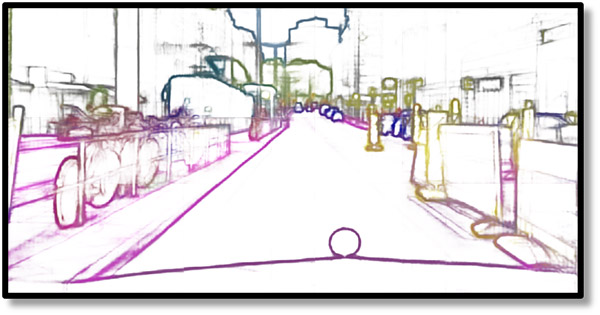}
\includegraphics[width=.193\textwidth]{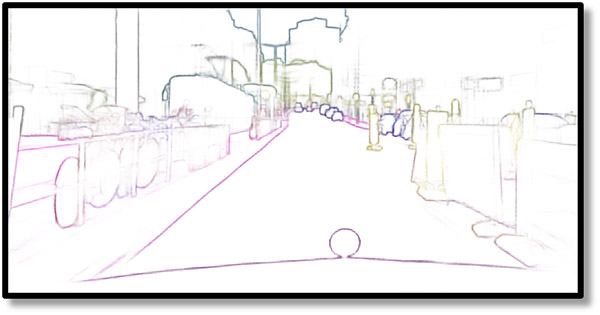}
\includegraphics[width=.193\textwidth]{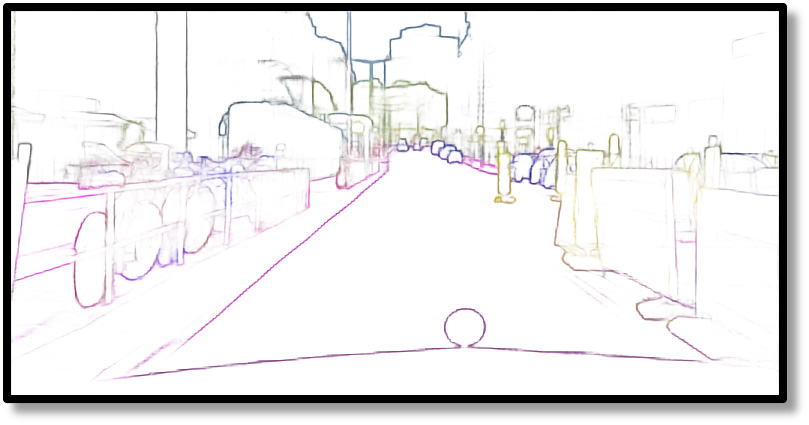}\\

\includegraphics[width=.193\textwidth]{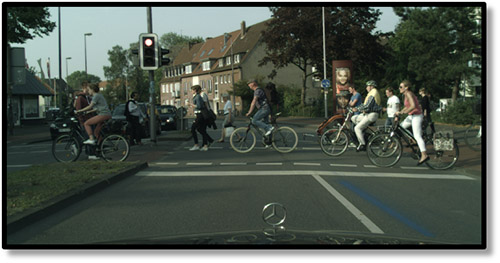}
\includegraphics[width=.193\textwidth]{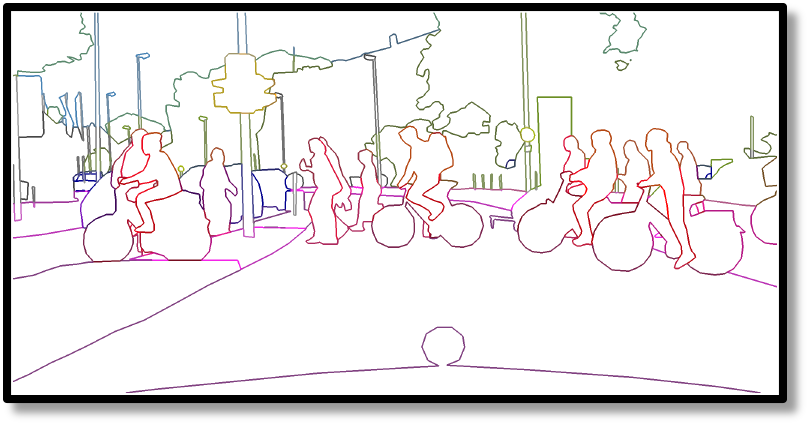}
\includegraphics[width=.193\textwidth]{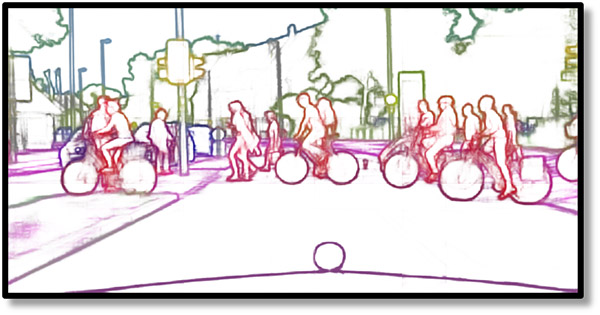}
\includegraphics[width=.193\textwidth]{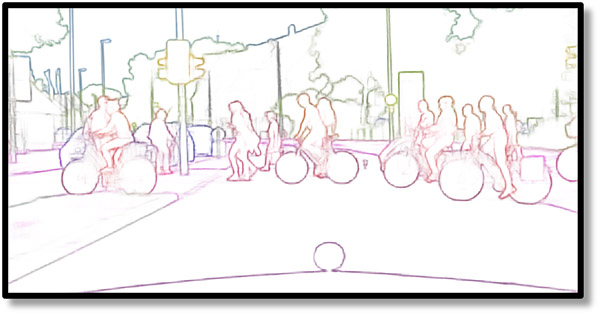}
\includegraphics[width=.193\textwidth]{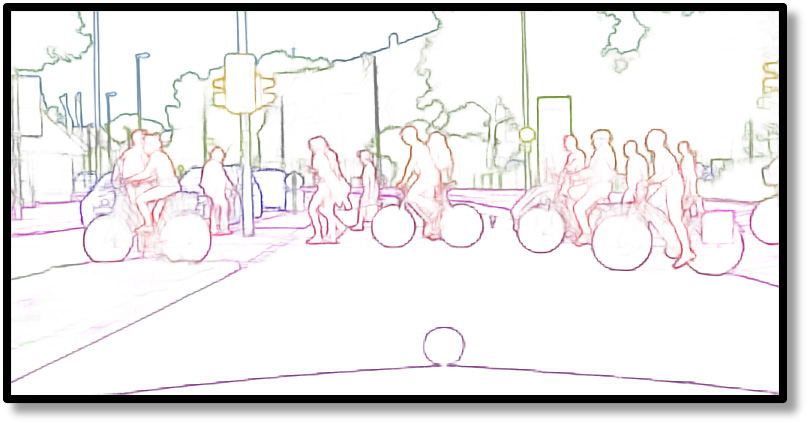}\\
\caption{Qualitative comparison among ground truth, CASENet, CASENet-S, and SEAL (ordering from left to right in the figure). Best viewed in color.}
\label{cityscapes}
\end{figure}

The Cityscapes dataset contains 2975 training images and 500 validation images. Following~\cite{yu2017casenet}, we train SEAL on the training set and test on the validation set, with the results visualized in Fig.~\ref{cityscapes} and quantified in Table~\ref{tb:cityscapes}. Again, SEAL overall outperforms all comparing baselines.

\subsubsection{Alignment visualization}
\begin{figure*}[t!]
\centering
\includegraphics[height=.193\textwidth]{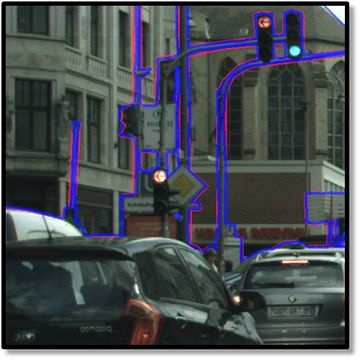}\label{Fig7a}
\includegraphics[height=.193\textwidth]{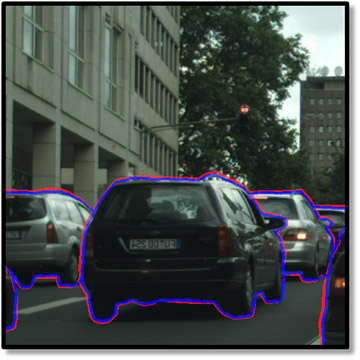}\label{Fig7b}
\includegraphics[height=.193\textwidth]{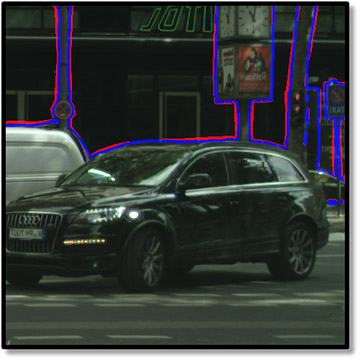}\label{Fig7c}
\includegraphics[height=.193\textwidth]{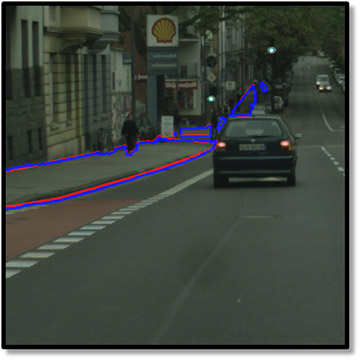}\label{Fig7d}
\includegraphics[height=.193\textwidth]{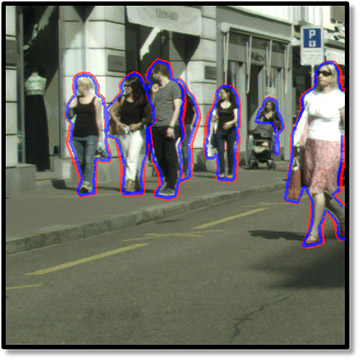}\label{Fig7e}
\caption{Edge alignment on Cityscapes. Color coding follows Fig.~\ref{Fig2}. Best viewed in color.}
\label{Fig7}
\end{figure*}

We show that misalignment can still be found on Cityscapes. Fig.~\ref{Fig7} shows misaligned labels and the corrections made by SEAL.

\section{Concluding remarks}
In this paper, we proposed SEAL: an end-to-end learning framework for joint edge alignment and learning. Our work considers a novel pixel-level noisy label learning problem, levering structured priors to address an open issue in edge learning. Extensive experiments demonstrate that the proposed framework is able to correct noisy labels and generate sharp edges with better quality.

\clearpage

\bibliographystyle{splncs04}
\bibliography{ref}

\begin{thebibliography}{10}
\providecommand{\url}[1]{\texttt{#1}}
\providecommand{\urlprefix}{URL }
\providecommand{\doi}[1]{https://doi.org/#1}

\bibitem{Arbelaez2011}
Arbel\'{a}ez, P., Maire, M., Fowlkes, C., Malik, J.: Contour detection and
  hierarchical image segmentation. IEEE Trans. PAMI  \textbf{33}(5),  898--916
  (2011)

\bibitem{Arbelaez2014}
Arbel\'{a}ez, P., Pont-Tuset, J., Barron, J., Marques, F., Malik, J.:
  Multiscale combinatorial grouping. In: CVPR (2014)

\bibitem{Bertasius2015_de}
Bertasius, G., Shi, J., Torresani, L.: Deepedge: A multiscale bifurcated deep
  network for top-down contour detection. In: CVPR (2015)

\bibitem{Bertasius2015_hfl}
Bertasius, G., Shi, J., Torresani, L.: High-for-low, low-for-high: Efficient
  boundary detection from deep object features and its applications to
  high-level vision. In: ICCV (2015)

\bibitem{bertasius2016semantic}
Bertasius, G., Shi, J., Torresani, L.: Semantic segmentation with boundary
  neural fields. In: CVPR (2016)

\bibitem{bilmes2004virtual}
Bilmes, J.: On virtual evidence and soft evidence in bayesian networks. Tech.
  rep. (2004)

\bibitem{Canny1986}
Canny, J.: A computational approach to edge detection. IEEE Trans. PAMI (6),
  679--698 (1986)

\bibitem{castrejon2017annotating}
Castrej{\'o}n, L., Kundu, K., Urtasun, R., Fidler, S.: Annotating object
  instances with a polygon-rnn. In: CVPR (2017)

\bibitem{Chen2016}
Chen, L.C., Barron, J.T., G.~Papandreou, K.M., Yuille, A.L.: Semantic image
  segmentation with task-specific edge detection using cnns and a
  discriminatively trained domain transform. In: CVPR (2016)

\bibitem{cherkassky1995implementing}
Cherkassky, B.V., Goldberg, A.V.: On implementing push-relabel method for the
  maximum flow problem. In: Int. Conf. on Integer Programming and Combinatorial
  Optimization (1995)

\bibitem{cordts2016cityscapes}
Cordts, M., Omran, M., Ramos, S., Rehfeld, T., Enzweiler, M., Benenson, R.,
  Franke, U., Roth, S., Schiele, B.: The \uppercase{C}ityscapes dataset for
  semantic urban scene understanding. In: CVPR (2016)

\bibitem{Dollar2006}
Dollar, P., Tu, Z., Belongie, S.: Supervised learning of edges and object
  boundaries. In: CVPR (2006)

\bibitem{dollar2015fast}
Doll{\'a}r, P., Zitnick, C.L.: Fast edge detection using structured forests.
  IEEE Trans. PAMI  \textbf{37}(8),  1558--1570 (2015)

\bibitem{pascal-voc-2011}
Everingham, M., Van~Gool, L., Williams, C.K.I., Winn, J., Zisserman, A.: The
  {PASCAL} {V}isual {O}bject {C}lasses {C}hallenge 2011 {(VOC2011)} {R}esults.
  http://www.pascal-network.org/challenges/VOC/voc2011/workshop/index.html

\bibitem{goldberg1995efficient}
Goldberg, A.V., Kennedy, R.: An efficient cost scaling algorithm for the
  assignment problem. SIAM J. Discrete Math.  (1993)

\bibitem{hancock1990edge}
Hancock, E.R., Kittler, J.: Edge-labeling using dictionary-based relaxation.
  IEEE Trans. PAMI  \textbf{12}(2),  165--181 (1990)

\bibitem{Hariharan2011}
Hariharan, B., Arbel\'{a}ez, P., Bourdev, L., Maji, S., Malik, J.: Semantic
  contours from inverse detectors. In: ICCV (2011)

\bibitem{He2016}
He, K., Zhang, X., Ren, S., Sun, J.: Deep residual learning for image
  recognition. In: CVPR (2016)

\bibitem{hoiem2005geometric}
Hoiem, D., Efros, A.A., Hebert, M.: Geometric context from a single image. In:
  ICCV (2005)

\bibitem{Hwang2015}
Hwang, J., Liu, T.L.: Pixel-wise deep learning for contour detection. In: ICLR
  (2015)

\bibitem{Karsch2013}
Karsch, K., Liao, Z., Rock, J., Barron, J.T., Hoiem, D.: Boundary cues for 3d
  object shape recovery. In: CVPR (2013)

\bibitem{khoreva2016weakly}
Khoreva, A., Benenson, R., Omran, M., Hein, M., Schiele, B.: Weakly supervised
  object boundaries. In: CVPR (2016)

\bibitem{Kittler1983}
Kittler, J.: On the accuracy of the sobel edge detector. Image and Vision
  Computing  \textbf{1}(1),  37--42 (1983)

\bibitem{kokkinos2016pushing}
Kokkinos, I.: Pushing the boundaries of boundary detection using deep learning
  (2016)

\bibitem{Konishi2003}
Konishi, S., Yuille, A.L., Coughlan, J.M., Zhu, S.C.: Statistical edge
  detection: Learning and evaluating edge cues. IEEE Trans. PAMI
  \textbf{25}(1),  57--74 (2003)

\bibitem{krizhevsky2012imagenet}
Krizhevsky, A., Sutskever, I., Hinton, G.E.: Imagenet classification with deep
  convolutional neural networks. In: NIPS (2012)

\bibitem{li2017learning}
Li, Y., Yang, J., Song, Y., Cao, L., Luo, J., Li, L.J.: Learning from noisy
  labels with distillation. In: CVPR (2017)

\bibitem{liao2007training}
Liao, L., Choudhury, T., Fox, D., Kautz, H.A.: Training conditional random
  fields using virtual evidence boosting. In: IJCAI (2007)

\bibitem{Lim2013}
Lim, J., Zitnick, C., Dollar, P.: Sketch tokens: A learned mid-level
  representation for contour and object detection. In: CVPR (2013)

\bibitem{liu2017richer}
Liu, Y., Cheng, M.M., Hu, X., Wang, K., Bai, X.: Richer convolutional features
  for edge detection. In: CVPR (2017)

\bibitem{Maire2014}
Maire, M., Yu, S.X., Perona, P.: Reconstructive sparse code transfer for
  contour detection and semantic labeling. In: ACCV (2014)

\bibitem{lineCurvedObjects}
Malik, J.: Interpreting line drawings of curved objects. Int. Journal of
  Computer Vision  \textbf{1}(1),  73--103 (1987)

\bibitem{malik1989recovering}
Malik, J., Maydan, D.: Recovering three-dimensional shape from a single image
  of curved objects. IEEE Trans. PAMI  \textbf{11}(6),  555--566 (1989)

\bibitem{Martin2004}
Martin, D.R., Fowlkes, C.C., Malik, J.: Learning to detect natural image
  boundaries using local brightness, color, and texture cues. IEEE Trans. PAMI
  \textbf{26}(5),  530--549 (2004)

\bibitem{patrini2017making}
Patrini, G., Rozza, A., Menon, A.K., Nock, R., Qu, L.: Making deep neural
  networks robust to label noise: a loss correction approach. In: CVPR (2017)

\bibitem{pearl1988probabilistic}
Pearl, J.: Probabilistic reasoning in intelligent systems: Networks of
  plausible inference (1988)

\bibitem{pinheiro2015learning}
Pinheiro, P.O., Collobert, R., Doll{\'a}r, P.: Learning to segment object
  candidates. In: NIPS (2015)

\bibitem{prasad2006learning}
Prasad, M., Zisserman, A., Fitzgibbon, A., Kumar, M.P., Torr, P.H.: Learning
  class-specific edges for object detection and segmentation. In: Computer
  Vision, Graphics and Image Processing (2006)

\bibitem{ren2008learning}
Ren, X., Fowlkes, C.C., Malik, J.: Learning probabilistic models for contour
  completion in natural images. Int. Journal of Computer Vision
  \textbf{77}(1-3),  47--63 (2008)

\bibitem{rupprecht2016deep}
Rupprecht, C., Huaroc, E., Baust, M., Navab, N.: Deep active contours. arXiv
  preprint arXiv:1607.05074  (2016)

\bibitem{russell2008labelme}
Russell, B.C., Torralba, A., Murphy, K.P., Freeman, W.T.: Labelme: a database
  and web-based tool for image annotation. IJCV  \textbf{77}(1-3),  157--173
  (2008)

\bibitem{Shan2014}
Shan, Q., Curless, B., Furukawa, Y., Hernandez, C., Seitz, S.: Occluding
  contours for multi-view stereo. In: CVPR (2014)

\bibitem{Simonyan2015}
Simonyan, K., Zisserman, A.: Very deep convolutional networks for large-scale
  image recognition. In: ICLR (2015)

\bibitem{mild-sugihara}
Sugihara, K.: Machine Interpretation of Line Drawings. MIT Press (1986)

\bibitem{vahdat2017toward}
Vahdat, A.: Toward robustness against label noise in training deep
  discriminative neural networks. In: NIPS (2017)

\bibitem{veit2017learning}
Veit, A., Alldrin, N., Chechik, G., Krasin, I., Gupta, A., Belongie, S.:
  Learning from noisy large-scale datasets with minimal supervision. In: CVPR
  (2017)

\bibitem{wang2018iterative}
Wang, Y., Liu, W., Ma, X., Bailey, J., Zha, H., Song, L., Xia, S.T.: Iterative
  learning with open-set noisy labels. In: CVPR (2018)

\bibitem{xiao2015learning}
Xiao, T., Xia, T., Yang, Y., Huang, C., Wang, X.: Learning from massive noisy
  labeled data for image classification. In: CVPR (2015)

\bibitem{Xie2015}
Xie, S., Tu, Z.: Holistically-nested edge detection. In: ICCV (2015)

\bibitem{yang2016object}
Yang, J., Price, B., Cohen, S., Lee, H., Yang, M.H.: Object contour detection
  with a fully convolutional encoder-decoder network. In: CVPR (2016)

\bibitem{yu2015generalized}
Yu, Z., Liu, W., Liu, W., Peng, X., Hui, Z., Kumar, B.V.: Generalized
  transitive distance with minimum spanning random forest. In: IJCAI (2015)

\bibitem{yu2017casenet}
Yu, Z., Feng, C., Liu, M.Y., Ramalingam, S.: Casenet: Deep category-aware
  semantic edge detection. In: CVPR (2017)

\bibitem{zitnick2014edge}
Zitnick, C.L., Doll{\'a}r, P.: Edge boxes: Locating object proposals from
  edges. In: ECCV (2014)

\end{thebibliography}

\section*{Appendix}
The main paper presents our ECCV 2018 camera ready submission. In the appendix, we further present the additional details and results that are not covered by the camera ready paper due to space constraints. We believe these details will benefit successful reproduction of the reported experiments.

Specifically, Section~\ref{details} contains implementation details of this work, including network/benchmark parameters, dataset preprocessing, computation cost, and visualization. Section~\ref{anno} illustrates an example of the re-annotation on the SBD test set. Finally, additional results on the SBD dataset and the Cityscapes dataset are included in Section~\ref{add_sbd} and~\ref{add_city}, respectively.

\setcounter{section}{0}
\renewcommand{\thesection}{\Alph{section}}
\section{Implementation details}\label{details}
\subsection{Network hyperparameters}
We use the code from~\cite{yu2017casenet}, and exactly follow~\cite{yu2017casenet} to set the network hyperparameters, including learning rate, gamma, momentum, decay, etc. As a result, the learning rate and gamma on SBD/Cityscapes are respectively set as $1.0\times 10^{-7}$/$5.0\times 10^{-8}$ and $0.1$/$0.2$. We keep the crop size as $472\times 472$ on Cityscapes, and unify the SBD crop size also as $472\times 472$. For any method involving supervision with the unweighted sigmoid cross-entropy loss, we unify the learning rates on SBD/Cityscapes as $5.0\times 10^{-8}$ and $2.5\times 10^{-8}$, while keeping other parameters the same as counter parts with reweighted loss.

\subsection{Network training iteration numbers}
Following~\cite{yu2017casenet}, we report the performance of all models on SBD at 22000 training iterations. For all models trained on cityscapes, the network training iteration number is empirically selected as 28000.

\subsection{SBD data split}\label{sec:sbd_data_split}
In this work, we further randomly sample 1000 images from the original SBD training set as a \textbf{validation set}, while treating the rest 7498 images as a new training set. In addition, the original SBD test set with 2857 images remains a held out \textbf{test set}. For the rest of the appendix, we will refer to the new training set with 7495 images as ``\textbf{training set}'', and the original SBD training set with 8498 images as ``\textbf{trainval set}'' for clarity.

To conduct parameter analysis and ablation study for the proposed framework, models with different parameters and modules are trained on the training set, and validated on the validation set. In addition, results reported on the SBD test set in the main paper correspond to models trained on the trainval set with parameters determined via parameter analysis.

\subsection{Data augmentation and training label generation}
We follow the \href{https://github.com/Chrisding/sbd-preprocess}{preprocessing code} of~\cite{yu2017casenet} to perform multi-scale data augmentation and generate slightly thicker edge labels for model training on SBD. In particular, we slightly modify the code to preserve instance-sensitive edges and augment both the SBD training and trainval set in Sec.~\ref{sec:sbd_data_split}, while keeping other implementation the same. We apply similar training label generation procedure to Cityscapes except removing the data augmentation.

Note that for evaluation under the ``Raw'' setting, raw predictions are matched with unthinned ground truths whose edge width is set to be the same as training edge labels. Under the ``Thin'' setting, on the other hand, the evaluation ground truth consists of single pixel wide edge labels.

\subsection{Computation cost}
The actual computation cost of SEAL optimization largely depends on the number of Assign steps in edge alignment. While more Assign steps tend to give slightly better performance, they also lead to the increase of computation cost. In practice, one Assign step is often enough and the readers may kindly refer to Section~\ref{sec:assign_steps} for more details.

Consider the case where SEAL optimization contains one Assign step and the hardware platform has 1 TitanXP GPU + 2 Xeon E5-2640v4 CPUs. On the SBD dataset, each backbone network learning iteration takes 7 seconds for iteration size of 10. In addition, the alignment of every 300 images takes about 180 seconds using all 40 CPU threads in parallel. Accordingly, the learning framework can process about 6000 iterations per day. Due to the denser edges in each image, the same framework can process 4000 iterations per day on Cityscapes.

\subsection{Benchmark parameters}
In both~\cite{Hariharan2011} and~\cite{Martin2004}, an important benchmark parameter is the matching distance tolerance which is the maximum slop allowed for correct matches of edges to ground truth during evaluation. The distance tolerance is often measured as proportion to the image diagonal. On the BSDS dataset~\cite{Martin2004,Arbelaez2011}, this parameter is by default set as 0.0075, while on SBD~\cite{Hariharan2011}, the parameter is increased to 0.02 to compensate the increased annotation noise.

In light of these previous works, we follow~\cite{Hariharan2011} to set the matching distance tolerance as 0.02 for evaluations using the original SBD annotations. We also decrease the tolerance to 0.0075 for evaluations using the re-annotated high quality SBD test labels. In addition, given the high label quality and the large image diagonals, we decrease the tolerance in Cityscapes experiments to 0.0035. This corresponds to tolerating 8 pixels approximately on Cityscapes images. Note that our Cityscapes benchmark is significantly stricter than~\cite{yu2017casenet} which followed~\cite{Hariharan2011} by setting the distance tolerance to 0.02 in their Cityscapes experiment.

Unlike cityscapes, SBD labels do not guarantee consistent conditions near image borders. On some images, the imperfect alignment between annotations and image borders leads to extra edges which do not appear on images with perfect alignment. Considering that such inconsistency may reduce the benchmark precision, we choose to ignore the evaluation of edges within 5 pixels to image borders in all SBD experiments. For cityscapes, no border pixels are ignored since there is no such issue.

\subsection{Color coding protocol}
For experiments on both SBD and Cityscapes, we follow the original color codings of PASCAL VOC and Cityscapes to visualize the semantic classes. In particular, the visualization of edge pixels is based on the following equation:
\begin{equation}\label{Eq:Col}
\mathbf{I} =
\left\{
\begin{aligned}
\mathbf{255} - (\sup_{c}P_{c})\frac{\sum_{c=1}^{C}P_{c}(\mathbf{255} - \mathbf{M}_{c})}{\sum_{c=1}^{C}P_{c}}, &\text{~~if~~} \sum_{c=1}^{C}P_{c}>0\\
\mathbf{255}~~~~~~~~~~~~~~~~~~~~~~~~~~~~, &\text{~~Otherwise}\\
\end{aligned}
\right.
\end{equation}
where $c$ indicates the class index, and $C$ is the number of classes. $P_{c}$ is the predicted edge probability of class $c$, while $\mathbf{M}_{c}$ is the RGB vector of class $c$ following the original color coding of the corresponding dataset. $\mathbf{255}\triangleq [255,255,255]^\top$.

\section{SBD test set re-annotation}\label{anno}
\begin{figure}[htb]
	\centering
	\includegraphics[width=1\textwidth]{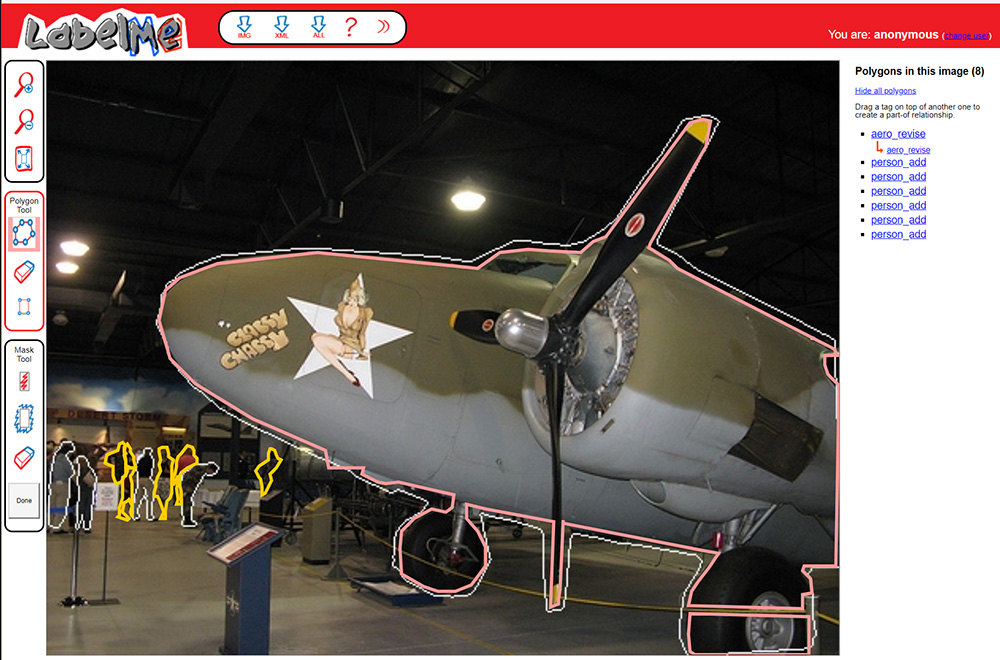}
	\caption{Illustration of SBD re-annotation with MIT LabelMe toolkit. In the image, white lines indicate the original SBD annotation, and colored polygons indicate re-annotated labels. One may notice the significant misalignment along the aeroplane boundary, as well as multiple persons with missing labels.}\label{Fig8}
\end{figure}

Although the SBD dataset provided instance-level segmentation labels with generally good qualities, we observe that a considerable portion of the labels exhibit the issues of large misalignment and missing annotation. This raises some concerns on the reliability of the evaluations based on the original labels since large distance tolerance may potentially compromise the measurement accuracy on localization and precision-recall. In light of these issues, we use the LabelMe~\cite{russell2008labelme} open annotation toolkit to generate a high quality subset of the SBD test set with 1059 images. Fig.~\ref{Fig8} illustrates an example of the re-annotation interface. Additional examples of the re-annotated edge labels versus the original labels are also shown in Fig.~\ref{align_vis}.

\section{Additional results on SBD}\label{add_sbd}
In this section, we report additional results on the SBD dataset.

\subsection{Parameter analysis and ablation study}
We report the results of different parameter configurations of $\sigma_x$, $\sigma_y$ and $\lambda$ on the SBD validation set to determine their values. Since we assume that our system does not have any knowledge on high quality annotations, we validate the parameters of SEAL using the original noisy SBD labels with $0.02$ distance tolerance. Table~\ref{tb:sbd_val} comprehensively reports the results corresponding to different parameter configurations. Note that under the ``Raw'' evaluation setting, the recall rate is guaranteed to increase monotonically as the threshold decreases, since no edge thinning is involved. As a result average precision (AP) is well-defined, and can be used as another metric to measure the system performance. In addition, $\lambda=0$ essentially means removing the Markov smoothness term in edge prior, while having $\sigma_x=\sigma_y=4$ indicates removing the kernel bias.

Results show that alignment without smoothness prior produces even sharper edges but worse results under the traditional ``Thin'' setting. On the other hand, when smoothness is imposed, decreasing $\sigma_y$ (less alignment flexibility) does not make great difference under the ``Thin'' setting, but considerably drops the performance under the ``Raw'' setting. One could see that biased kernel and Markov prior both benefit the performance. As a result of the parameter analysis, we choose $\sigma_x=1, \sigma_y=4, \lambda=0.02$ since this configuration gives the best trade-off under both ``Thin'' and ``Raw'' settings. The selected parameters are used to train SEAL on the SBD trainval set, with the performance on the test set reported and compared with other baselines in the main paper.

\begin{table*}[t!]
	\centering
	\caption{Evaluation of different SEAL models on the SBD validation set with $0.02$ distance tolerance. Results are measured by $\%$.\label{tb:sbd_val}}
	\resizebox{\textwidth}{!}{\begin{tabular}{c|c|c|c|c|c|c|c|c|c|c|c|c|c|c|c|c|c|c|c|c|c|c}
			Metric & $\sigma_x, \sigma_y, \lambda$ & aero & bike & bird & boat & bottle & bus & car & cat & chair & cow & table & dog & horse & mbike & person & plant & sheep & sofa & train & tv & ~mean~\\
			\hline \hline
			\multirow{7}{0.08\linewidth}{\centering{MF\\(Thin)}}
			& 4, 4, 0 & 86.1 & 74.5 & 81.8 & 59.4 & 67.5 & 82.8 & 79.7 & 82.6 & 55.3 & 76.5 & 47.5 & 81.1 & 83.1 & 75.2 & 80.3 & 54.1 & 76.3 & 49.1 & 74.3 & 65.4 & 71.6\\
			& 1, 4, 0 & \textbf{87.0} & 75.9 & 83.1 & 61.7 & 67.9 & 83.6 & 81.1 & 83.6 & 56.2 & \textbf{77.4} & 47.6 & 81.7 & 84.0 & 76.8 & 81.4 & 55.8 & 76.7 & 48.6 & 76.2 & 66.9 & 72.7\\
			& 1, 4, 0.01 & 86.2 & 77.4 & 83.5 & 61.3 & 69.8 & 84.9 & 82.2 & 84.8 & 56.3 & 76.1 & 47.6 & 82.2 & 83.8 & 77.7 & 81.9 & \textbf{58.5} & 76.8 & 49.5 & 77.8 & 68.0 & 73.3\\
			& \textbf{1, 4, 0.02} & 86.5 & 77.4 & \textbf{83.7} & 61.4 & 68.7 & 84.9 & 82.1 & \textbf{84.9} & 56.1 & 76.1 & 47.6 & 82.8 & 83.5 & \textbf{77.8} & 81.9 & 58.0 & 77.1 & \textbf{50.4} & \textbf{77.9} & \textbf{68.3} & \textbf{73.4}\\
			& \textbf{1, 4, 0.04} & 86.6 & 77.5 & 83.2 & \textbf{62.1} & \textbf{69.9} & 84.7 & 81.8 & 84.5 & 56.0 & 76.2 & 48.2 & \textbf{82.9} & \textbf{84.7} & 77.7 & 82.0 & 58.1 & 77.3 & 49.8 & \textbf{77.9} & 67.8 & \textbf{73.4}\\
			& \textbf{1, 3, 0.02} & \textbf{87.0} & \textbf{78.0} & 83.3 & 61.7 & 68.9 & \textbf{85.0} & 82.1 & 84.6 & 56.6 & 76.2 & \textbf{48.5} & 82.3 & 84.2 & 77.6 & \textbf{82.1} & 58.0 & 77.2 & 49.7 & 77.5 & 67.7 & \textbf{73.4}\\
			& \textbf{1, 2, 0.02} & 86.8 & 77.9 & 83.4 & 61.0 & 68.8 & 84.9 & 81.7 & 84.2 & \textbf{57.0} & 77.3 & 47.4 & 82.2 & 83.8 & 77.7 & 82.0 & \textbf{58.5} & \textbf{77.6} & 50.3 & 77.6 & 68.1 & \textbf{73.4}\\
			\hline \hline
			\multirow{7}{0.08\linewidth}{\centering{MF\\(Raw)}}
			& 4, 4, 0 & 87.7 & 76.9 & 83.0 & 59.7 & \textbf{67.2} & 84.2 & 80.6 & 83.1 & 55.7 & \textbf{78.0} & \textbf{48.2} & 81.8 & 84.3 & \textbf{77.9} & 81.8 & 55.9 & \textbf{77.0} & \textbf{49.7} & \textbf{76.5} & 66.8 & 72.8\\
			& \textbf{1, 4, 0} & \textbf{88.1} & \textbf{77.0} & \textbf{83.3} & \textbf{60.8} & 66.8 & \textbf{84.3} & \textbf{81.0} & \textbf{83.4} & \textbf{56.0} & 77.9 & 48.0 & \textbf{82.0} & \textbf{84.4} & 77.6 & \textbf{82.0} & \textbf{56.8} & 76.2 & 48.8 & 76.4 & \textbf{67.7} & \textbf{72.9}\\
			& 1, 4, 0.01 & 84.8 & 71.7 & 81.1 & 56.6 & 65.6 & 82.8 & 78.2 & 80.2 & 54.3 & 74.6 & 44.9 & 79.3 & 81.7 & 73.6 & 79.1 & 54.5 & 74.1 & 49.1 & 73.2 & 64.8 & 70.2\\
			& 1, 4, 0.02 & 84.5 & 71.3 & 80.6 & 56.5 & 64.5 & 82.8 & 78.0 & 80.3 & 53.9 & 74.9 & 44.7 & 78.9 & 81.2 & 73.7 & 78.9 & 53.4 & 74.1 & 49.5 & 72.9 & 64.9 & 70.0\\
			& 1, 4, 0.04 & 84.5 & 71.1 & 80.4 & 56.5 & 65.4 & 82.5 & 78.1 & 79.7 & 53.7 & 73.7 & 44.3 & 79.0 & 82.0 & 72.9 & 79.0 & 53.4 & 73.3 & 49.1 & 72.9 & 64.7 & 69.8\\
			& 1, 3, 0.02 & 83.4 & 70.7 & 79.9 & 54.8 & 64.4 & 82.5 & 77.8 & 79.4 & 53.8 & 72.8 & 45.5 & 78.3 & 81.5 & 72.9 & 78.8 & 52.6 & 73.6 & 49.3 & 72.6 & 64.0 & 69.4\\
			& 1, 2, 0.02 & 81.7 & 69.1 & 79.1 & 53.9 & 63.9 & 81.5 & 76.8 & 77.4 & 53.6 & 73.0 & 43.2 & 77.1 & 80.7 & 70.9 & 77.8 & 51.5 & 72.5 & 49.4 & 71.3 & 63.9 & 68.4\\
			\hline \hline
			\multirow{7}{0.08\linewidth}{\centering{AP\\(Raw)}}
			& 4, 4, 0 & 71.1 & 75.2 & 67.2 & 56.0 & 63.4 & 70.3 & 75.8 & 72.0 & \textbf{53.8} & 76.4 & \textbf{42.5} & 71.4 & 78.5 & 75.7 & 76.9 & 54.1 & 73.1 & 41.8 & 68.3 & 61.5 & 66.2\\
			& 1, 4, 0 & 78.9 & \textbf{77.9} & 76.0 & \textbf{59.7} & 65.1 & 76.8 & 80.5 & 79.4 & 53.3 & \textbf{78.8} & 42.2 & 77.8 & 83.4 & \textbf{78.7} & 81.5 & \textbf{54.9} & 75.9 & 40.6 & 73.4 & \textbf{64.4} & 70.0\\
			& \textbf{1, 4, 0.01} & 87.7 & 75.1 & \textbf{84.8} & 54.6 & \textbf{66.7} & 86.0 & \textbf{82.8} & \textbf{85.1} & 52.0 & 77.9 & 39.1 & 82.8 & 85.7 & 77.5 & 83.7 & 51.9 & \textbf{77.3} & 42.5 & \textbf{76.0} & 62.4 & \textbf{71.6}\\
			& 1, 4, 0.02 & 87.8 & 74.9 & 84.3 & 53.8 & 65.5 & 85.7 & 82.5 & \textbf{85.1} & 51.1 & 78.2 & 38.2 & 82.9 & 85.3 & 77.9 & 83.4 & 51.5 & 76.7 & 42.3 & 75.6 & 62.1 & 71.2\\
			& 1, 4, 0.04 & \textbf{88.0} & 74.1 & 84.0 & 54.7 & 66.2 & 85.9 & 82.4 & 84.4 & 51.2 & 77.3 & 38.7 & \textbf{83.0} & \textbf{86.4} & 76.9 & \textbf{83.9} & 50.5 & 76.2 & 42.4 & 75.5 & 61.5 & 71.2\\
			& 1, 3, 0.02 & 87.6 & 74.3 & 83.9 & 53.3 & 64.7 & \textbf{86.1} & 82.2 & 84.2 & 52.0 & 76.9 & 39.9 & 82.6 & 85.6 & 76.7 & 83.6 & 49.6 & 76.4 & 41.9 & 75.1 & 61.2 & 70.9\\
			& 1, 2, 0.02 & 86.0 & 72.7 & 83.0 & 51.6 & 64.2 & 85.2 & 81.0 & 82.7 & 50.8 & 78.1 & 38.2 & 81.1 & 85.4 & 75.2 & 82.9 & 48.3 & 75.3 & \textbf{43.0} & 74.1 & 60.9 & 70.0\\
	\end{tabular}}
\end{table*}

To better study the behavior of SEAL, we also include an ablation study by evaluating the above models on the re-annotated SBD test set with $0.0075$ distance tolerance. We use re-annotated labels since their high quality can better capture the algorithm performance with higher precision. The results are reported in Table~\ref{tb:sbd_val2}, where one can draw conclusions similar to the evaluation results on the SBD validation set. Note that this experiment is purely for ablation study and is independent from parameter selection.

\begin{table*}[t!]
	\centering
	\caption{Evaluation of different SEAL models on the re-annotated SBD test set with $0.0075$ distance tolerance. Results are measured by $\%$.\label{tb:sbd_val2}}
	\resizebox{\textwidth}{!}{\begin{tabular}{c|c|c|c|c|c|c|c|c|c|c|c|c|c|c|c|c|c|c|c|c|c|c}
			Metric & $\sigma_x, \sigma_y, \lambda$ & aero & bike & bird & boat & bottle & bus & car & cat & chair & cow & table & dog & horse & mbike & person & plant & sheep & sofa & train & tv & ~mean~\\
			\hline \hline
			\multirow{7}{0.08\linewidth}{\centering{MF\\(Thin)}} 
			& 4, 4, 0    & 76.0 & 61.9 & 72.6 & 49.1 & 63.6 & 74.3 & 66.5 & 74.2 & 48.1 & 68.6 & 38.6 & 74.9 & 72.2 & 62.7 & 75.0 & 48.3 & 72.1 & 49.6 & 67.9 & 53.7 & 63.5\\
			& 1, 4, 0    & 77.6 & 64.1 & 75.3 & 51.4 & 65.1 & 76.8 & 67.9 & 76.3 & 49.6 & 70.5 & 39.4 & 76.5 & 75.0 & 64.3 & 76.8 & \textbf{49.8} & 72.5 & 49.9 & 70.5 & 54.8 & 65.2\\
			& 1, 4, 0.01 & 77.7 & 65.3 & 76.0 & 52.0 & \textbf{68.4} & 79.4 & 70.4 & 78.7 & \textbf{50.0} & 71.0 & 40.5 & 77.7 & 74.9 & 65.9 & 78.3 & 48.9 & 73.7 & \textbf{51.3} & 73.5 & 57.2 & 66.5\\
			& \textbf{1, 4, 0.02} & 77.6 & \textbf{65.7} & 76.1 & 52.1 & \textbf{68.4} & 79.8 & 70.6 & \textbf{79.1} & 49.8 & 71.1 & 40.7 & \textbf{78.2} & 75.3 & \textbf{66.1} & 78.2 & 49.4 & \textbf{74.1} & 51.1 & 73.5 & 57.4 & \textbf{66.7}\\
			& 1, 4, 0.04 & \textbf{78.1} & 64.7 & \textbf{76.4} & 51.6 & 68.1 & \textbf{80.1} & 70.6 & 78.3 & 49.5 & 71.0 & \textbf{41.2} & 77.6 & \textbf{75.4} & 65.7 & 78.3 & 48.4 & 73.4 & 51.0 & 73.4 & 57.2 & 66.5\\
			& 1, 3, 0.02 & 77.5 & 64.8 & 76.0 & \textbf{52.5} & 67.8 & 79.5 & 71.0 & 78.6 & 49.6 & 71.9 & 40.7 & 77.8 & 75.3 & 65.6 & \textbf{78.4} & 47.8 & 73.1 & 51.2 & \textbf{73.9} & \textbf{57.6} & 66.5\\
			& 1, 2, 0.02 & 76.6 & 64.0 & 75.9 & 52.3 & 68.1 & \textbf{80.1} & \textbf{71.1} & 78.8 & 49.4 & \textbf{72.1} & 40.4 & 77.2 & 75.1 & 65.6 & 78.1 & 47.1 & 73.3 & 50.6 & 73.7 & 57.4 & 66.4\\
			\hline \hline
			\multirow{7}{0.08\linewidth}{\centering{MF\\(Raw)}}
			& 4, 4, 0 & 78.2 & 65.1 & 76.0 & 52.9 & 64.1 & 76.6 & 69.5 & 76.7 & \textbf{51.3} & 70.4 & 40.3 & 76.8 & 75.2 & 65.3 & 76.8 & 51.7 & \textbf{74.0} & \textbf{49.6} & 71.4 & 56.2 & 65.9\\
			& \textbf{1, 4, 0} & \textbf{78.7} & \textbf{65.7} & \textbf{77.5} & \textbf{53.2} & 65.5 & \textbf{77.4} & \textbf{69.9} & \textbf{77.8} & \textbf{51.3} & \textbf{71.0} & \textbf{40.5} & \textbf{77.7} & \textbf{76.6} & \textbf{66.2} & \textbf{77.6} & \textbf{51.8} & 73.8 & 49.3 & \textbf{72.3} & 56.2 & \textbf{66.5}\\
			& 1, 4, 0.01 & 75.5 & 59.8 & 75.8 & 50.7 & \textbf{65.8} & 75.9 & 68.4 & 75.5 & 50.0 & 68.6 & 39.6 & 74.6 & 72.9 & 62.9 & 74.4 & 47.9 & 72.4 & 48.5 & 70.4 & 56.5 & 64.3\\
			& 1, 4, 0.02 & 74.8 & 60.2 & 75.2 & 50.7 & 65.5 & 76.2 & 68.1 & 75.5 & 49.2 & 67.9 & 39.1 & 74.3 & 73.0 & 62.1 & 74.1 & 48.1 & 72.4 & 48.8 & 69.8 & \textbf{57.3} & 64.1\\
			& 1, 4, 0.04 & 75.3 & 59.8 & 75.5 & 50.3 & 65.2 & 76.3 & 68.1 & 74.9 & 49.1 & 67.5 & 39.6 & 74.0 & 72.9 & 61.5 & 73.9 & 48.4 & 71.7 & 48.2 & 70.0 & 56.5 & 63.9\\
			& 1, 3, 0.02 & 73.8 & 59.0 & 74.7 & 50.0 & 65.6 & 75.4 & 68.0 & 74.9 & 49.0 & 68.0 & 38.6 & 73.9 & 72.2 & 61.6 & 73.8 & 46.4 & 70.9 & 48.6 & 69.7 & 56.4 & 63.5\\
			& 1, 2, 0.02 & 72.2 & 57.3 & 73.4 & 49.1 & 64.3 & 74.5 & 67.3 & 73.3 & 48.1 & 67.6 & 38.1 & 71.9 & 71.3 & 60.4 & 72.5 & 45.5 & 70.8 & 47.6 & 68.3 & 55.7 & 62.5\\
			\hline \hline
			\multirow{7}{0.08\linewidth}{\centering{AP\\(Raw)}}
			& 4, 4, 0 & 67.1 & 64.3 & 62.3 & 48.1 & 58.5 & 64.0 & 61.8 & 65.9 & \textbf{48.0} & 63.8 & \textbf{34.0} & 66.0 & 67.9 & 61.7 & 72.3 & 47.7 & 66.4 & 44.7 & 64.4 & 51.0 & 59.0\\
			& 1, 4, 0 & 74.4 & \textbf{67.2} & 71.1 & \textbf{50.1} & 63.3 & 71.1 & 66.3 & 74.9 & 47.6 & 67.4 & \textbf{34.0} & 74.5 & 75.1 & 65.1 & 78.2 & \textbf{48.0} & 70.1 & 44.2 & 71.0 & 51.8 & 63.3\\
			& \textbf{1, 4, 0.01} & \textbf{80.9} & 63.4 & 79.1 & 48.4 & \textbf{68.1} & 79.2 & \textbf{70.1} & \textbf{80.4} & 46.9 & 68.1 & 33.6 & \textbf{79.6} & 75.9 & \textbf{64.4} & \textbf{80.4} & 43.6 & 72.6 & 43.8 & 73.9 & 54.3 & \textbf{65.3}\\
			& \textbf{1, 4, 0.02} & 80.1 & 64.0 & 78.8 & 48.4 & 67.6 & 79.6 & 69.8 & 80.3 & 46.4 & 67.6 & 33.3 & 79.3 & 76.1 & 64.1 & 80.2 & 43.8 & \textbf{73.2} & \textbf{44.5} & 73.6 & \textbf{55.0} & \textbf{65.3}\\
			& 1, 4, 0.04 & 80.5 & 63.2 & \textbf{79.5} & 47.8 & 66.7 & \textbf{79.8} & 69.9 & 79.7 & 46.4 & 67.9 & 33.8 & 78.9 & \textbf{76.2} & 63.2 & 80.1 & 43.8 & 72.3 & 43.9 & 73.0 & 53.9 & 65.0\\
			& 1, 3, 0.02 & 79.4 & 62.1 & 78.5 & 47.7 & 67.4 & 78.3 & 69.8 & 80.1 & 46.2 & \textbf{69.3} & 31.9 & 78.6 & 75.5 & 63.1 & 79.9 & 42.2 & 71.9 & 44.4 & \textbf{73.8} & 54.1 & 64.7\\
			& 1, 2, 0.02 & 78.2 & 60.4 & 77.6 & 46.5 & 66.8 & 78.2 & 69.3 & 78.1 & 45.6 & 68.7 & 31.8 & 76.7 & 74.9 & 62.6 & 78.7 & 40.8 & 70.6 & 43.2 & 72.4 & 54.0 & 63.8\\
	\end{tabular}}
\end{table*}

\subsection{Average precision on the SBD test}
We additionally present the average precision scores on the original and the re-annotated SBD test sets under the ``Raw'' setting. The results are reported in Table~\ref{tb:sbd_ap}, where one can observe the considerable correlation between average precision and maximum F-Measure.

\begin{table*}[t!]
	\centering
	\caption{AP scores on the SBD test set under the ``Raw'' setting. ``Orig.'' and ``R.A.'' denote ``Original'' and ``Re-annotated'', respectively. Scores are measured by $\%$.\label{tb:sbd_ap}}
	\resizebox{\textwidth}{!}{\begin{tabular}{c|l|c|c|c|c|c|c|c|c|c|c|c|c|c|c|c|c|c|c|c|c|c}
			Label & Method	& aero & bike & bird & boat & bottle & bus & car & cat & chair & cow & table & dog & horse & mbike & person & plant & sheep & sofa & train & tv & mean \\
			\hline \hline
			\multirow{4}{0.08\linewidth}{\centering{Orig.}}
			& CASENet   & 72.1 & 63.2 & 62.0 & 42.3 & 52.7 & 68.8 & 62.1 & 72.6 & 49.4 & 63.1 & 34.1 & 68.2 & 74.2 & 64.8 & 75.9 & 38.1 & 67.3 & 41.8 & 63.6 & 49.8 & 59.3\\
			& CASENet-S & 82.6 & 69.3 & 83.4 & 57.2 & 66.7 & 81.1 & 74.1 & 80.9 & 53.6 & 71.2 & 41.5 & 77.6 & 81.5 & 72.2 & 82.8 & 43.4 & 78.9 & 46.6 & 74.8 & 60.5 & 69.0\\
			& CASENet-C & 86.0 & 70.2 & 84.2 & 58.6 & 67.1 & 81.2 & 74.9 & 83.3 & 50.9 & \textbf{76.9} & 40.1 & 80.6 & 82.5 & 73.9 & 83.5 & \textbf{47.5} & 80.5 & 47.4 & 75.3 & 61.5 & 70.3\\
			& SEAL      & \textbf{87.0} & \textbf{73.9} & \textbf{85.7} & \textbf{62.0} & \textbf{70.0} & \textbf{84.4} & \textbf{77.7} & \textbf{85.2} & \textbf{55.4} & 76.6 & \textbf{43.2} & \textbf{82.4} & \textbf{83.7} & \textbf{75.7} & \textbf{85.2} & 47.1 & \textbf{81.4} & \textbf{48.5} & \textbf{78.3} & \textbf{64.2} & \textbf{72.4}\\
			\hline \hline
			\multirow{4}{0.08\linewidth}{\centering{R.A.}}
			& CASENet   & 65.2 & 51.1 & 55.2 & 34.7 & 47.4 & 63.4 & 54.5 & 66.2 & 41.7 & 53.0 & 26.4 & 64.0 & 66.1 & 48.1 & 69.4 & 34.9 & 60.2 & 38.2 & 58.6 & 40.8 & 52.0\\
			& CASENet-S & 75.5 & 57.6 & 75.1 & 45.7 & 62.8 & 75.4 & 66.3 & 76.2 & 45.7 & 65.7 & 31.1 & 73.9 & 73.0 & 59.9 & 76.4 & 39.3 & 70.3 & 42.2 & 70.1 & 50.6 & 61.6\\
			& CASENet-C & \textbf{81.1} & 59.5 & 77.6 & 46.1 & 62.6 & 75.4 & 65.9 & 78.8 & 43.2 & \textbf{70.9} & 32.3 & 78.4 & 75.0 & 62.6 & 77.9 & \textbf{46.8} & 72.0 & 44.9 & 69.5 & 50.7 & 63.6\\
			& SEAL      & \textbf{81.1} & \textbf{63.6} & \textbf{79.5} & \textbf{49.8} & \textbf{67.1} & \textbf{79.7} & \textbf{70.0} & \textbf{81.0} & \textbf{47.4} & 70.5 & \textbf{34.5} & \textbf{79.7} & \textbf{76.5} & \textbf{63.5} & \textbf{80.4} & 45.2 & \textbf{74.0} & \textbf{45.4} & \textbf{74.2} & \textbf{54.1} & \textbf{65.9}\\
	\end{tabular}}
\end{table*}

\subsection{Precision-recall curves on SBD test}
\begin{figure}[tbh]
	\centering
	\includegraphics[width=.244\textwidth]{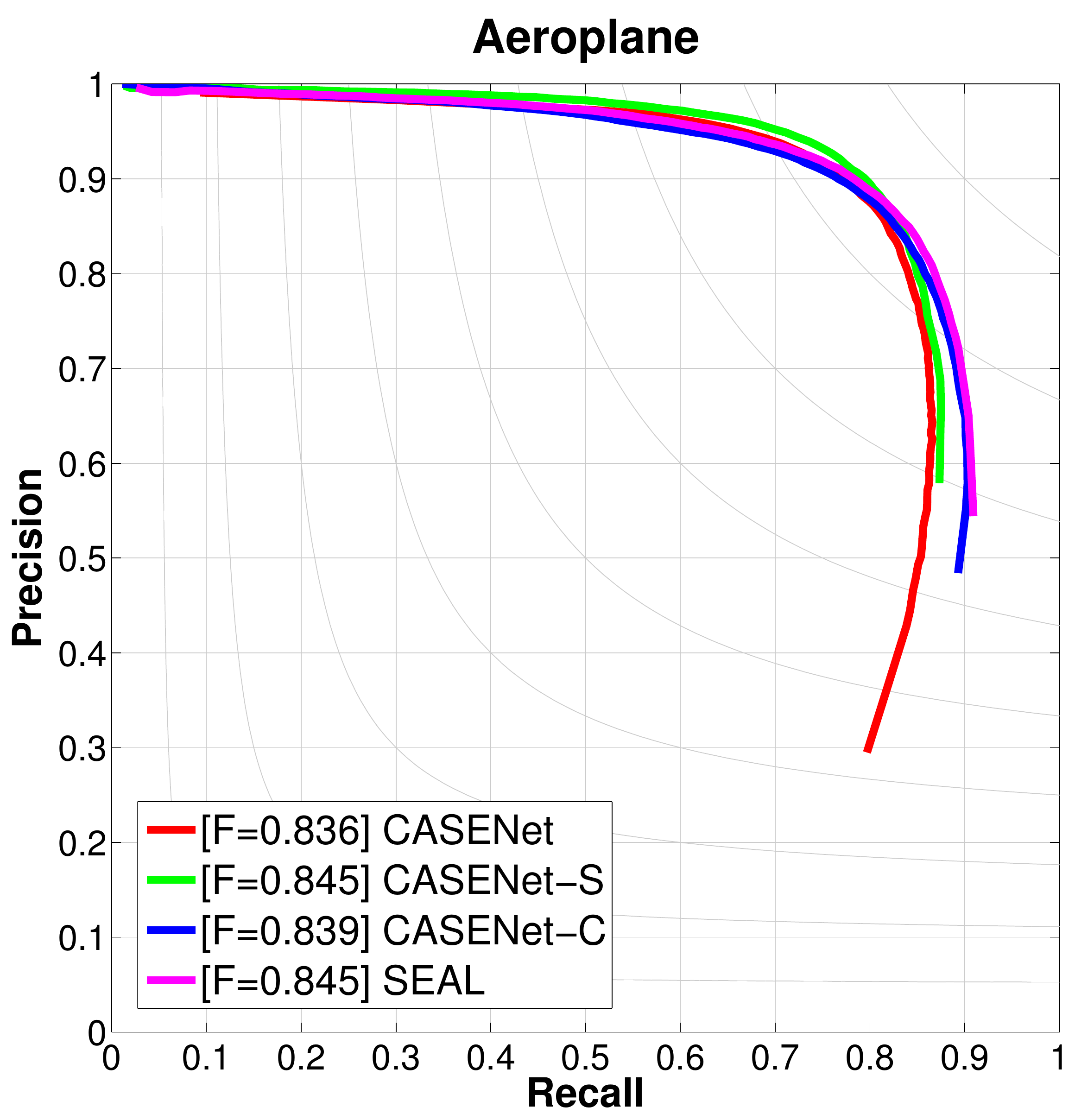}
	\includegraphics[width=.244\textwidth]{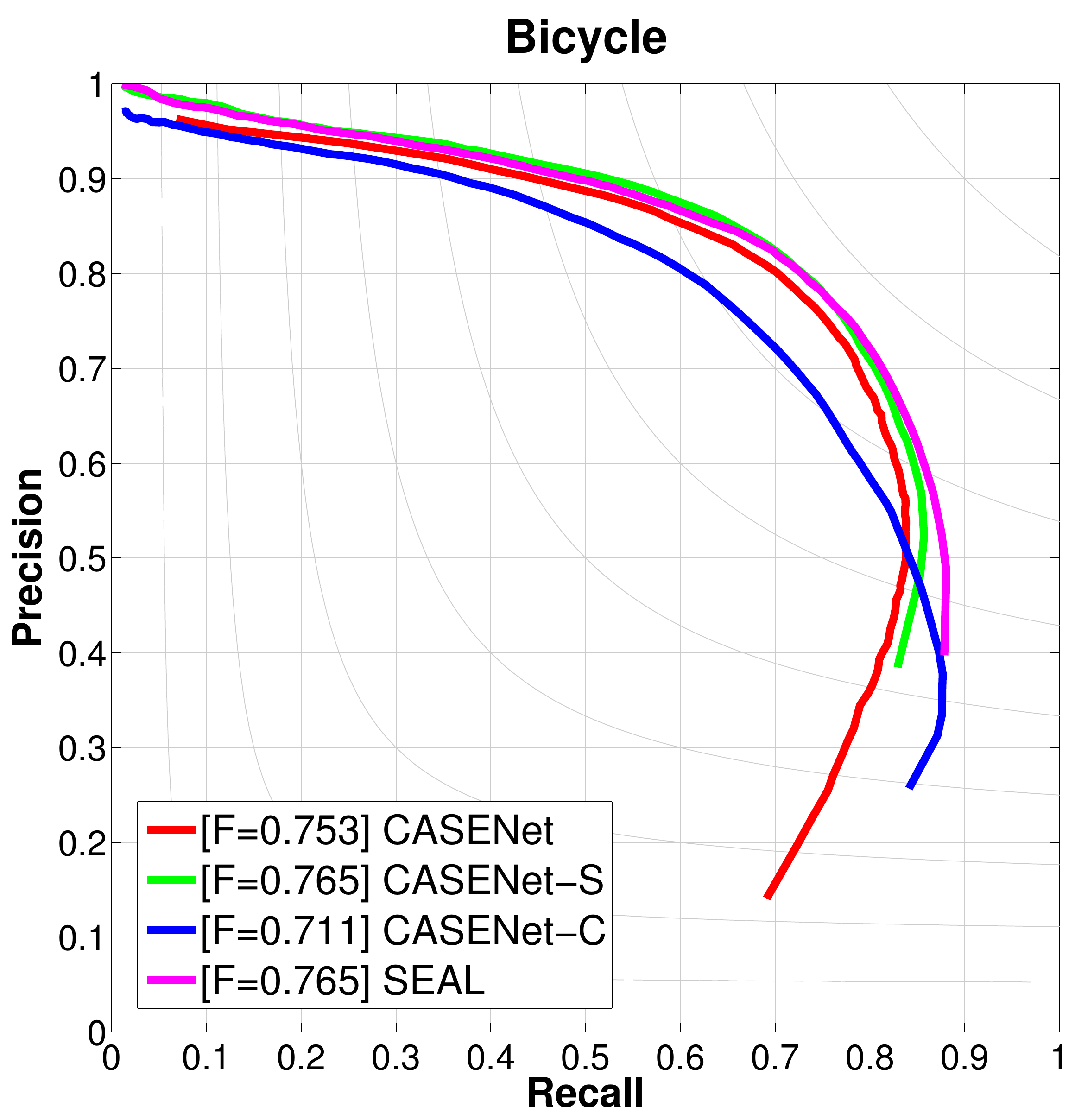}
	\includegraphics[width=.244\textwidth]{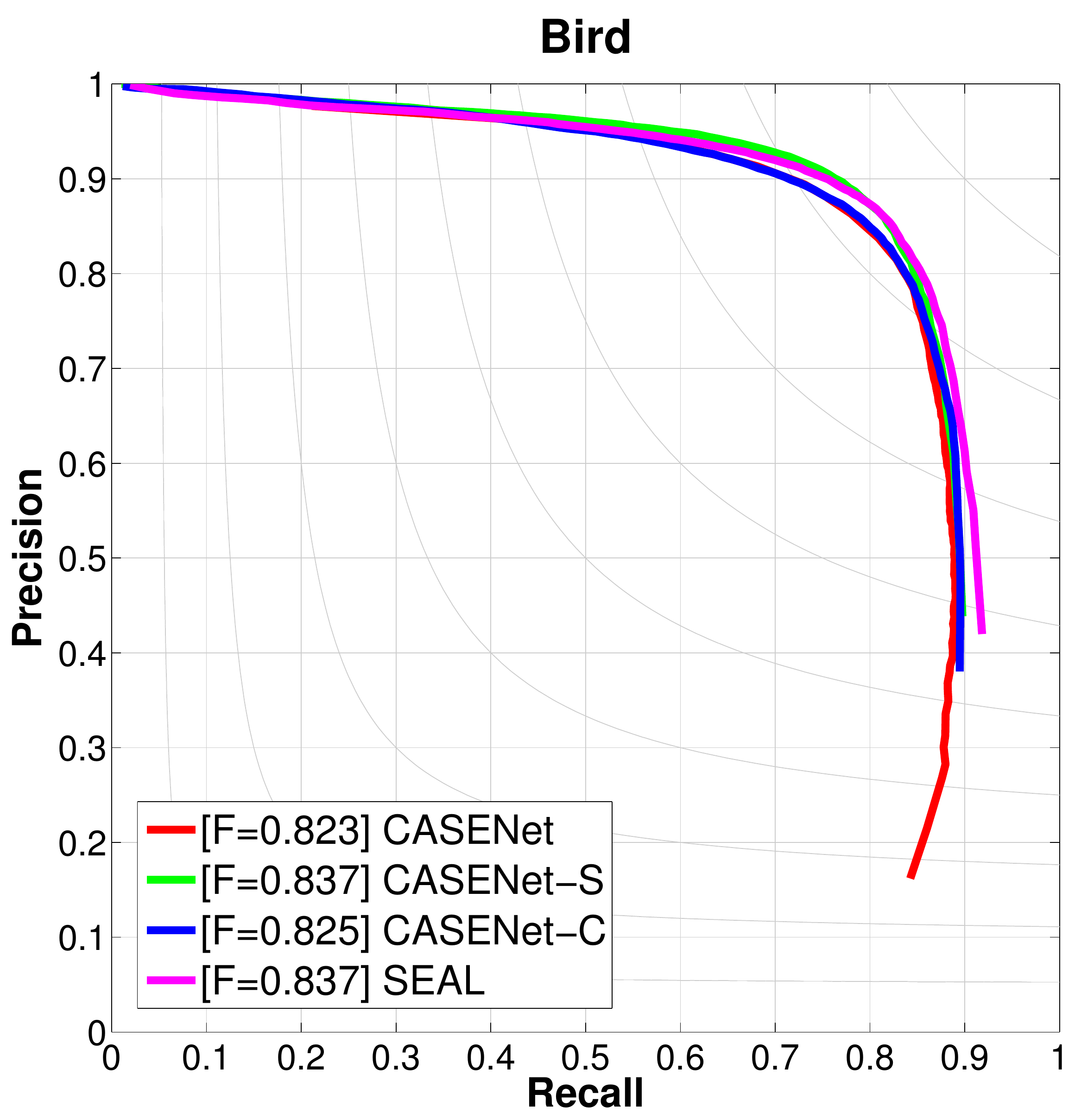}
	\includegraphics[width=.244\textwidth]{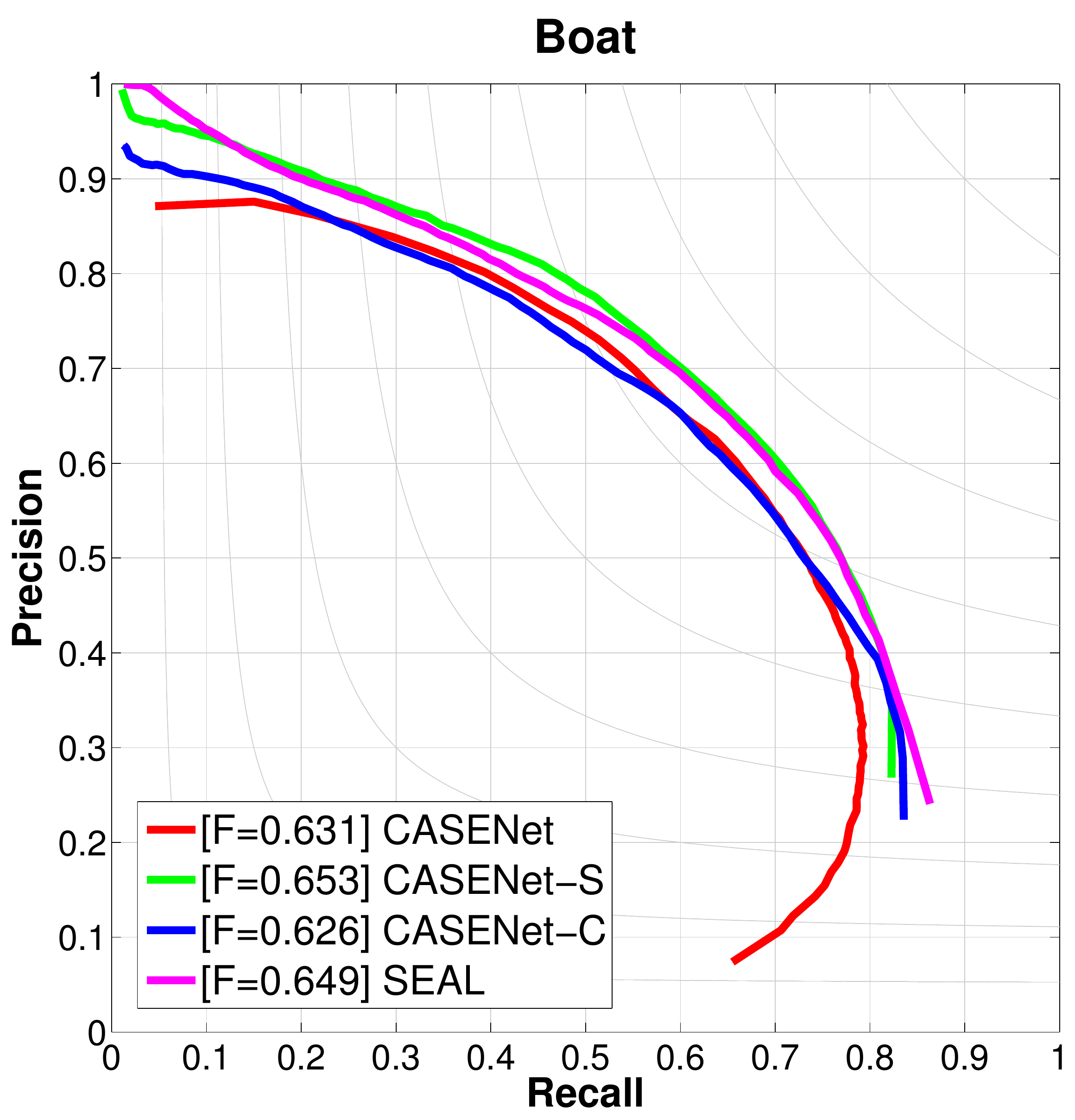}\\
	
	\includegraphics[width=.244\textwidth]{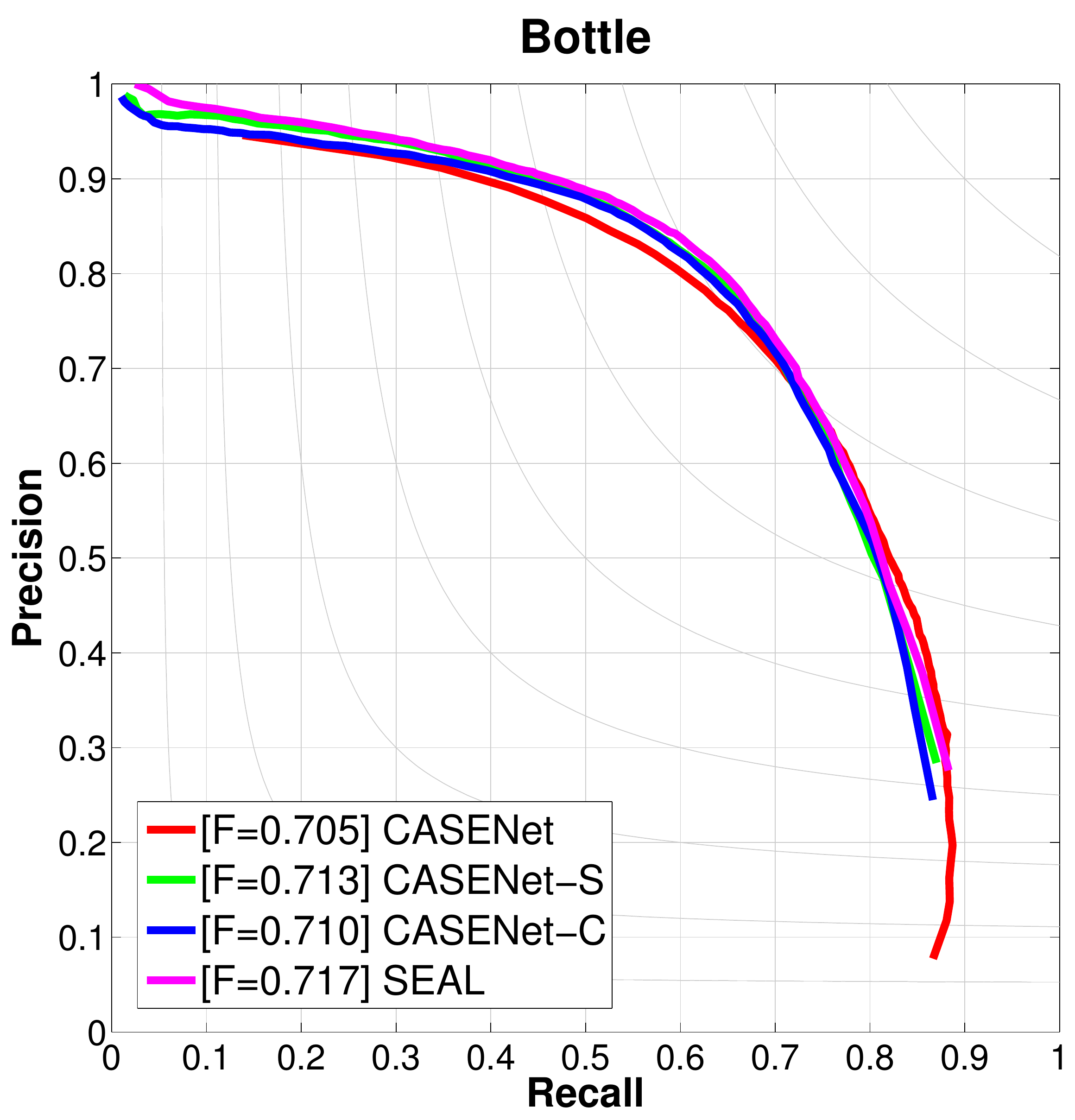}
	\includegraphics[width=.244\textwidth]{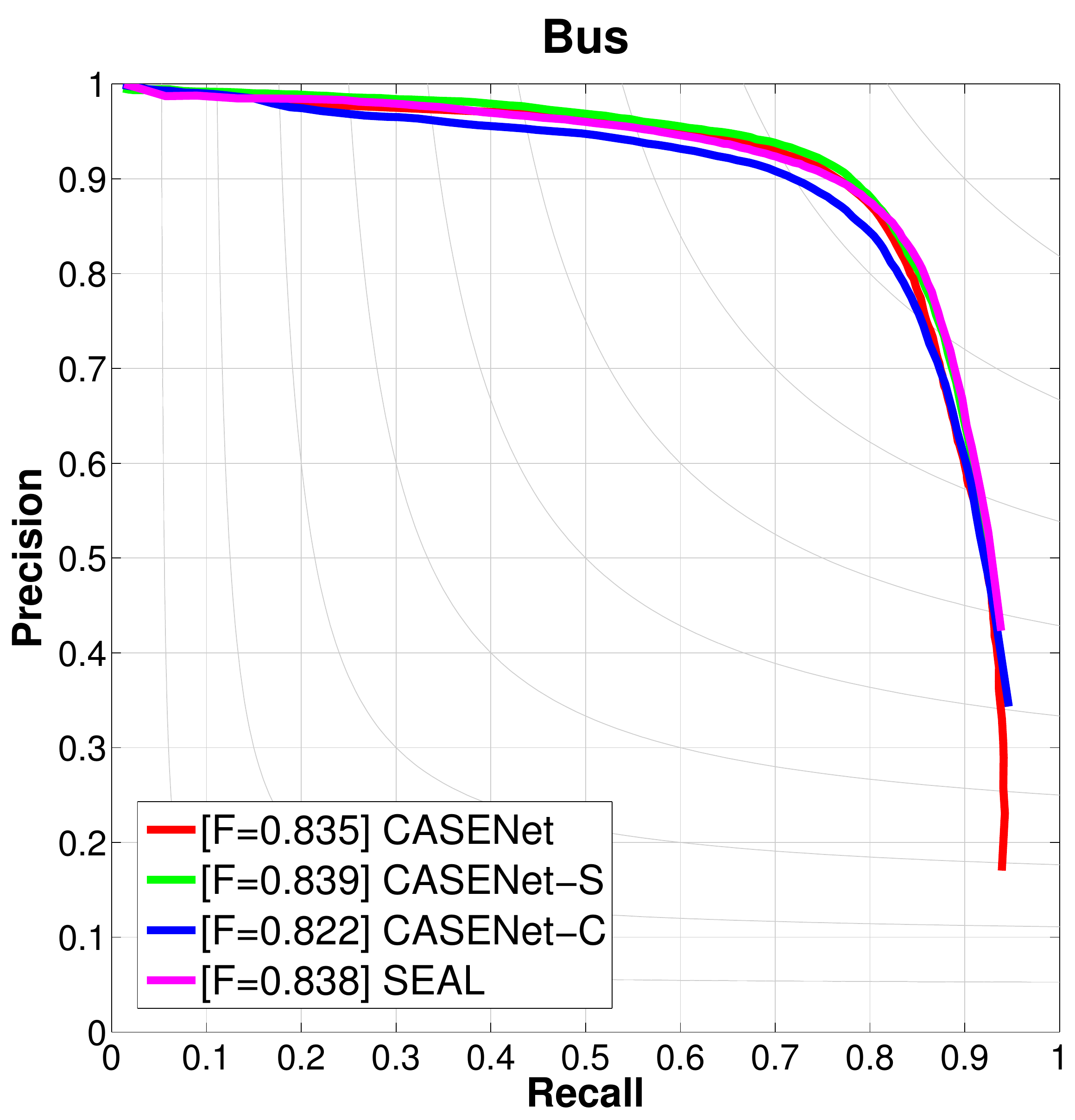}
	\includegraphics[width=.244\textwidth]{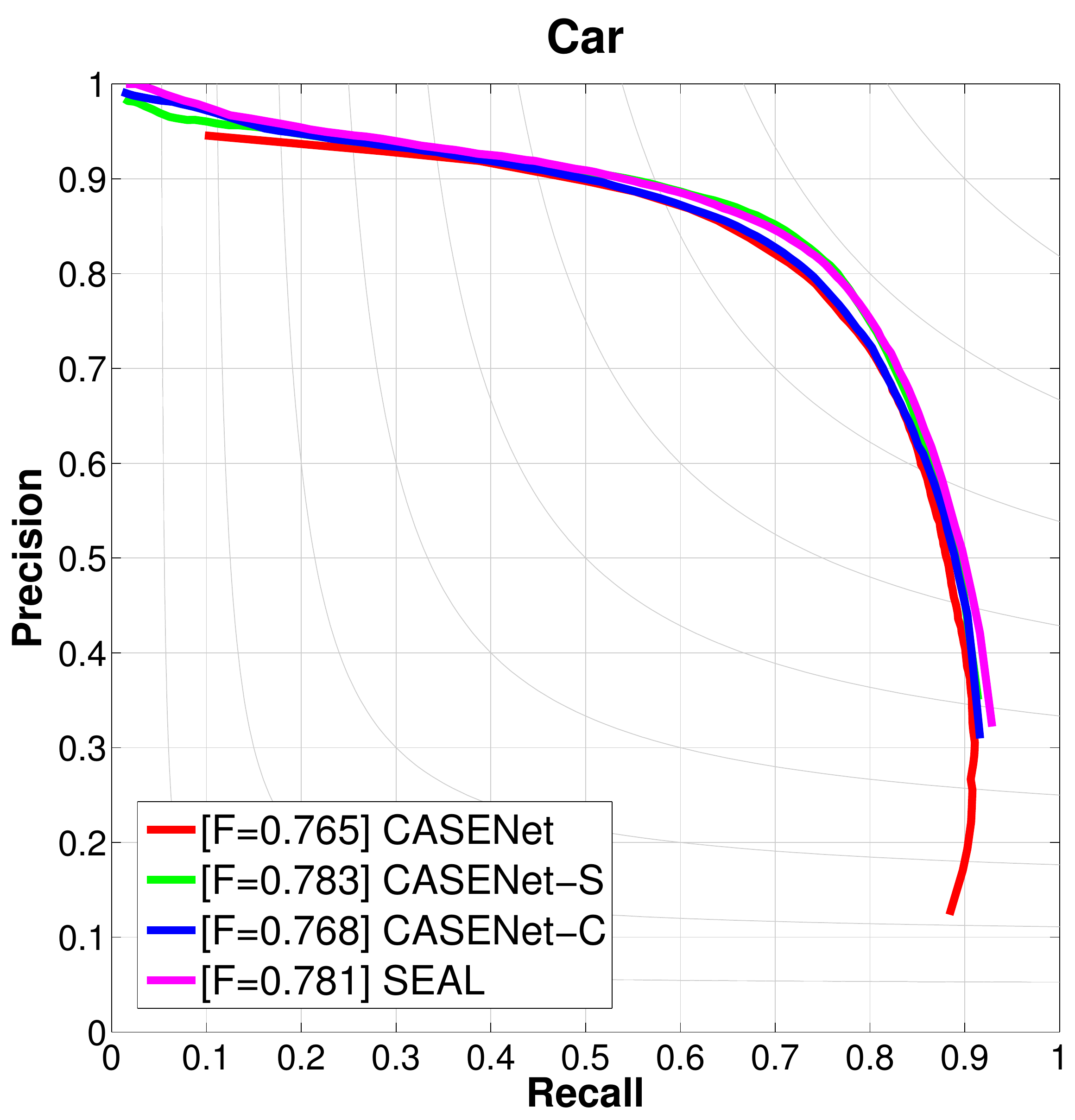}
	\includegraphics[width=.244\textwidth]{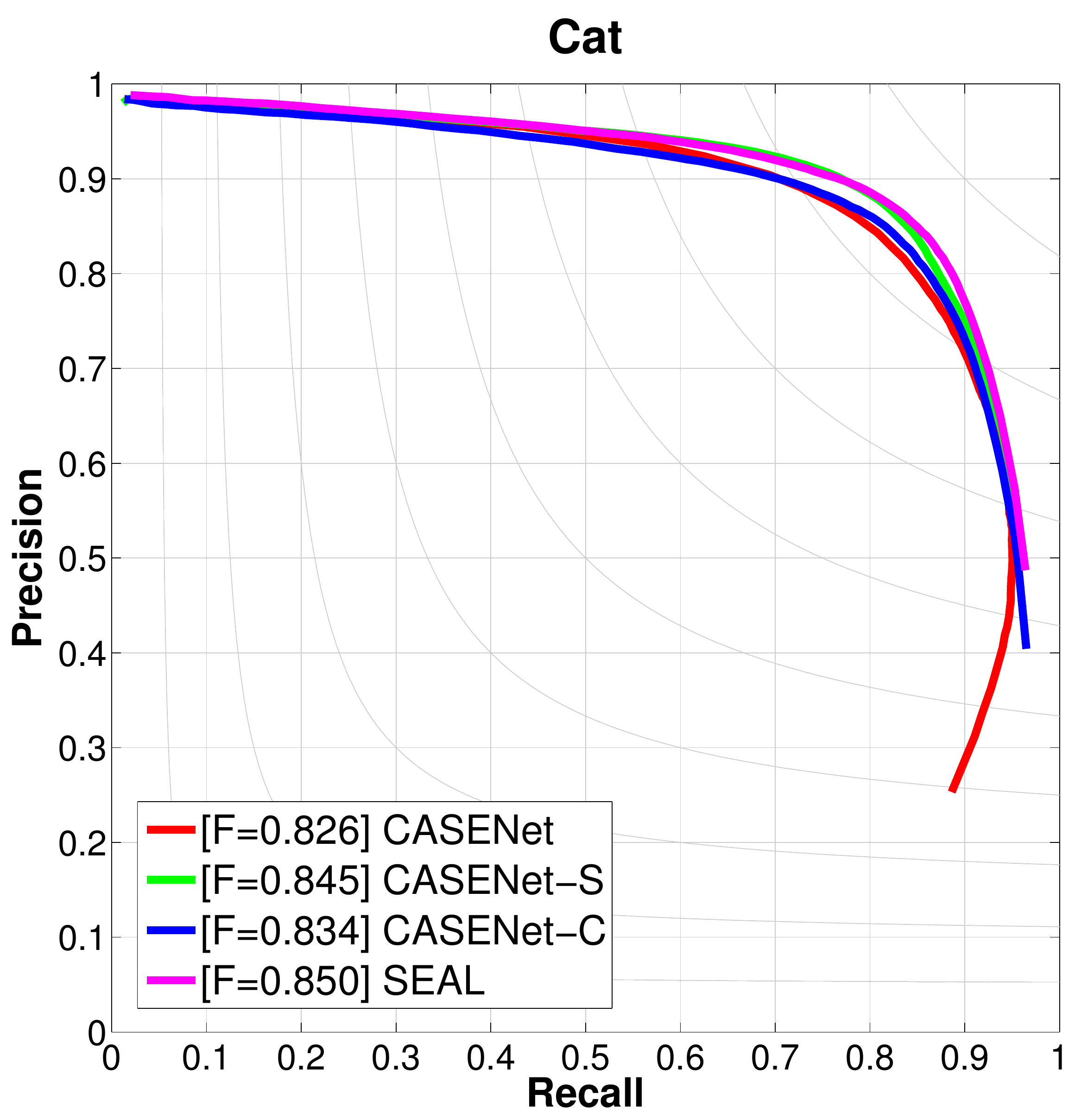}\\
	
	\includegraphics[width=.244\textwidth]{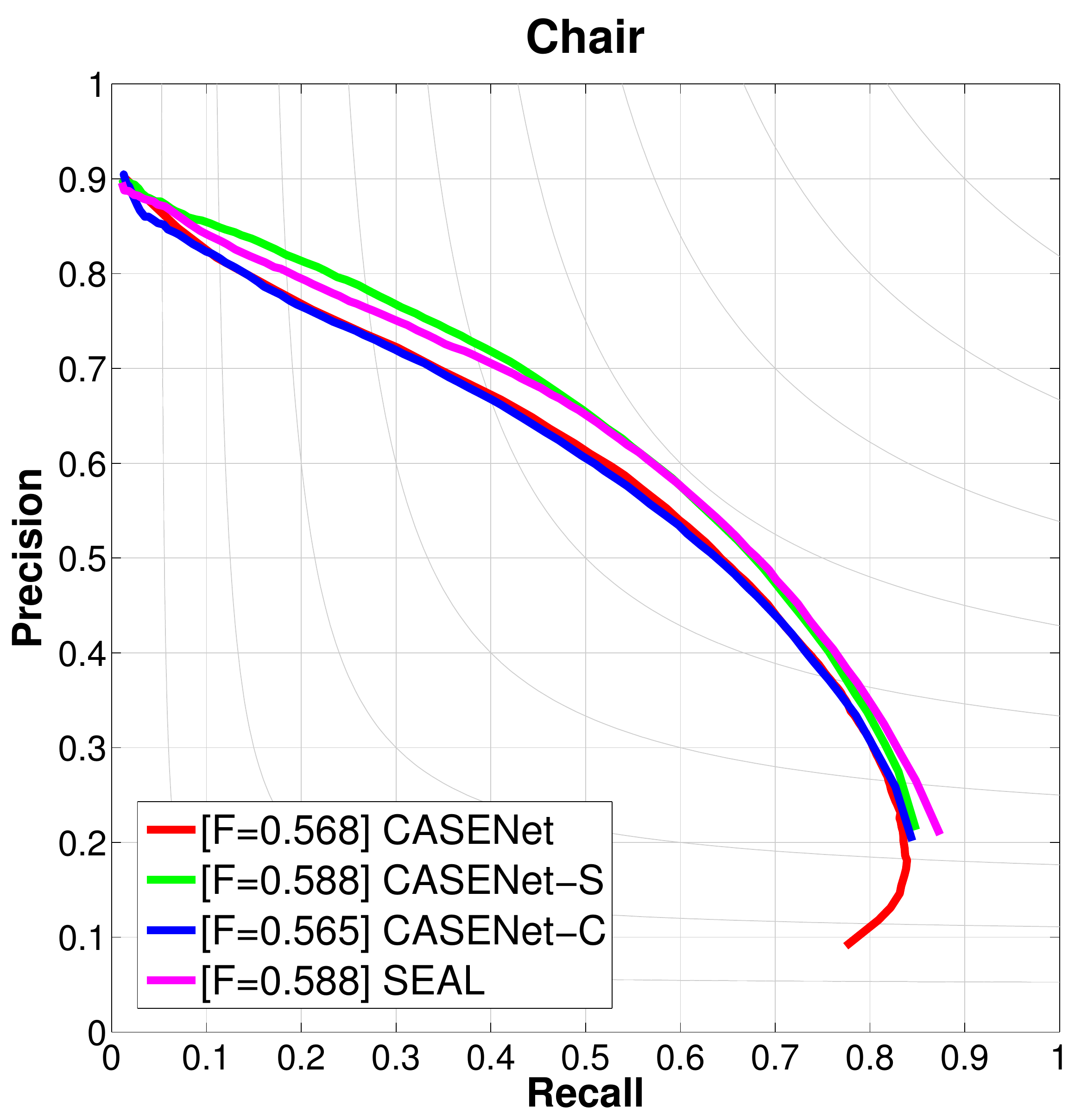}
	\includegraphics[width=.244\textwidth]{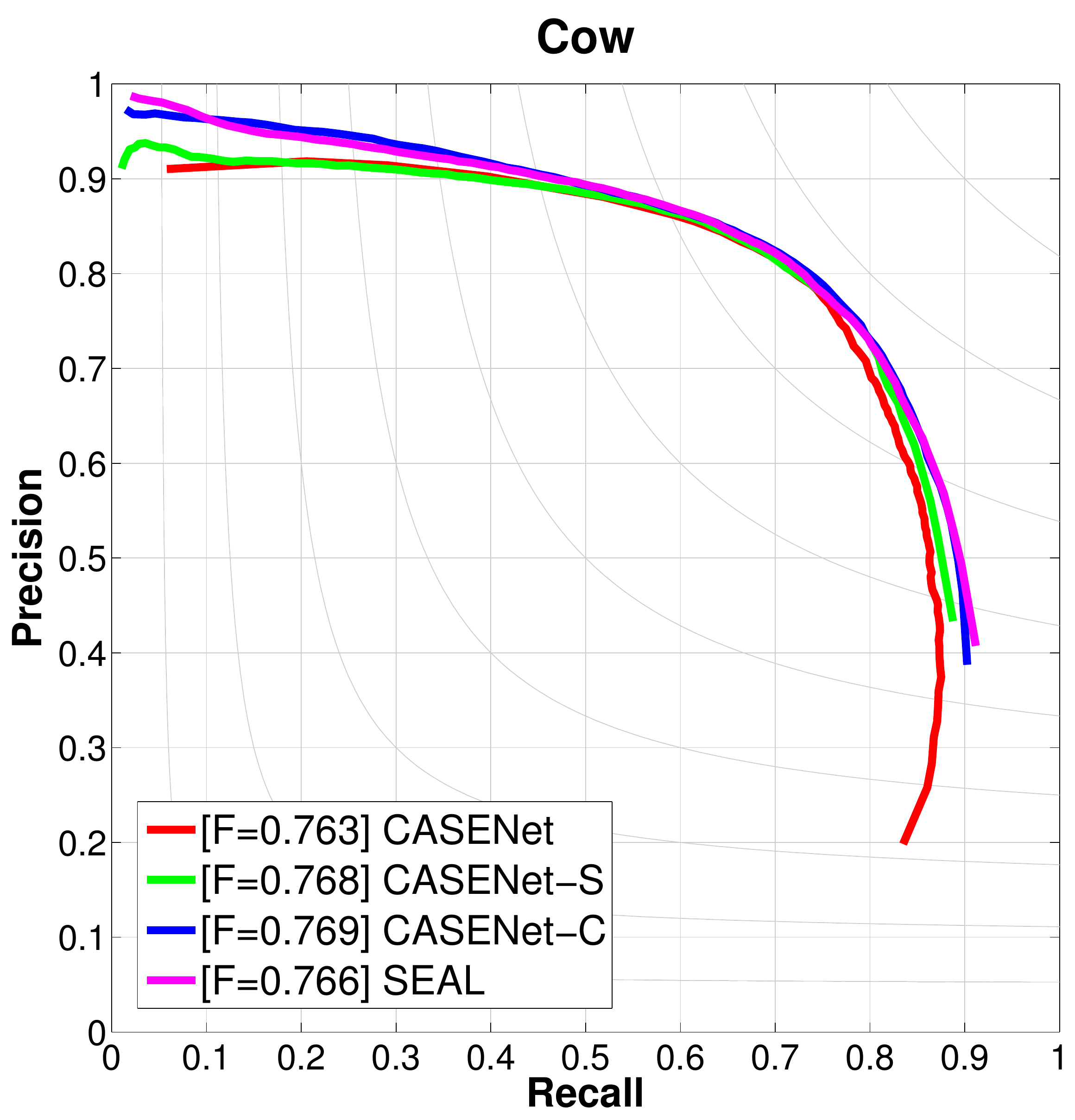}
	\includegraphics[width=.244\textwidth]{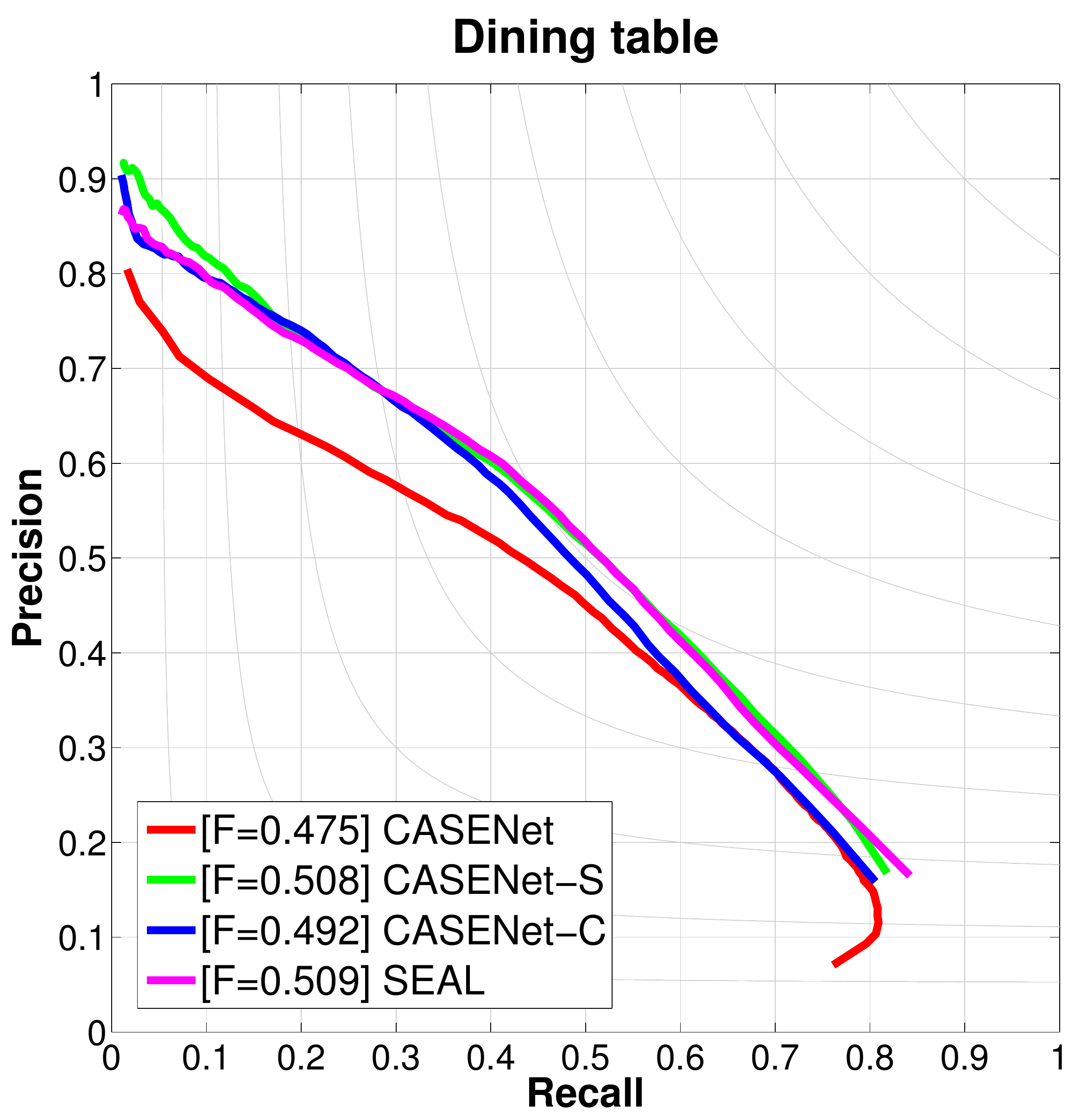}
	\includegraphics[width=.244\textwidth]{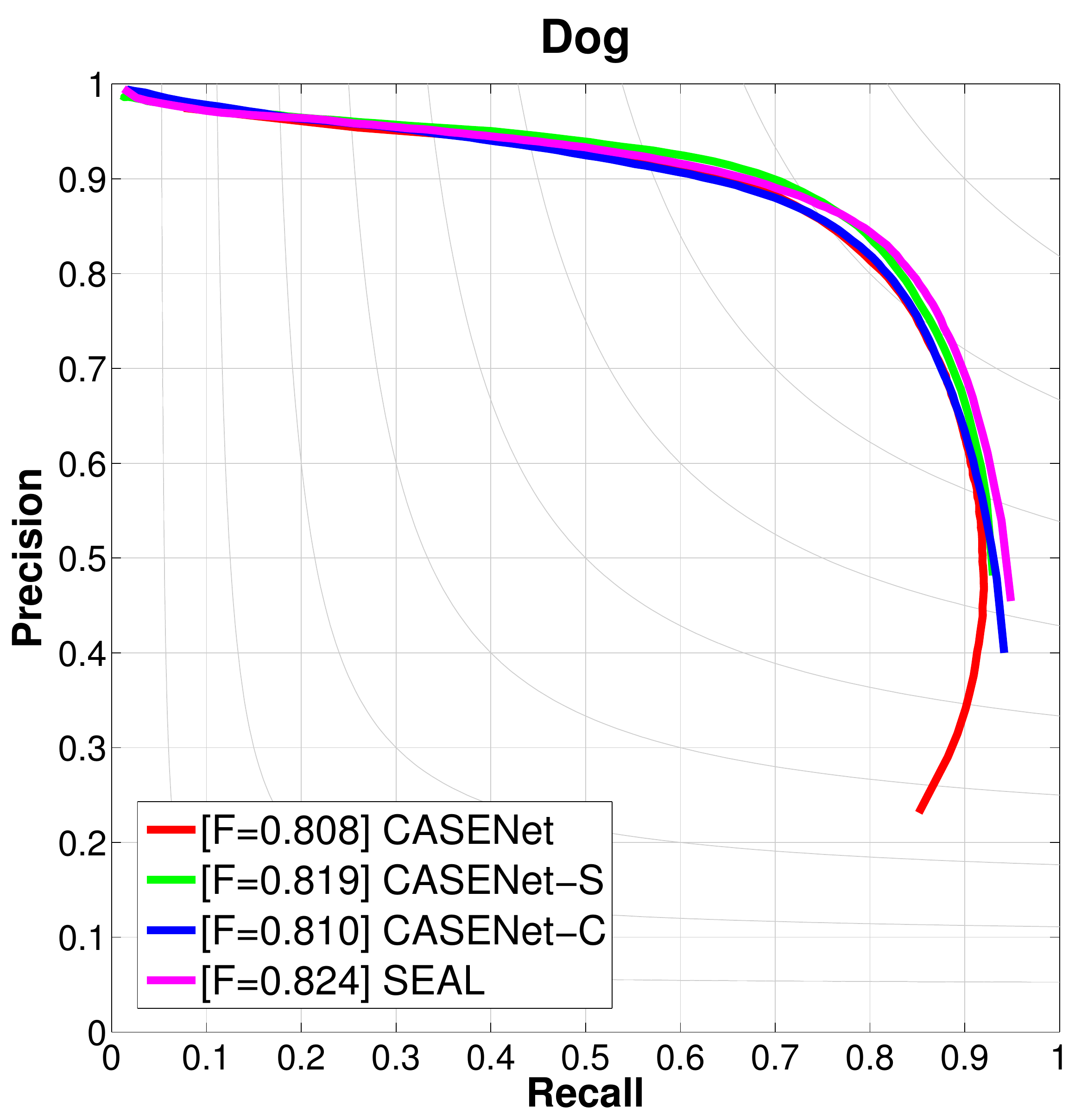}\\
	
	\includegraphics[width=.244\textwidth]{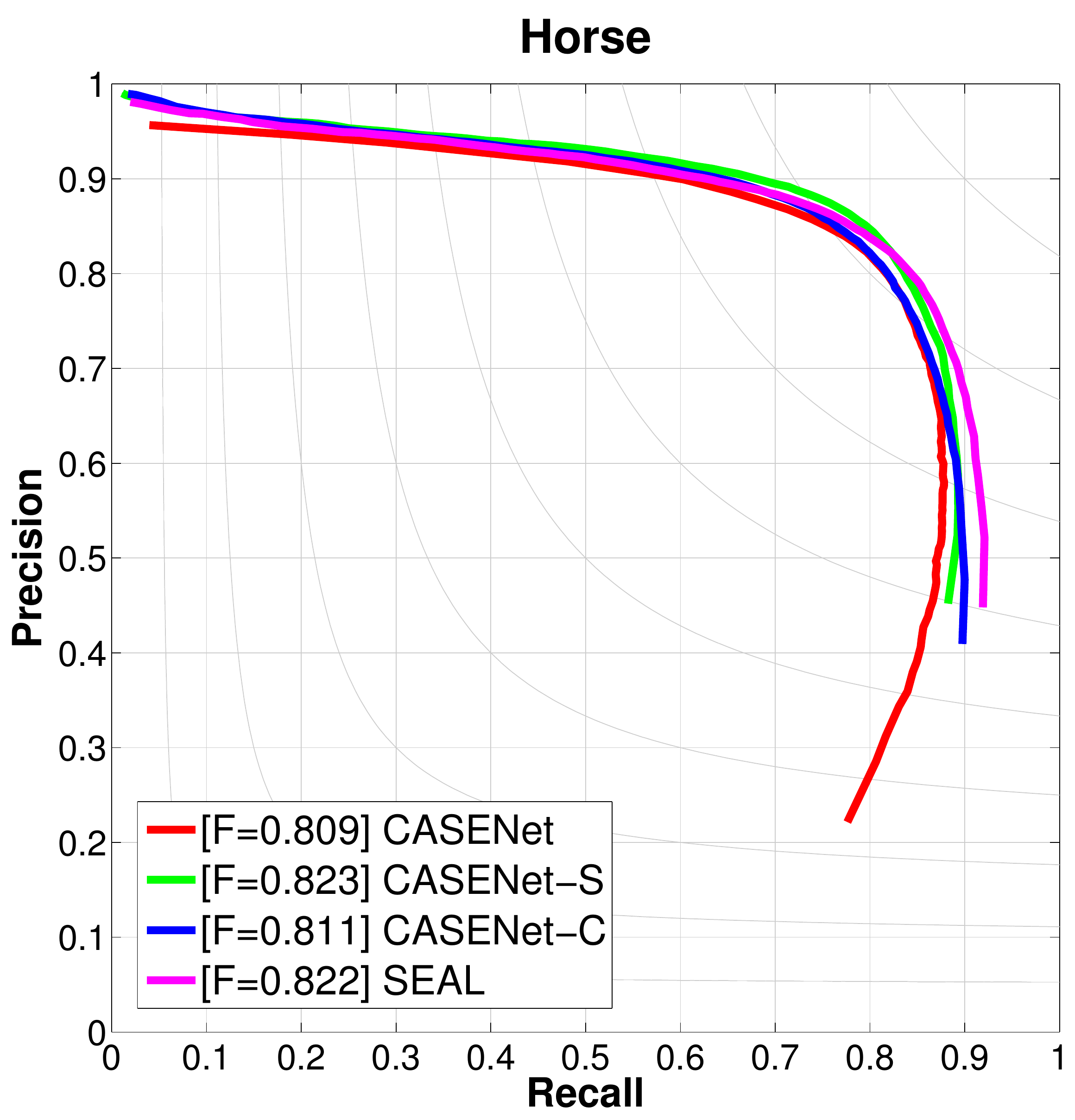}
	\includegraphics[width=.244\textwidth]{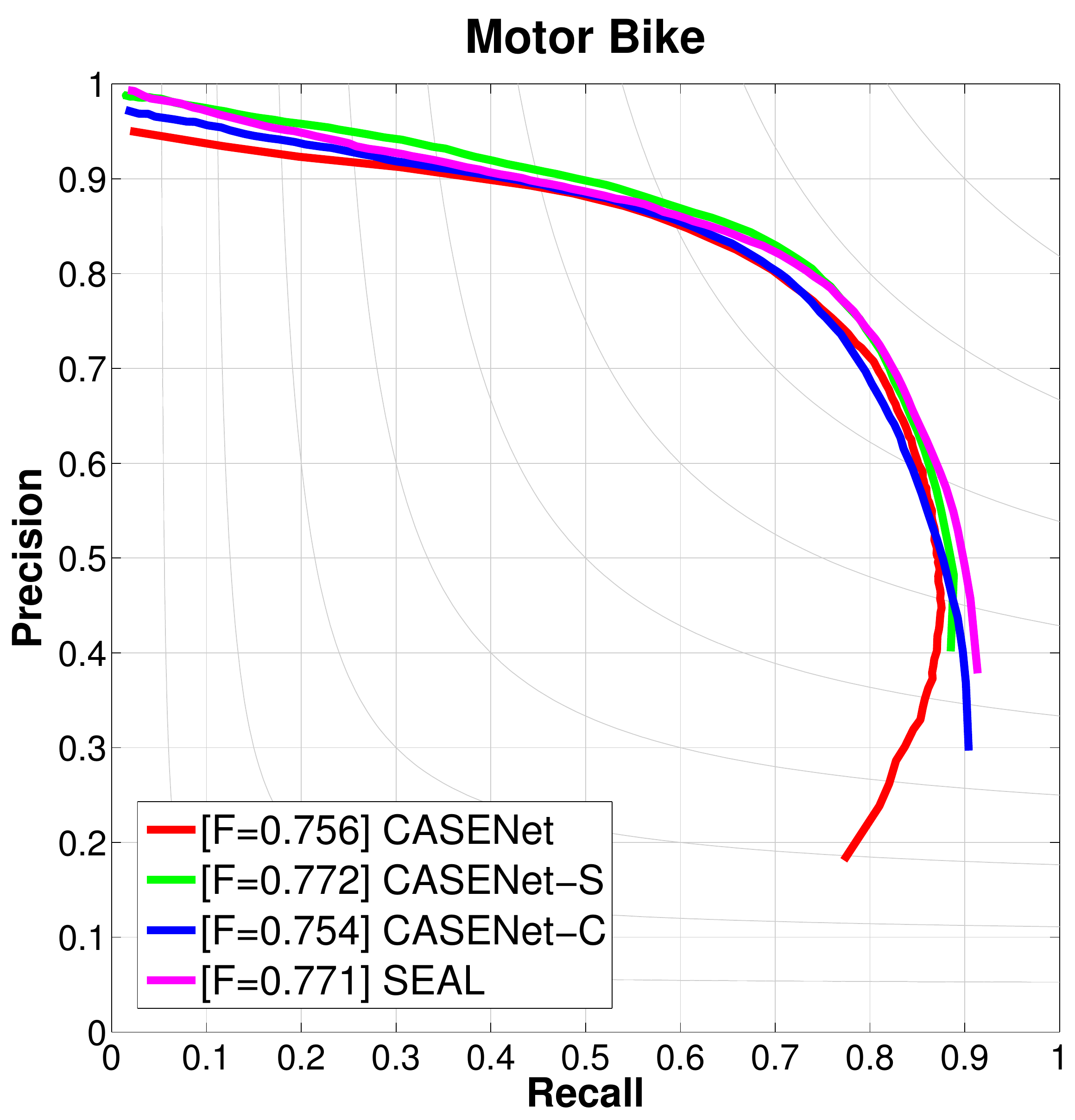}
	\includegraphics[width=.244\textwidth]{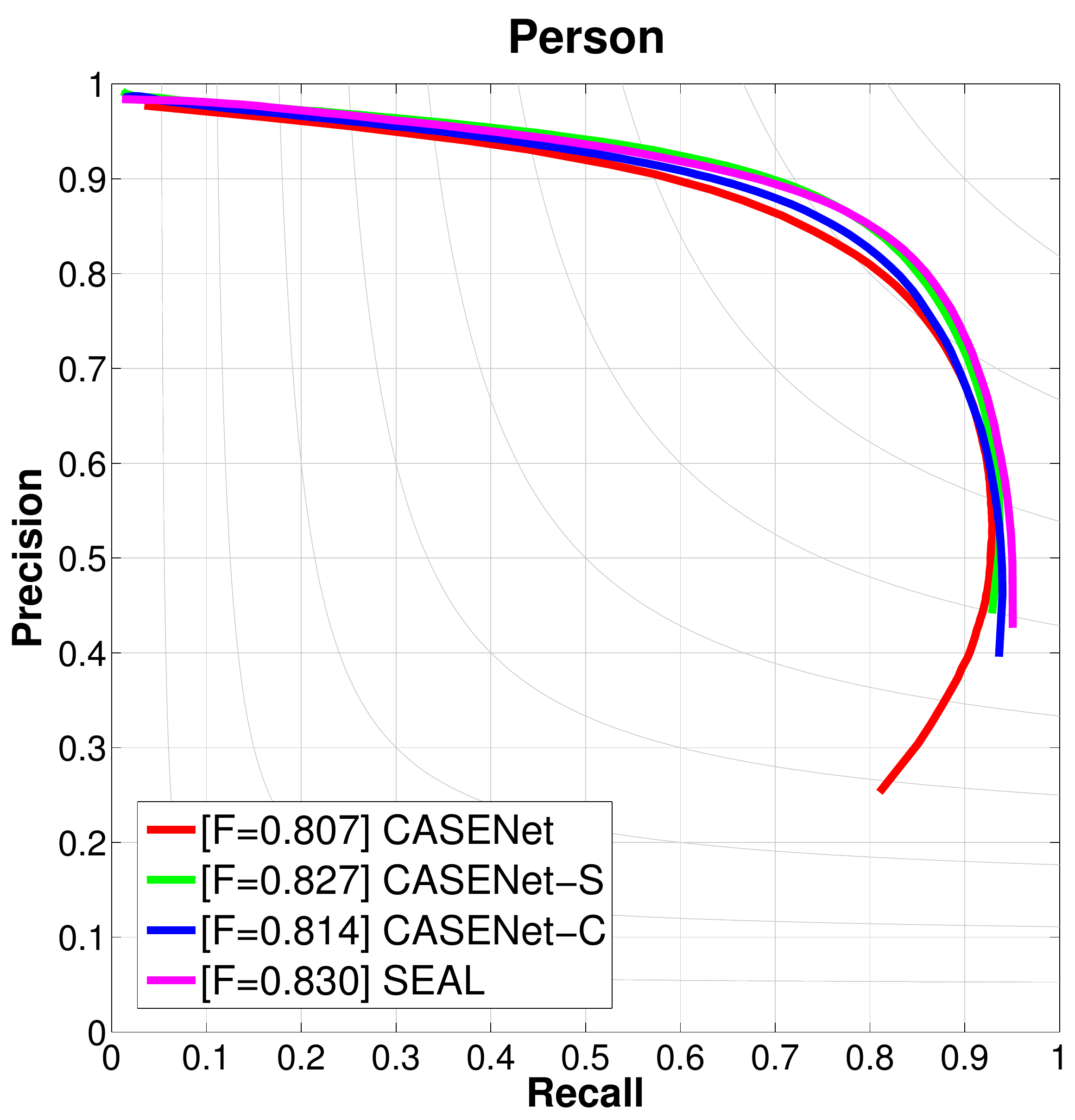}
	\includegraphics[width=.244\textwidth]{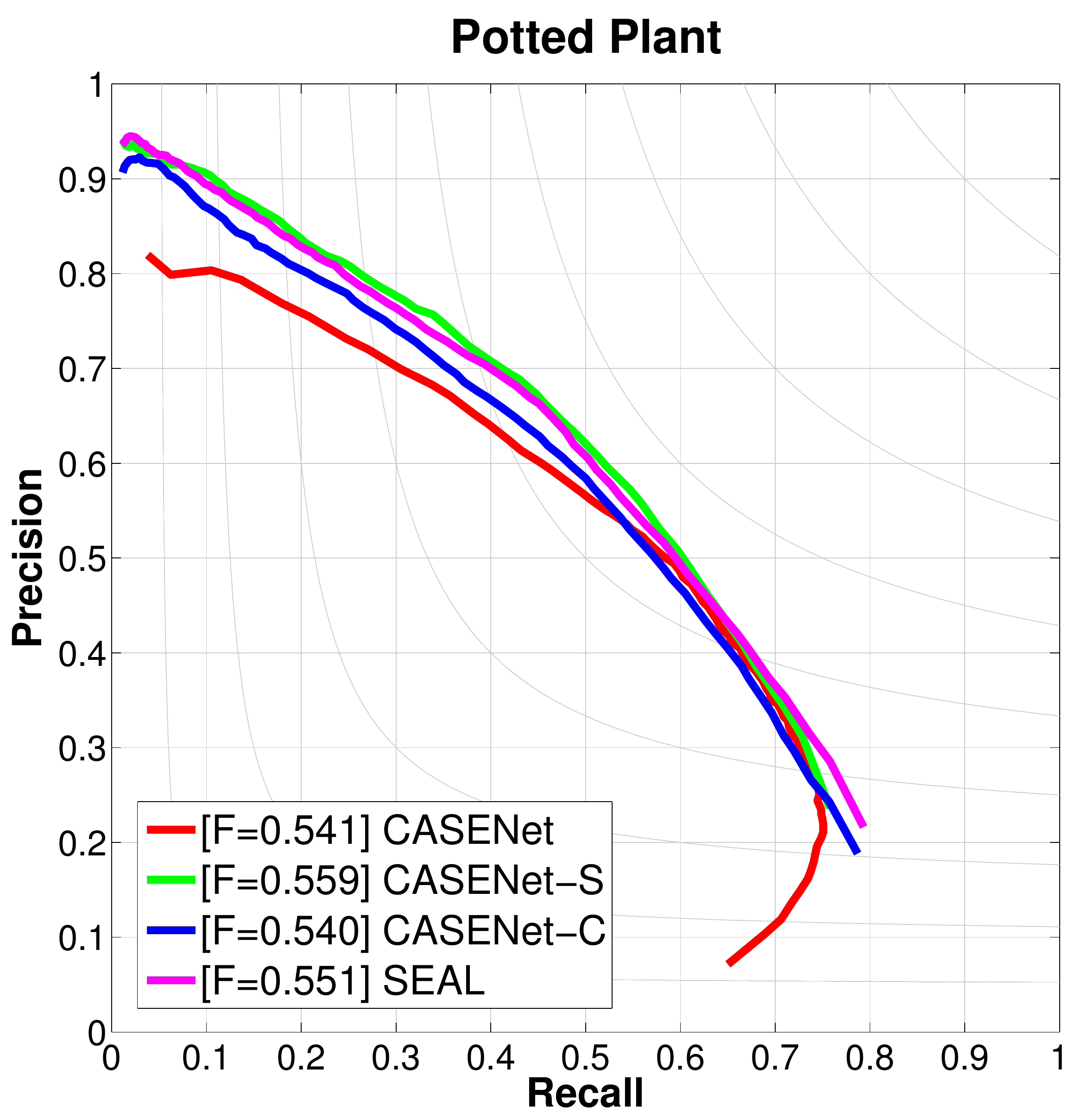}\\
	
	\includegraphics[width=.244\textwidth]{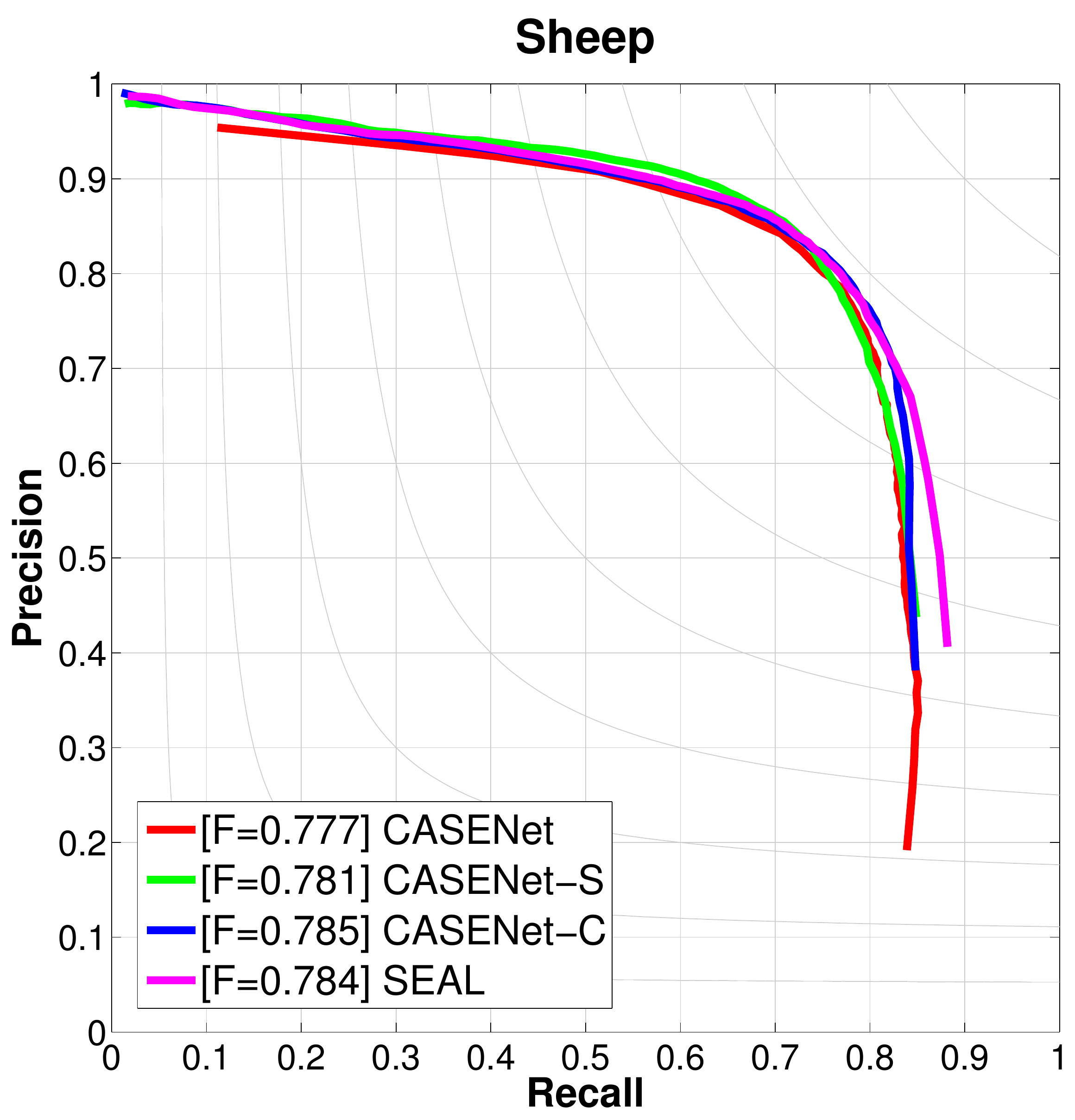}
	\includegraphics[width=.244\textwidth]{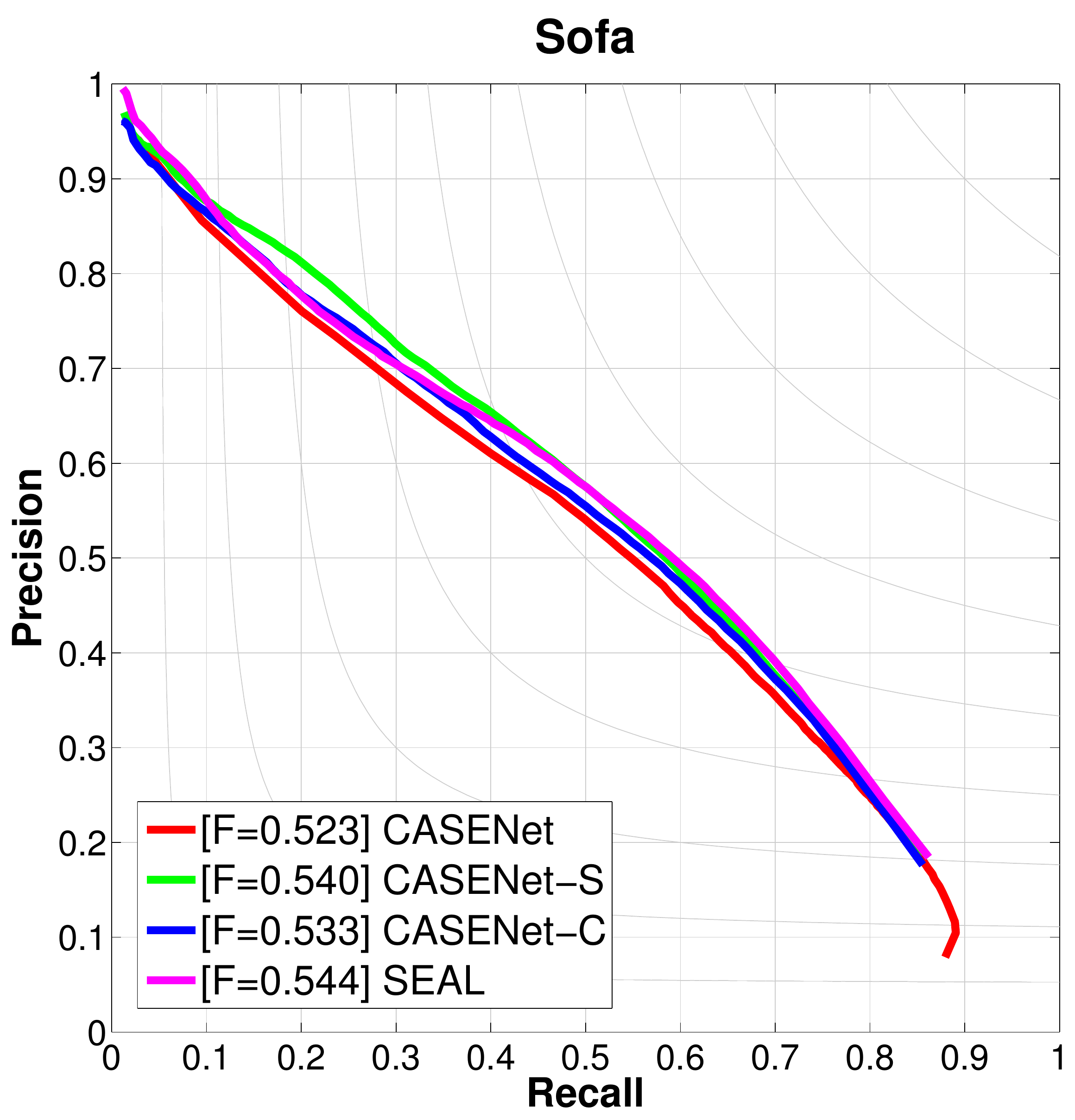}
	\includegraphics[width=.244\textwidth]{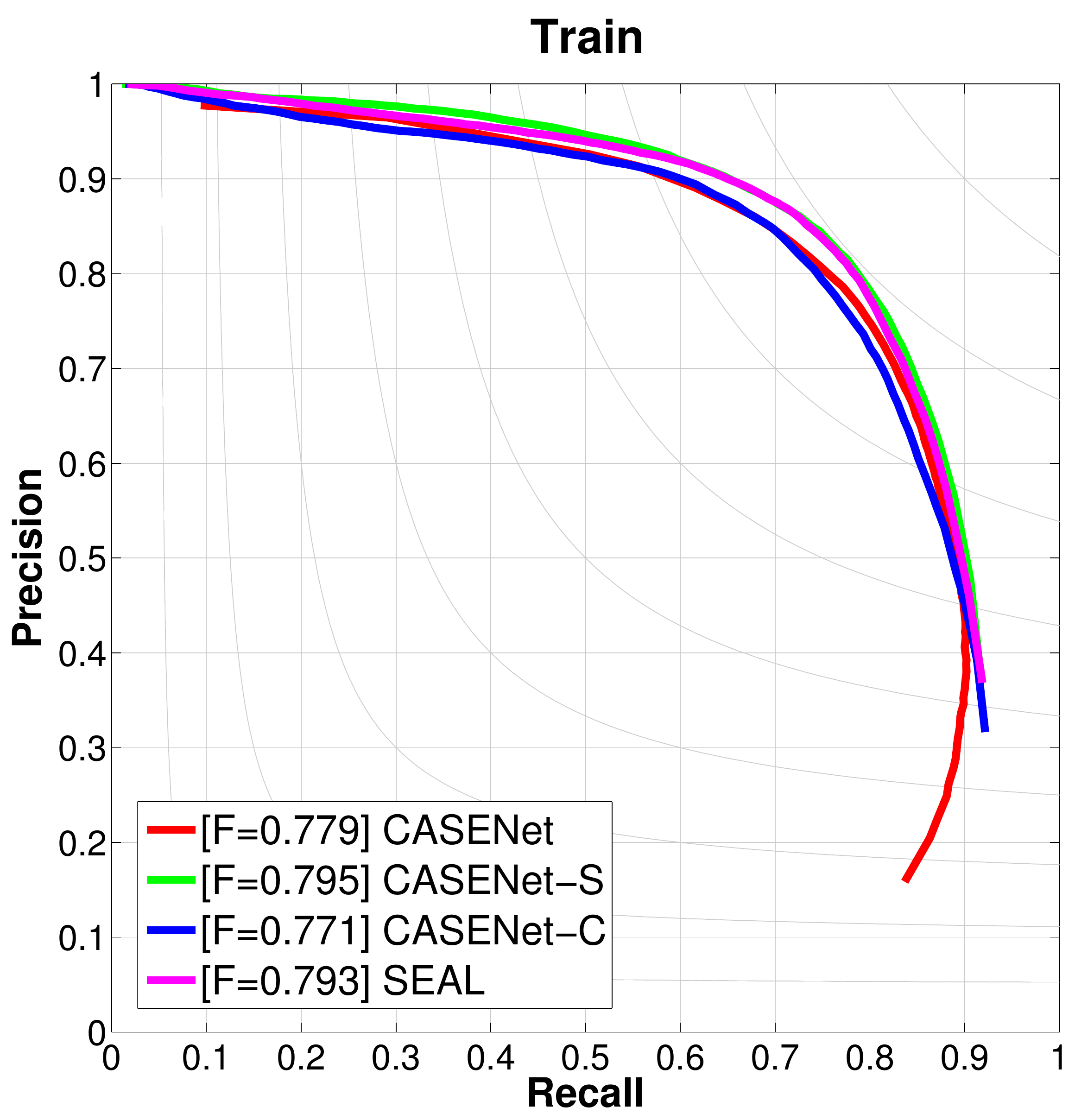}
	\includegraphics[width=.244\textwidth]{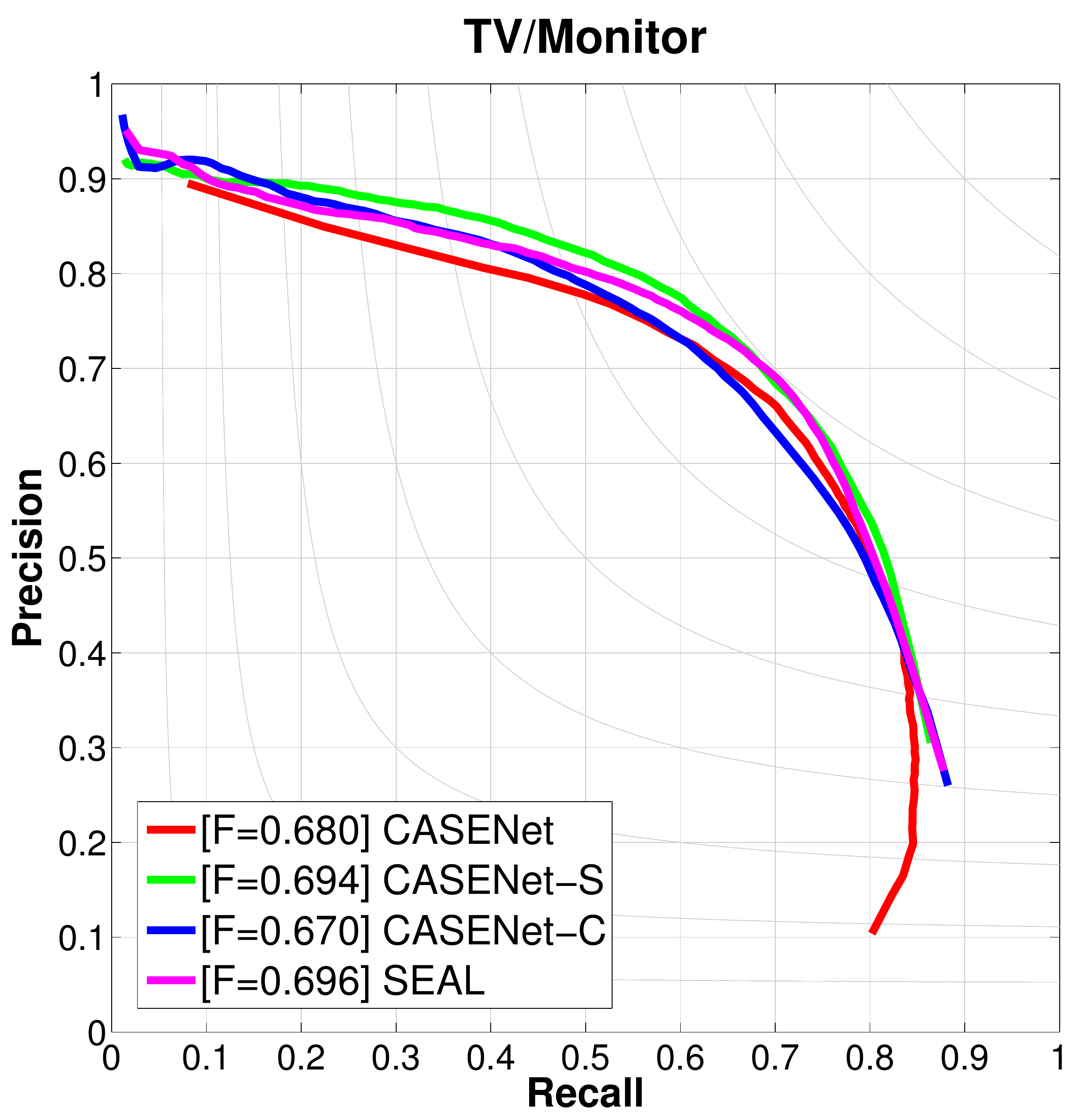}\\
	\caption{Class-wise precision-recall curves of SEAL and comparing baselines on the original SBD test set under the ``Thin'' setting.}\label{pr_sbd_orig_thin}
\end{figure}

\begin{figure}[tbh]
	\centering
	\includegraphics[width=.244\textwidth]{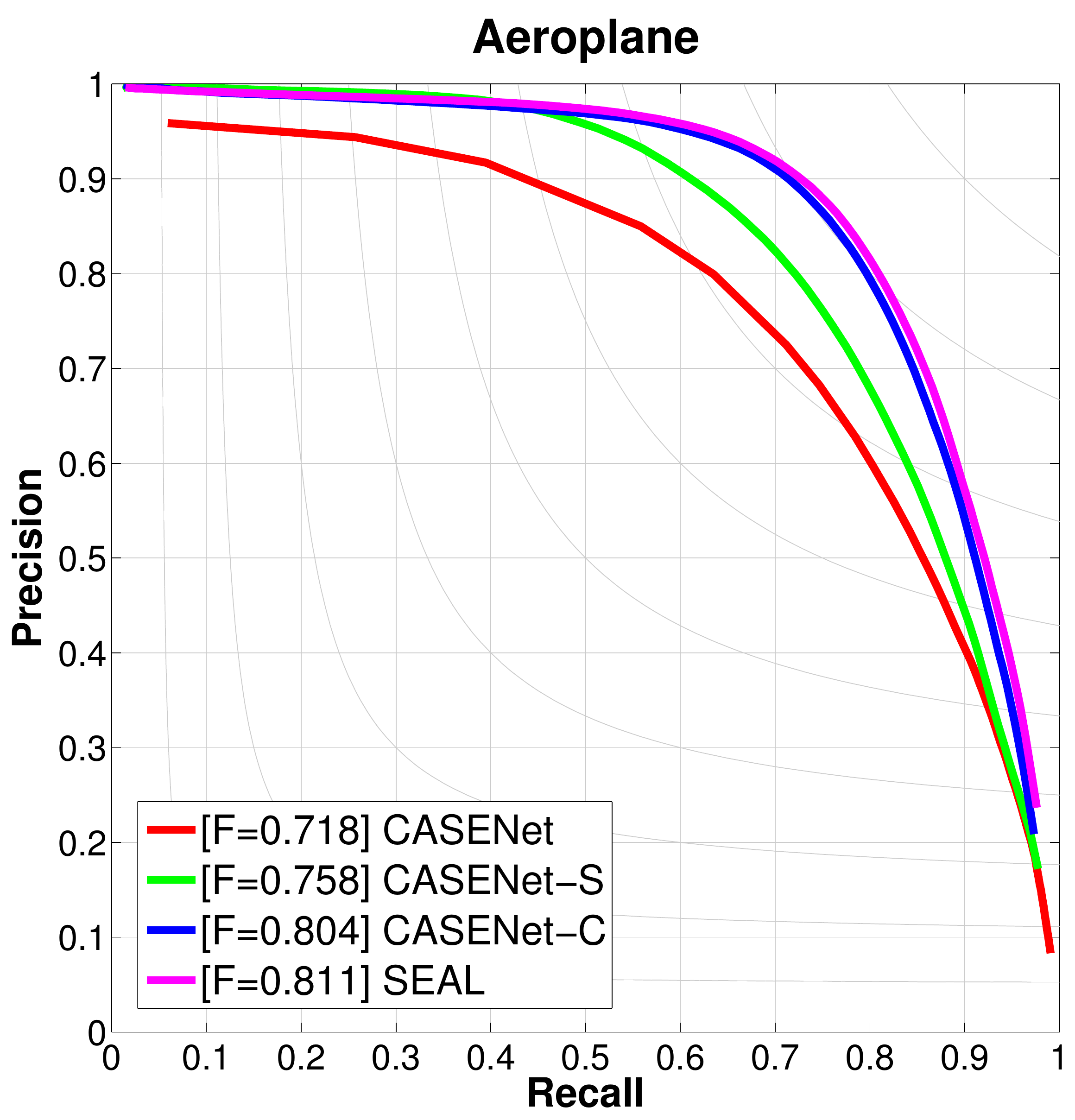}
	\includegraphics[width=.244\textwidth]{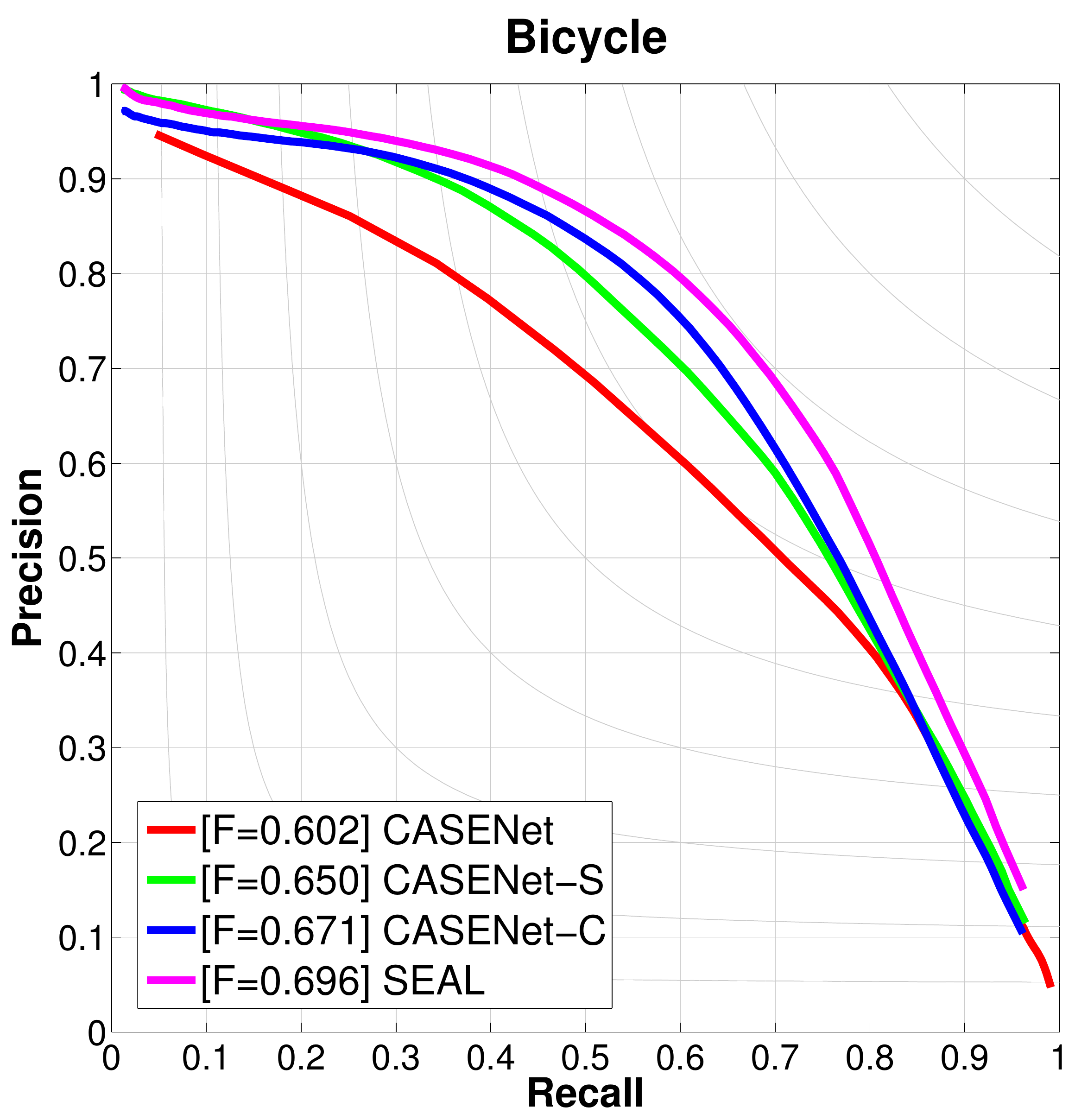}
	\includegraphics[width=.244\textwidth]{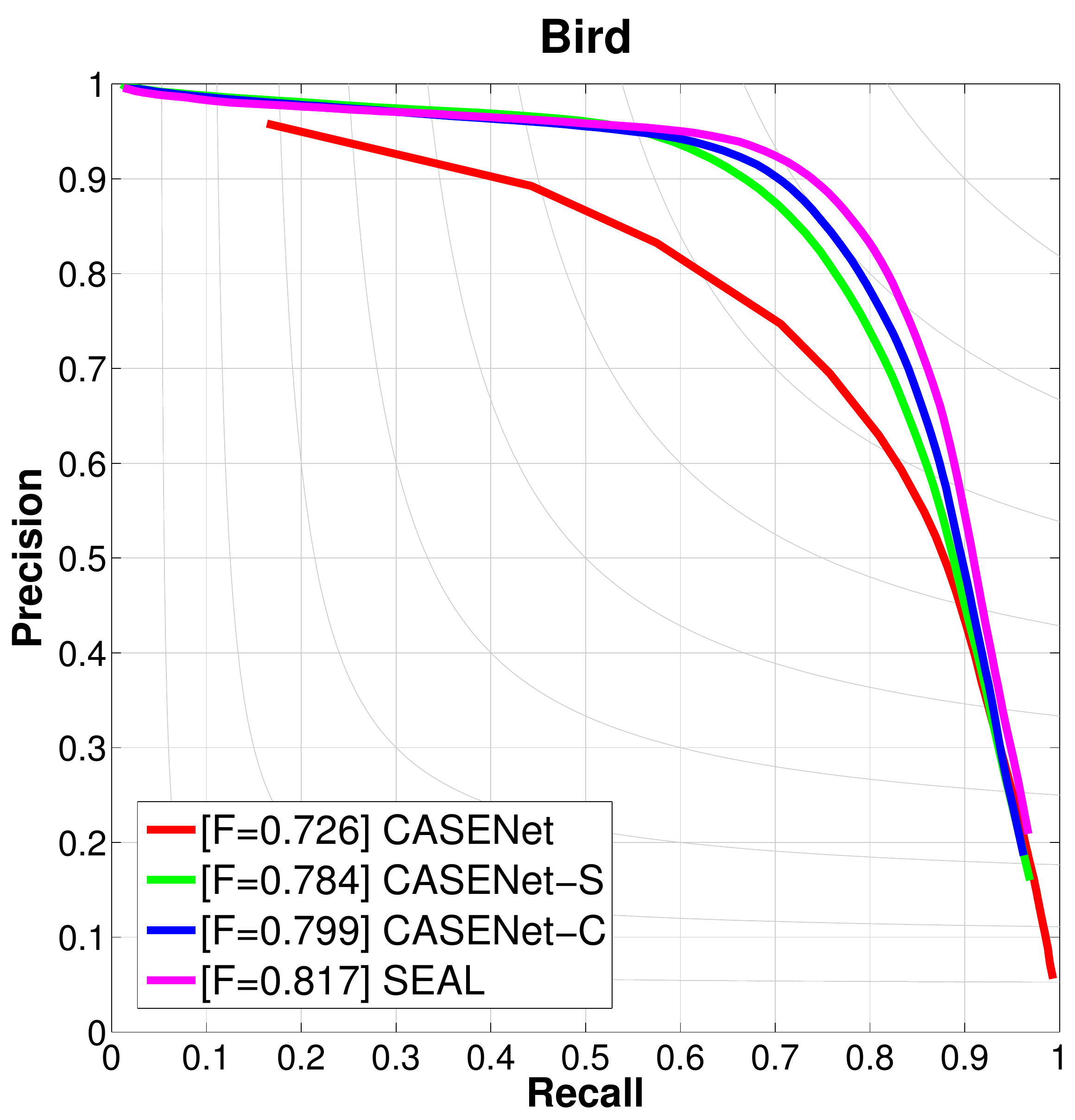}
	\includegraphics[width=.244\textwidth]{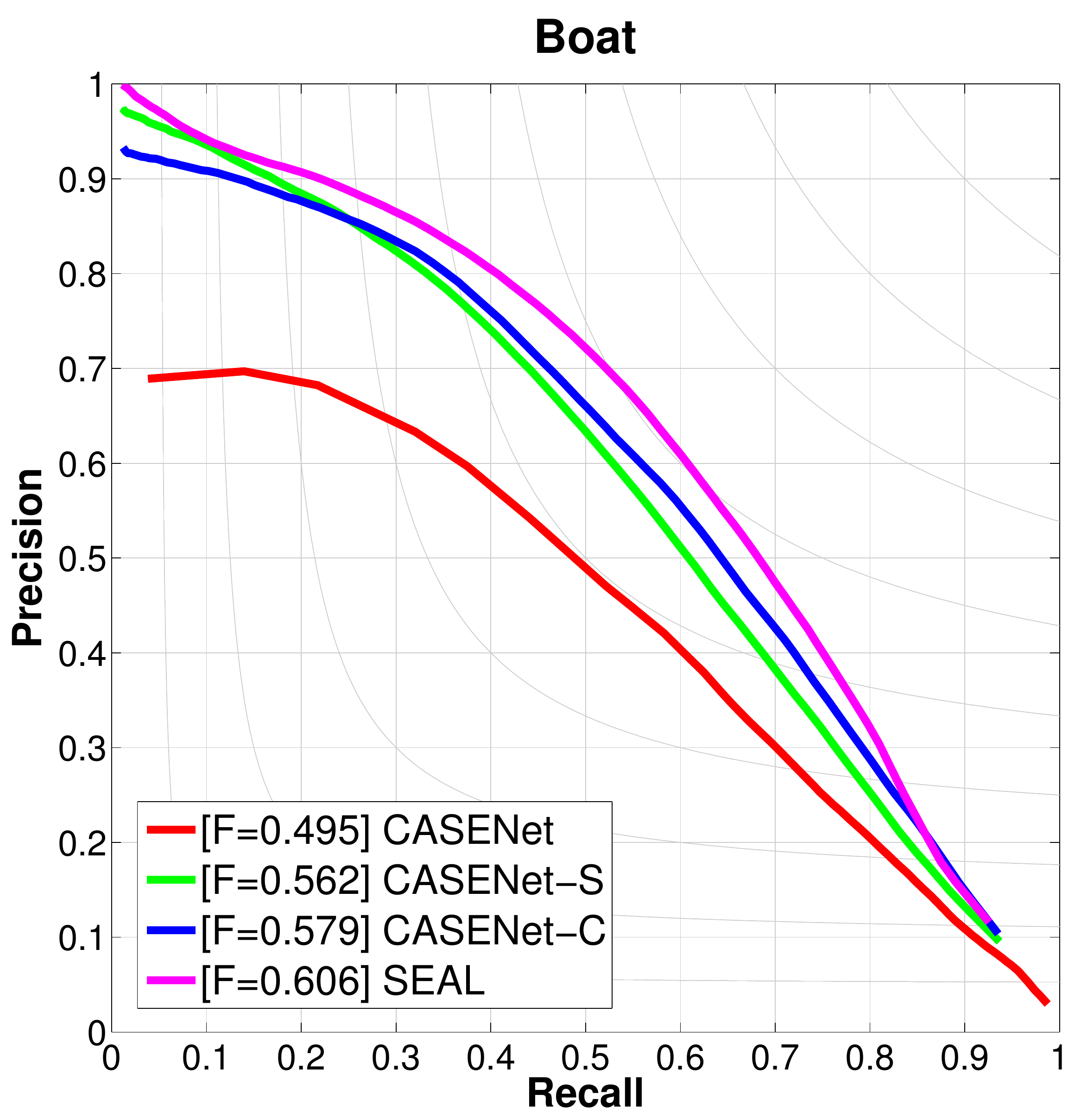}\\
	
	\includegraphics[width=.244\textwidth]{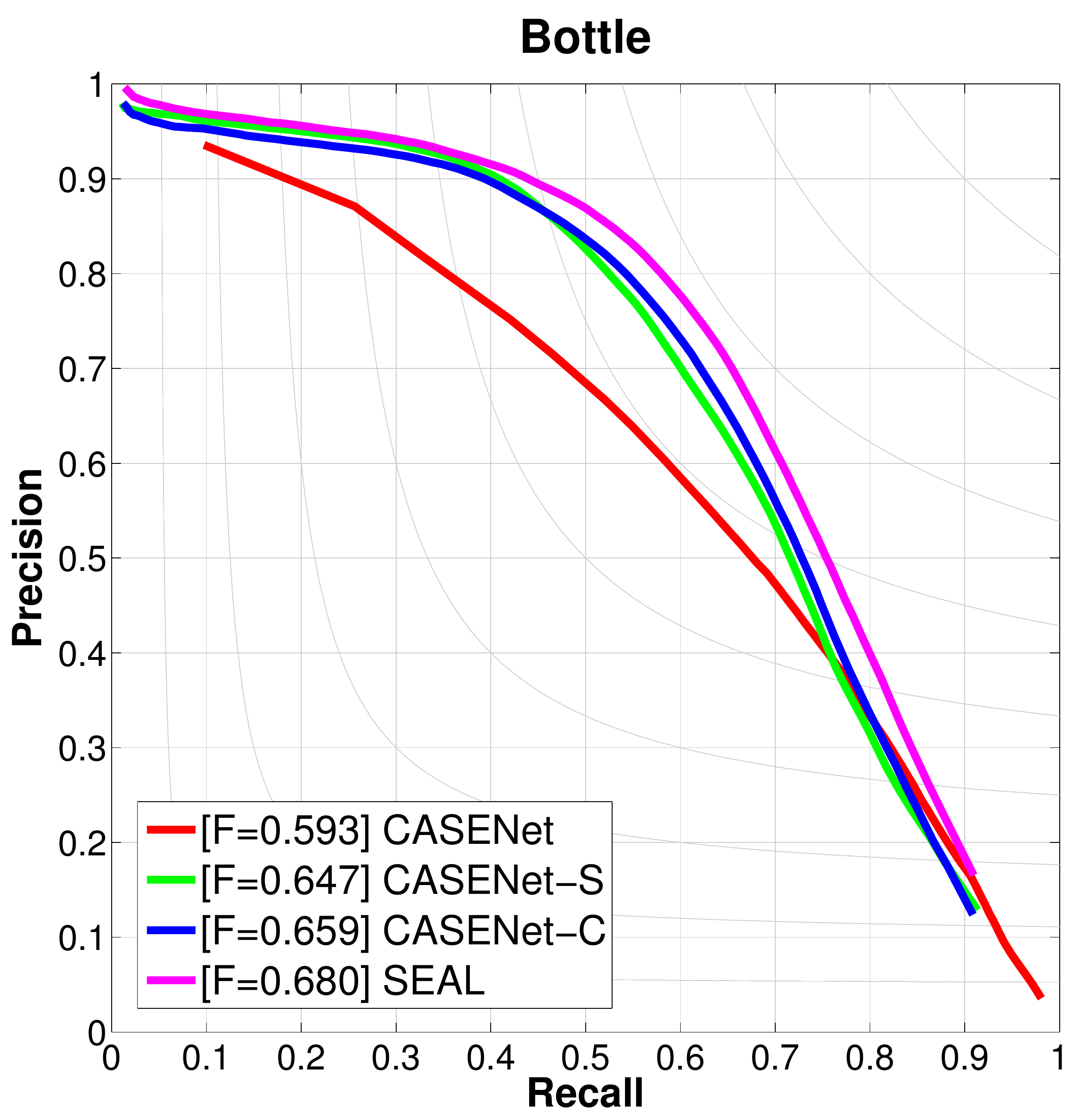}
	\includegraphics[width=.244\textwidth]{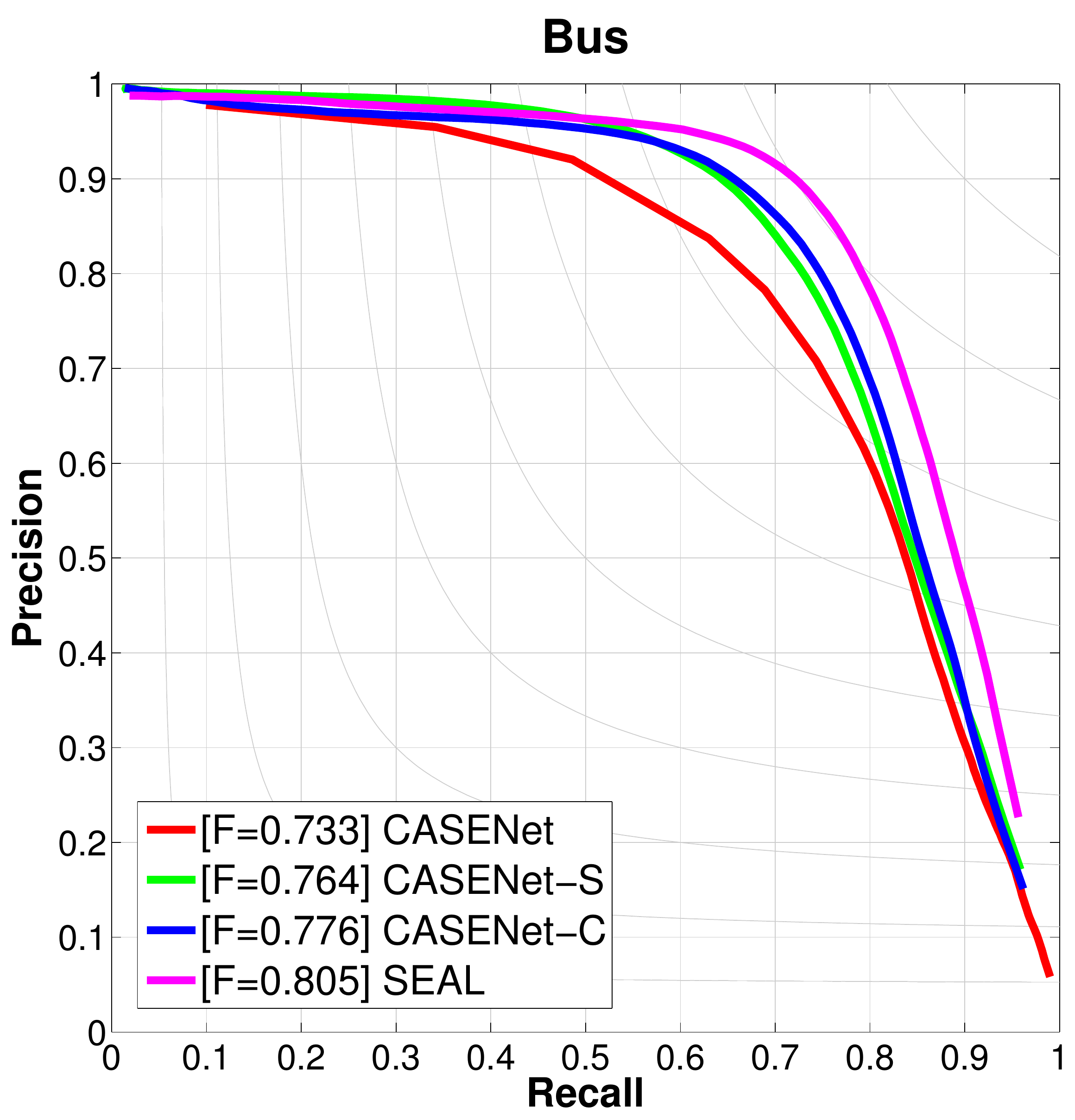}
	\includegraphics[width=.244\textwidth]{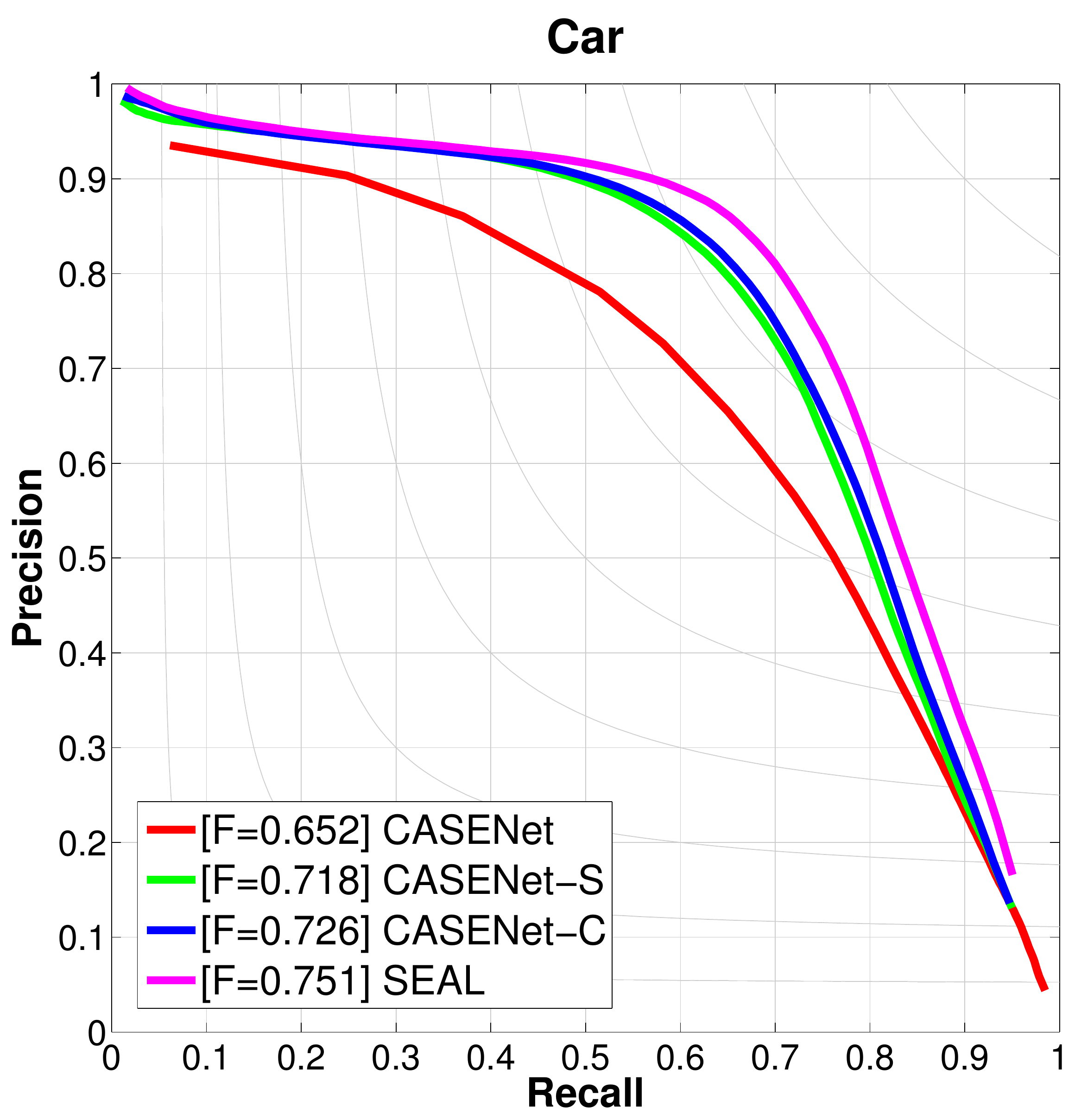}
	\includegraphics[width=.244\textwidth]{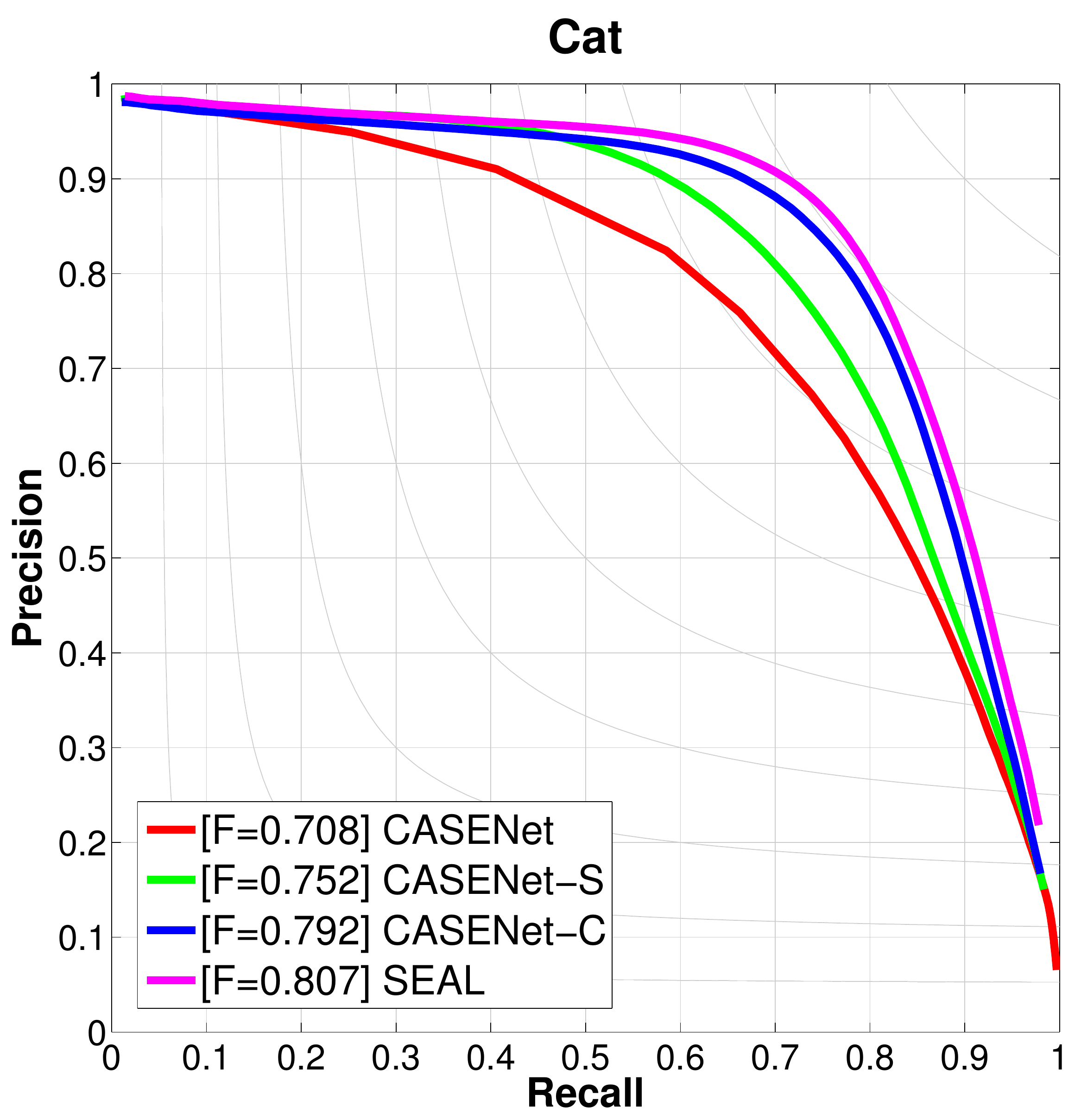}\\
	
	\includegraphics[width=.244\textwidth]{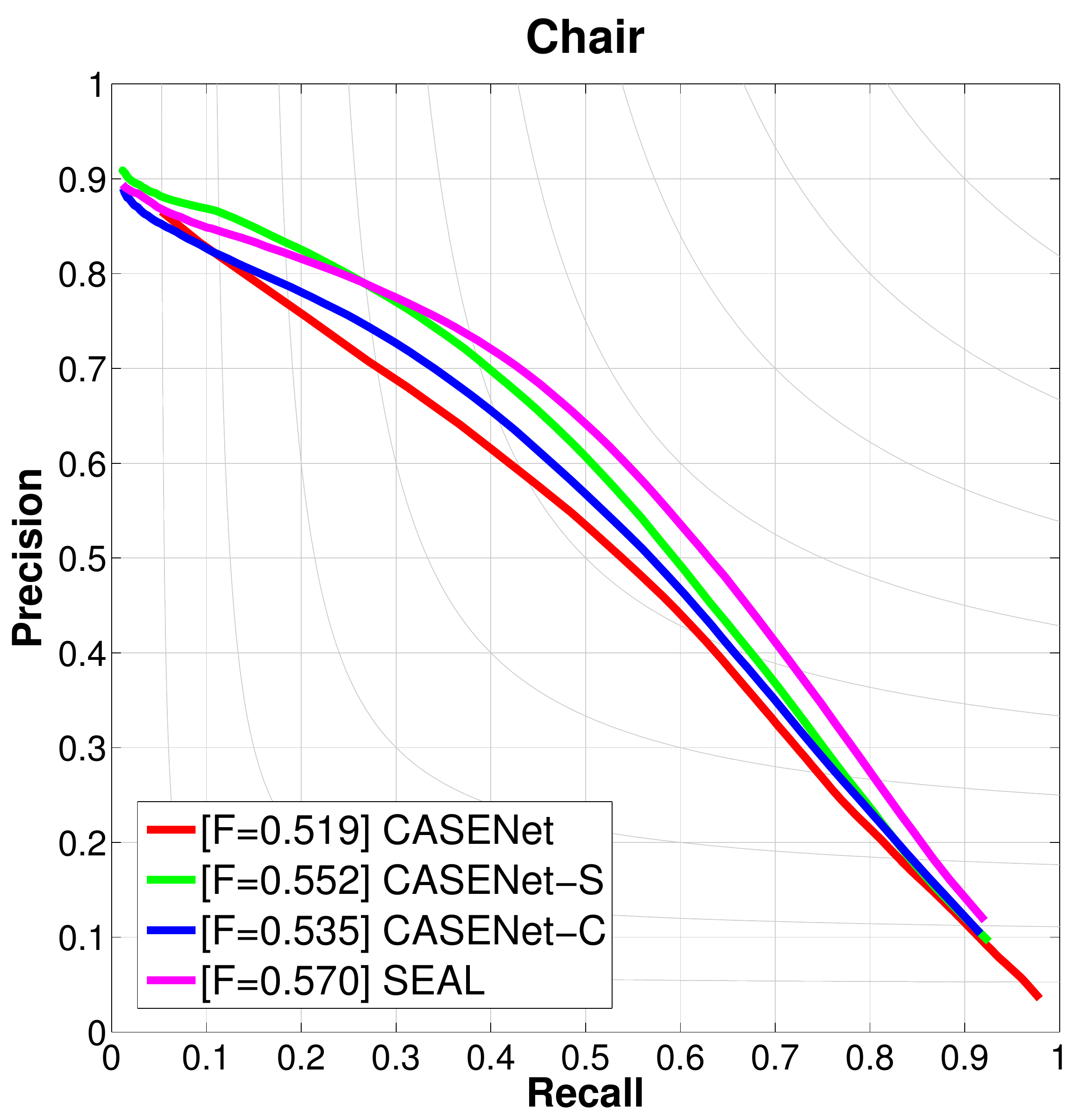}
	\includegraphics[width=.244\textwidth]{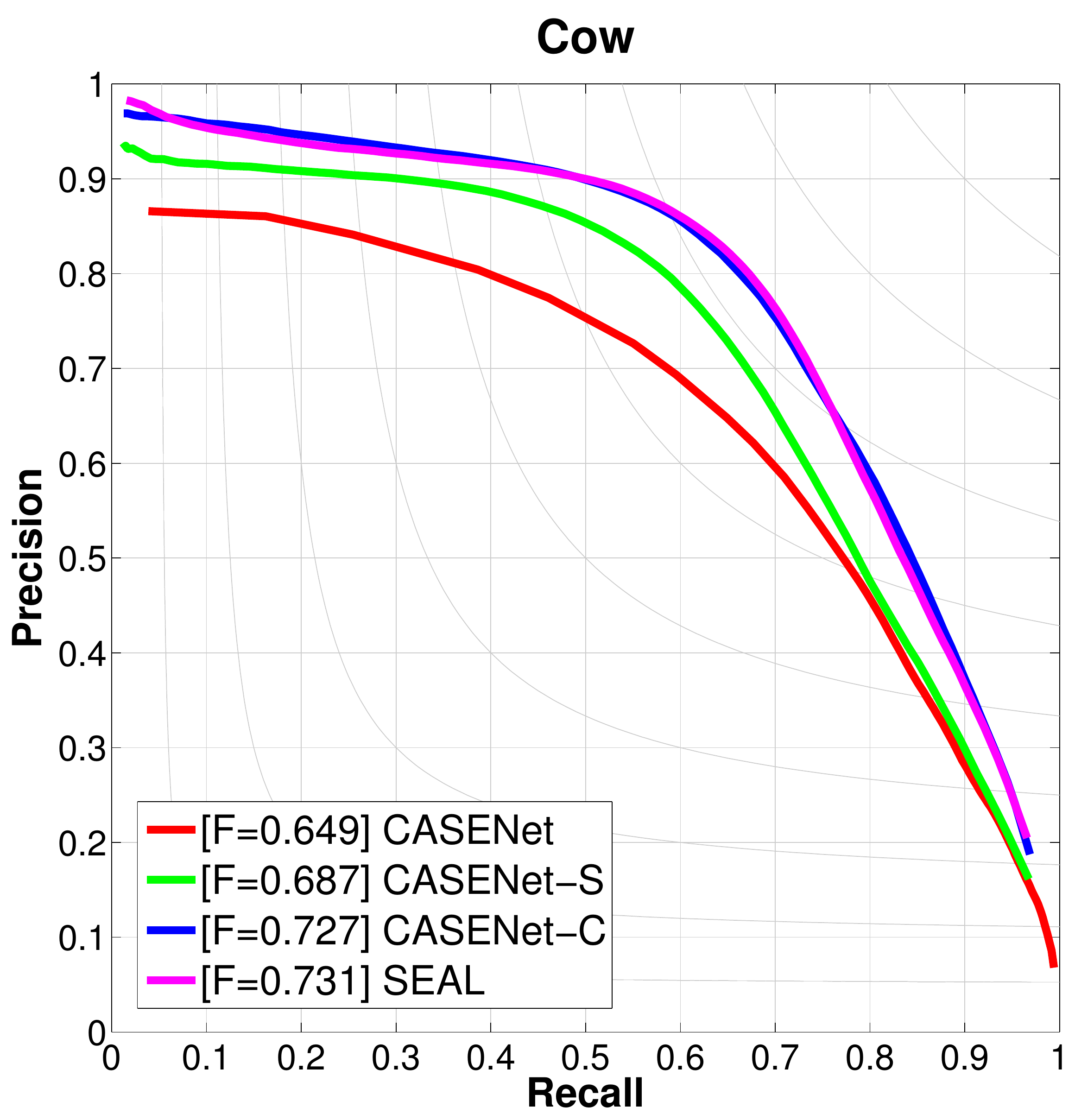}
	\includegraphics[width=.244\textwidth]{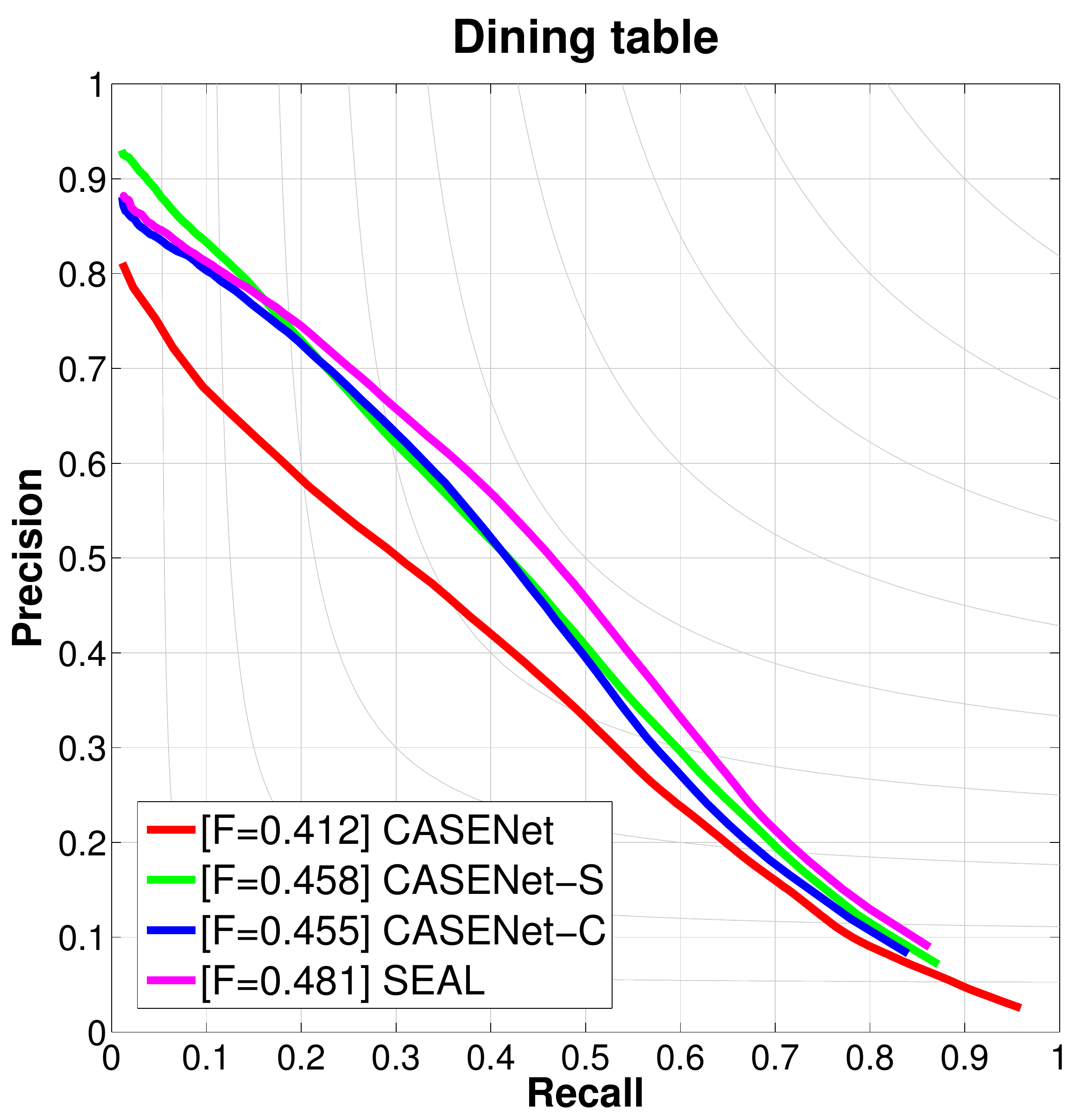}
	\includegraphics[width=.244\textwidth]{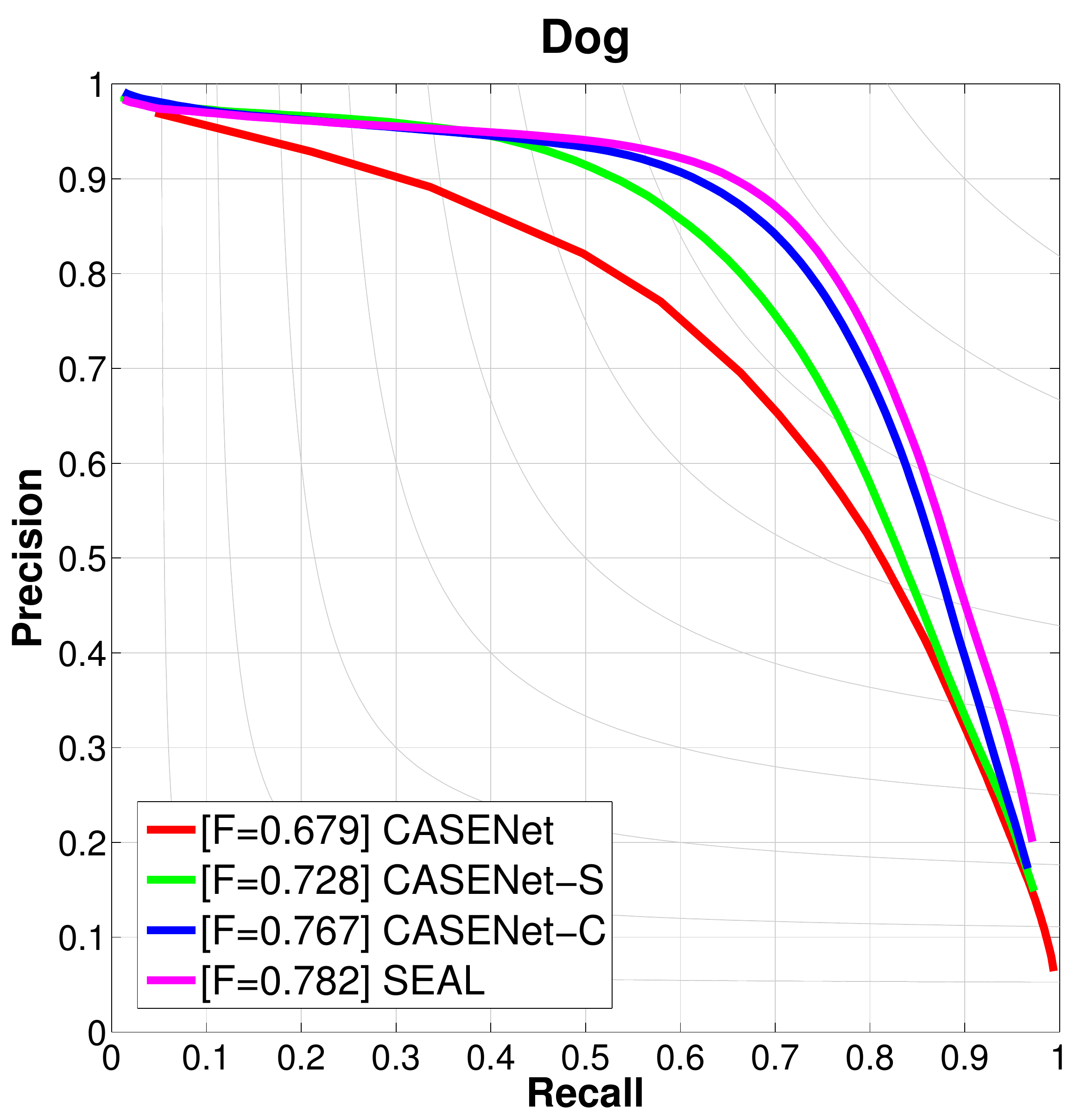}\\
	
	\includegraphics[width=.244\textwidth]{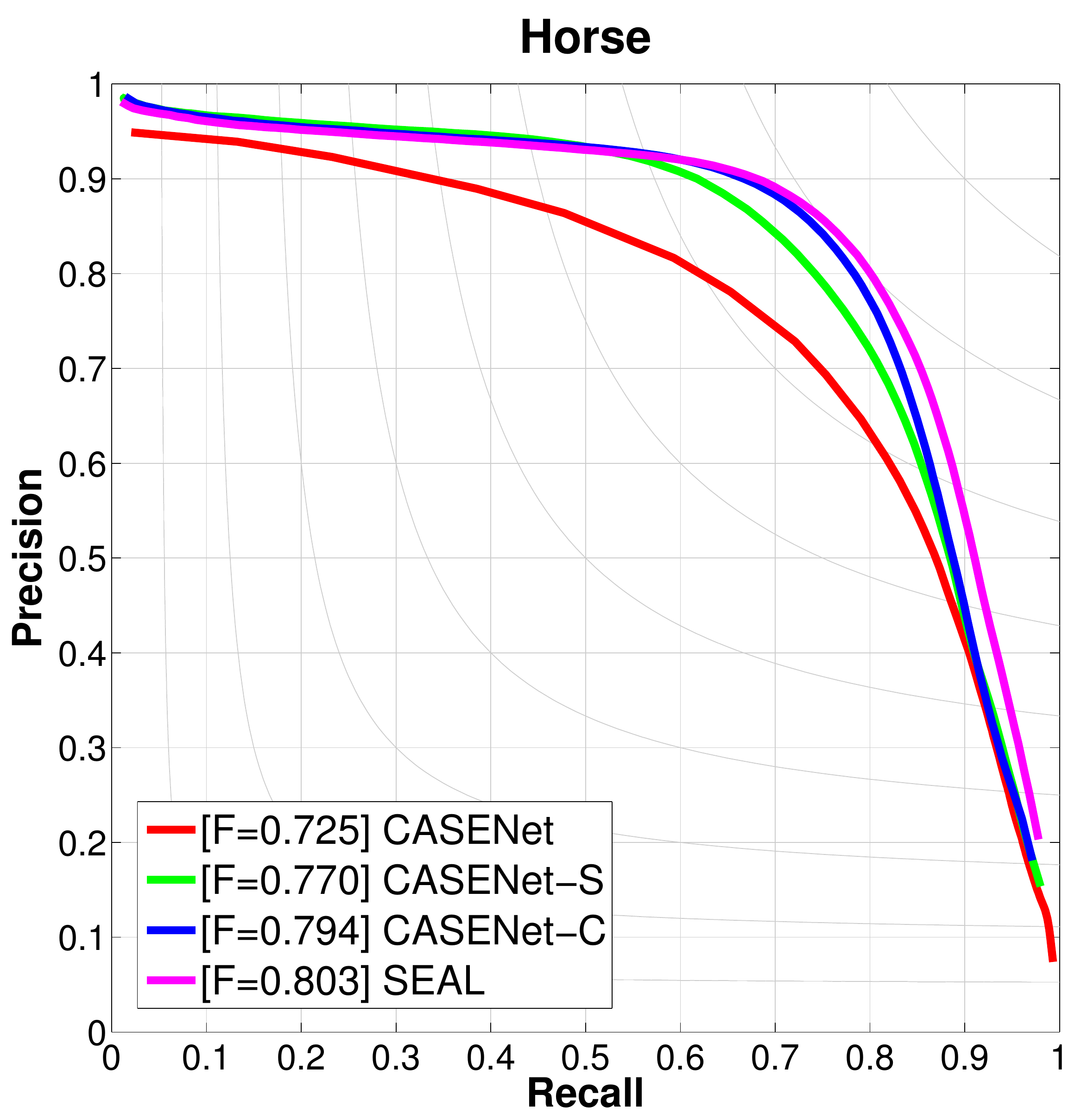}
	\includegraphics[width=.244\textwidth]{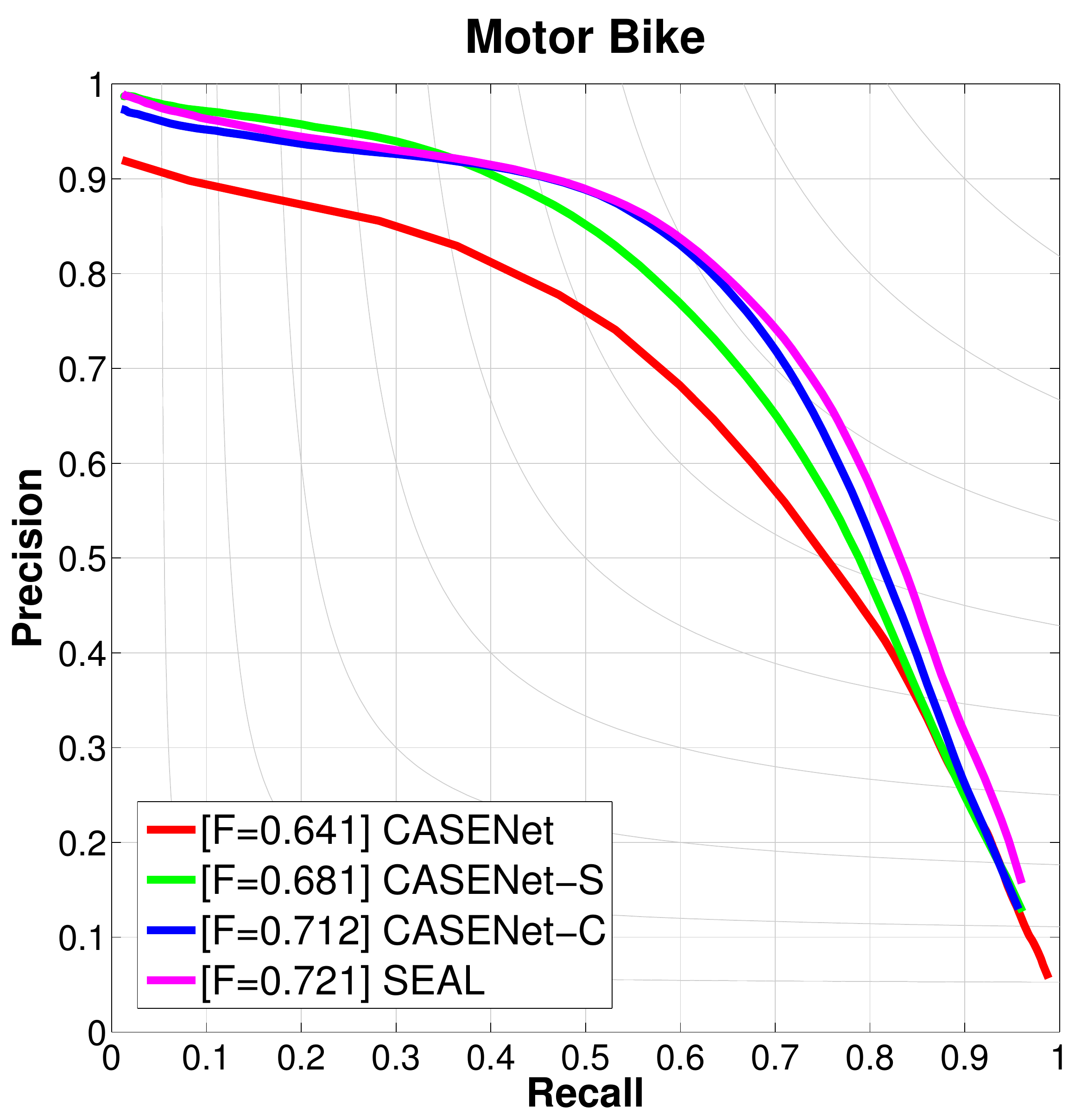}
	\includegraphics[width=.244\textwidth]{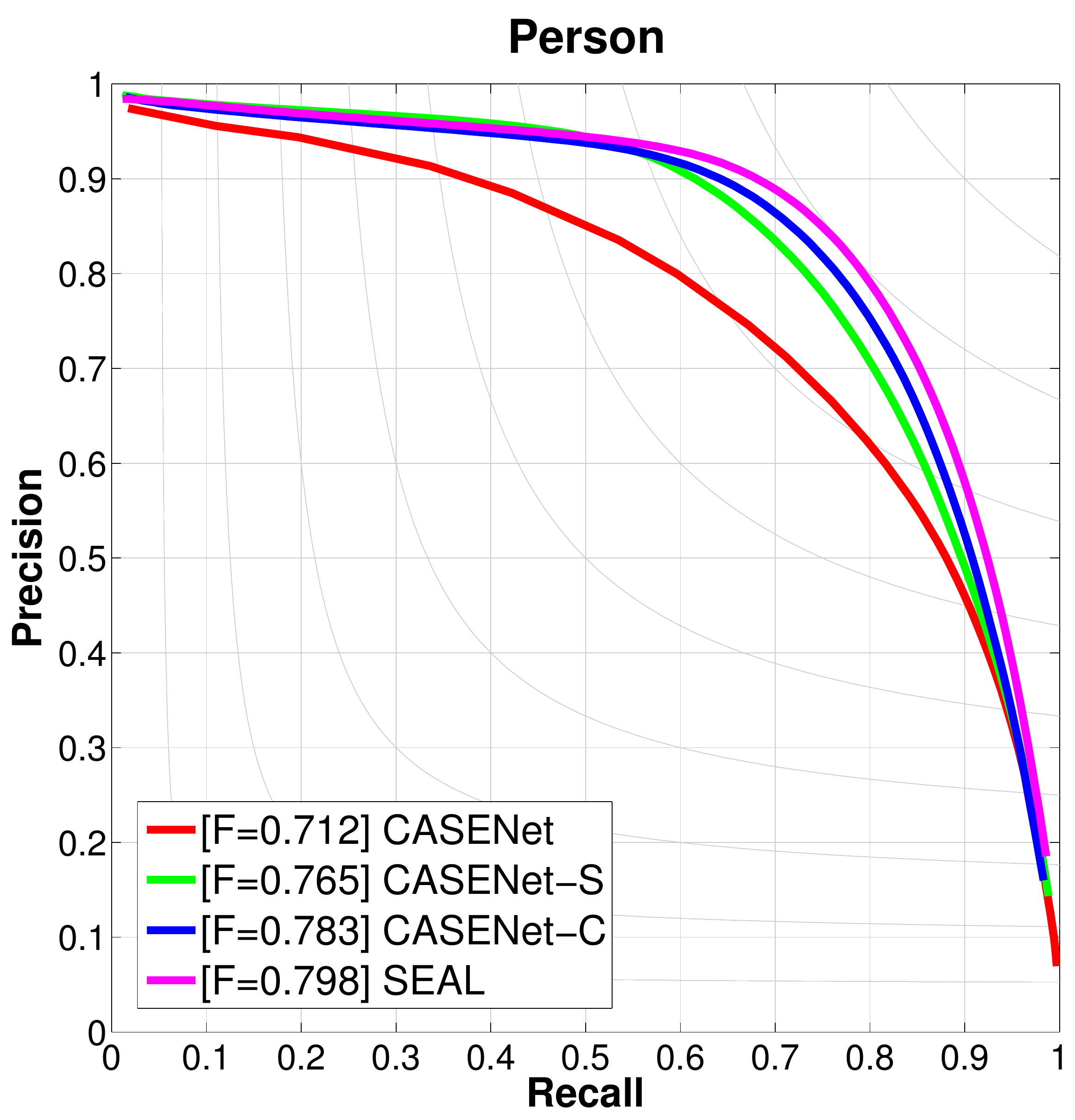}
	\includegraphics[width=.244\textwidth]{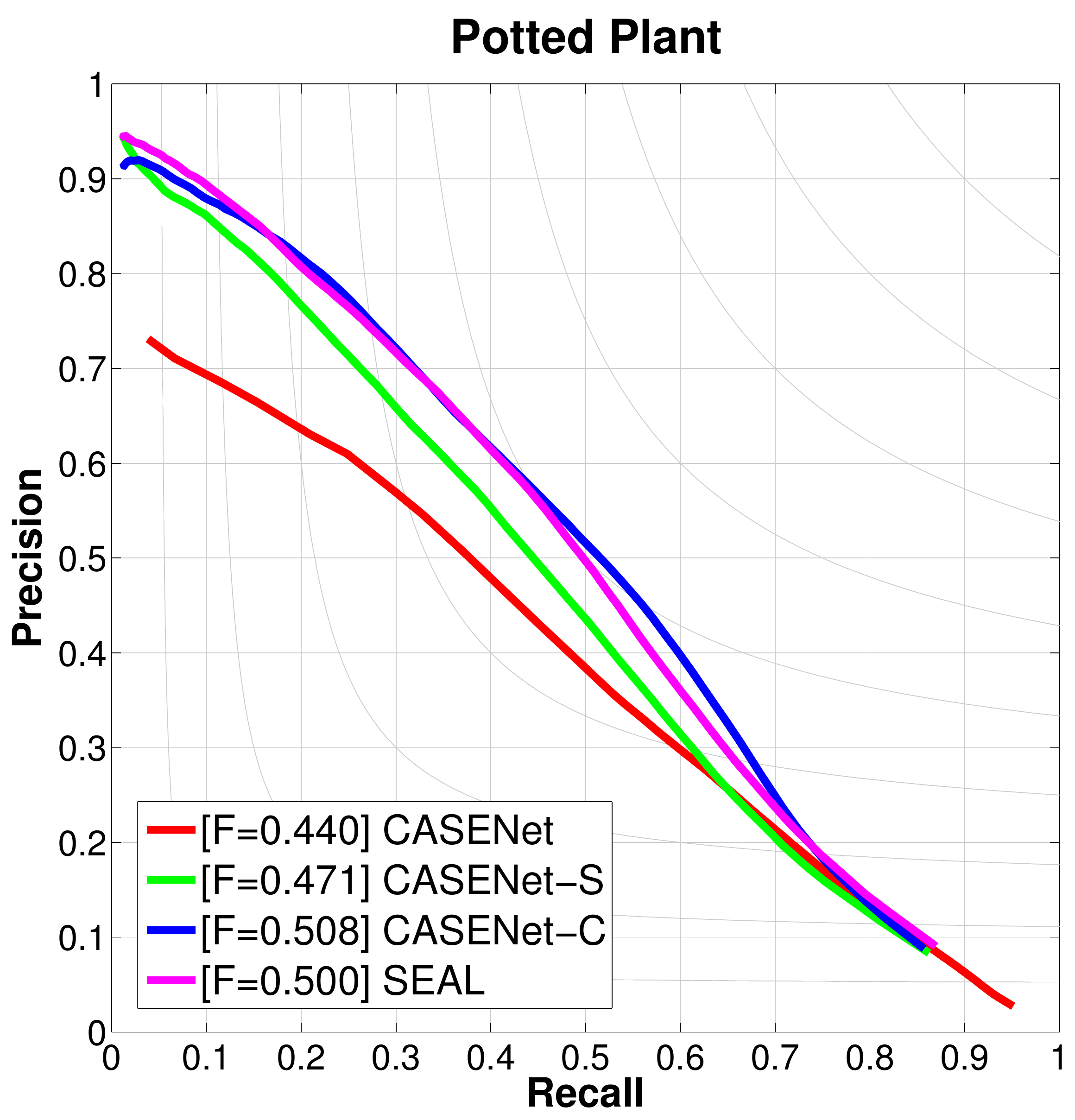}\\
	
	\includegraphics[width=.244\textwidth]{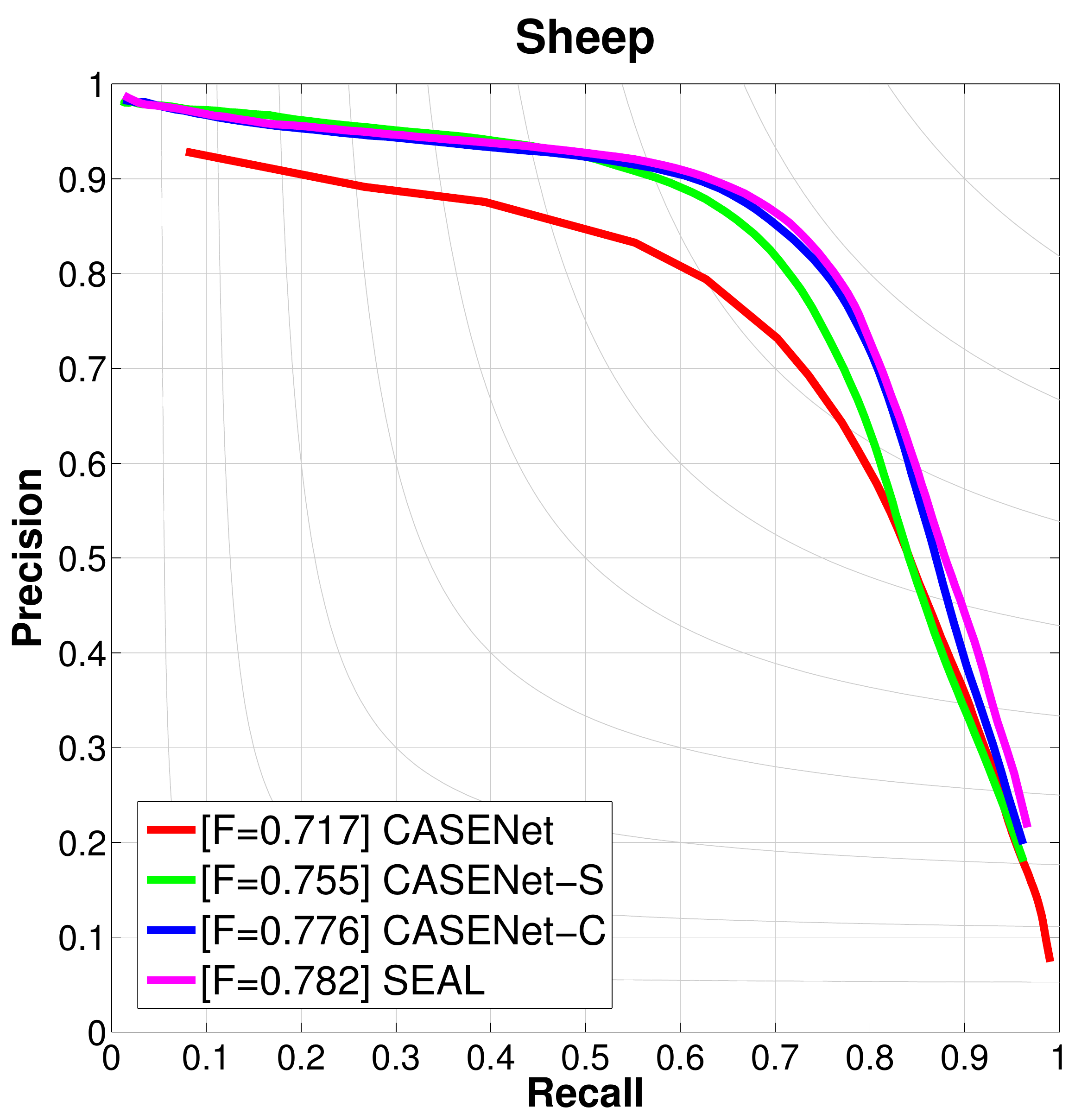}
	\includegraphics[width=.244\textwidth]{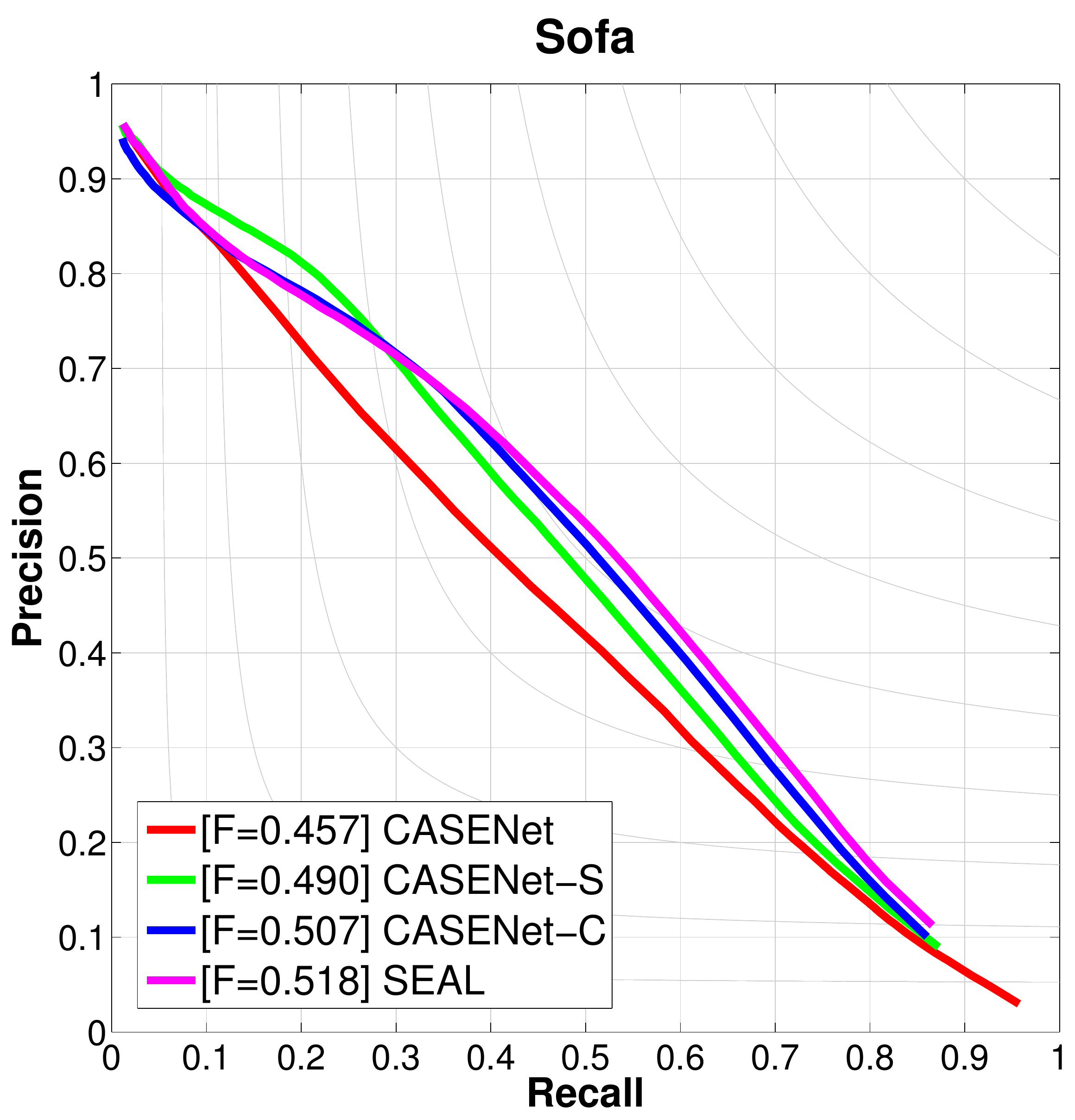}
	\includegraphics[width=.244\textwidth]{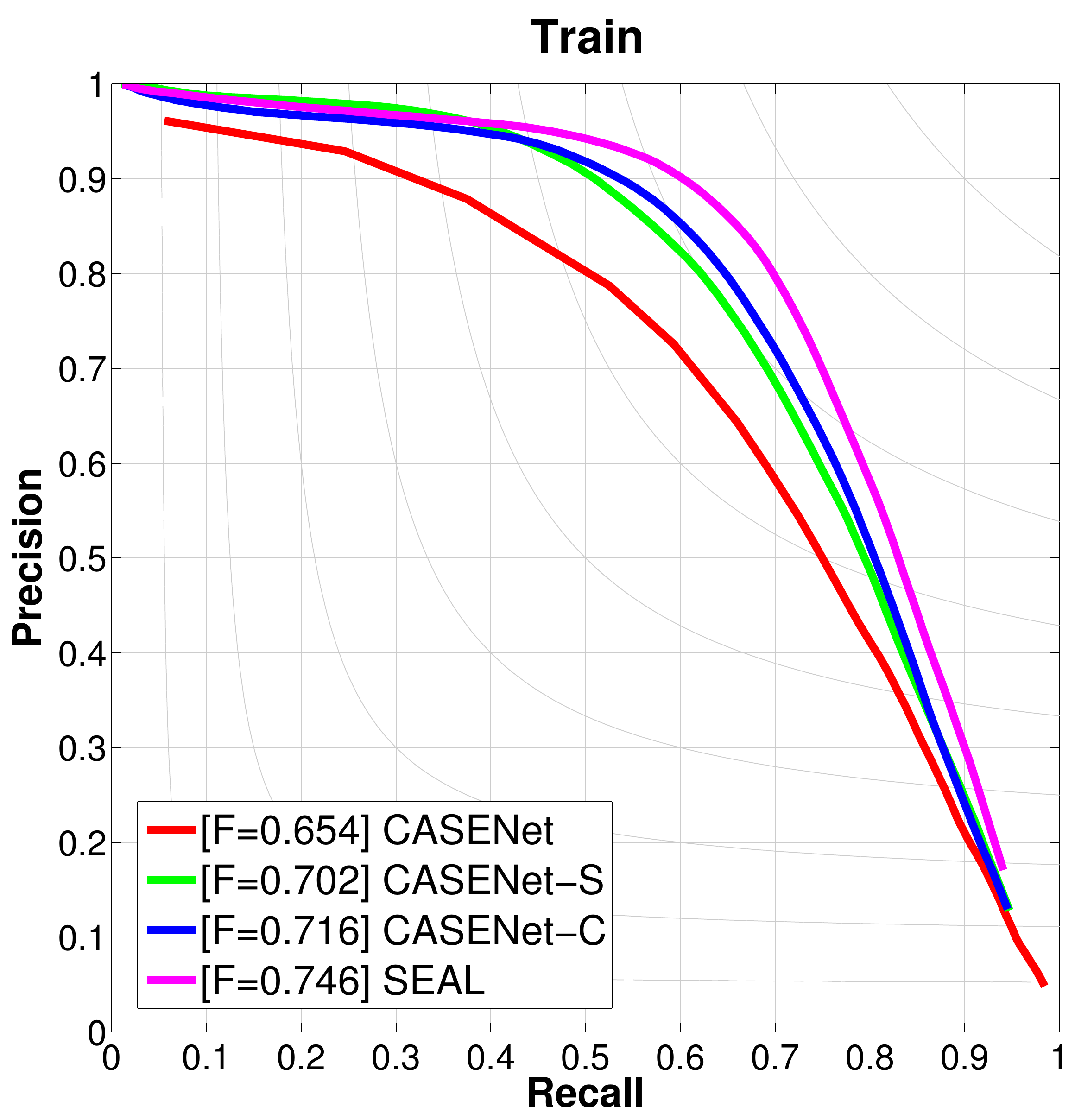}
	\includegraphics[width=.244\textwidth]{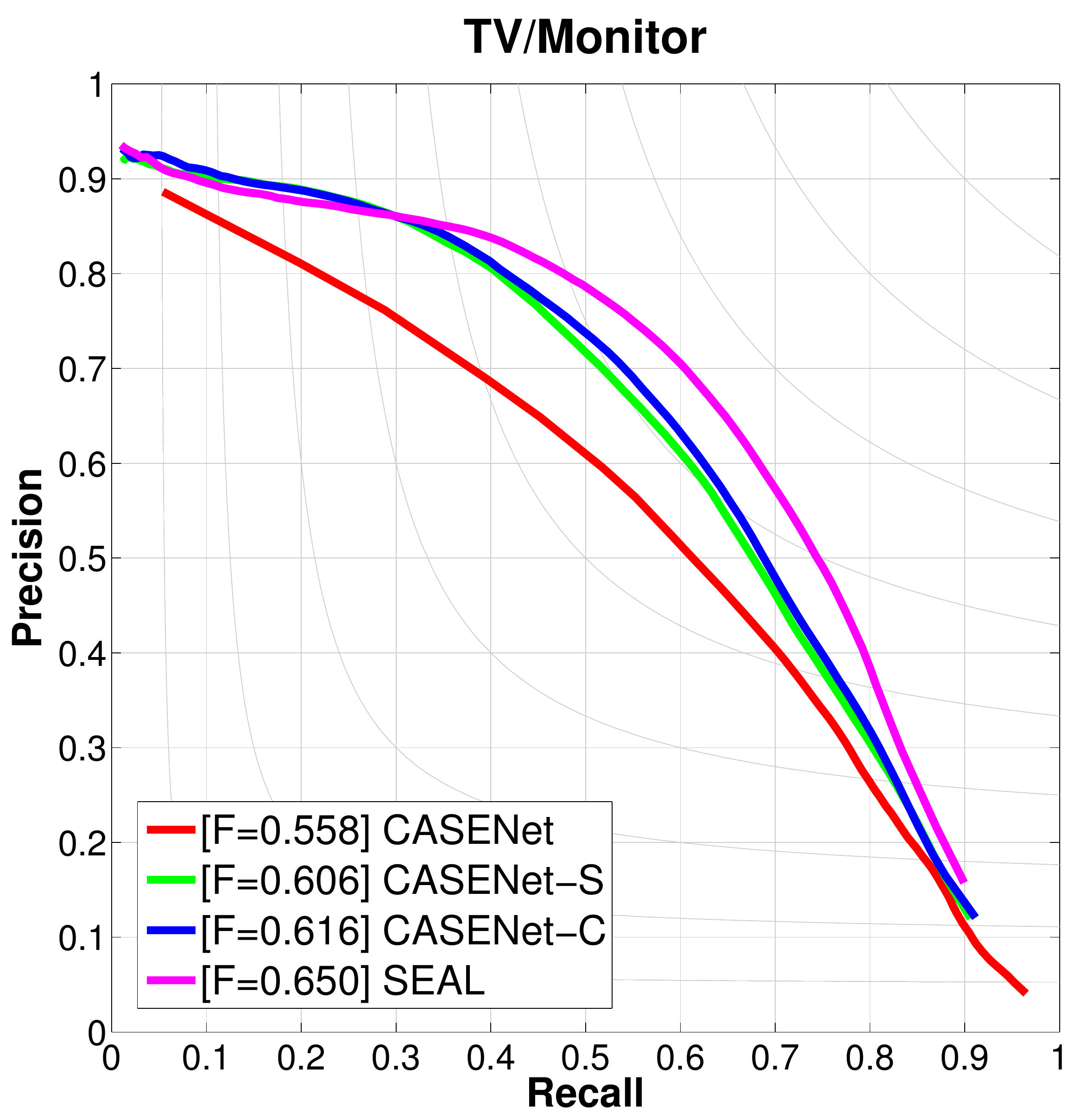}\\
	\caption{Class-wise precision-recall curves of SEAL and comparing baselines on the original SBD test set under the ``Raw'' setting.}\label{pr_sbd_orig_raw}
\end{figure}

\begin{figure}[tbh]
	\centering
	\includegraphics[width=.244\textwidth]{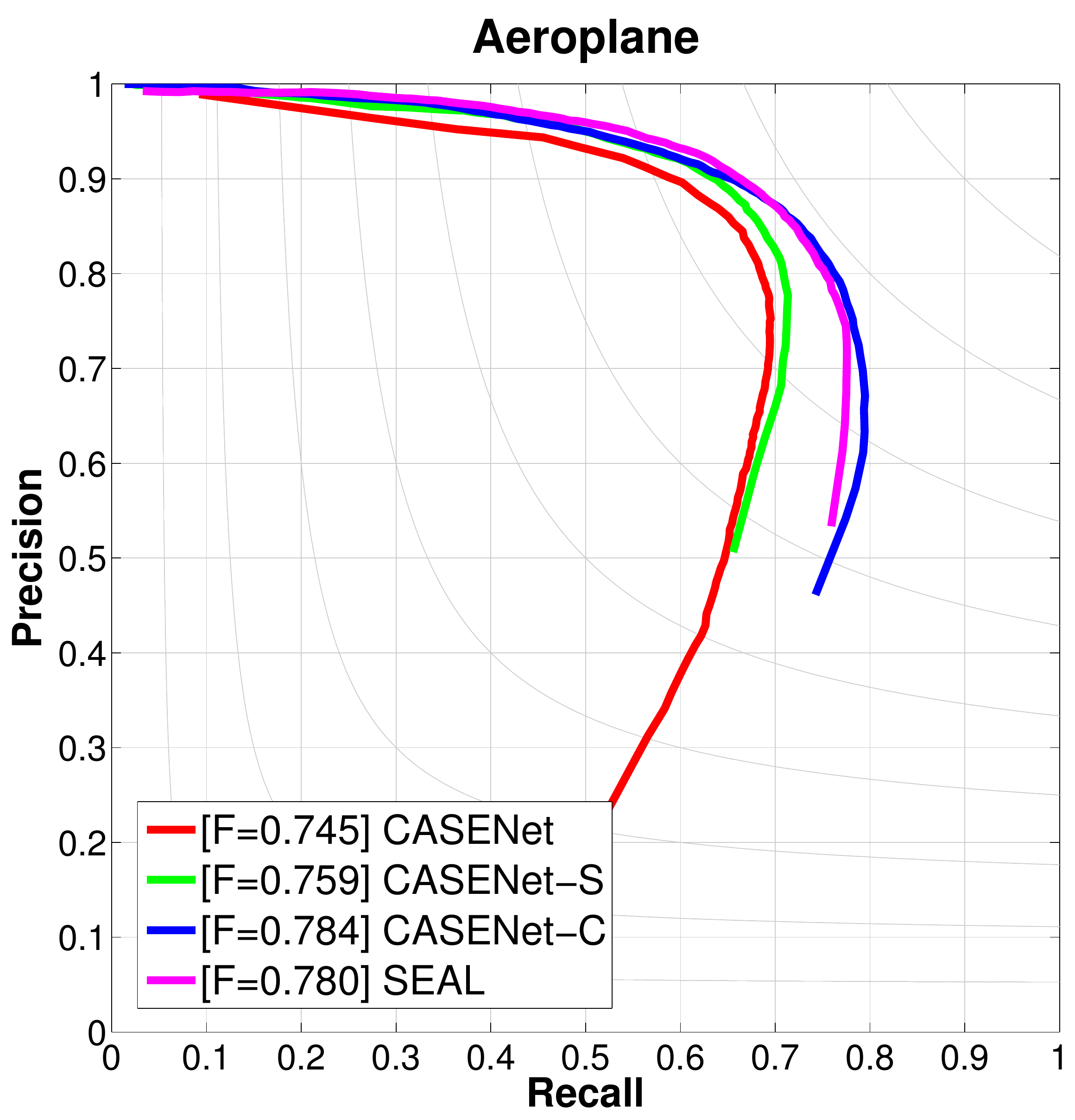}
	\includegraphics[width=.244\textwidth]{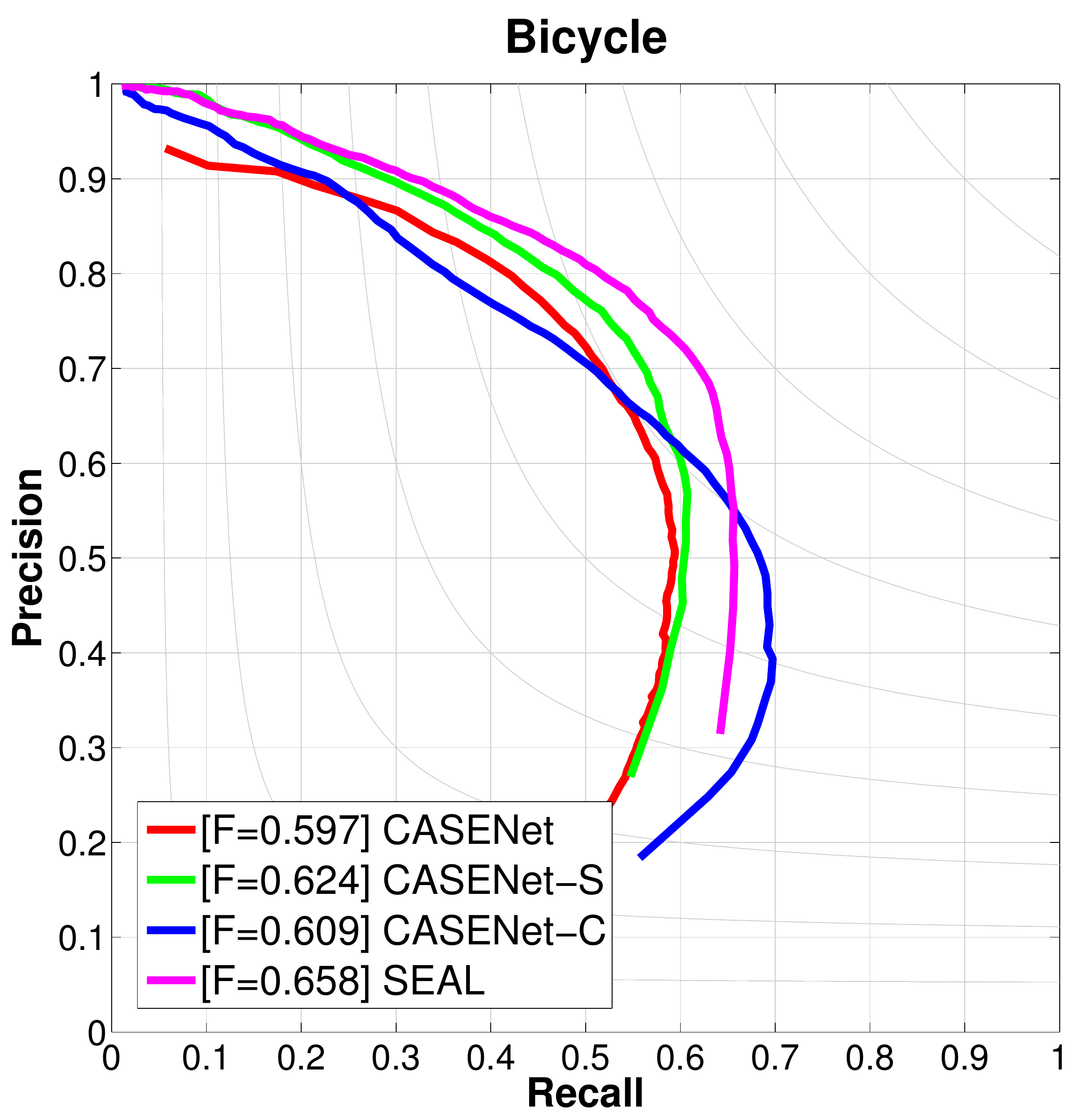}
	\includegraphics[width=.244\textwidth]{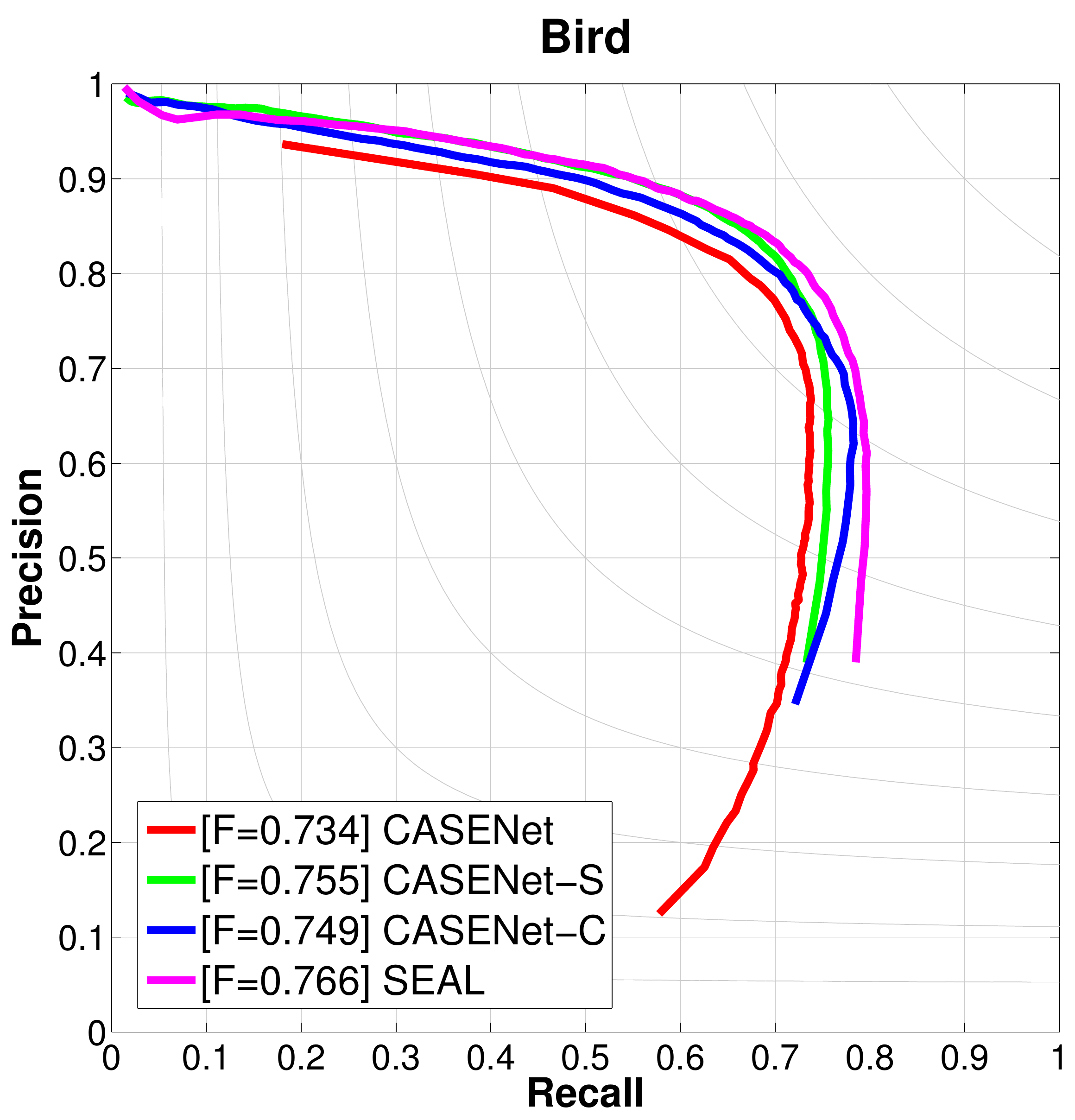}
	\includegraphics[width=.244\textwidth]{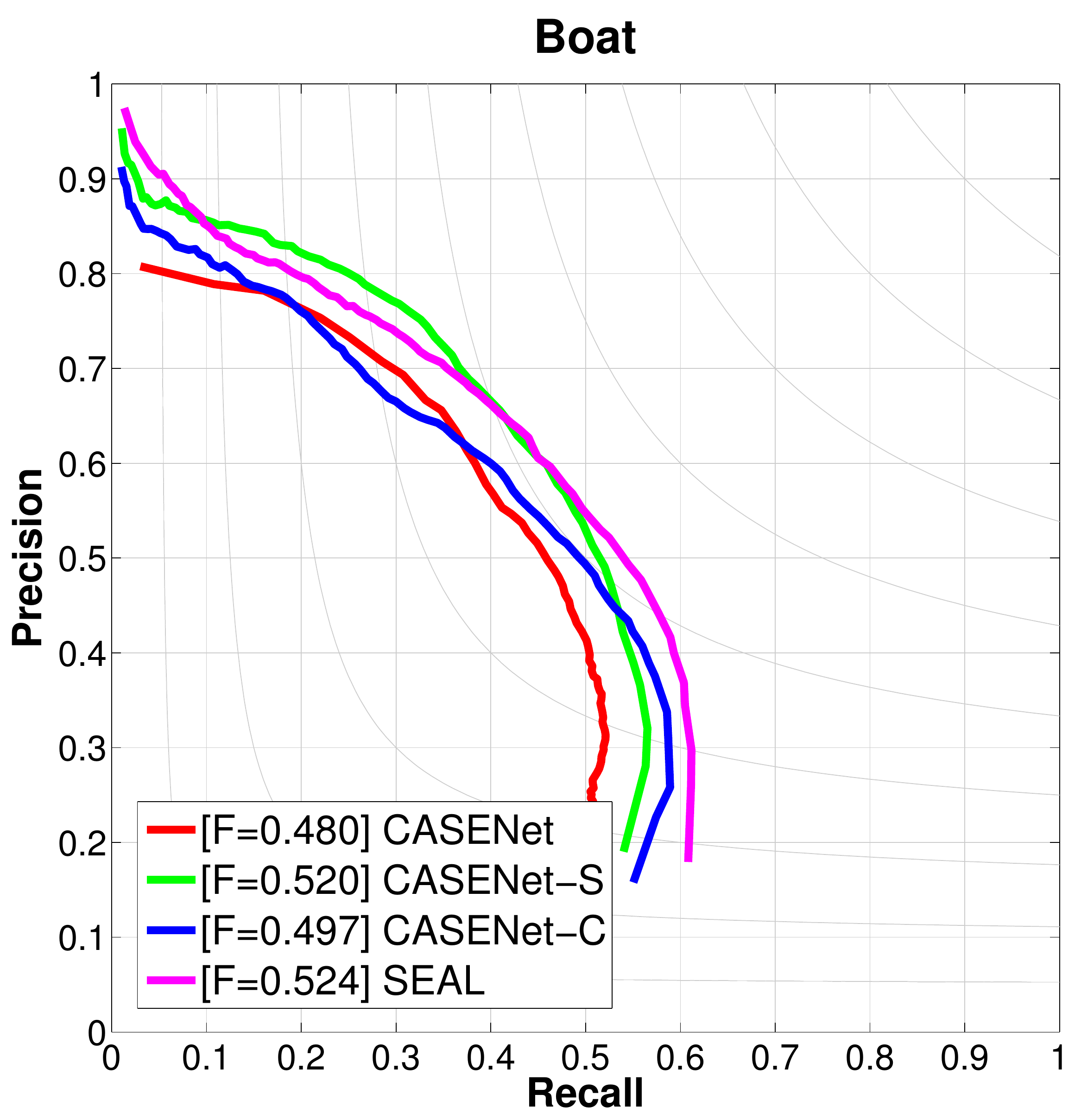}\\
	
	\includegraphics[width=.244\textwidth]{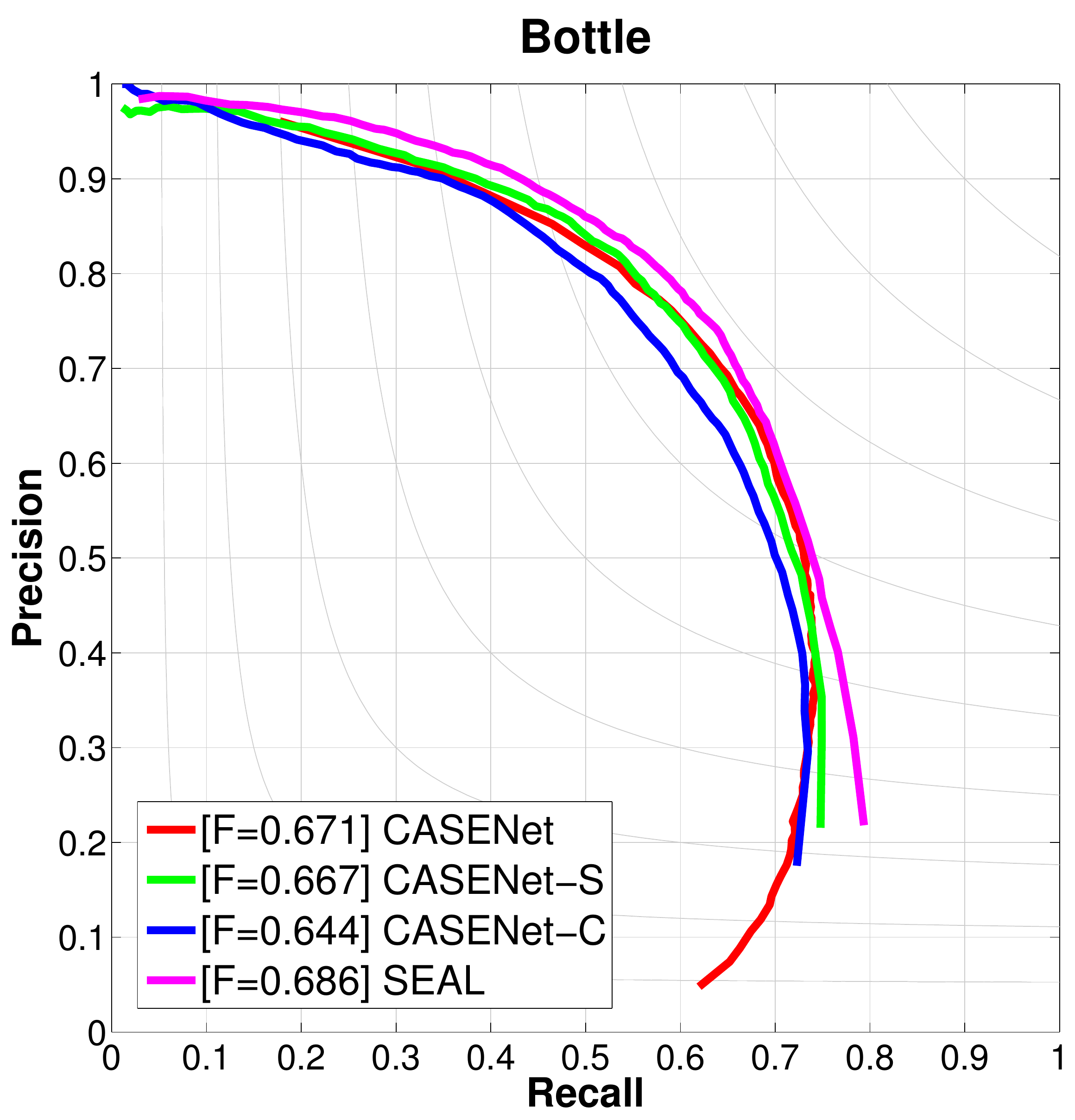}
	\includegraphics[width=.244\textwidth]{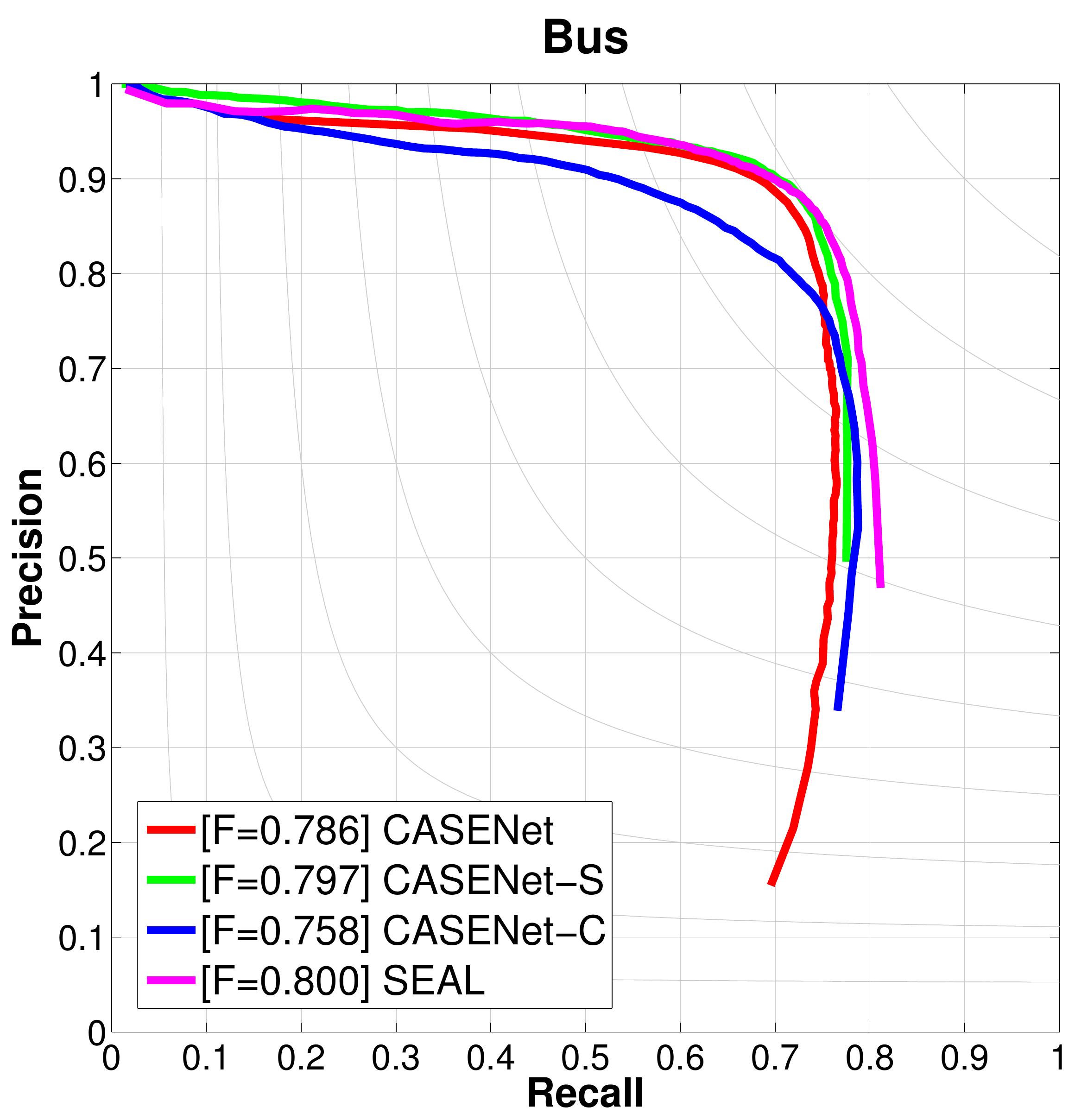}
	\includegraphics[width=.244\textwidth]{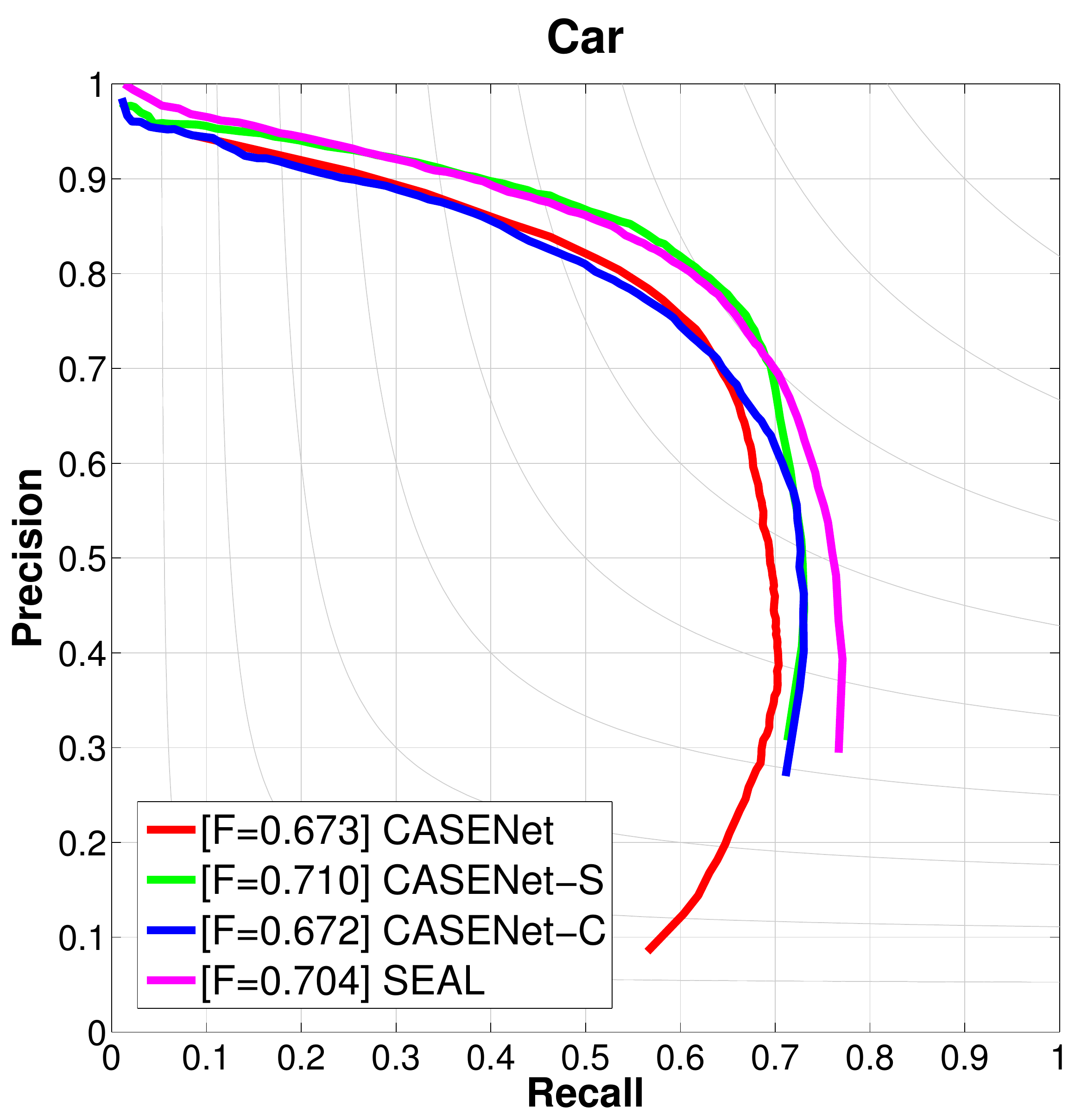}
	\includegraphics[width=.244\textwidth]{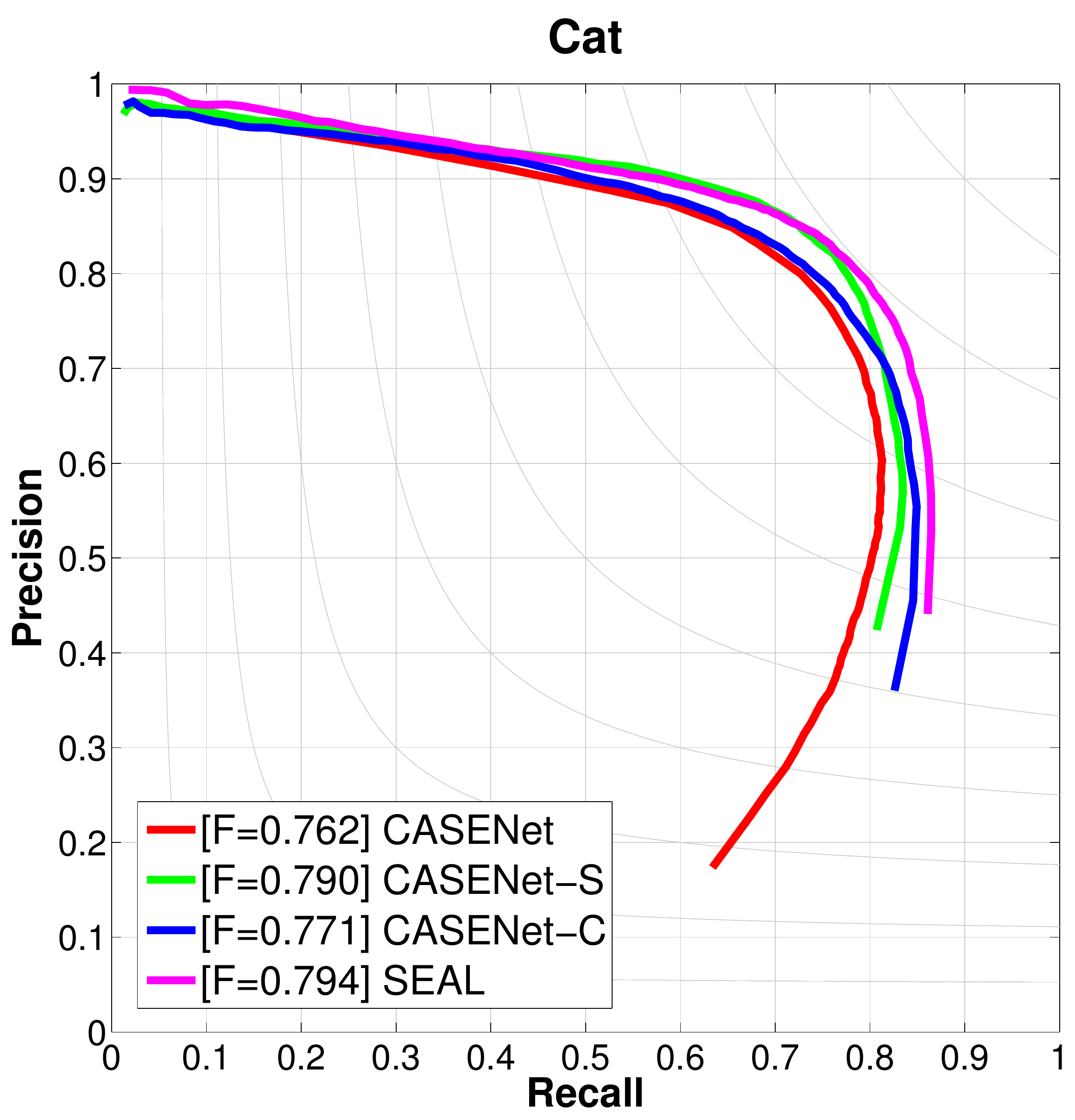}\\
	
	\includegraphics[width=.244\textwidth]{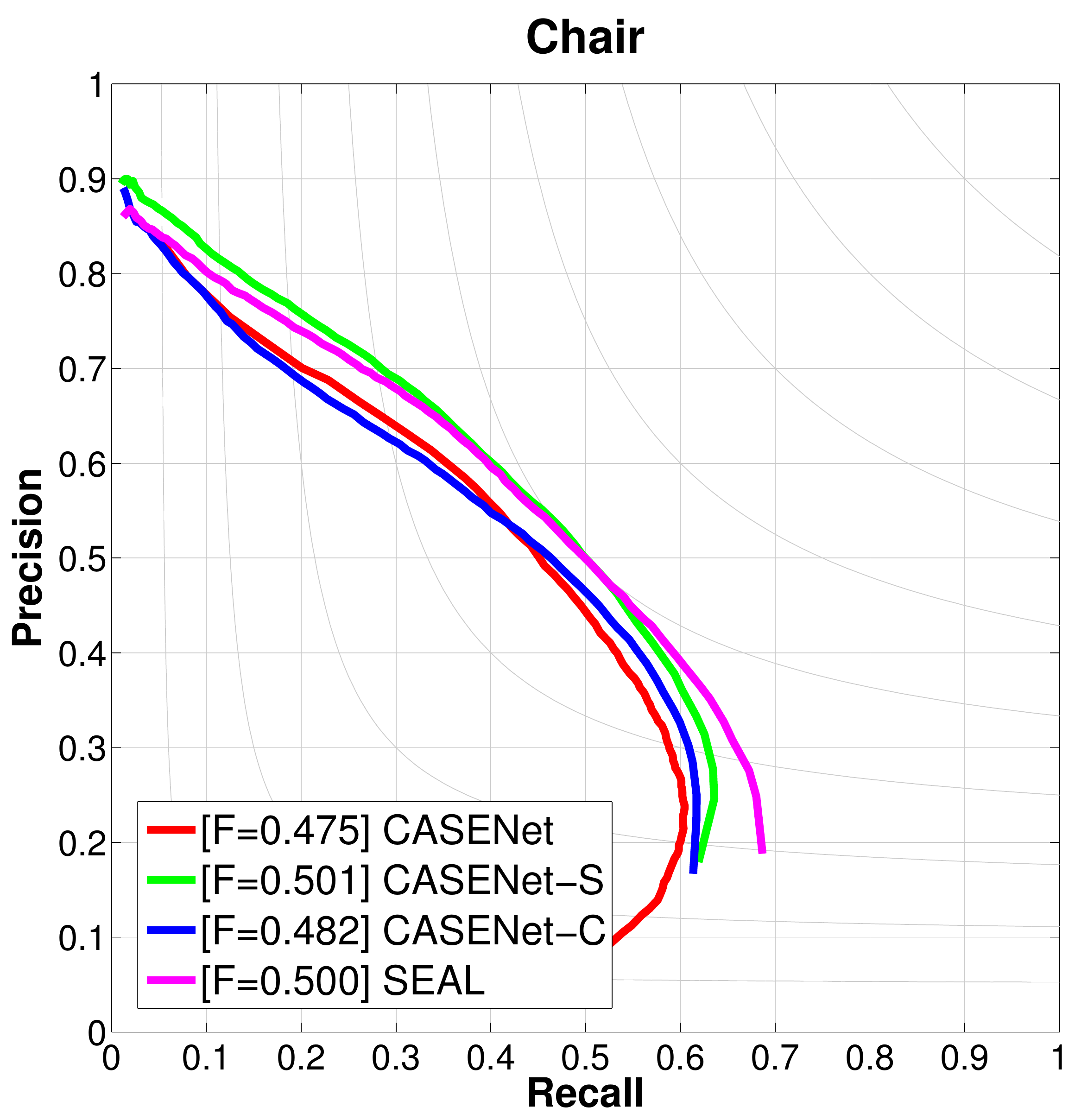}
	\includegraphics[width=.244\textwidth]{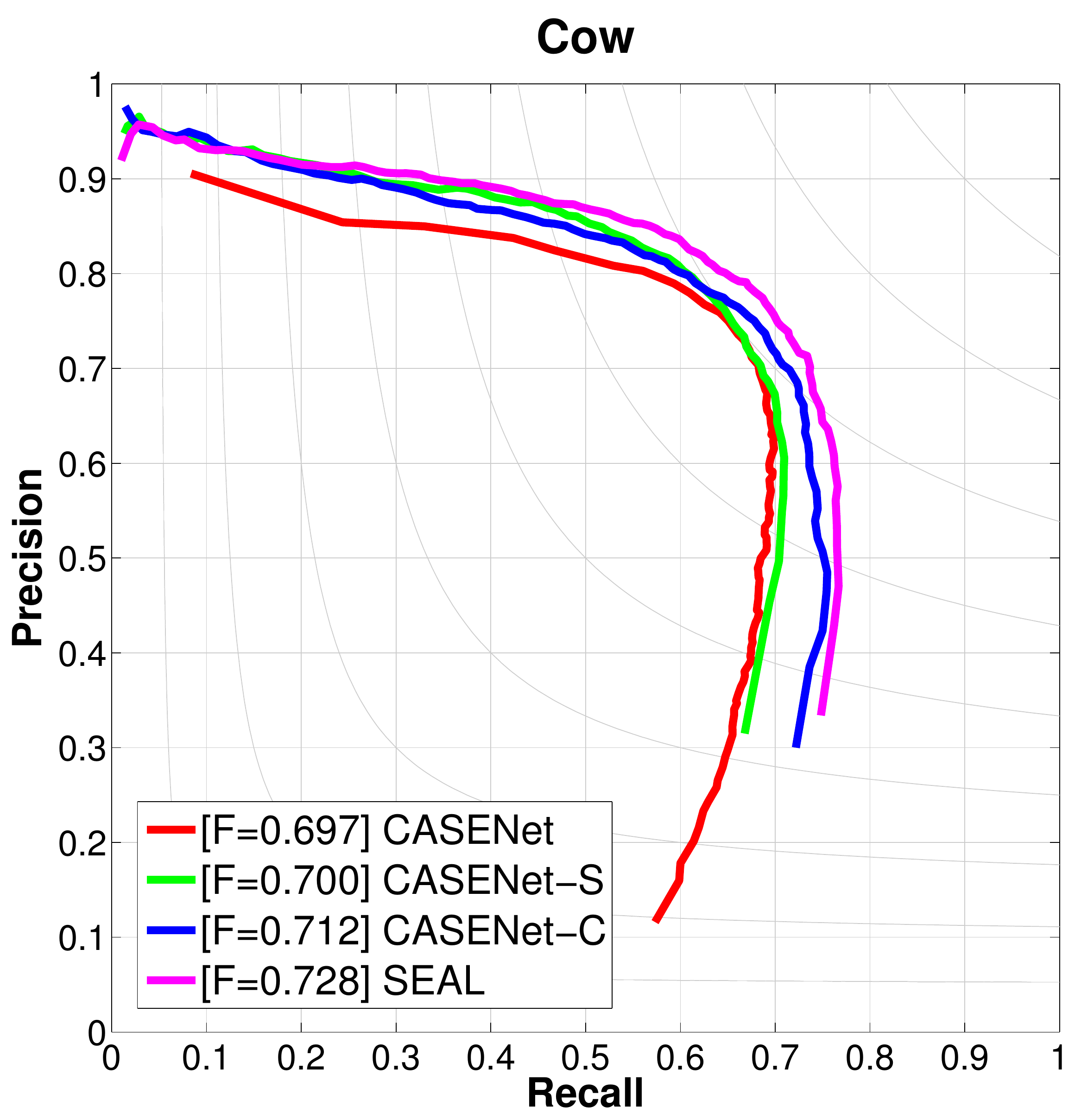}
	\includegraphics[width=.244\textwidth]{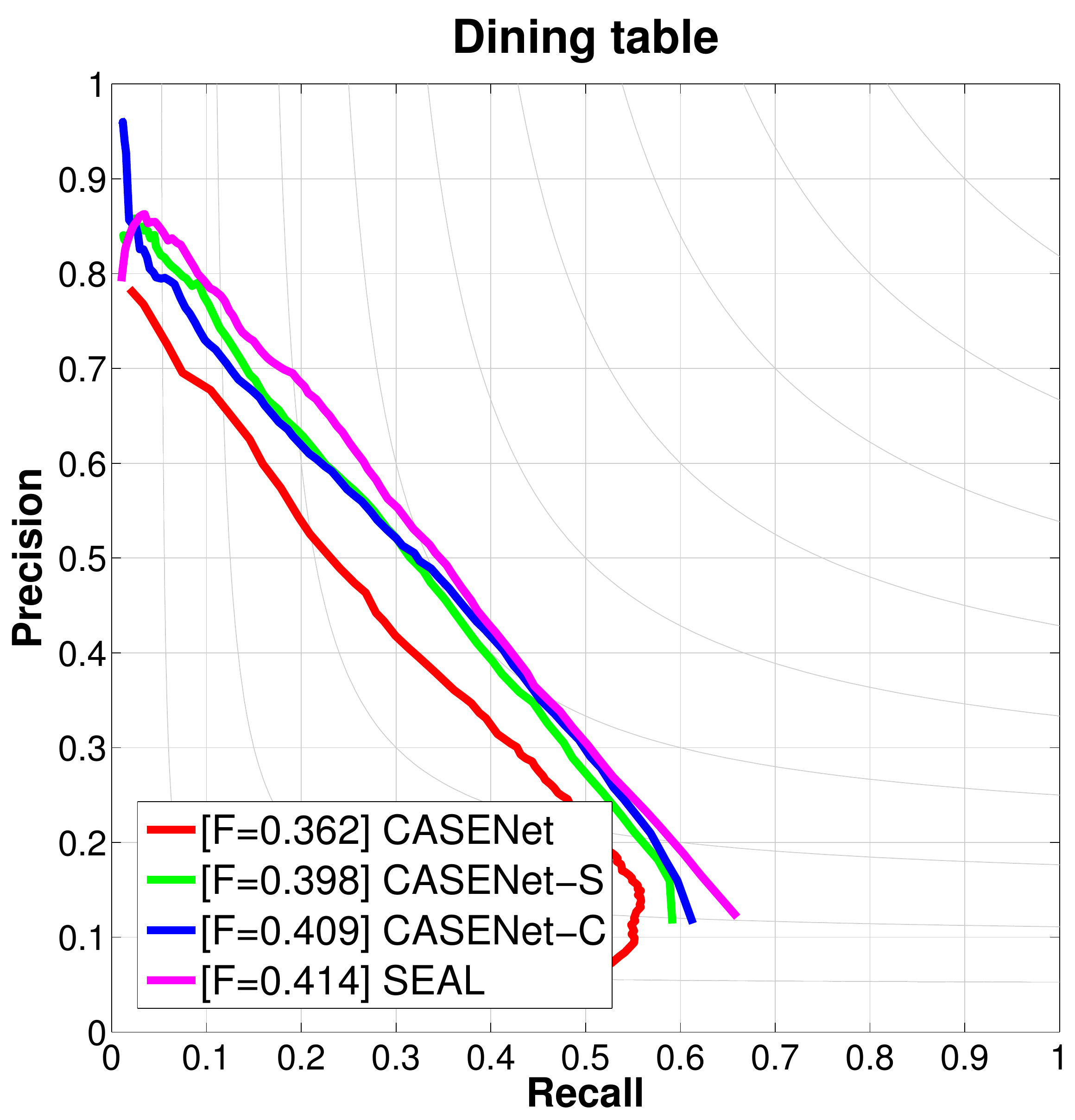}
	\includegraphics[width=.244\textwidth]{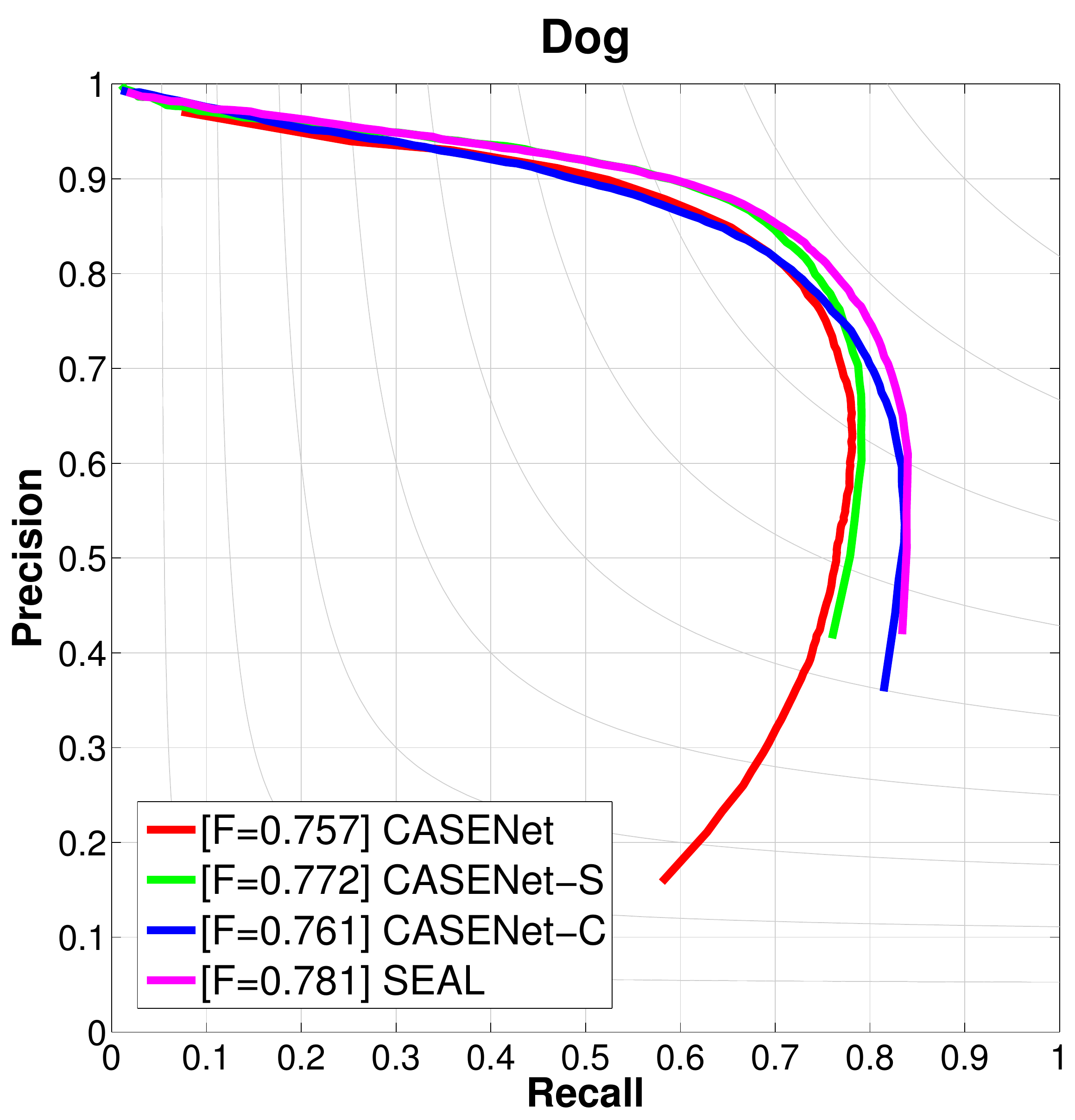}\\
	
	\includegraphics[width=.244\textwidth]{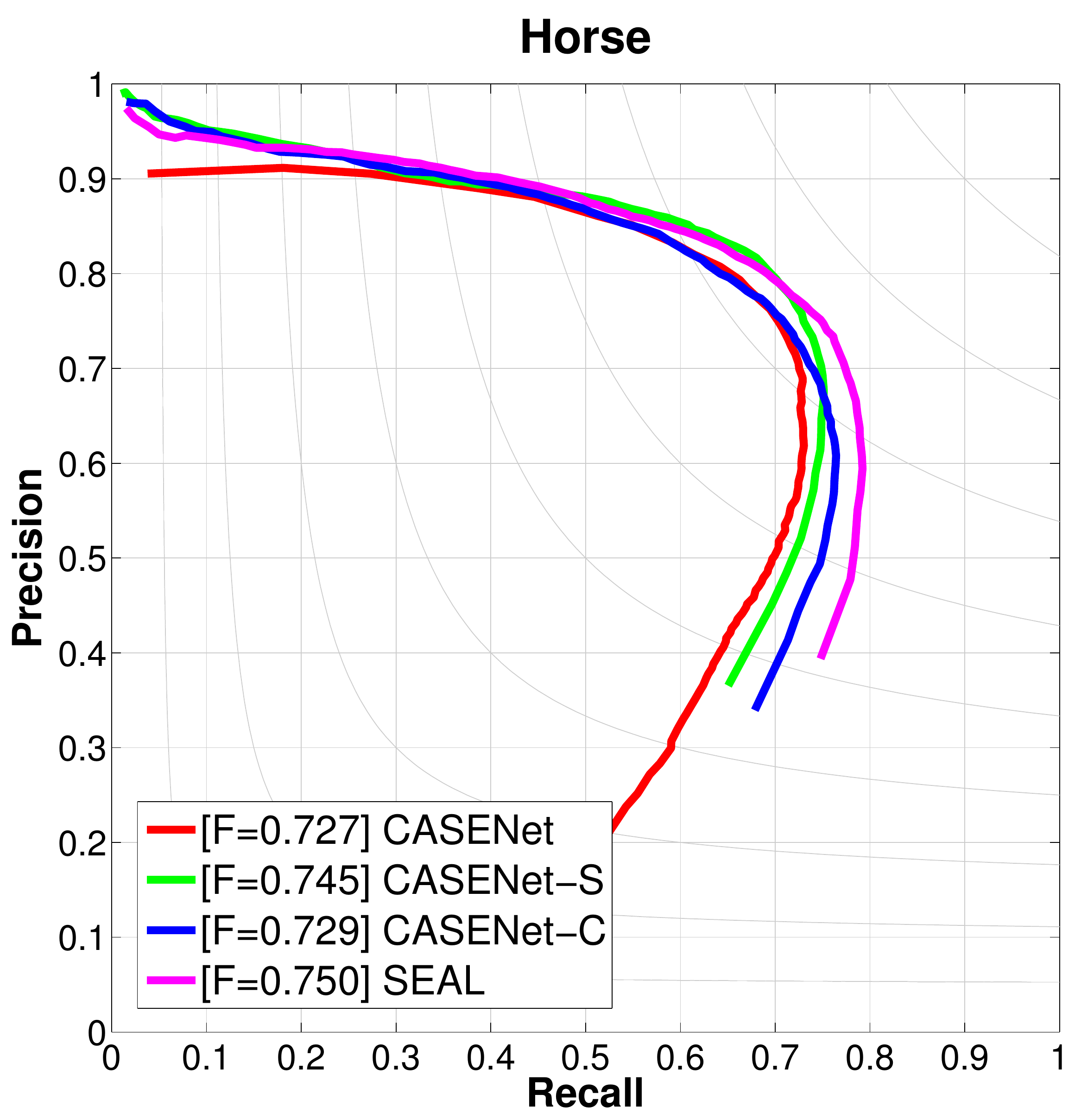}
	\includegraphics[width=.244\textwidth]{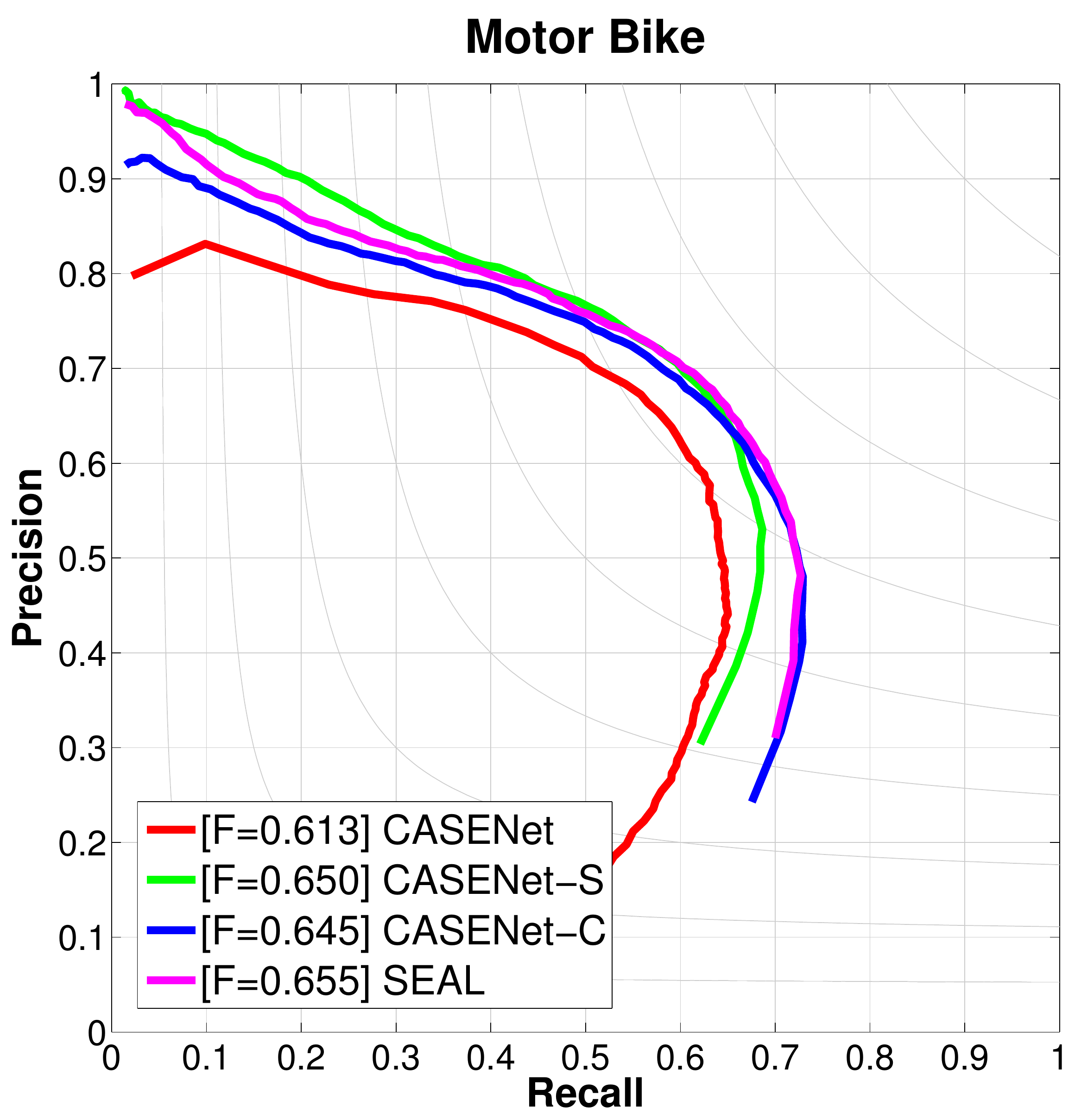}
	\includegraphics[width=.244\textwidth]{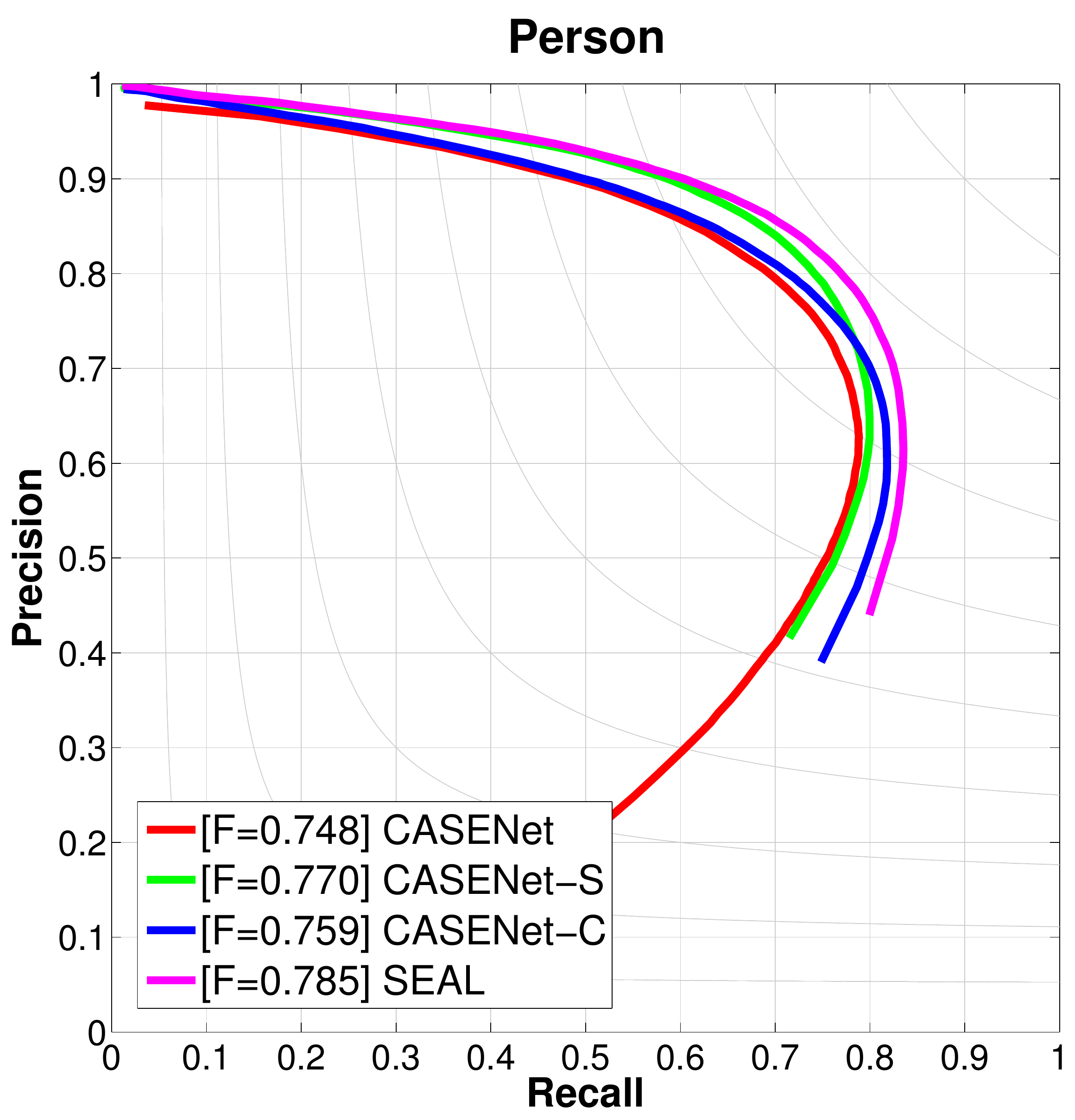}
	\includegraphics[width=.244\textwidth]{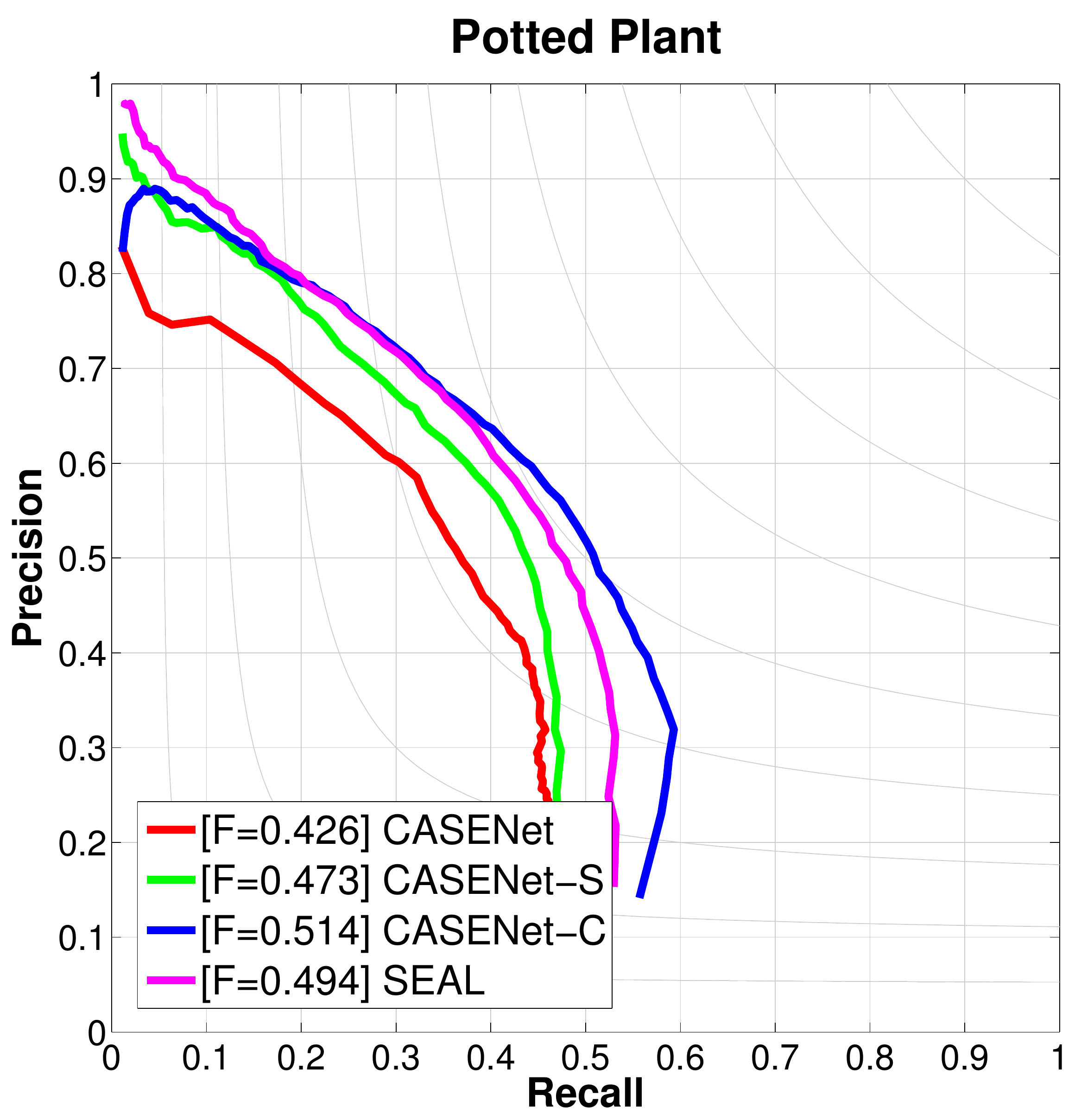}\\
	
	\includegraphics[width=.244\textwidth]{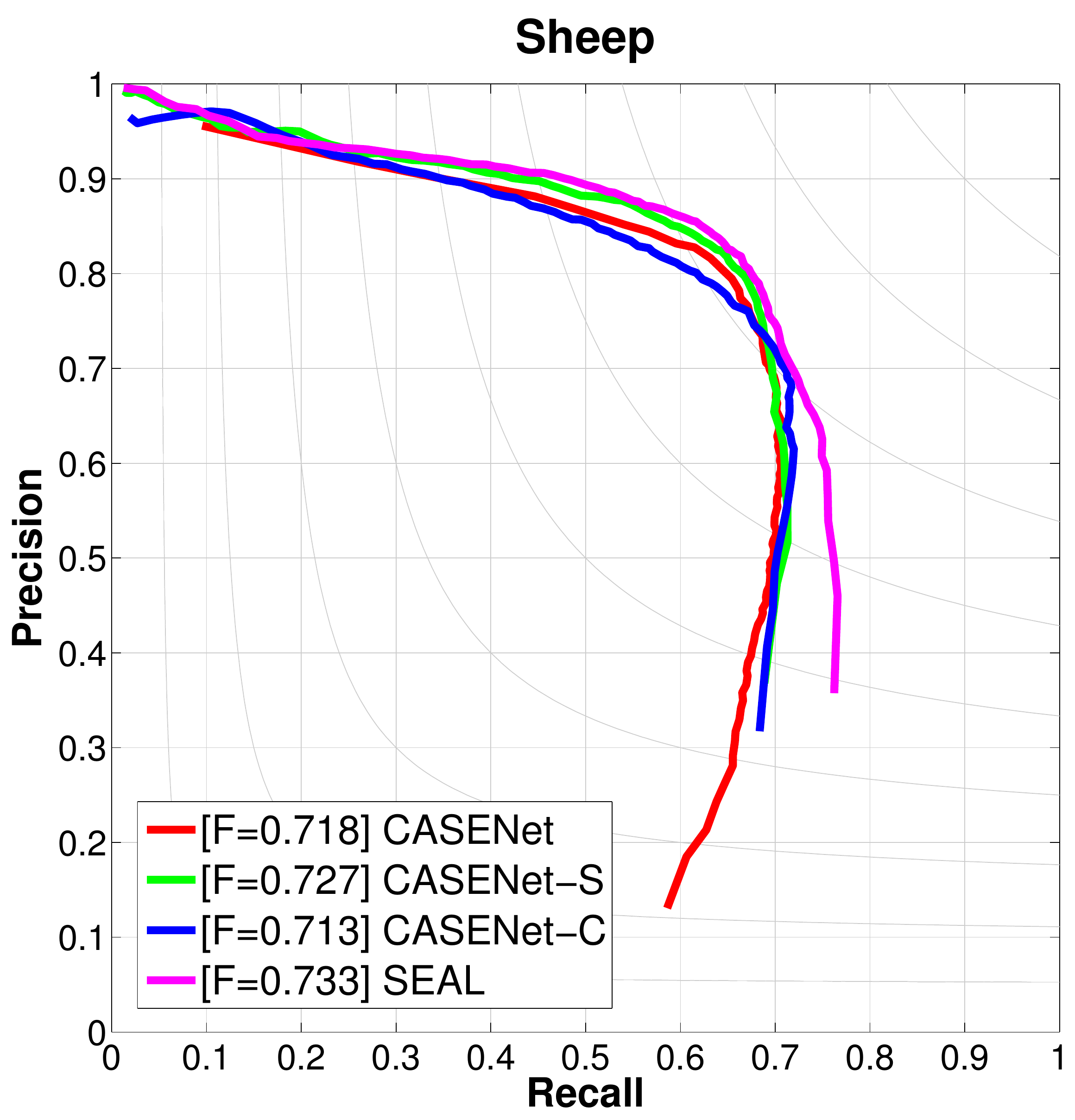}
	\includegraphics[width=.244\textwidth]{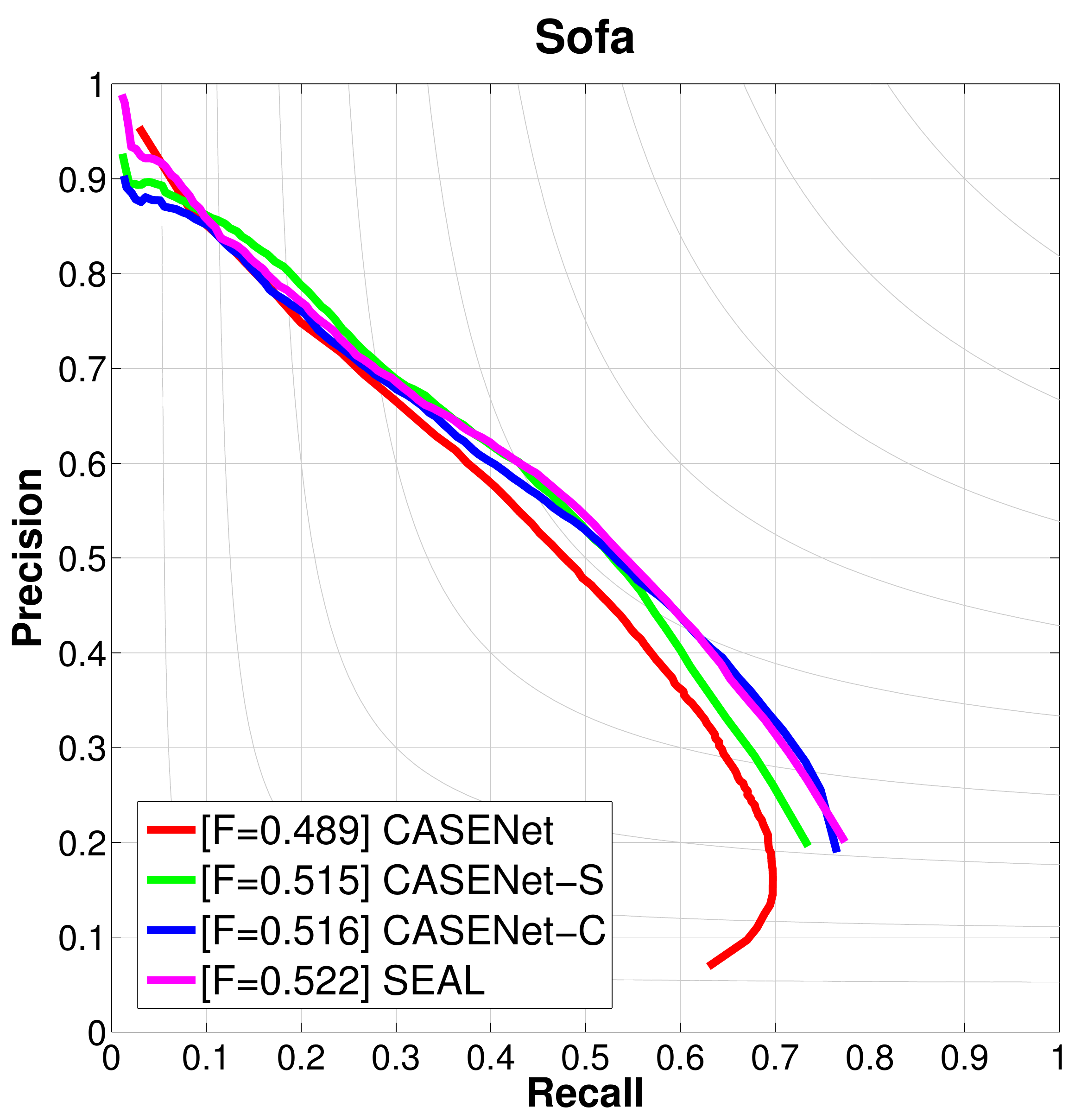}
	\includegraphics[width=.244\textwidth]{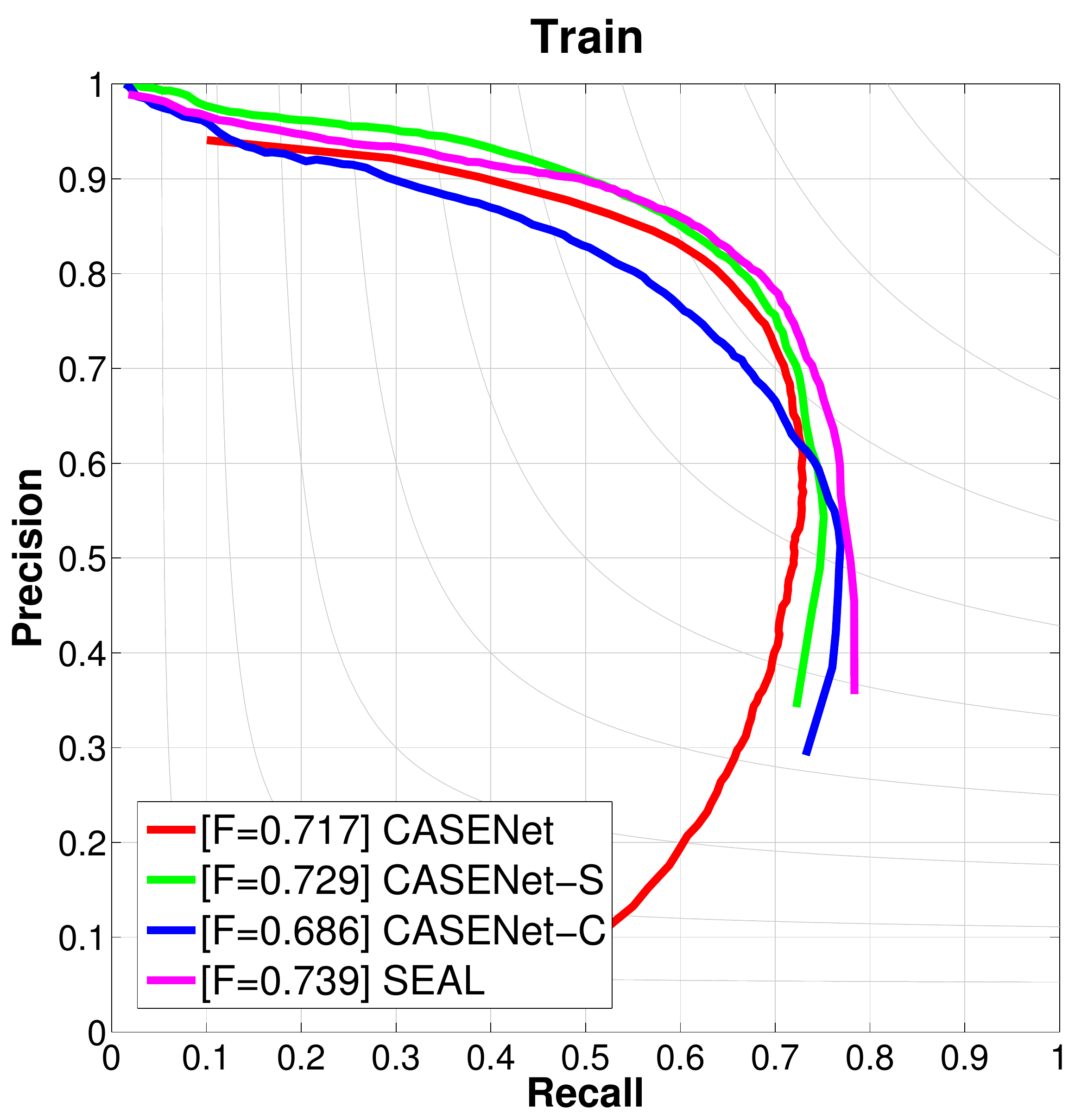}
	\includegraphics[width=.244\textwidth]{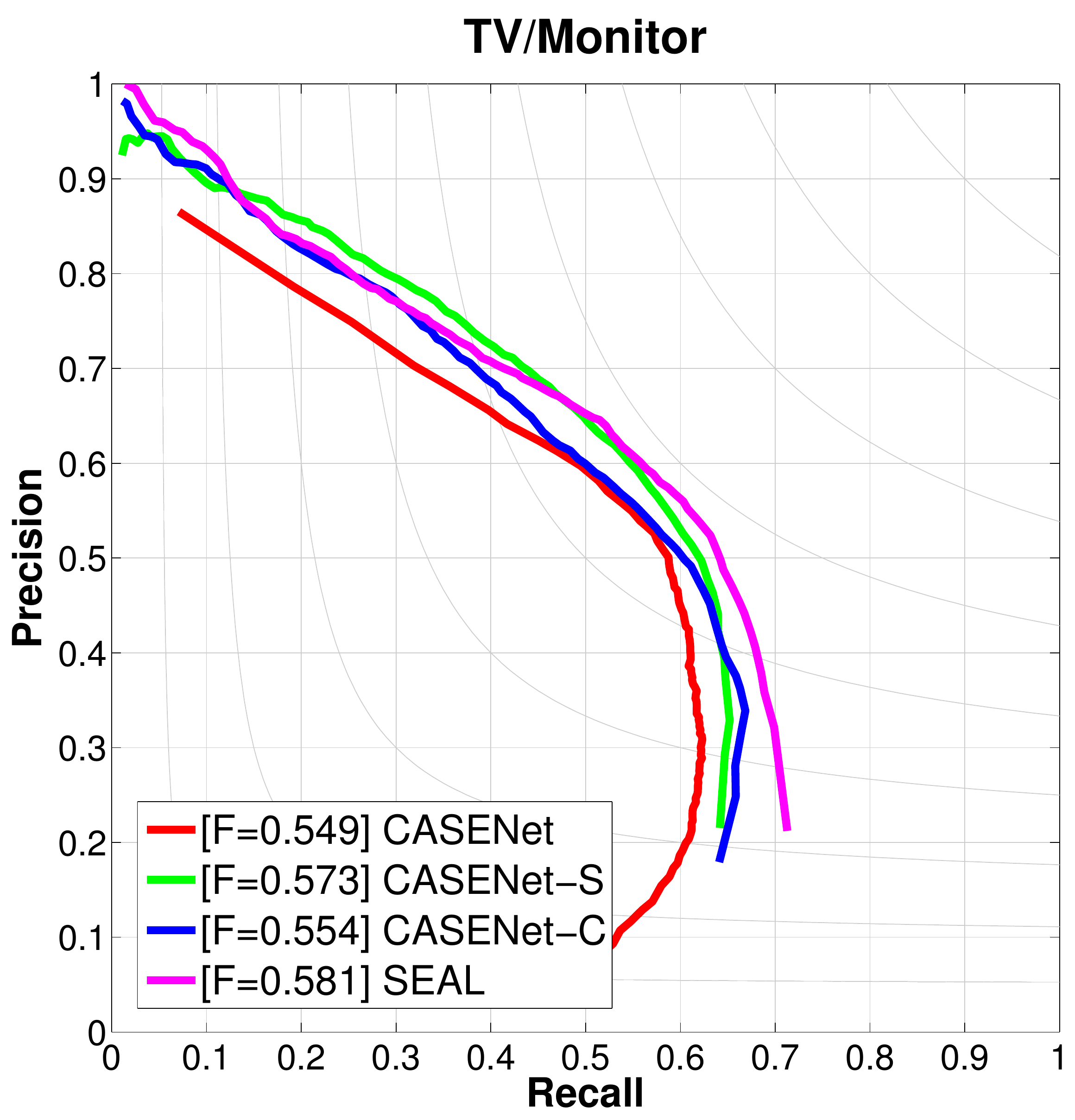}\\
	\caption{Class-wise precision-recall curves of SEAL and comparing baselines on the re-annotated SBD test set under the ``Thin'' setting.}\label{pr_sbd_refine_thin}
\end{figure}

\begin{figure}[tbh]
	\centering
	\includegraphics[width=.244\textwidth]{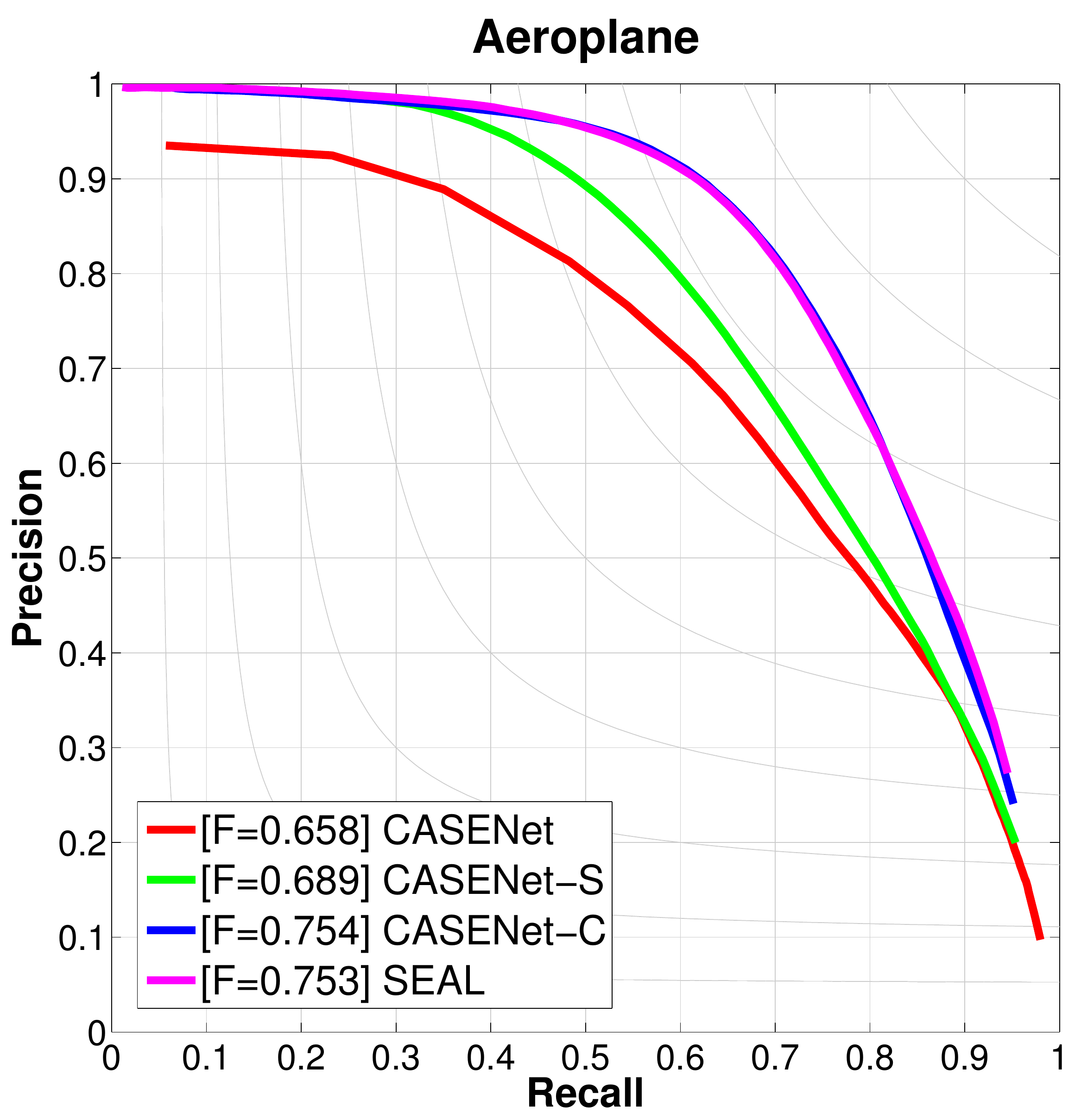}
	\includegraphics[width=.244\textwidth]{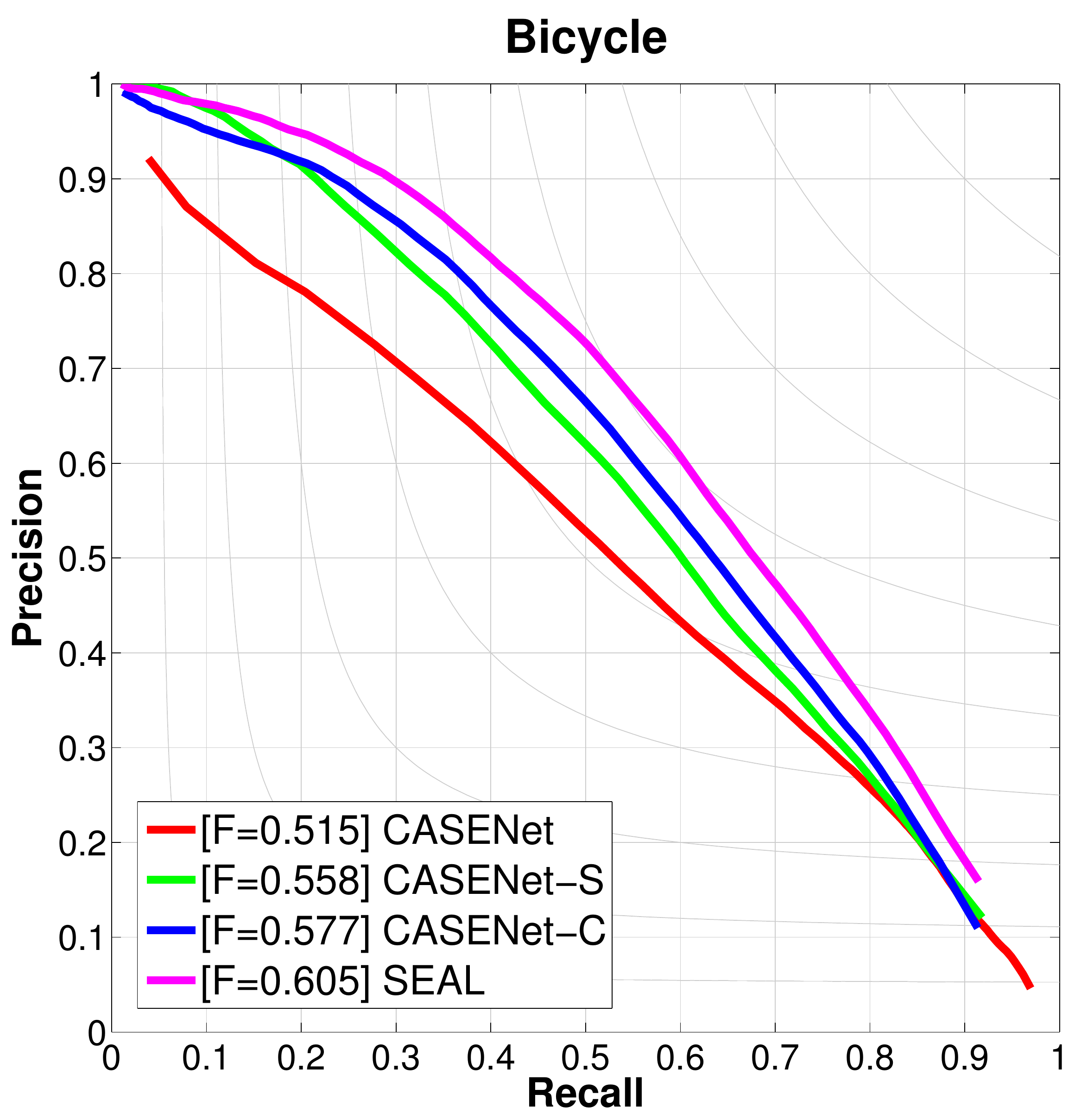}
	\includegraphics[width=.244\textwidth]{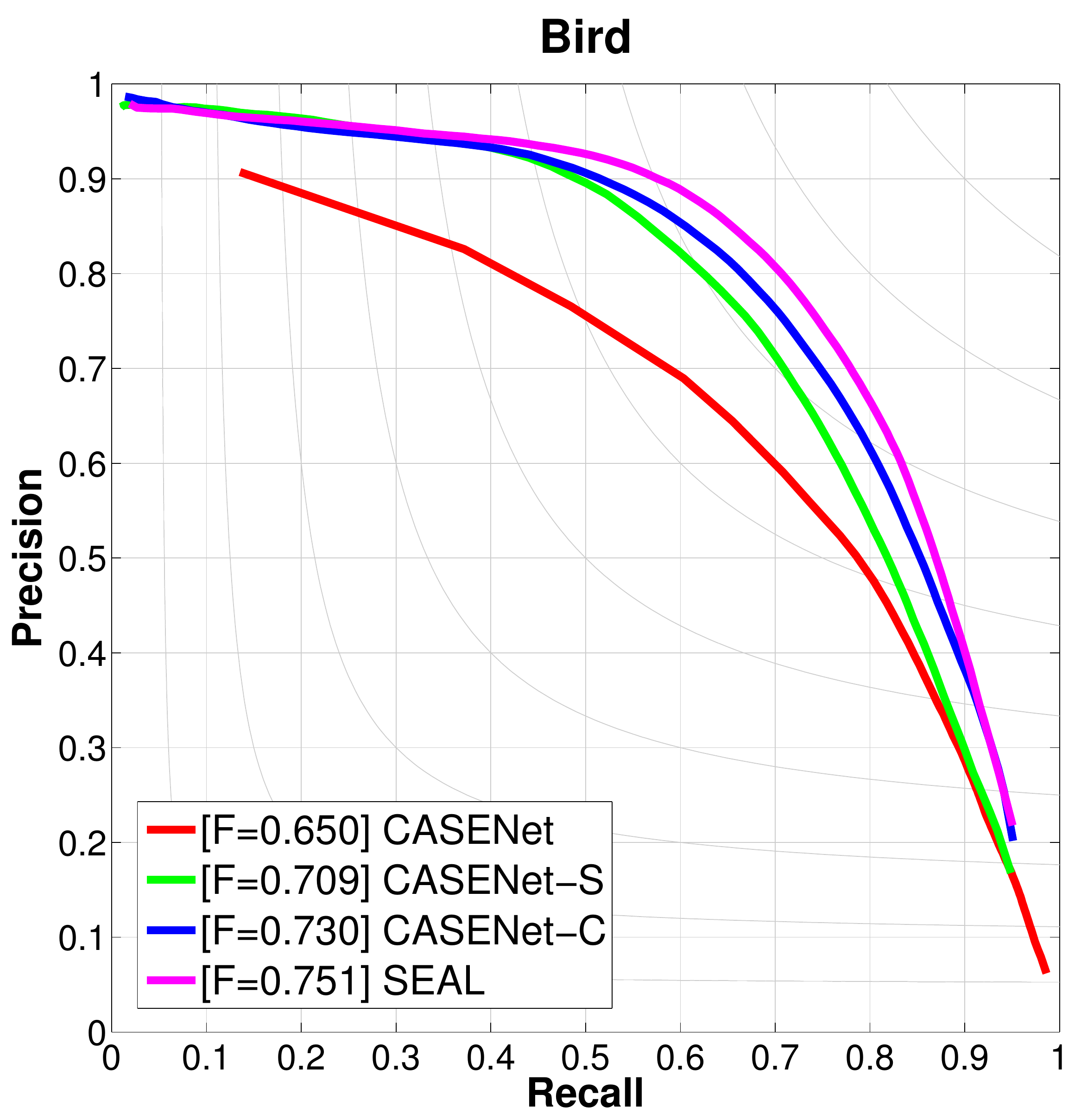}
	\includegraphics[width=.244\textwidth]{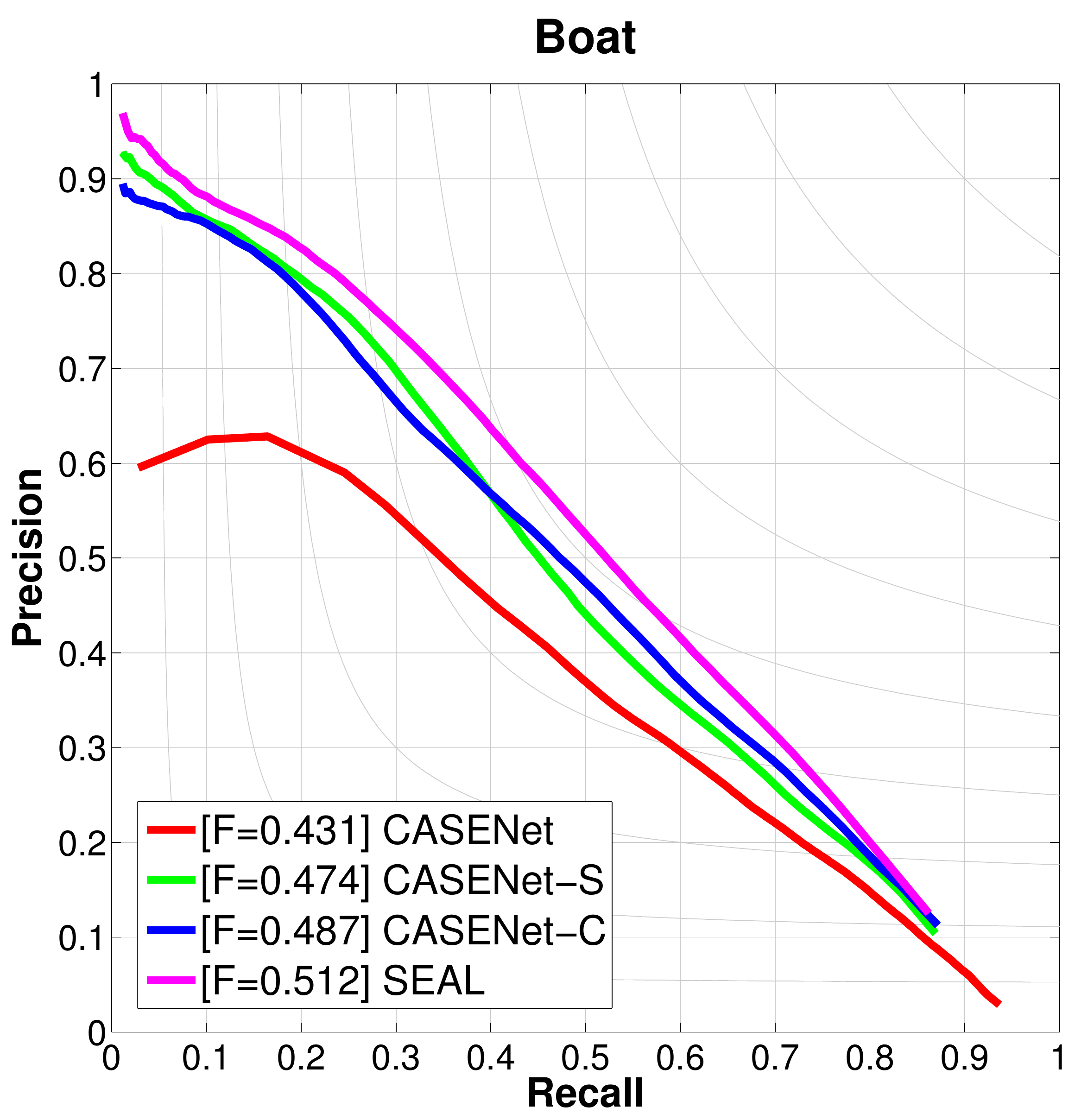}\\
	
	\includegraphics[width=.244\textwidth]{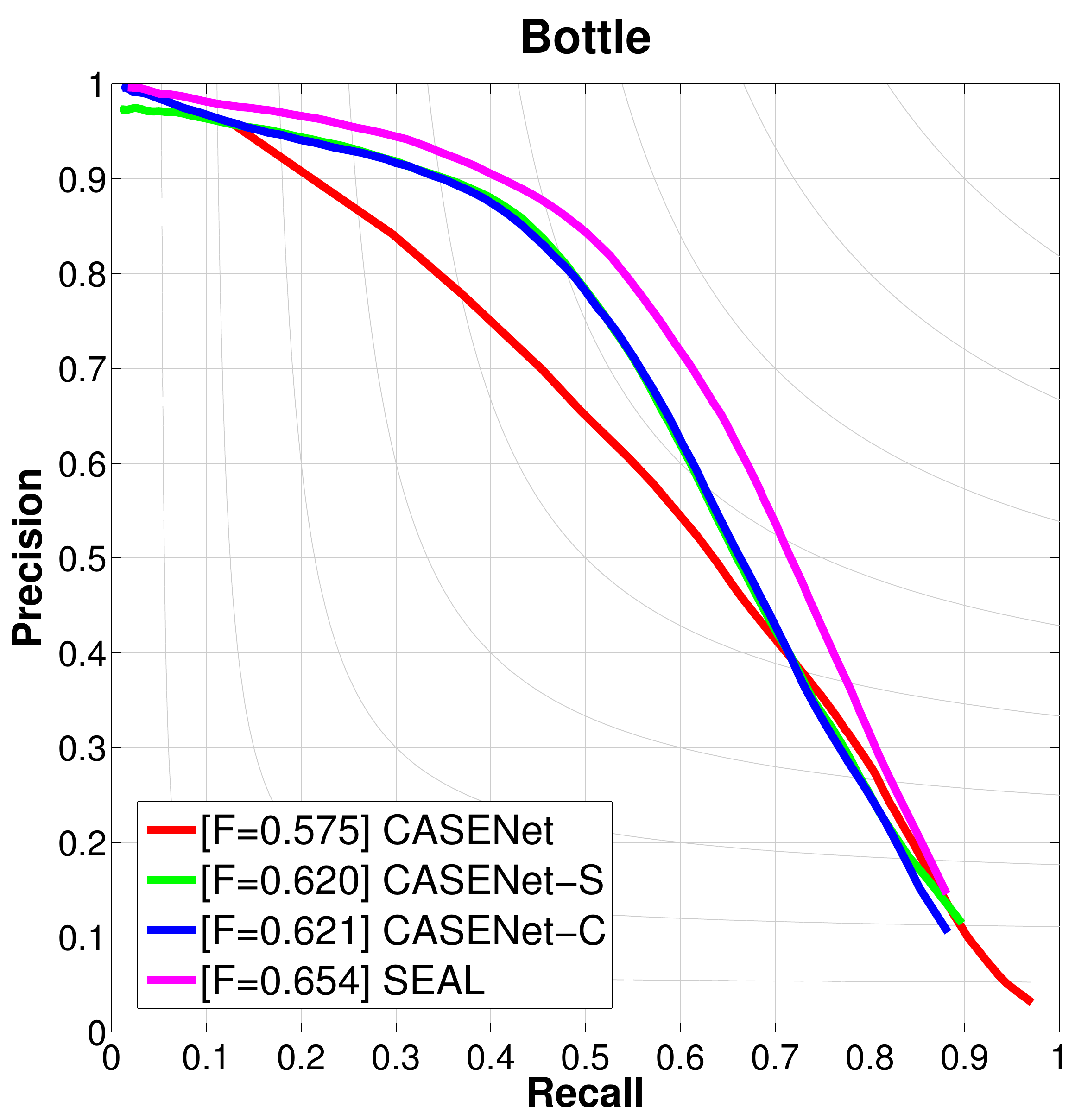}
	\includegraphics[width=.244\textwidth]{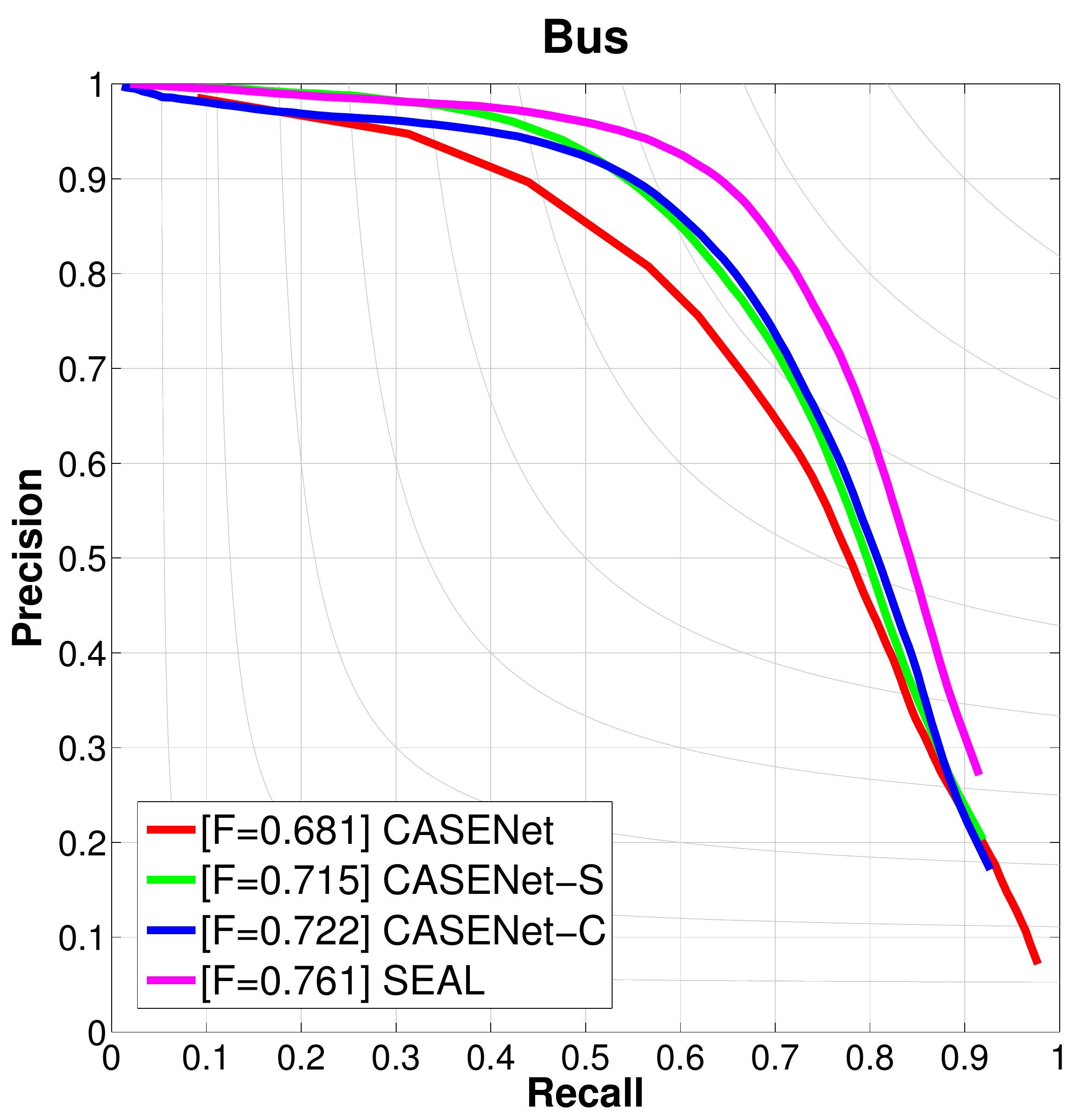}
	\includegraphics[width=.244\textwidth]{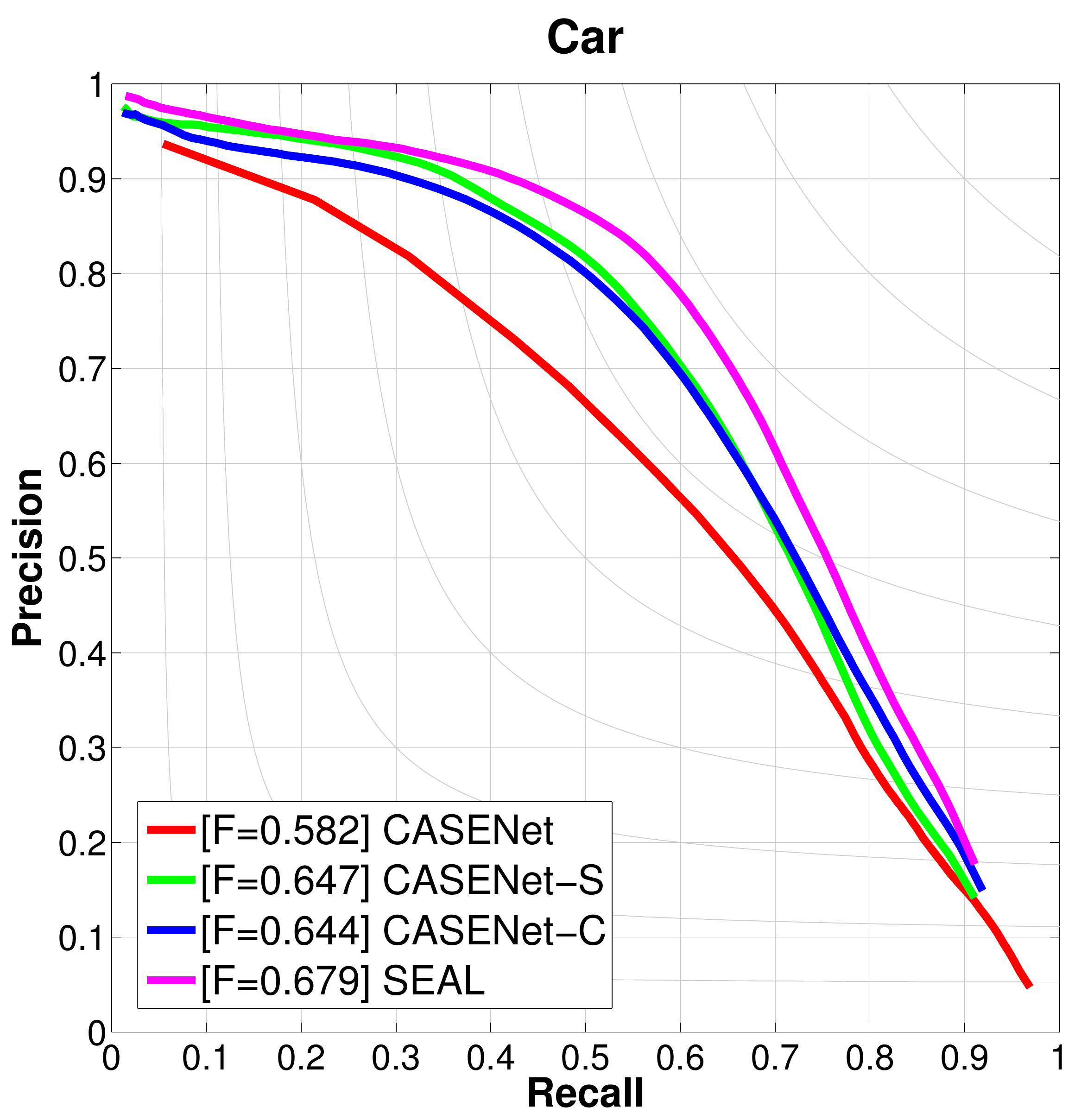}
	\includegraphics[width=.244\textwidth]{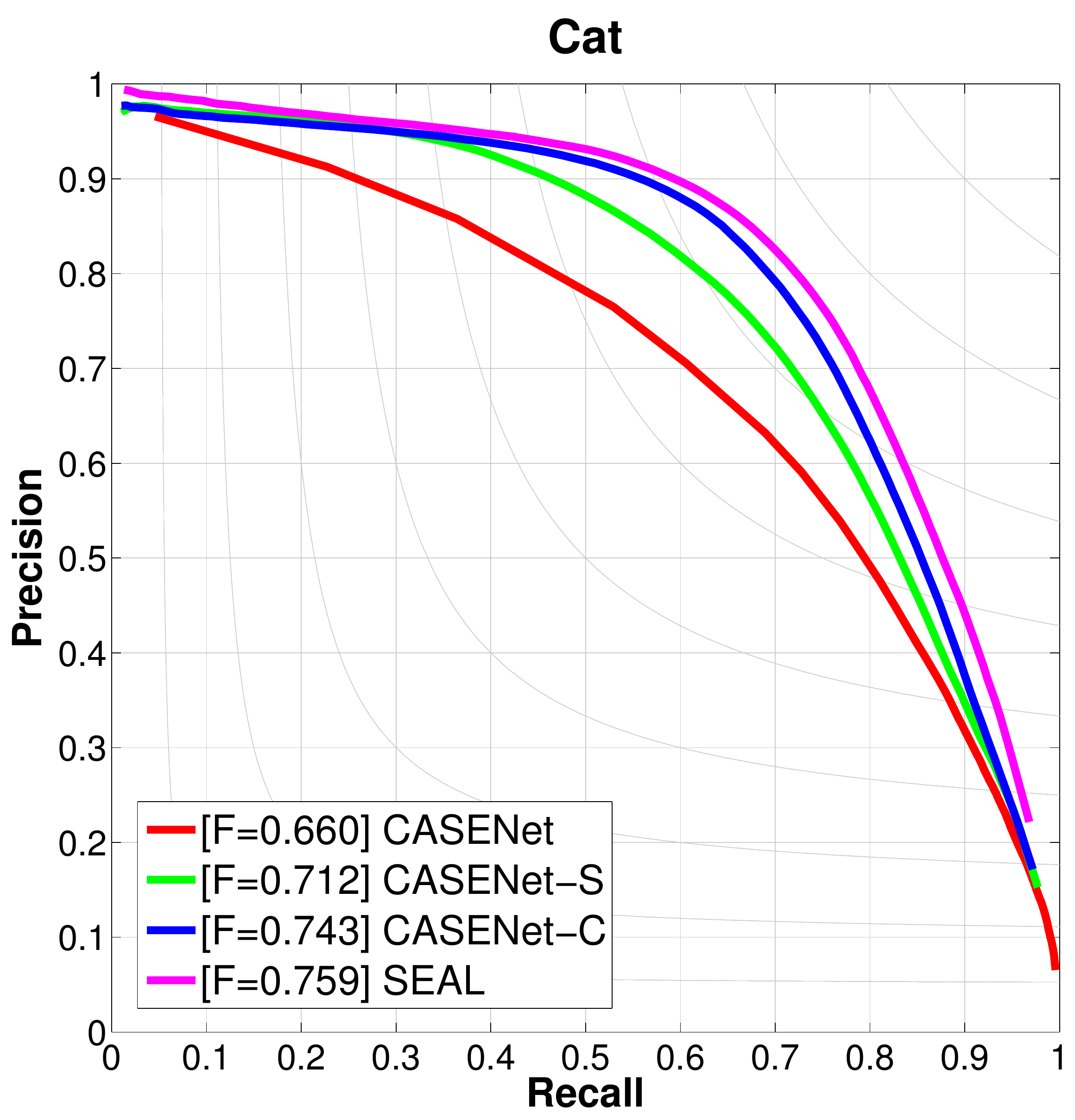}\\
	
	\includegraphics[width=.244\textwidth]{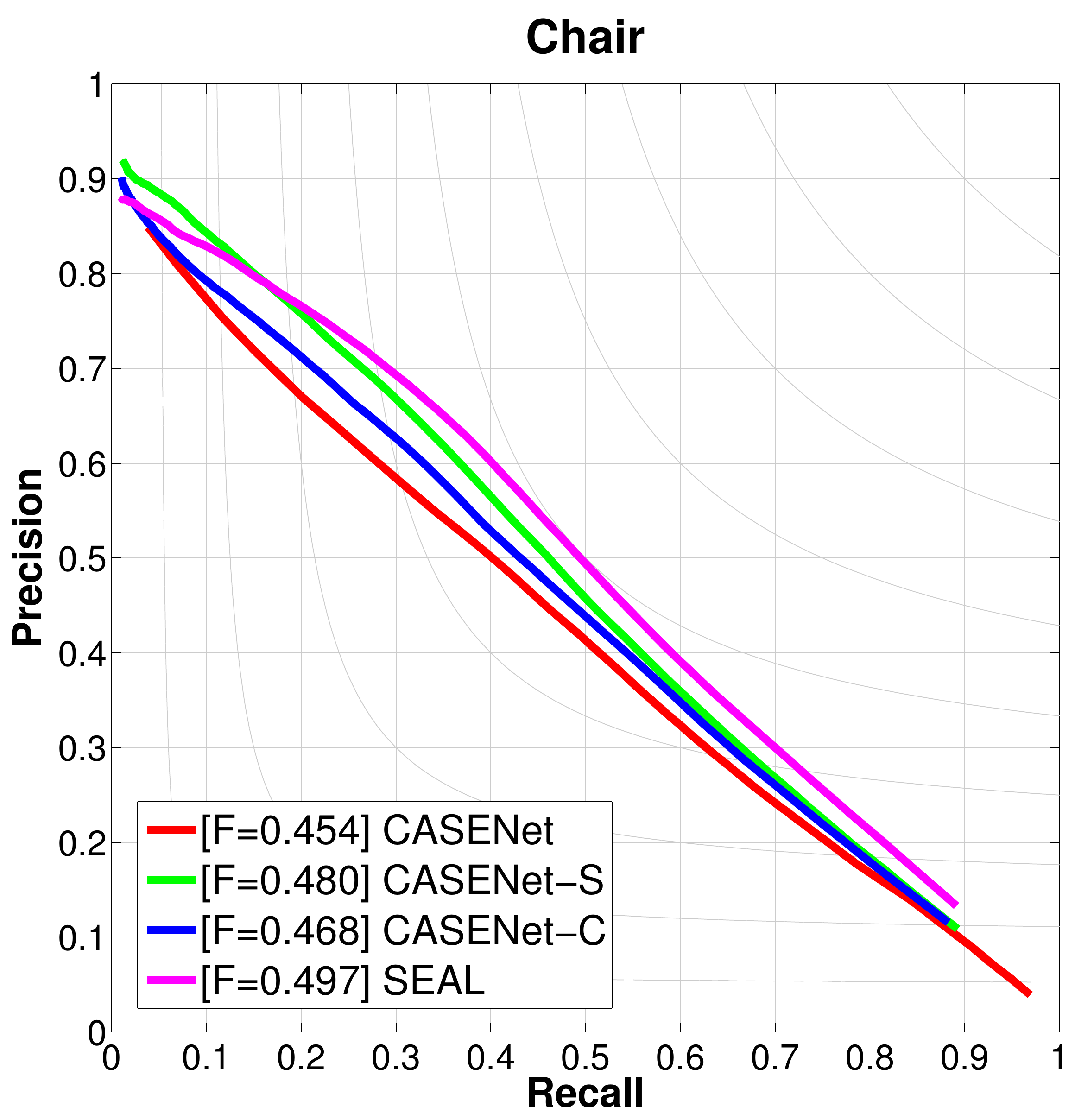}
	\includegraphics[width=.244\textwidth]{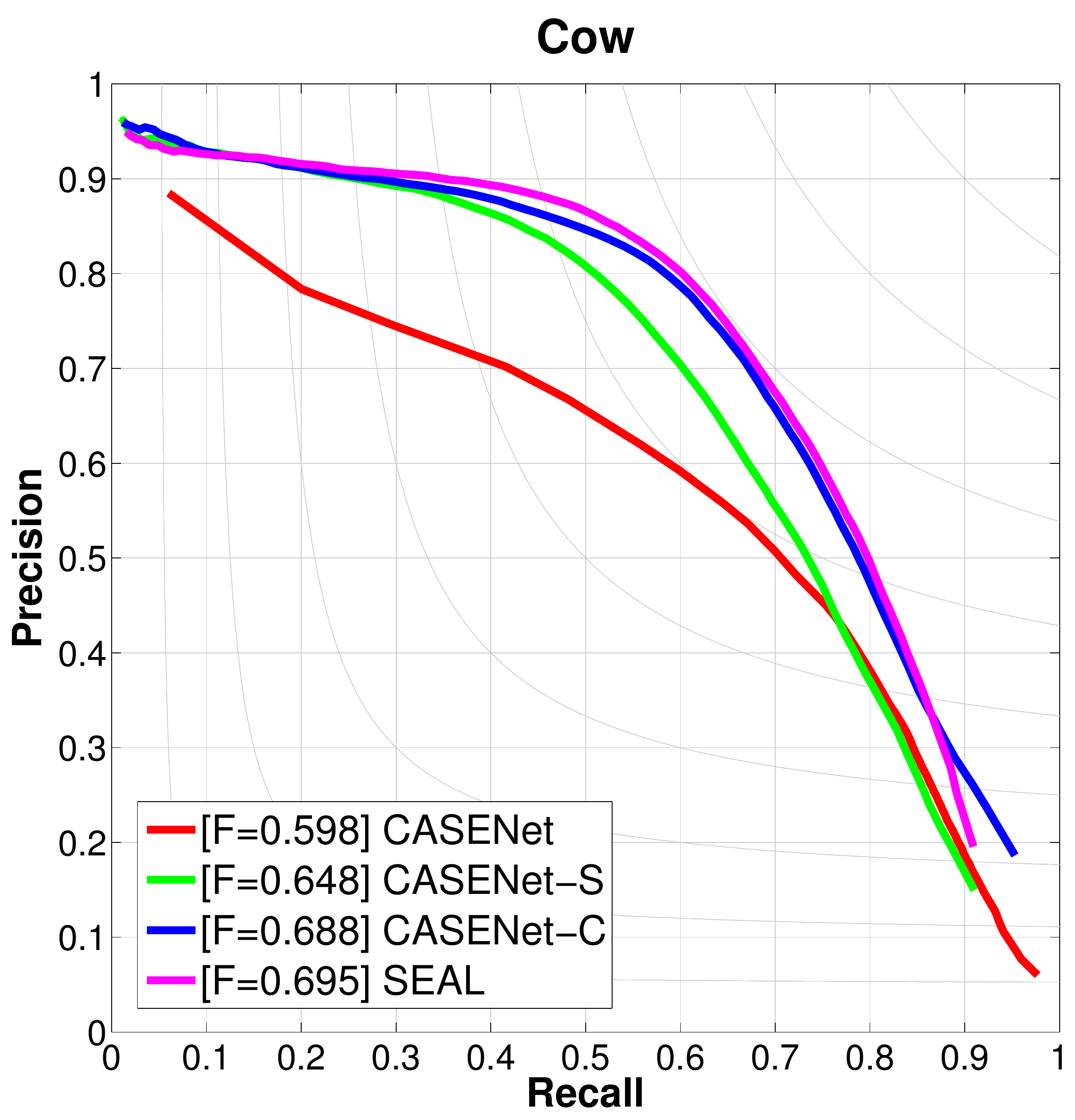}
	\includegraphics[width=.244\textwidth]{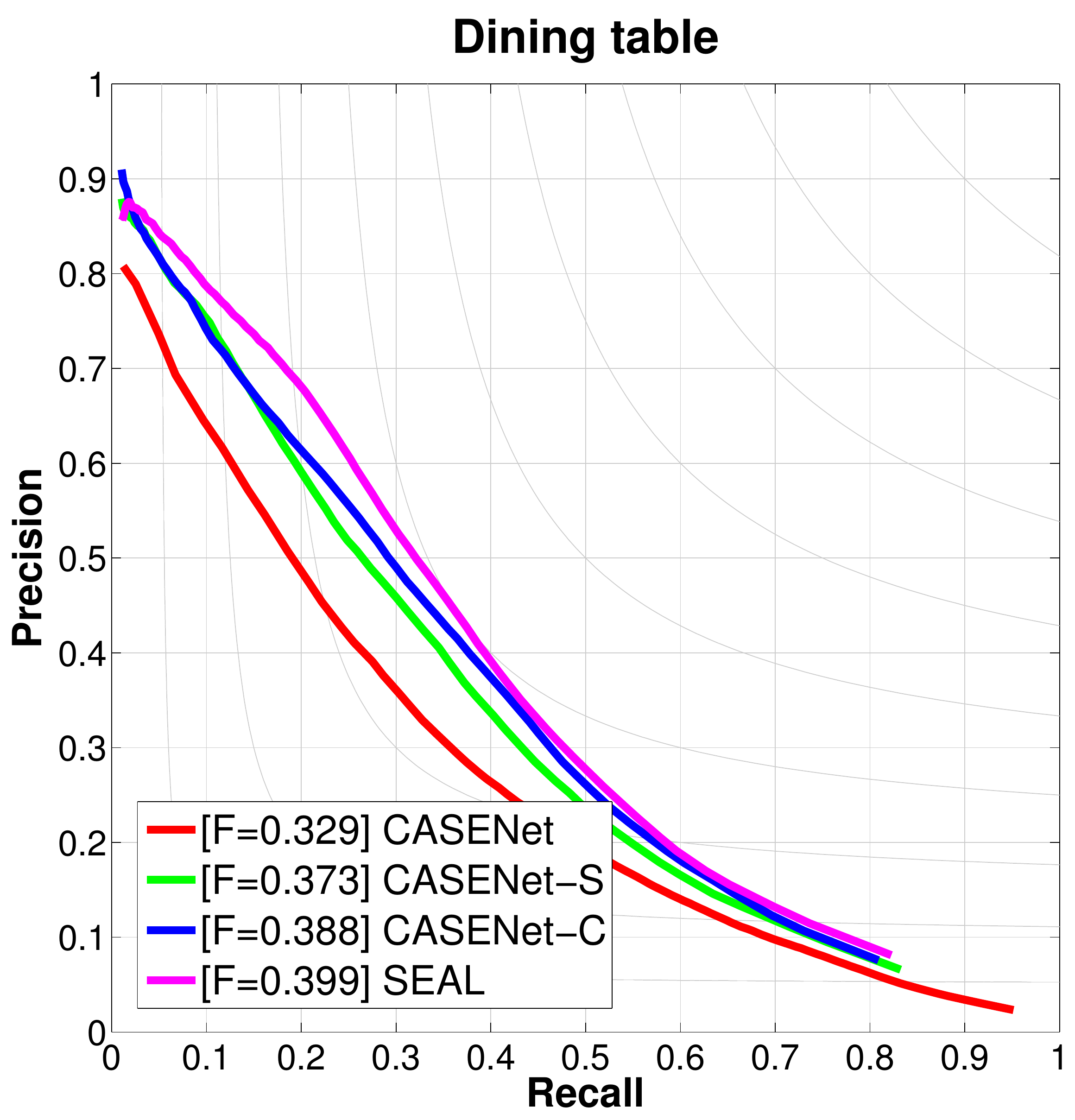}
	\includegraphics[width=.244\textwidth]{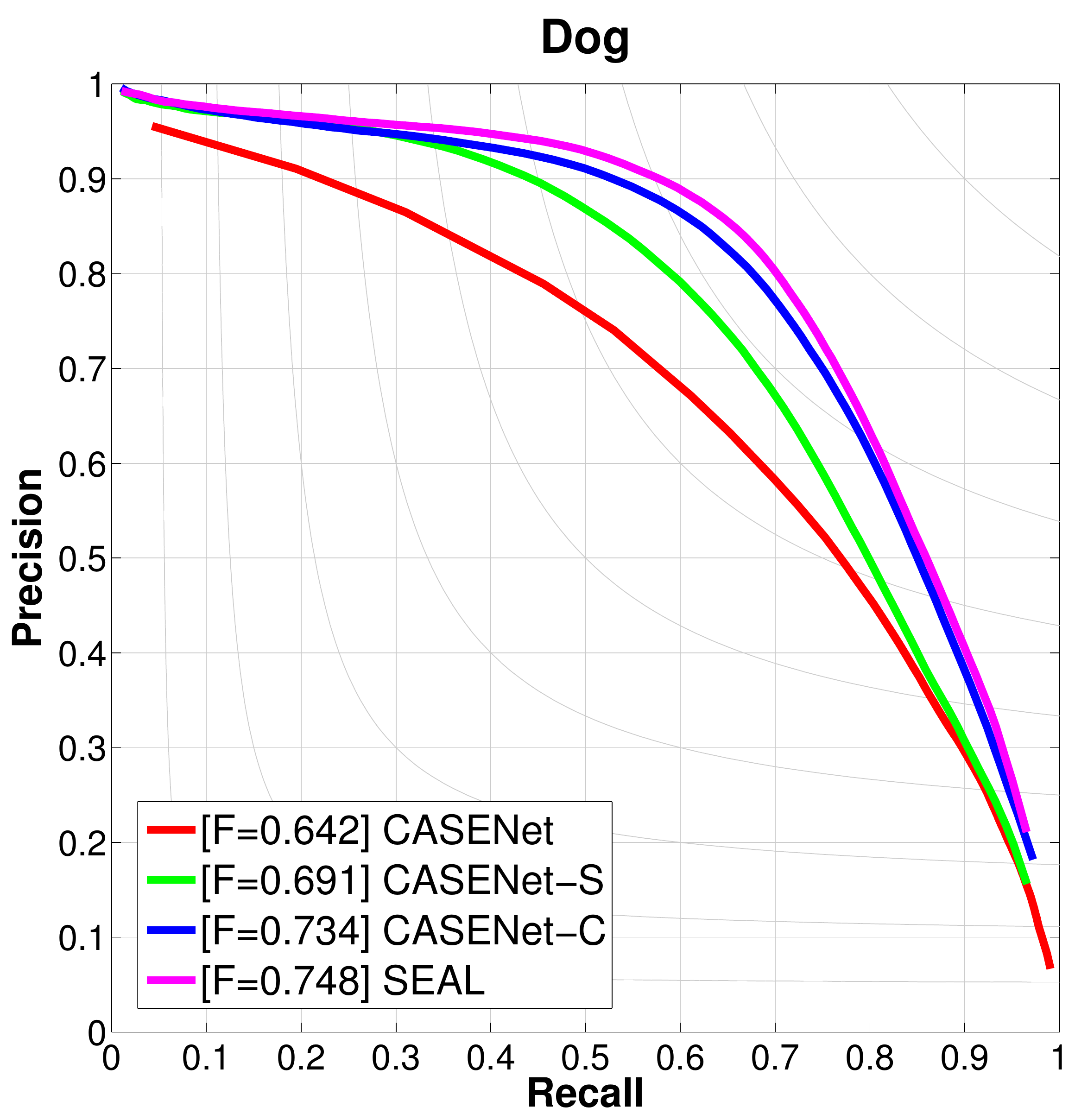}\\
	
	\includegraphics[width=.244\textwidth]{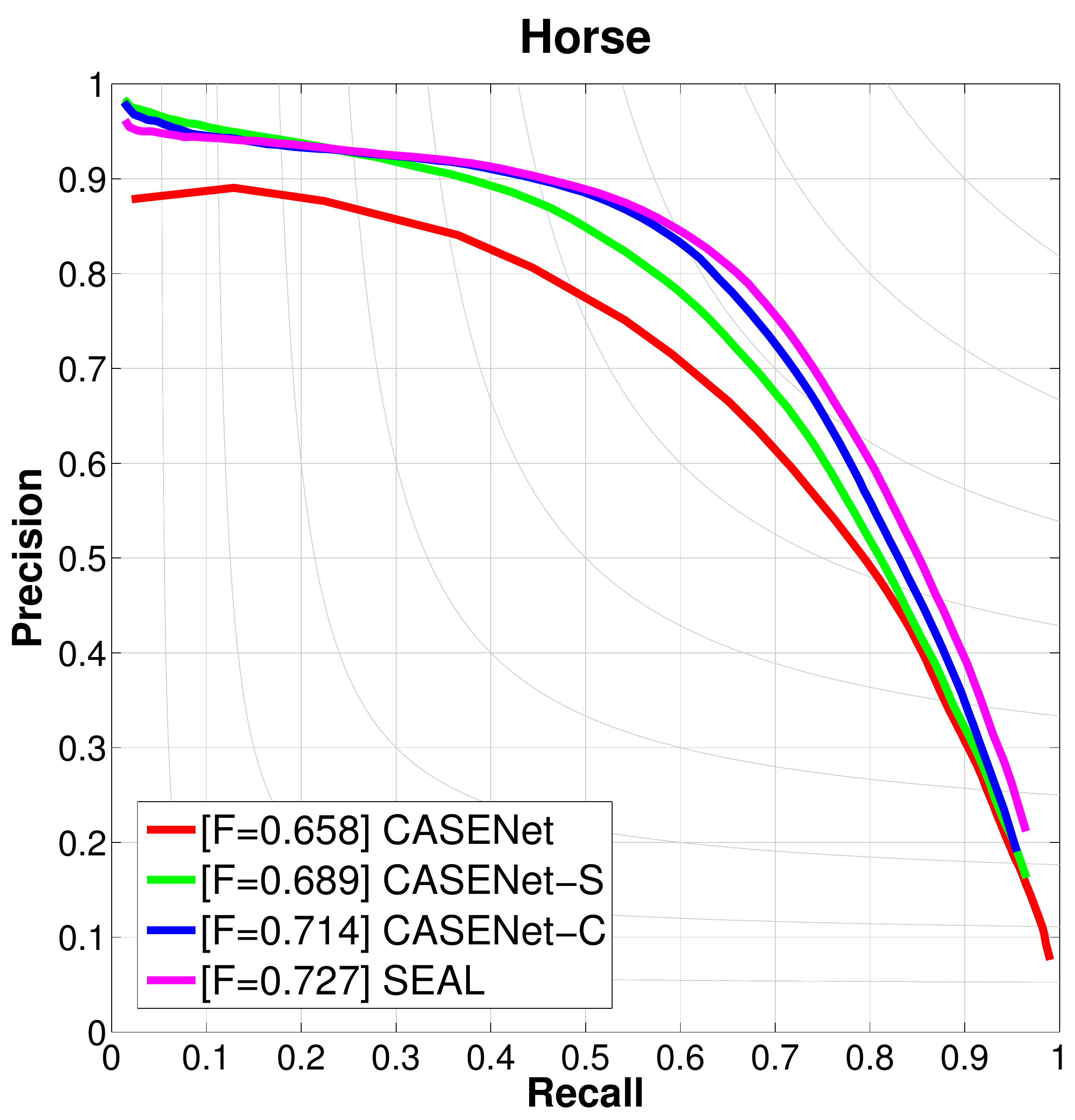}
	\includegraphics[width=.244\textwidth]{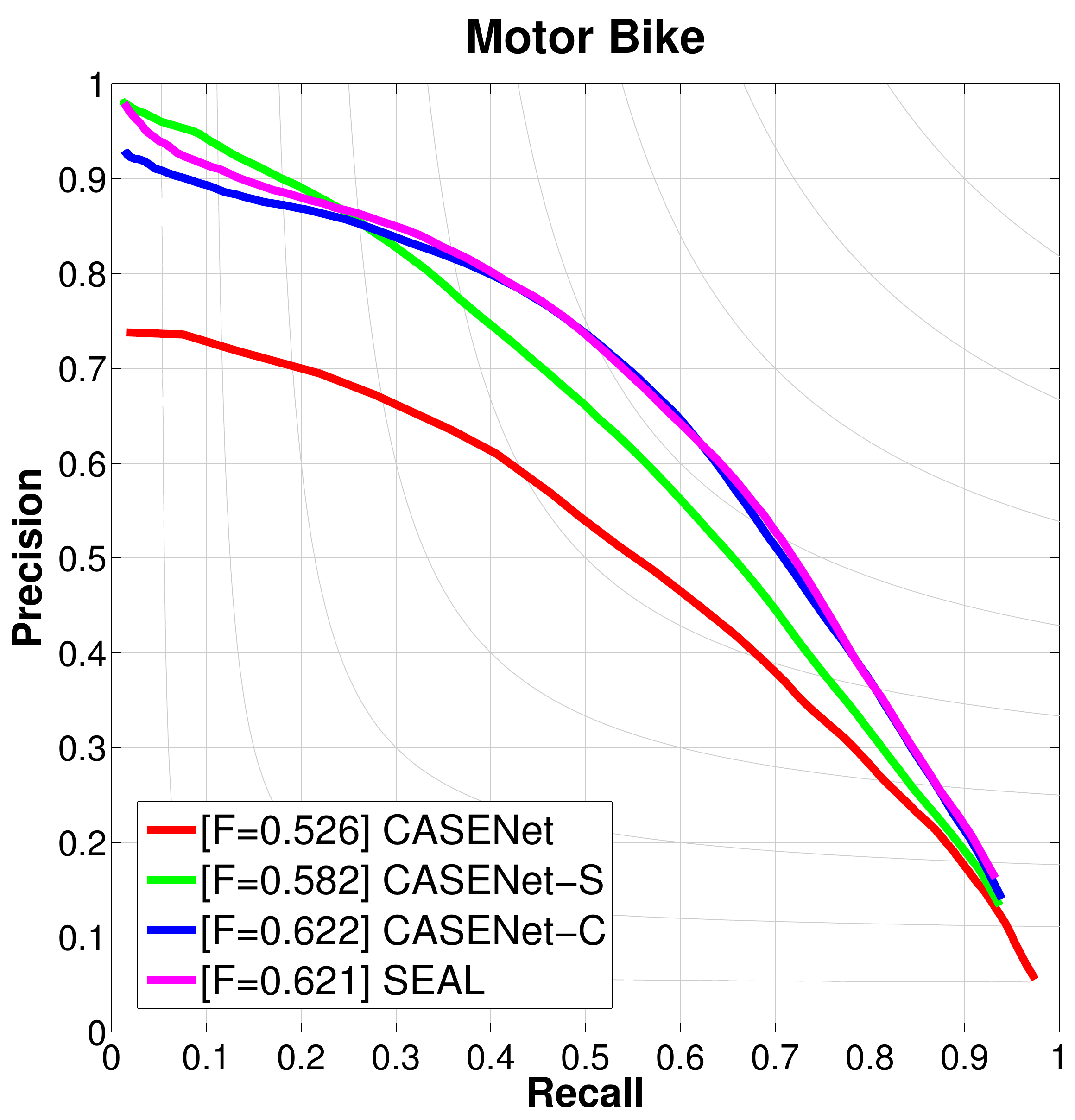}
	\includegraphics[width=.244\textwidth]{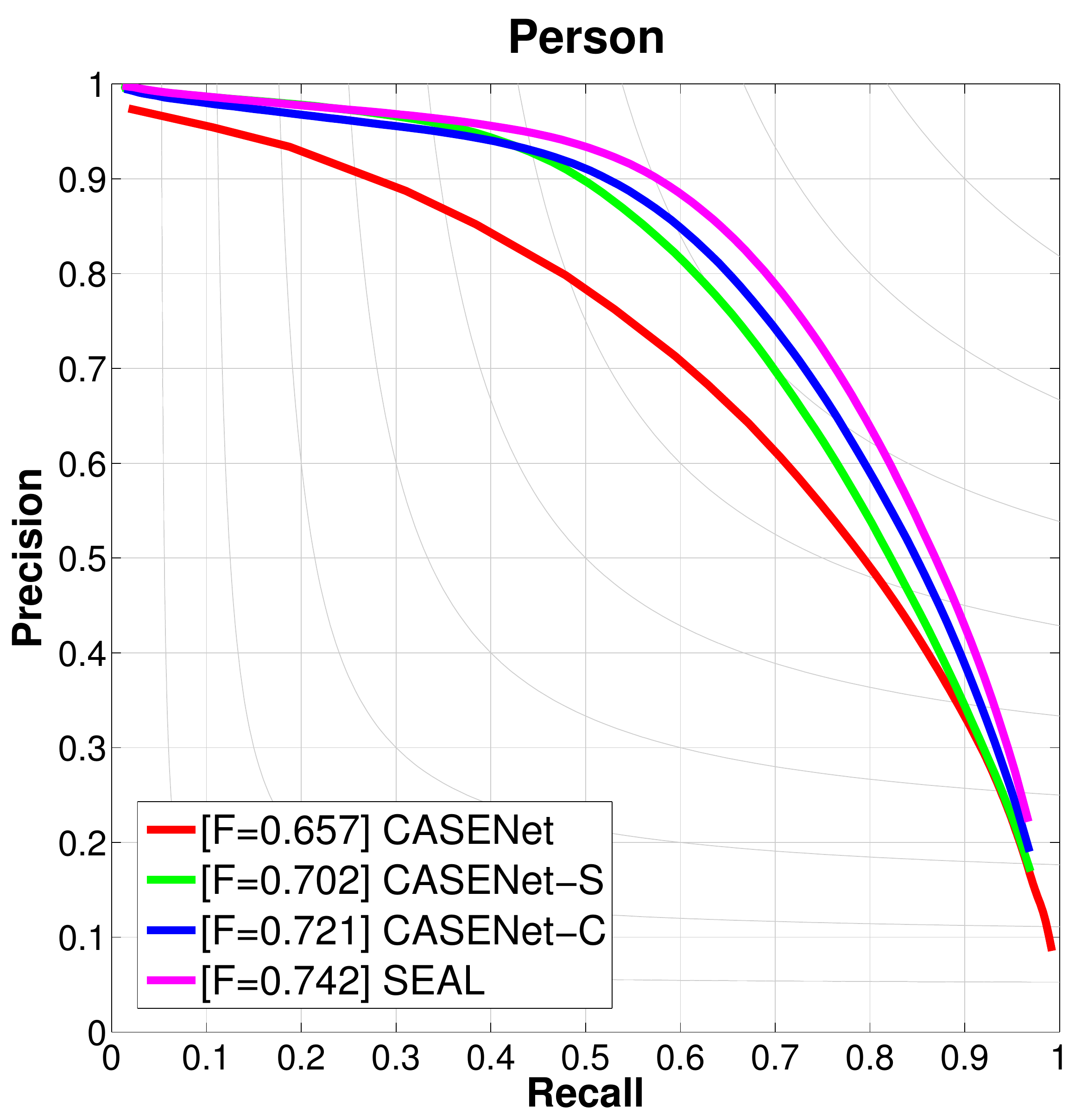}
	\includegraphics[width=.244\textwidth]{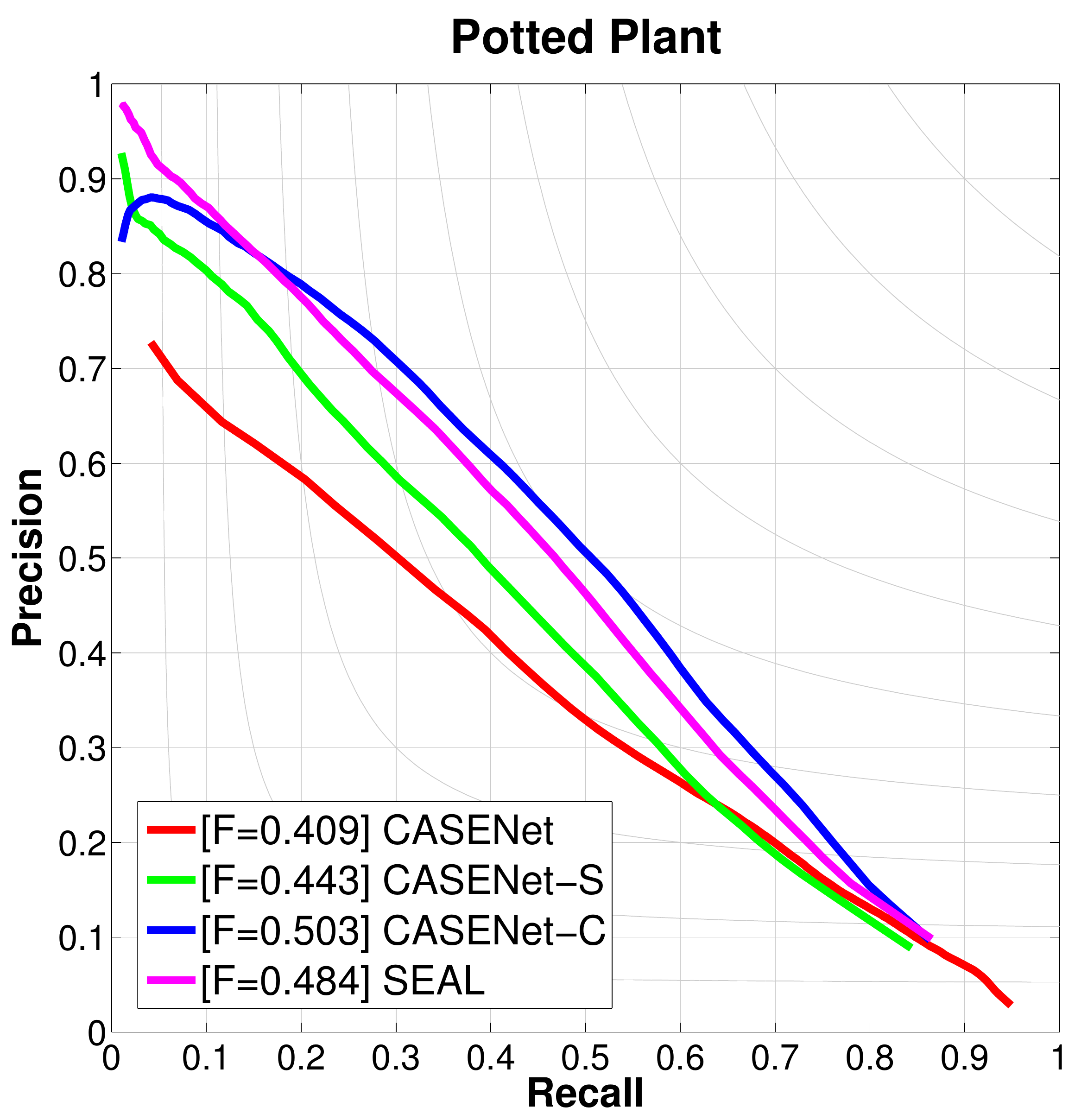}\\
	
	\includegraphics[width=.244\textwidth]{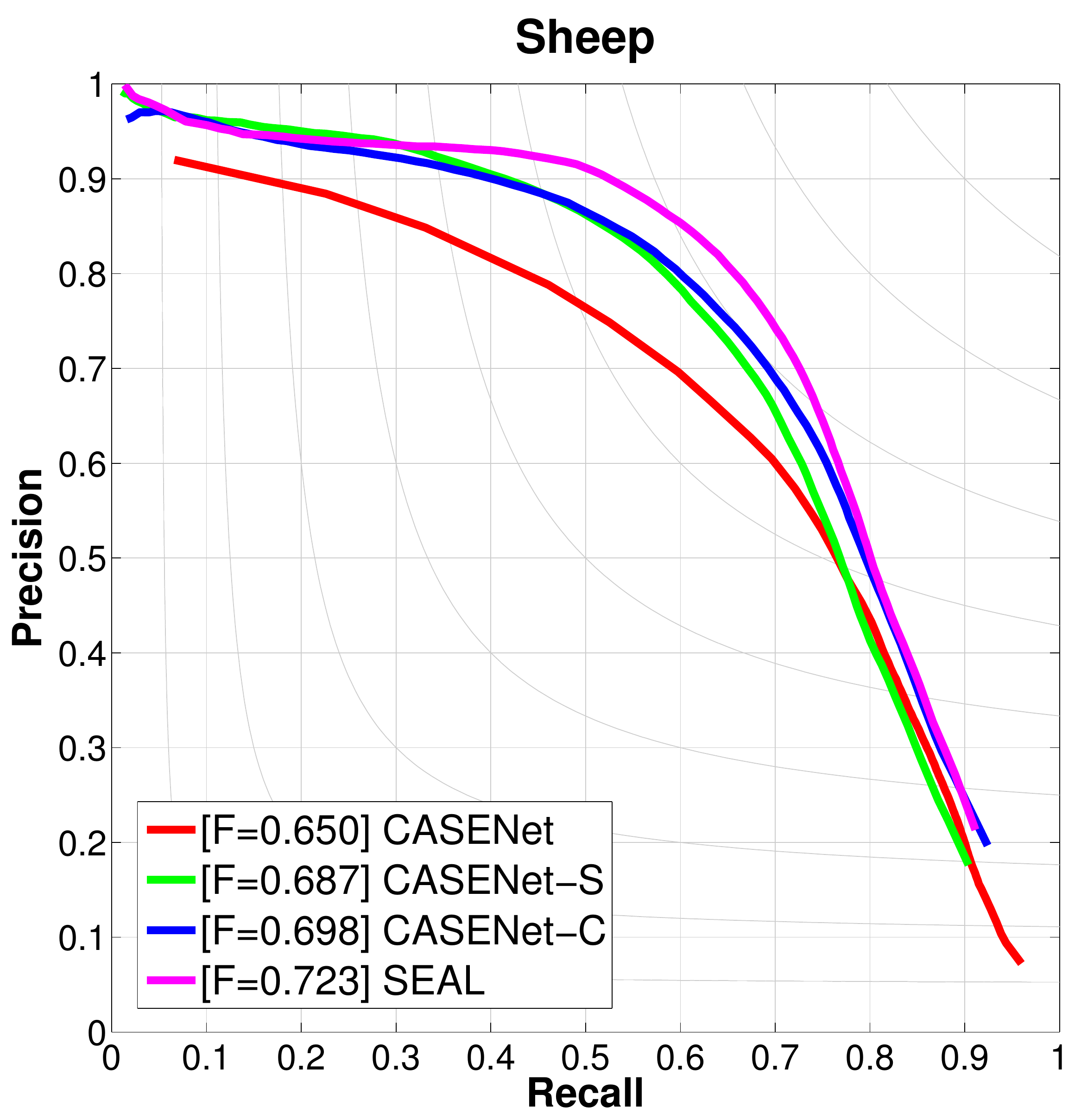}
	\includegraphics[width=.244\textwidth]{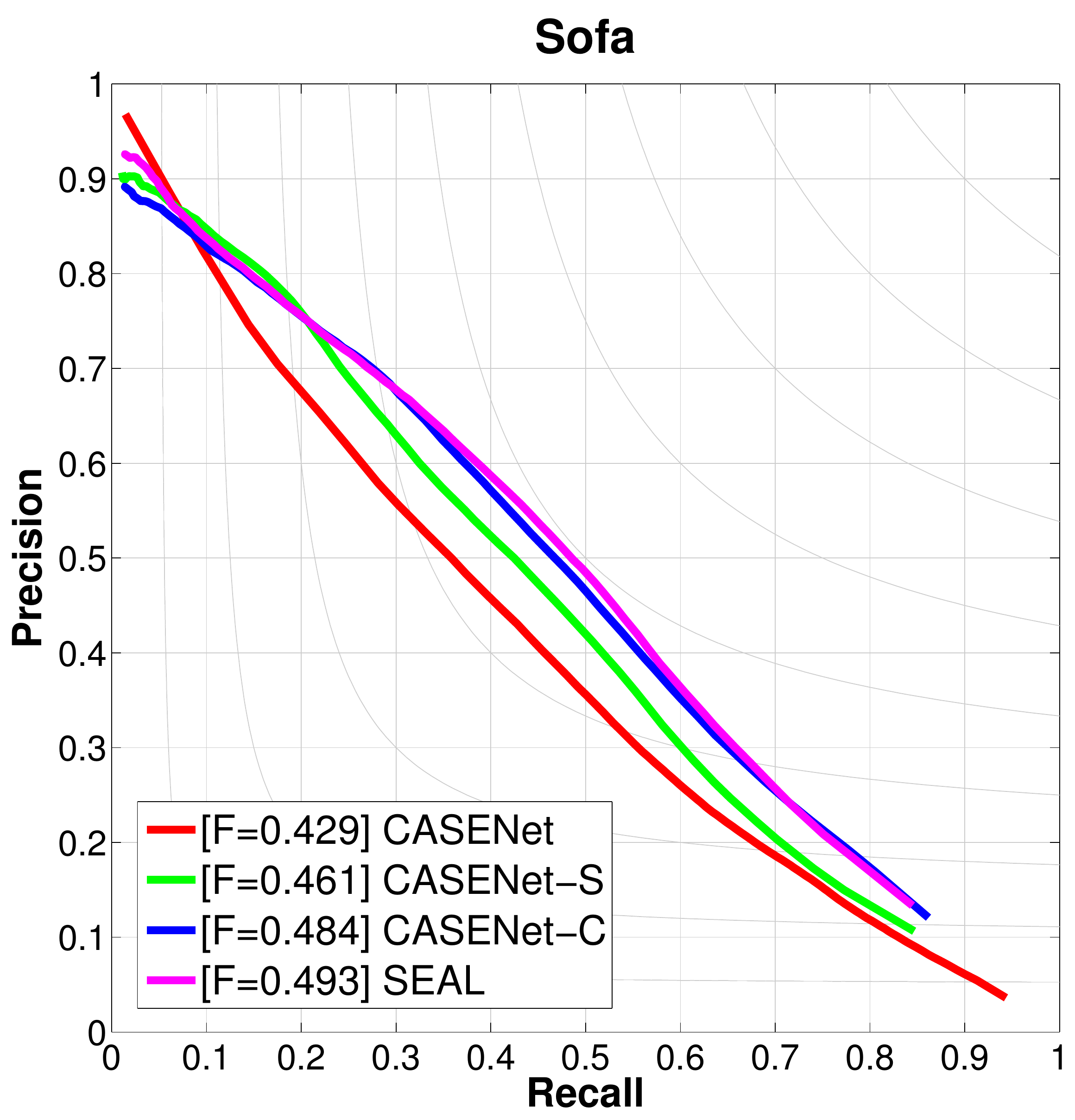}
	\includegraphics[width=.244\textwidth]{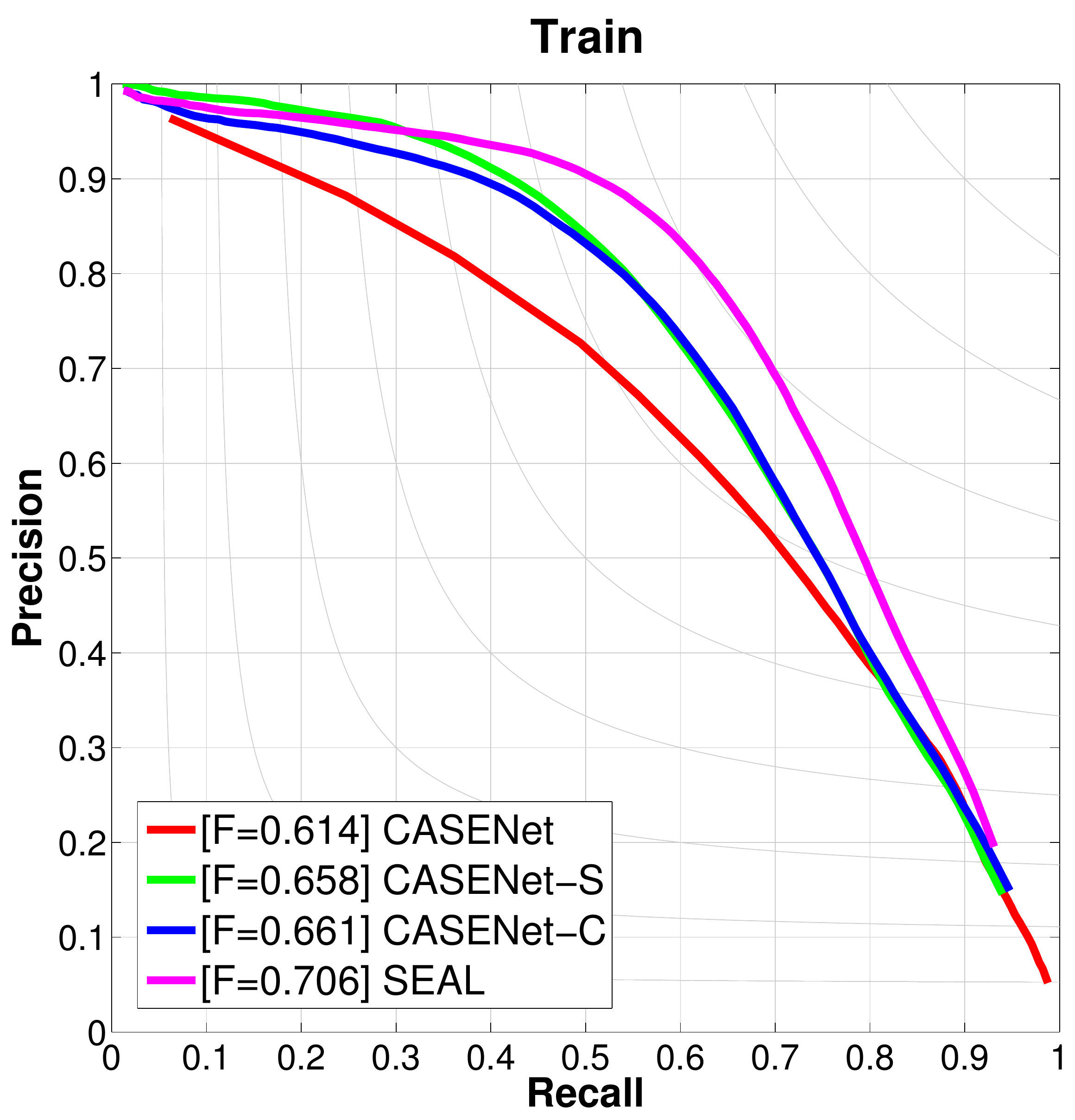}
	\includegraphics[width=.244\textwidth]{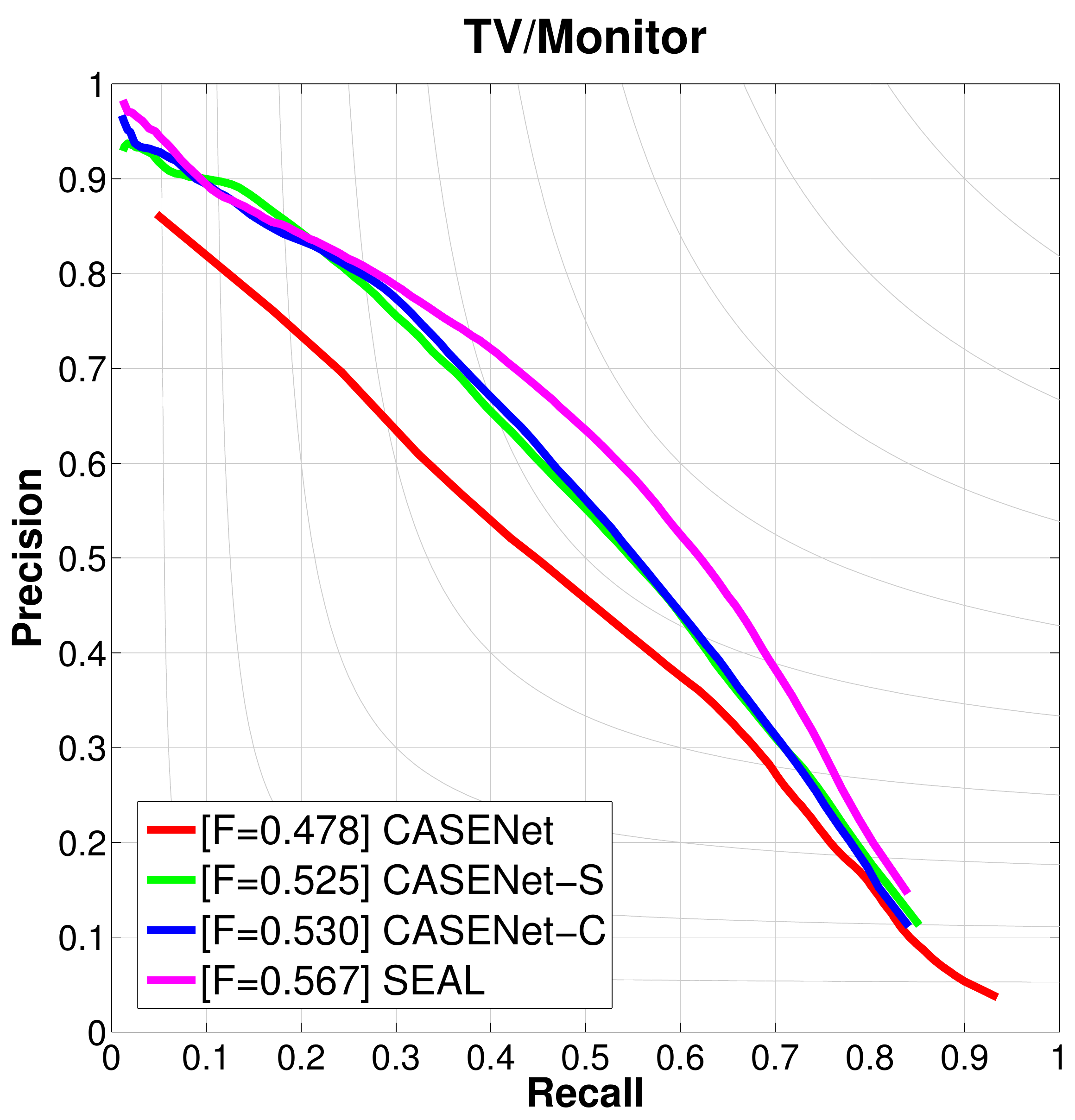}\\
	\caption{Class-wise precision-recall curves of SEAL and comparing baselines on the re-annotated SBD test set under the ``Raw'' setting.}\label{pr_sbd_refine_raw}
\end{figure}

Besides MF and AP scores, we also attach the class-wise precision-recall curves of SEAL and comparing baselines on the SBD test set. Fig.~\ref{pr_sbd_orig_thin} - \ref{pr_sbd_refine_raw} illustrate the precision-recall curves corresponding to the four evaluation settings reported in the main paper (Original/Re-annotated and ``Thin''/ ``Raw'').

\subsection{End quality versus number of Assign steps}\label{sec:assign_steps}
The iterated conditional mode like optimization in edge alignment involves repeated iterations of ``Assign'' and ``Update'' steps. The Assign step minimizes the assignment cost whose neighbor assignment vectors in the pairwise term is conditioned on previous assignment, whereas the Update step updates the neighbor assignment vectors with the newly solved assignment results.

\setlength{\intextsep}{0pt}
\begin{wraptable}[6]{r}{0.3\textwidth}
	\centering
	\caption{Results vs. number of Assign steps.\label{tb:assign}}
	\resizebox{0.3\textwidth}{!}{\begin{tabular}{c|c|c|c|c|c}
			Metric & Assign \# & 15000 & 22000 & 25000 & 30000\\
			\hline \hline
			\multirow{2}{0.3\linewidth}{\centering{MF\\(Thin)}}
			& 1 & 73.18 & 73.21 & 73.26 & 73.25 \\
			& 2 & 73.03 & 73.35 & 73.38 & 73.43 \\
			\hline \hline
			\multirow{2}{0.3\linewidth}{\centering{MF\\(Raw)}}
			& 1 & 70.06 & 70.17 & 70.18 & 70.16 \\
			& 2 & 69.79 & 69.98 & 69.98 & 70.00 \\
	\end{tabular}}
\end{wraptable}
The number of Assign steps in our experiments are by default set to two (one Update step). To study the end quality of edge learning at different iterations, we fix other SEAL parameters and present results under different numbers of Assign step and network training iterations in Table \ref{tb:assign}. The results are evaluated on SBD validation set with 0.02 tolerance. We observe that two Assign steps tends to improve the quality in ``Thin'' setting. This motivates us to finalize the number as two in our experiments. However, the difference between having one or two Assign steps is marginal. In practice, one may choose to have one Assign step for the benefit of training speed up.

\subsection{Visualization of ground truth refinement}
\begin{figure}[tbh]
	\centering
	\resizebox{1.0\textwidth}{!}{
		\begin{tabular}{@{}cccccccccc@{}}
			\cellcolor{sbd_color_1}\textcolor{white}{~~~~aero~~~~} &
			\cellcolor{sbd_color_2}\textcolor{white}{~~~~bike~~~~} &
			\cellcolor{sbd_color_3}\textcolor{white}{~~~~bird~~~~} &
			\cellcolor{sbd_color_4}\textcolor{white}{~~~~boat~~~~} &
			\cellcolor{sbd_color_5}\textcolor{white}{~~~~bottle~~~~} &
			\cellcolor{sbd_color_6}\textcolor{white}{~~~~bus~~~~} &
			\cellcolor{sbd_color_7}\textcolor{white}{~~~~c~ar~~~~} &
			\cellcolor{sbd_color_8}\textcolor{white}{~~~~cat~~~~} &
			\cellcolor{sbd_color_9}\textcolor{white}{~~~~chair~~~~} &
			\cellcolor{sbd_color_10}\textcolor{white}{~~~~cow~~~~} \\
			\cellcolor{sbd_color_11}\textcolor{white}{~table~} &
			\cellcolor{sbd_color_12}\textcolor{white}{~dog~} &
			\cellcolor{sbd_color_13}\textcolor{white}{~horse~} &
			\cellcolor{sbd_color_14}\textcolor{white}{~mbike~} &
			\cellcolor{sbd_color_15}\textcolor{white}{~person~} &
			\cellcolor{sbd_color_16}\textcolor{white}{~plant~} &
			\cellcolor{sbd_color_17}\textcolor{white}{~sheep~} &
			\cellcolor{sbd_color_18}\textcolor{white}{~sofa~} &
			\cellcolor{sbd_color_19}\textcolor{white}{~train~} &
			\cellcolor{sbd_color_20}\textcolor{white}{~tv~}
		\end{tabular}
	}\\
	\vspace{0.05cm}
	\includegraphics[width=0.244\textwidth]{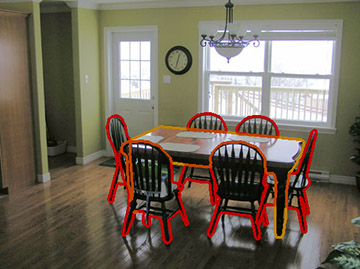}
	\includegraphics[width=0.244\textwidth]{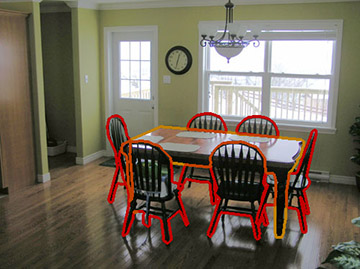}
	\includegraphics[width=0.244\textwidth]{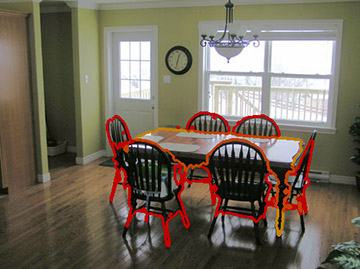}
	\includegraphics[width=0.244\textwidth]{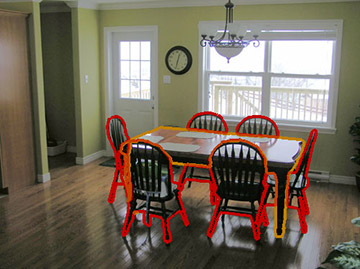}\\
	\vspace{0.05cm}
	\includegraphics[width=0.244\textwidth]{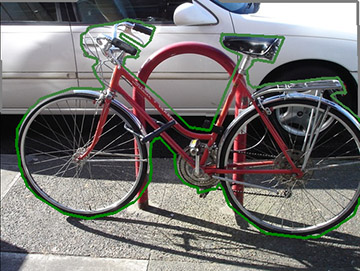}
	\includegraphics[width=0.244\textwidth]{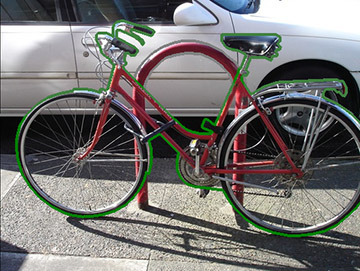}
	\includegraphics[width=0.244\textwidth]{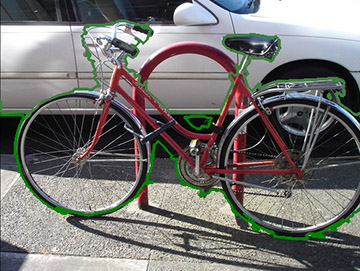}
	\includegraphics[width=0.244\textwidth]{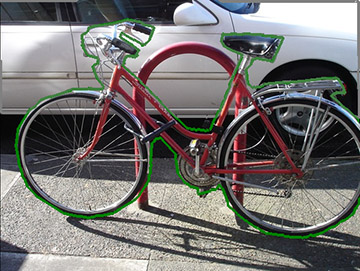}\\
	\vspace{0.05cm}
	\includegraphics[width=0.244\textwidth]{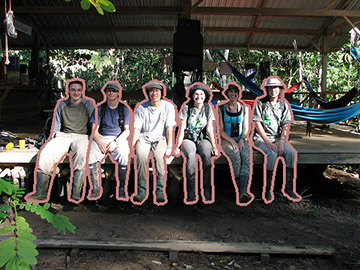}
	\includegraphics[width=0.244\textwidth]{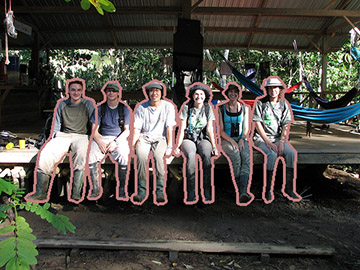}
	\includegraphics[width=0.244\textwidth]{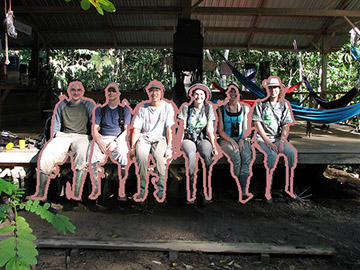}
	\includegraphics[width=0.244\textwidth]{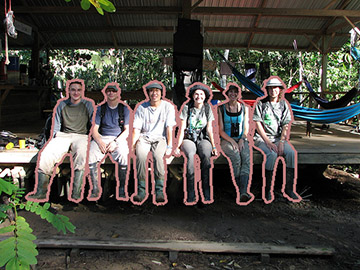}\\
	\vspace{0.05cm}
	\includegraphics[width=0.244\textwidth]{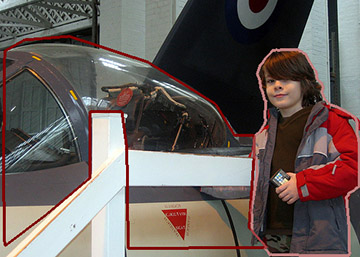}
	\includegraphics[width=0.244\textwidth]{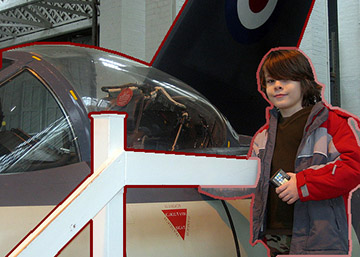}
	\includegraphics[width=0.244\textwidth]{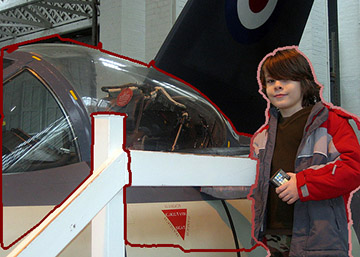}
	\includegraphics[width=0.244\textwidth]{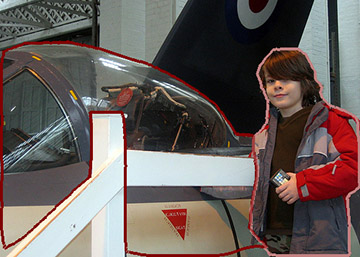}\\
	\vspace{0.05cm}
	\includegraphics[width=0.244\textwidth]{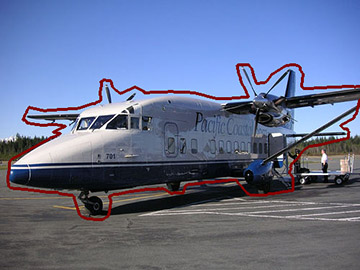}
	\includegraphics[width=0.244\textwidth]{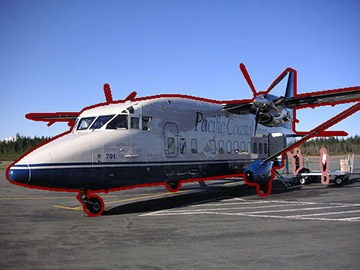}
	\includegraphics[width=0.244\textwidth]{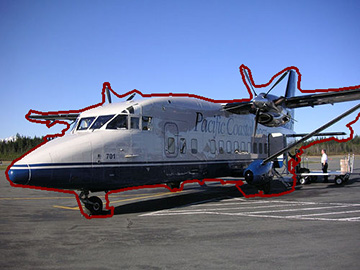}
	\includegraphics[width=0.244\textwidth]{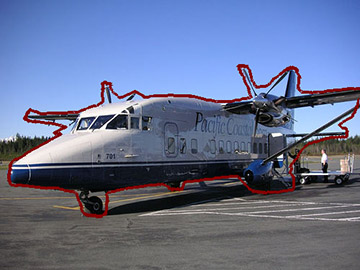}\\
	\vspace{0.05cm}
	\includegraphics[width=0.244\textwidth]{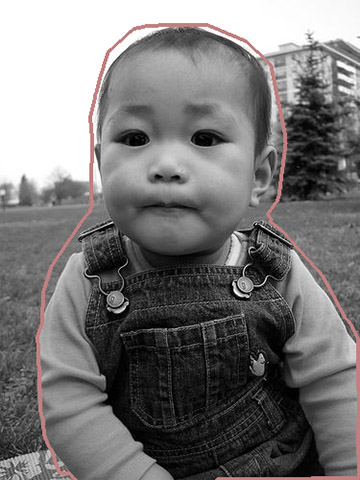}
	\includegraphics[width=0.244\textwidth]{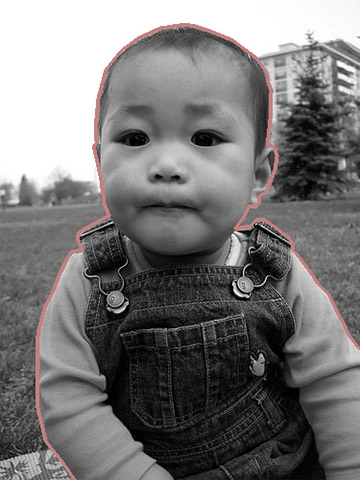}
	\includegraphics[width=0.244\textwidth]{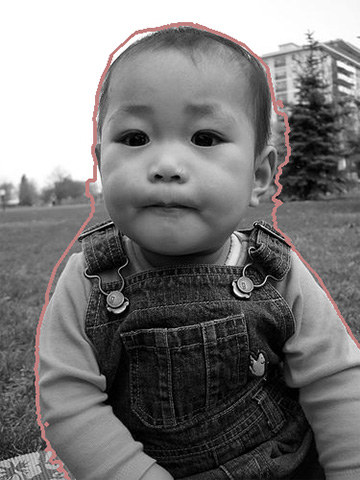}
	\includegraphics[width=0.244\textwidth]{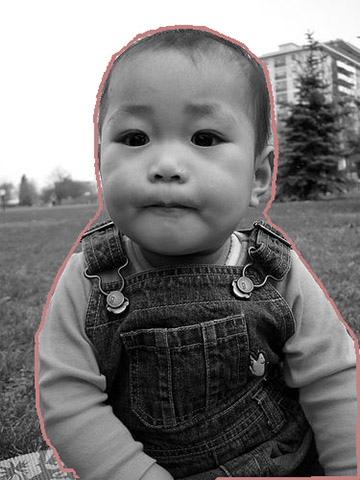}\\
	\caption{Examples of annotations and aligned edge labels learned by different methods on SBD test set images. From left to right: original ground truth, re-annotated high quality ground truth, edge labels aligned via dense CRF, edge labels aligned via SEAL.}\label{align_vis}
\end{figure}

We visualize some results of SEAL aligned edges and those of comparing methods in Fig.~\ref{align_vis}, in addition to the reported quantitative results in the main paper. We observe that CRF tends to smooth out thin structures such as human/chair legs, while these delicate structures are important in edge learning. In addition, CRF sometimes also introduces noisy boundaries because of the limited representation power of low-level features. This partly explains why dense-CRF overall does not even match the quality of the original labels. 

\section{Additional results on Cityscapes}\label{add_city}
\begin{figure}[tbh]
	\raggedright
	\includegraphics[width=.244\textwidth]{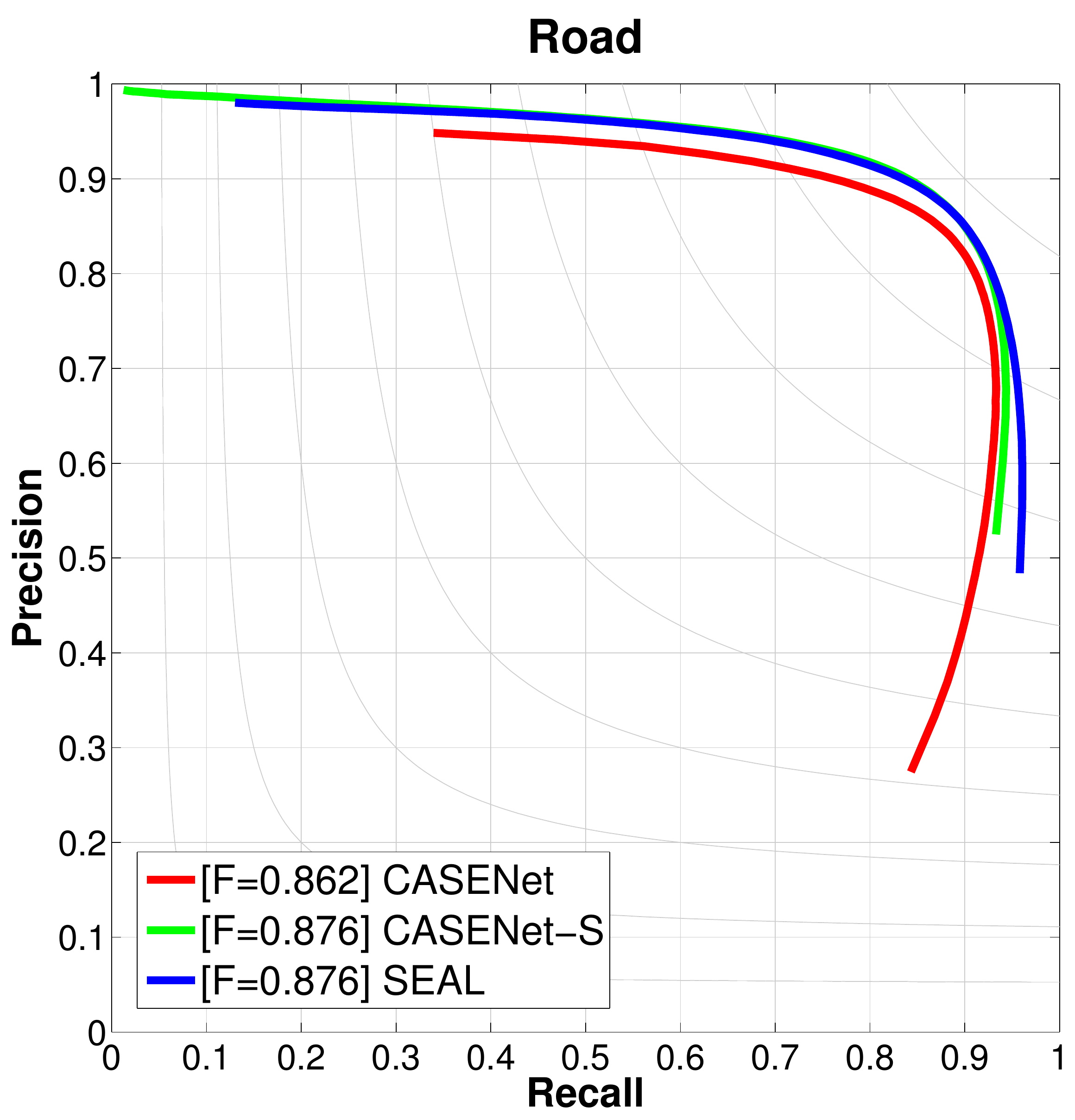}
	\includegraphics[width=.244\textwidth]{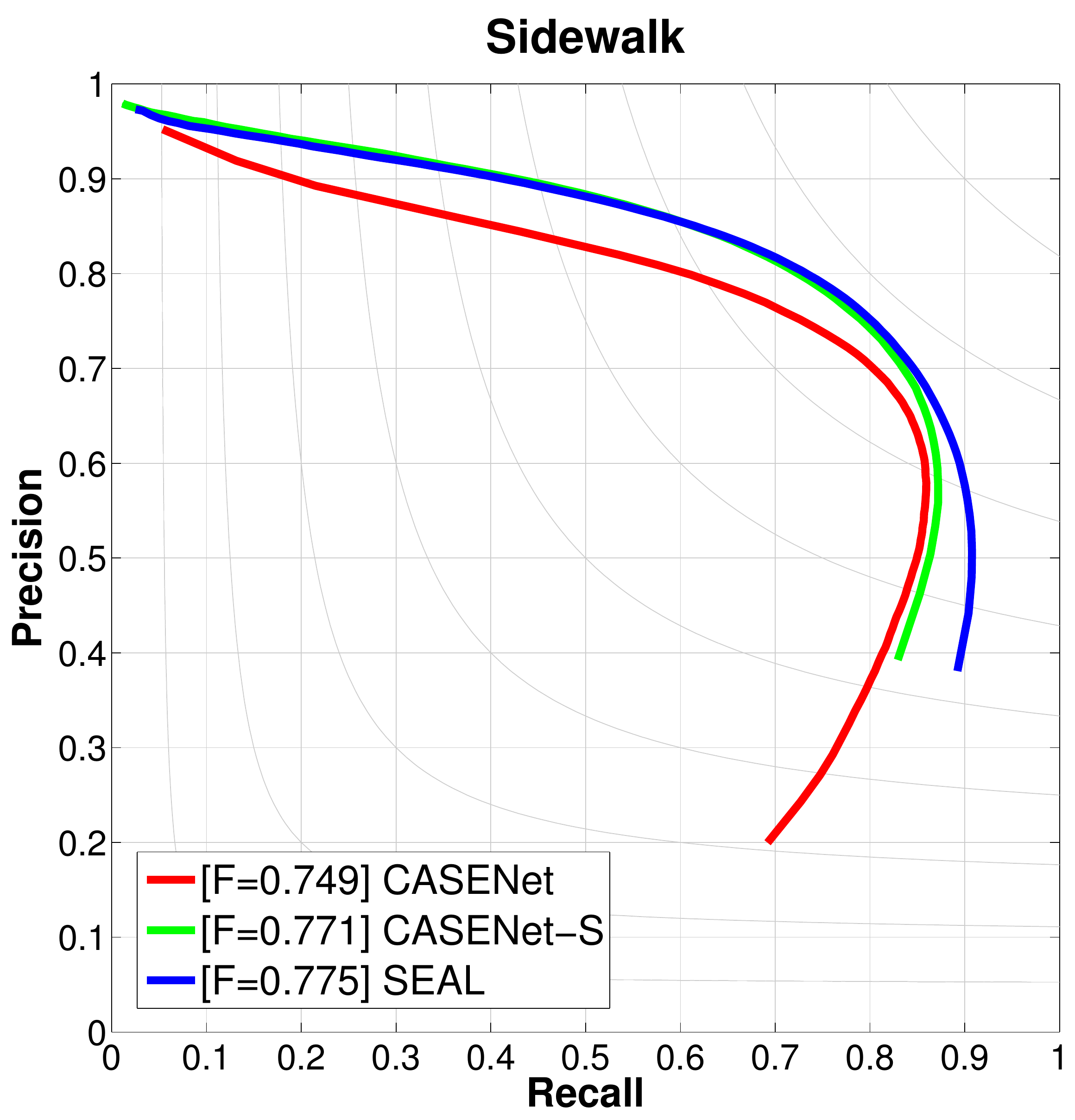}
	\includegraphics[width=.244\textwidth]{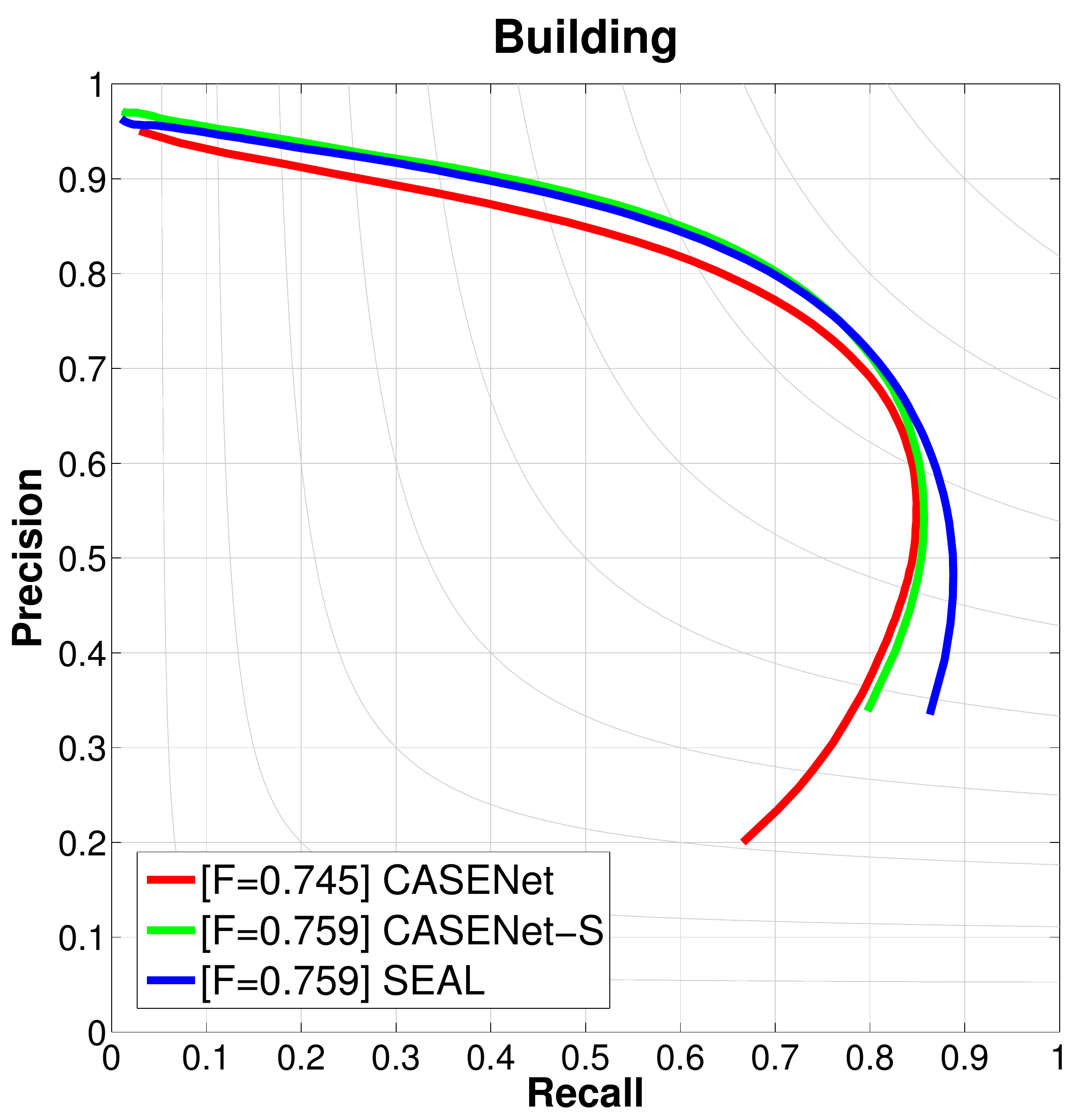}
	\includegraphics[width=.244\textwidth]{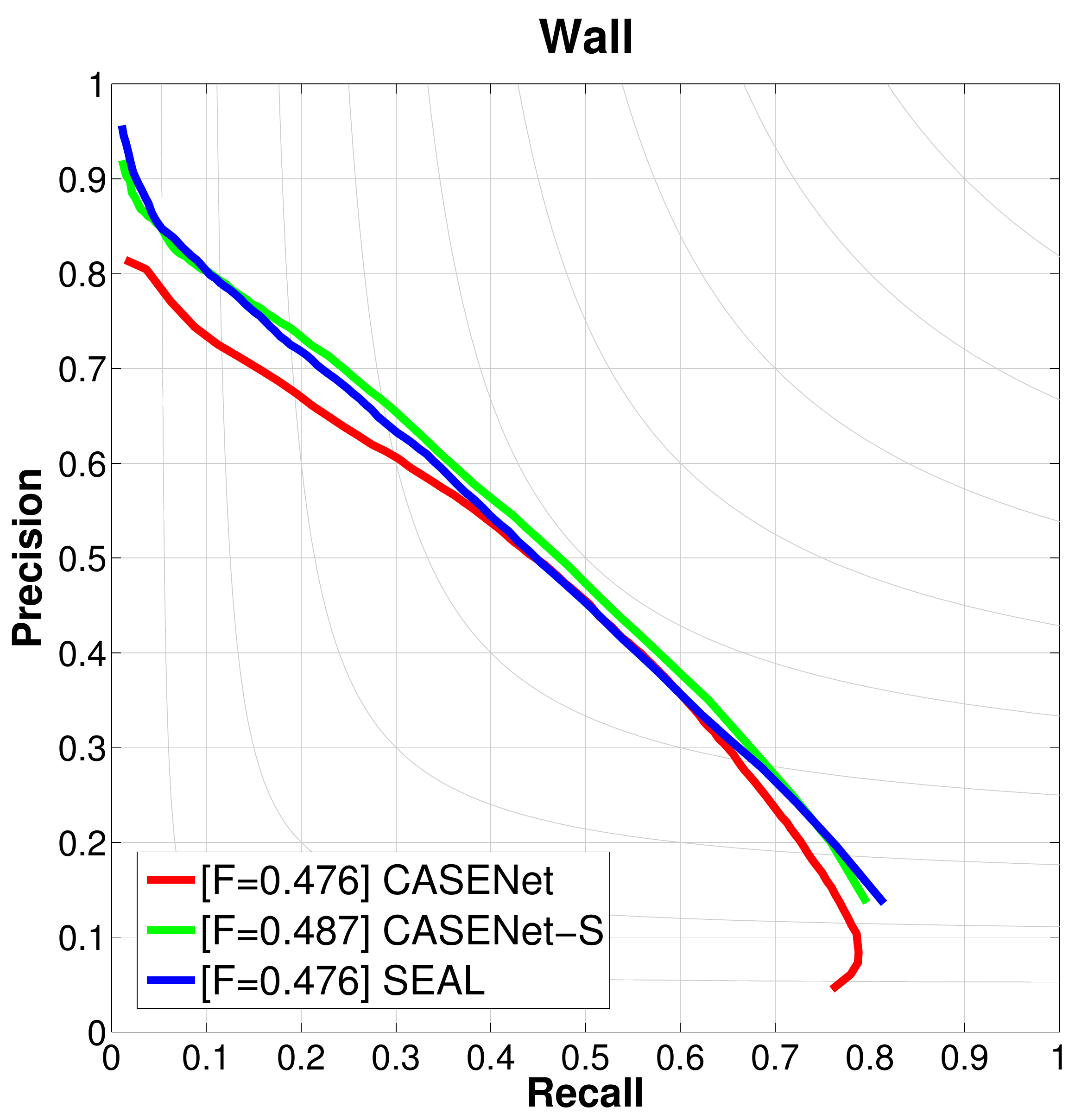}\\
	
	\includegraphics[width=.244\textwidth]{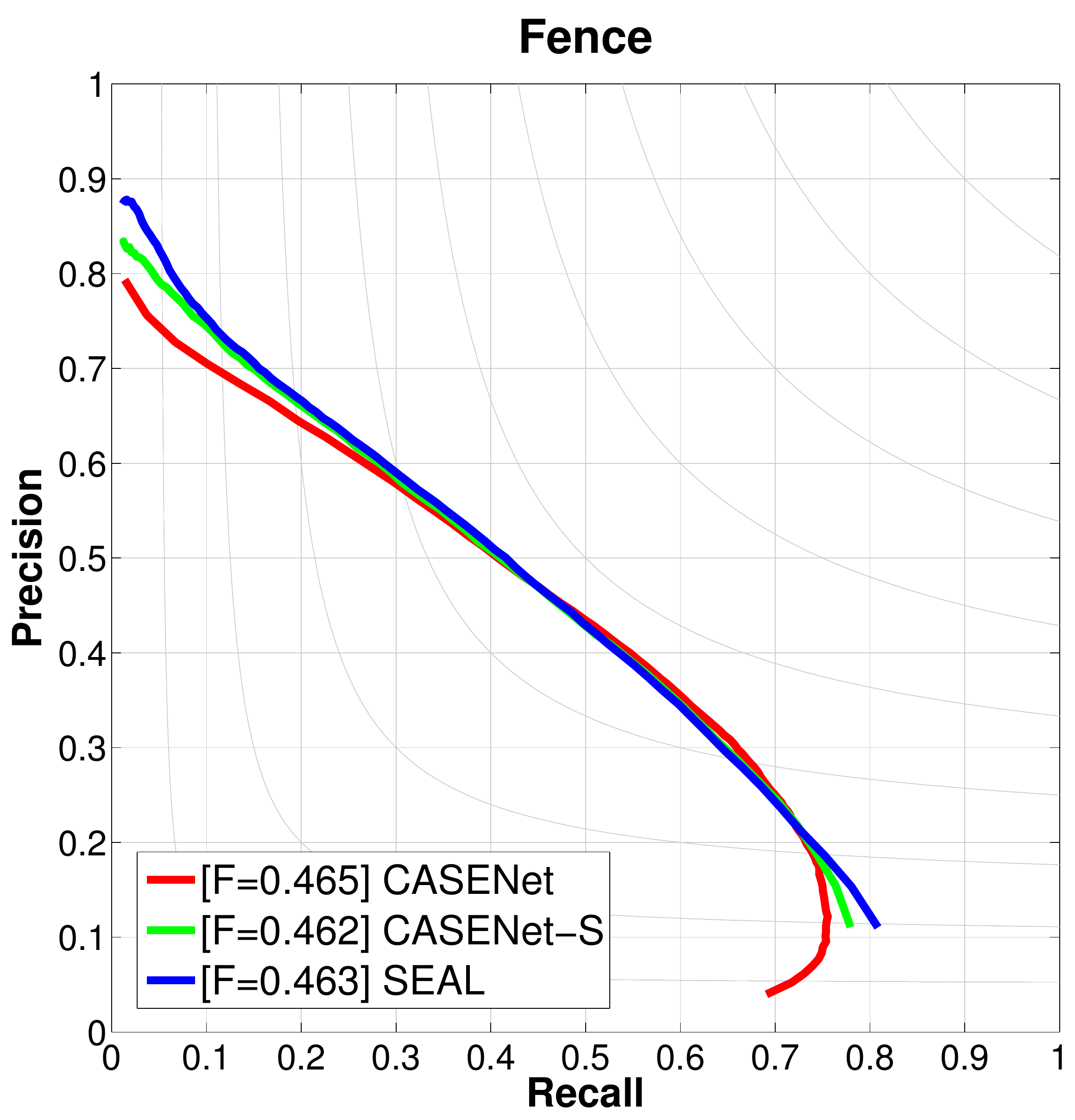}
	\includegraphics[width=.244\textwidth]{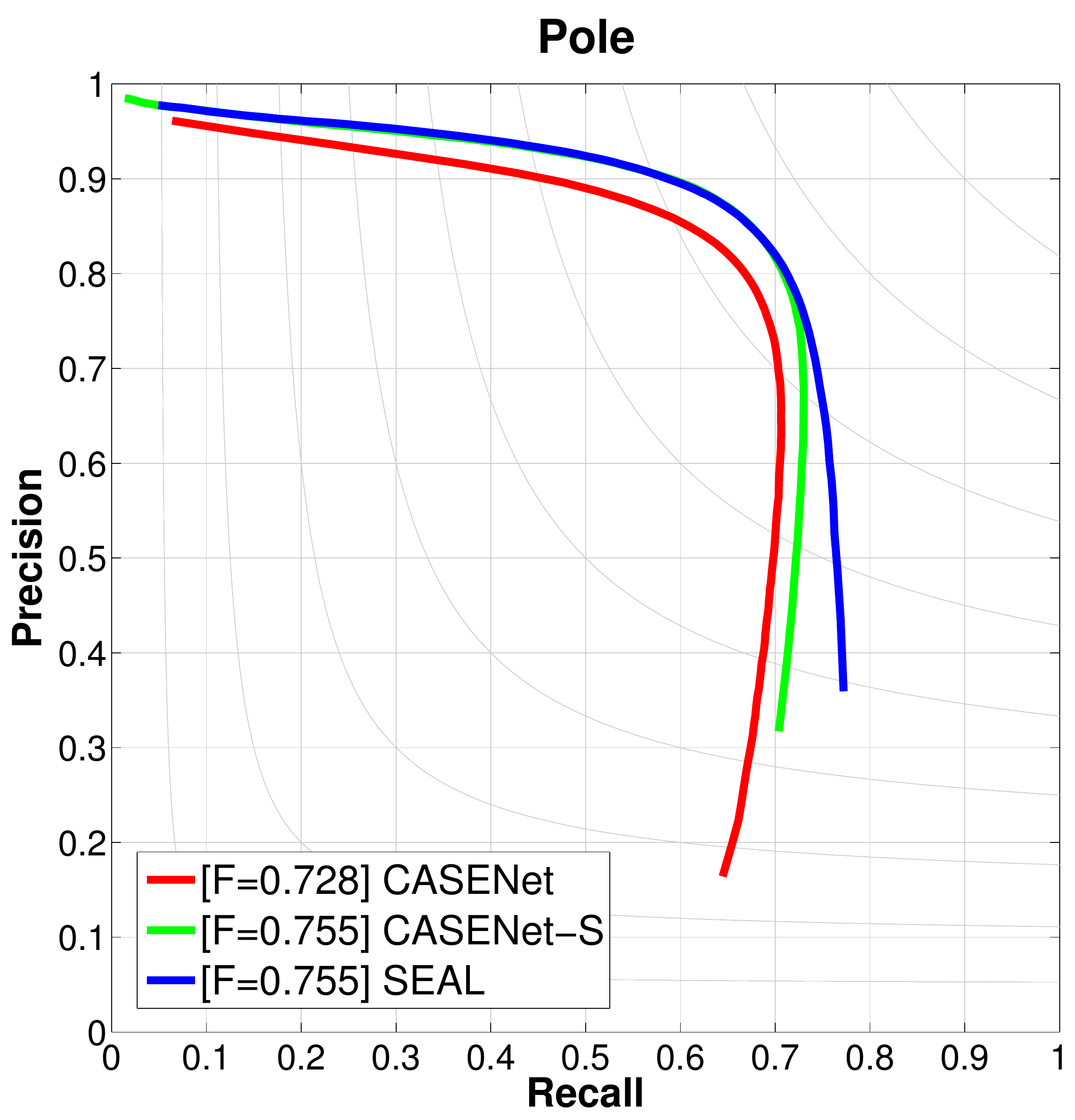}
	\includegraphics[width=.244\textwidth]{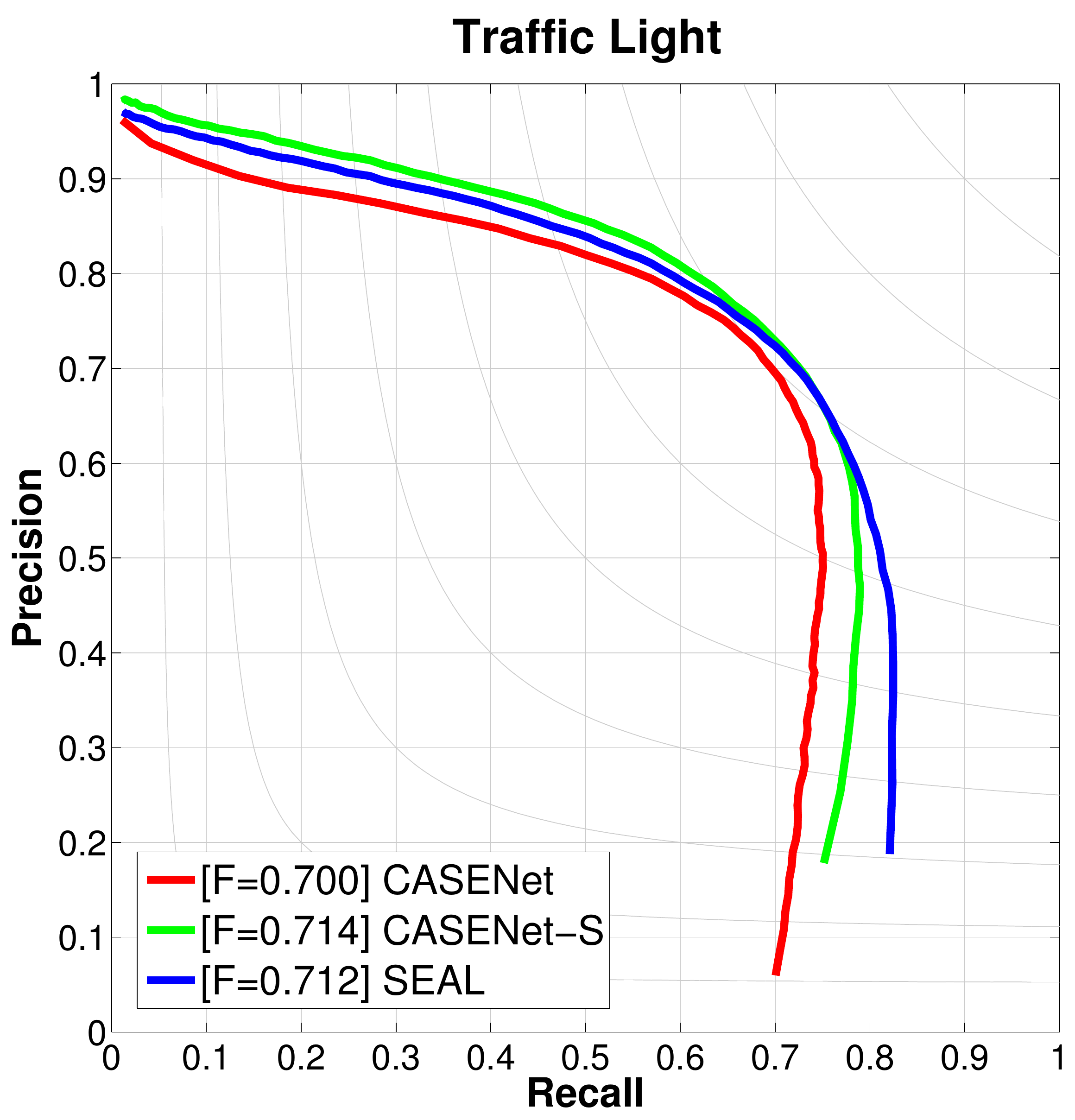}
	\includegraphics[width=.244\textwidth]{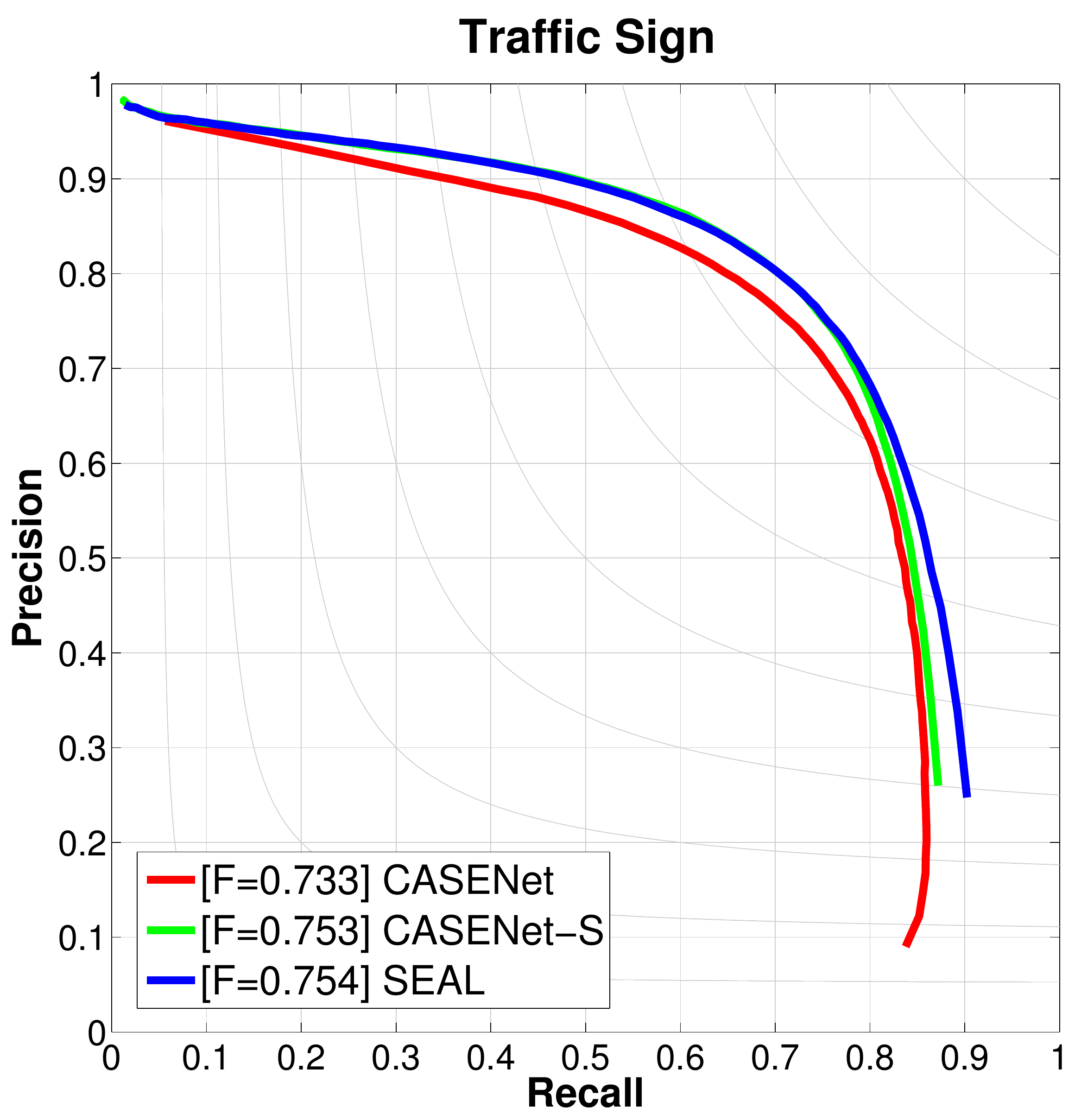}\\
	
	\includegraphics[width=.244\textwidth]{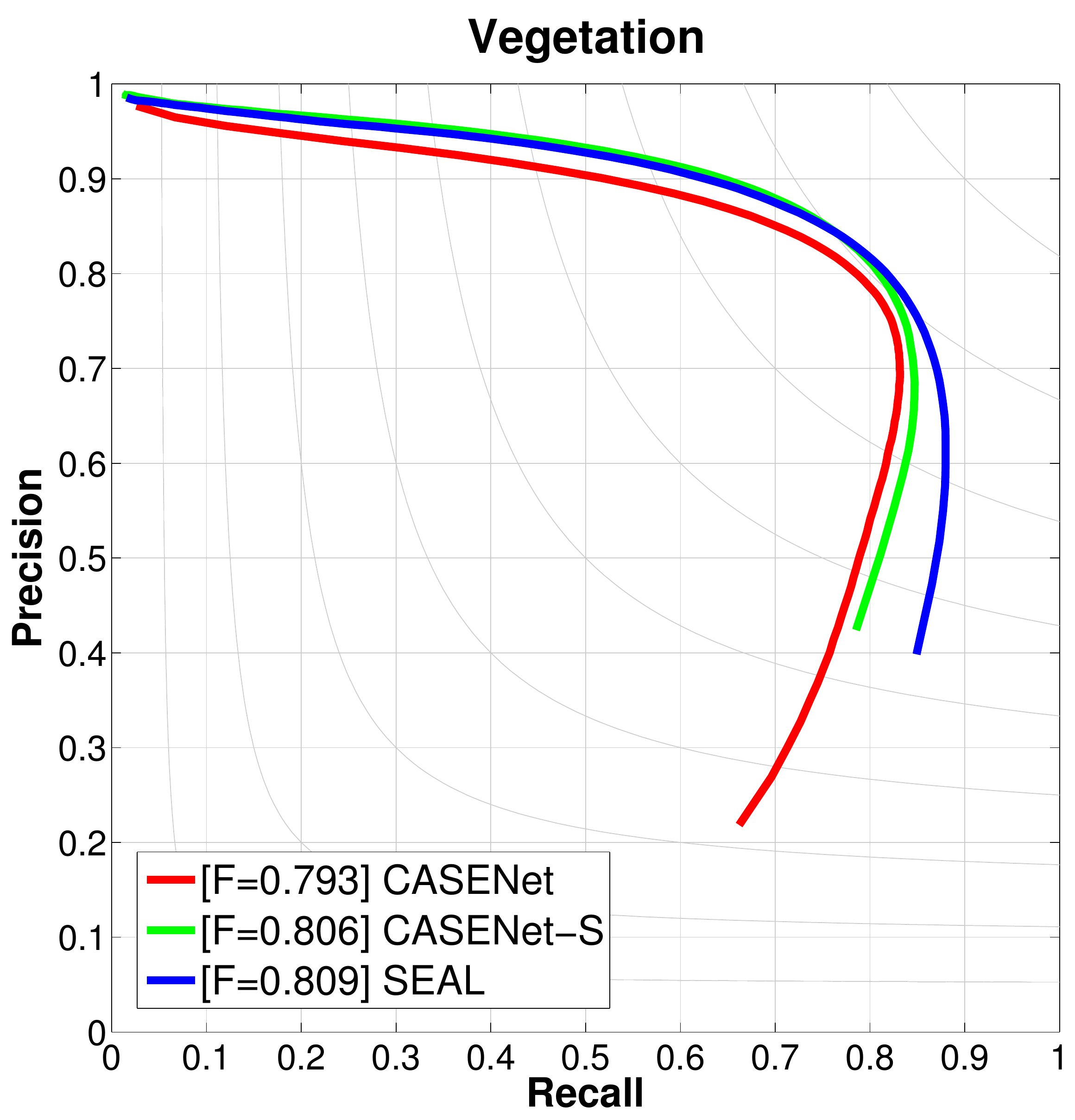}
	\includegraphics[width=.244\textwidth]{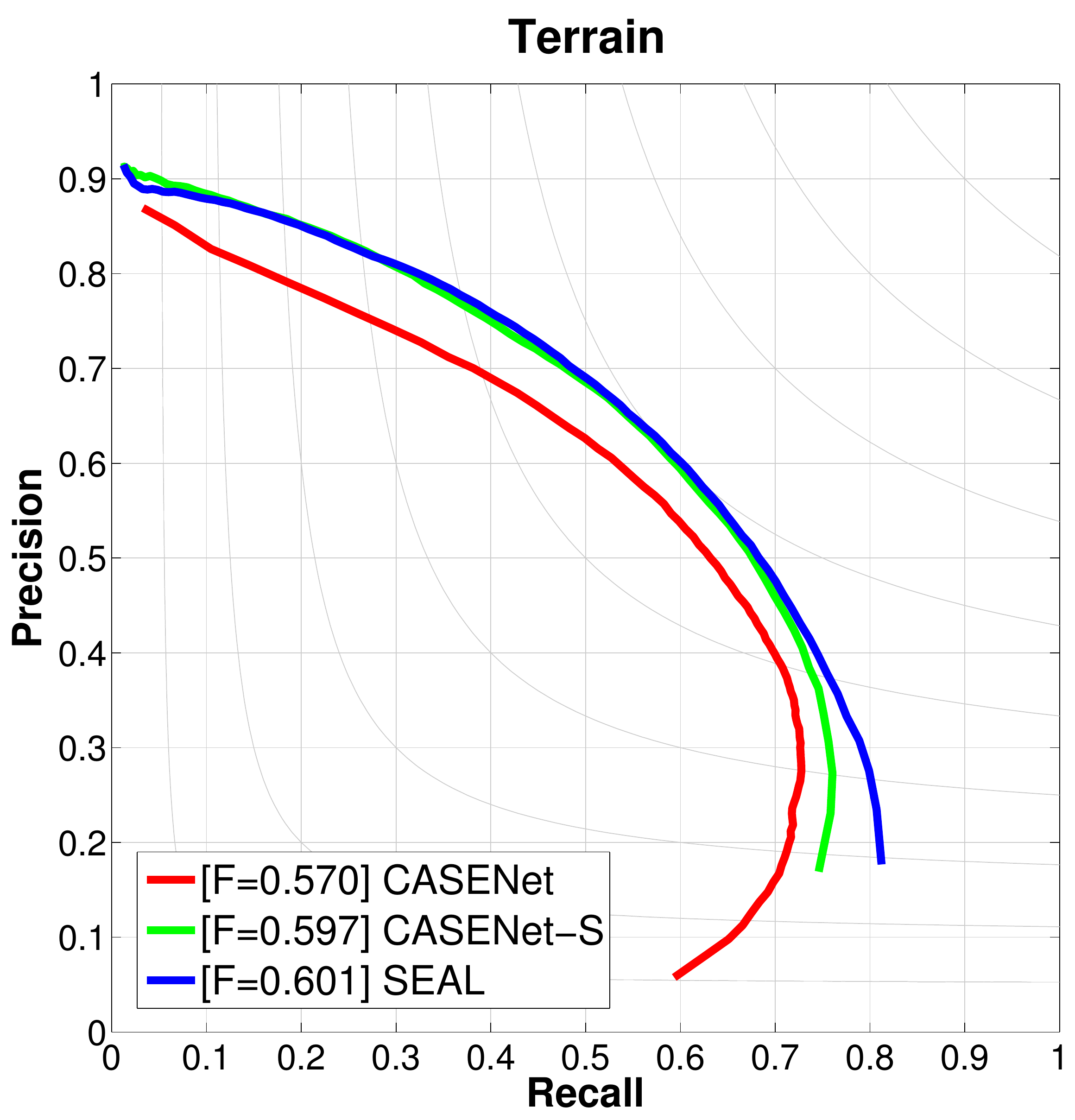}
	\includegraphics[width=.244\textwidth]{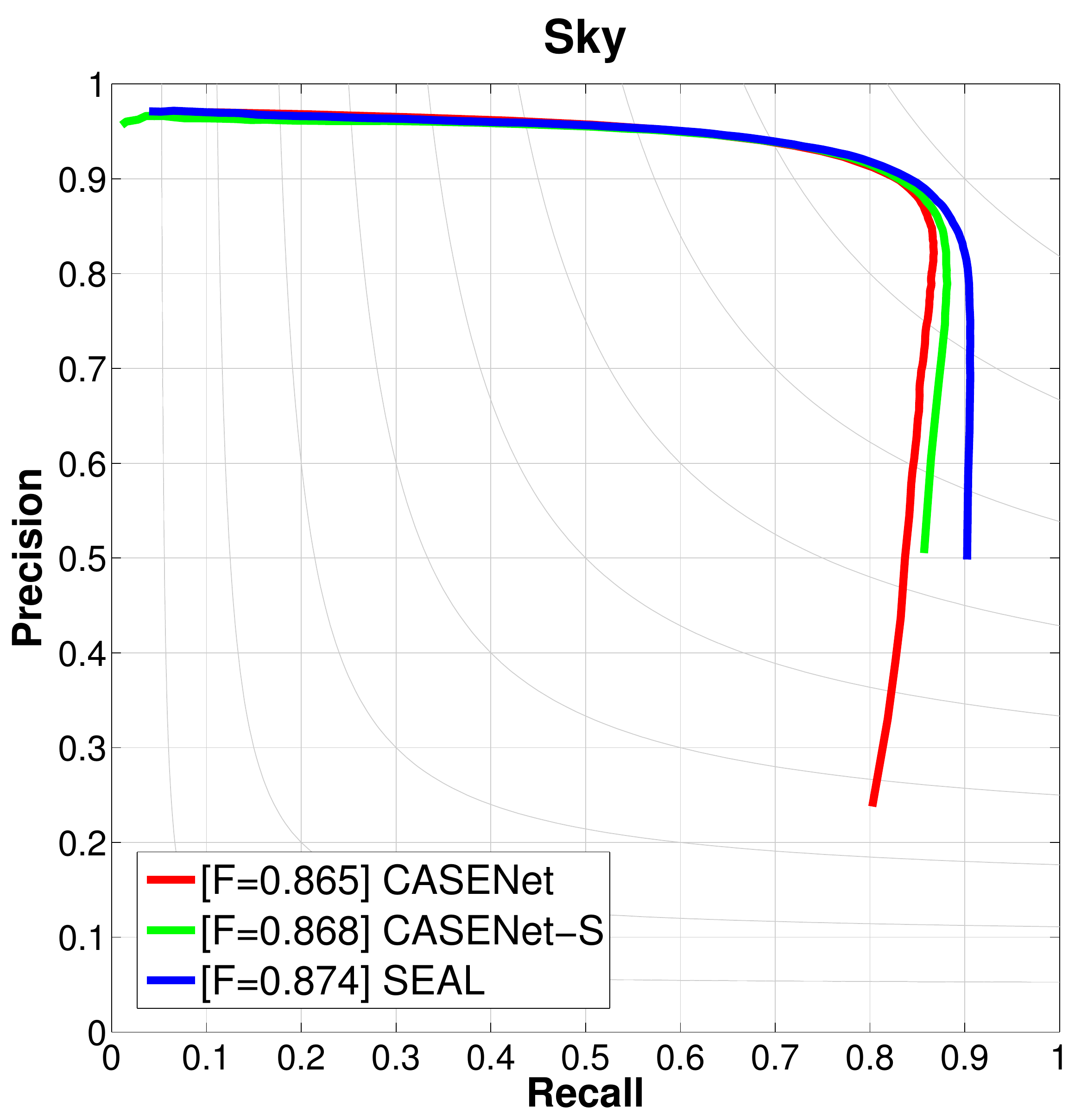}
	\includegraphics[width=.244\textwidth]{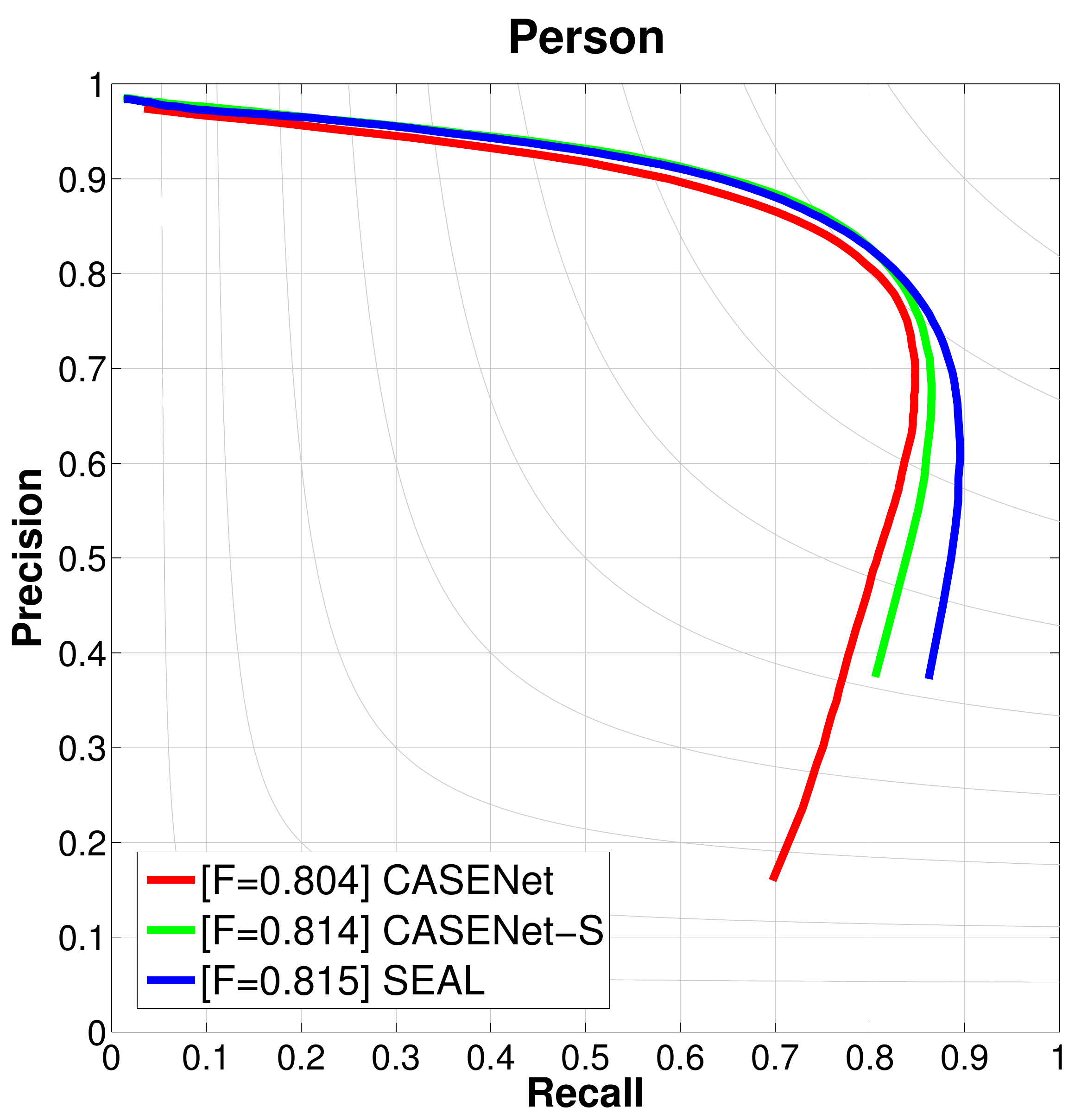}\\
	
	\includegraphics[width=.244\textwidth]{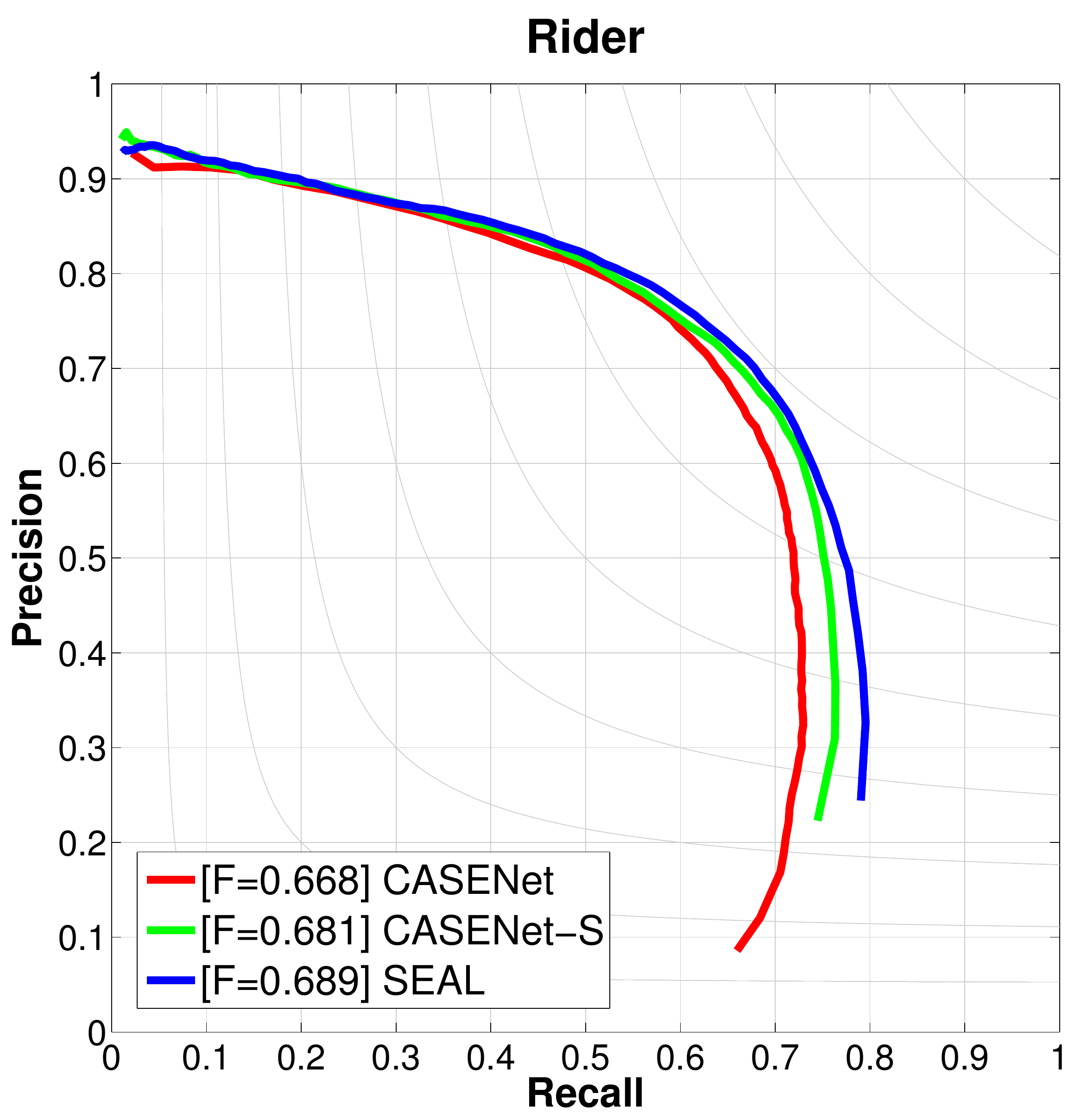}
	\includegraphics[width=.244\textwidth]{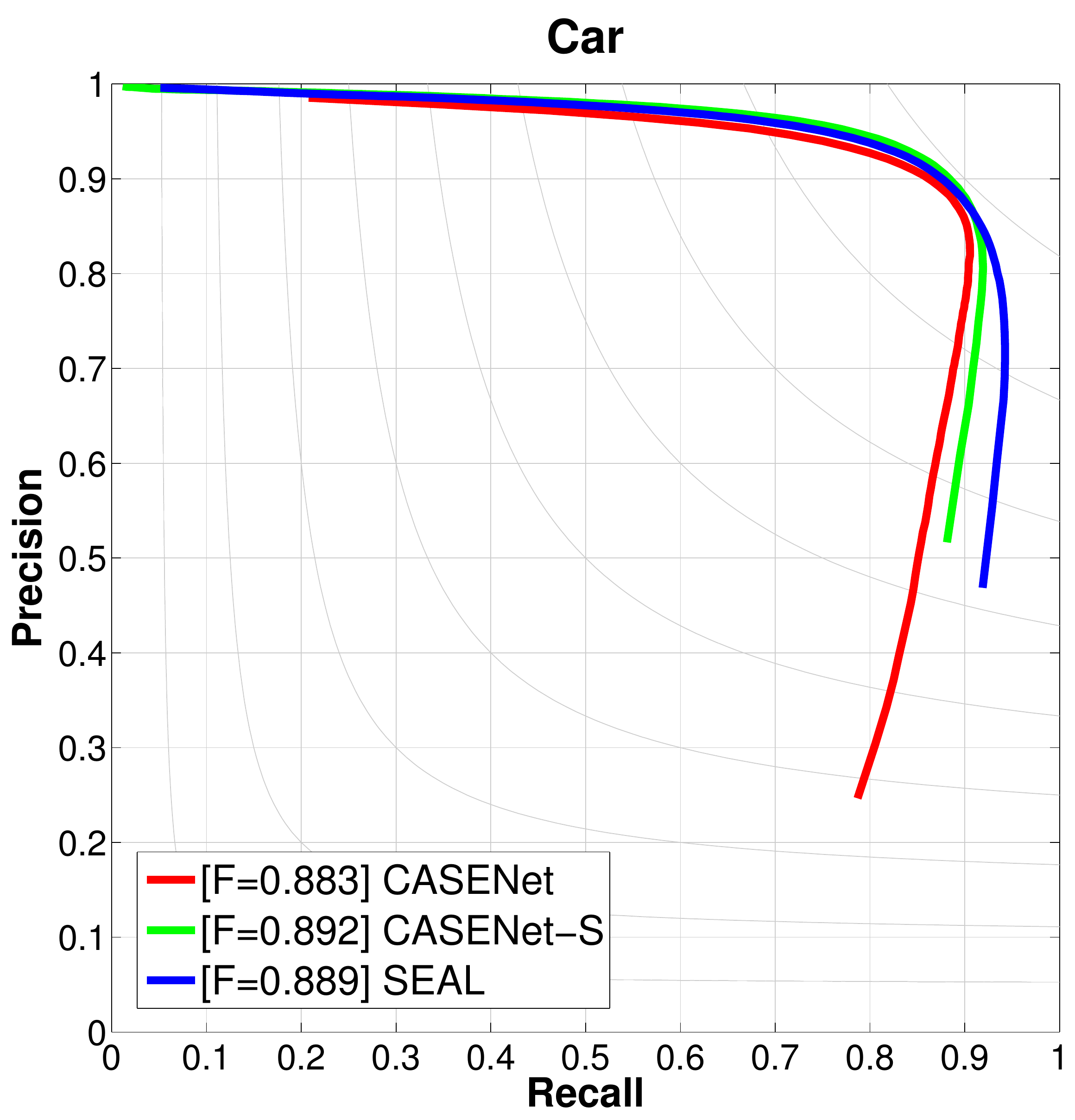}
	\includegraphics[width=.244\textwidth]{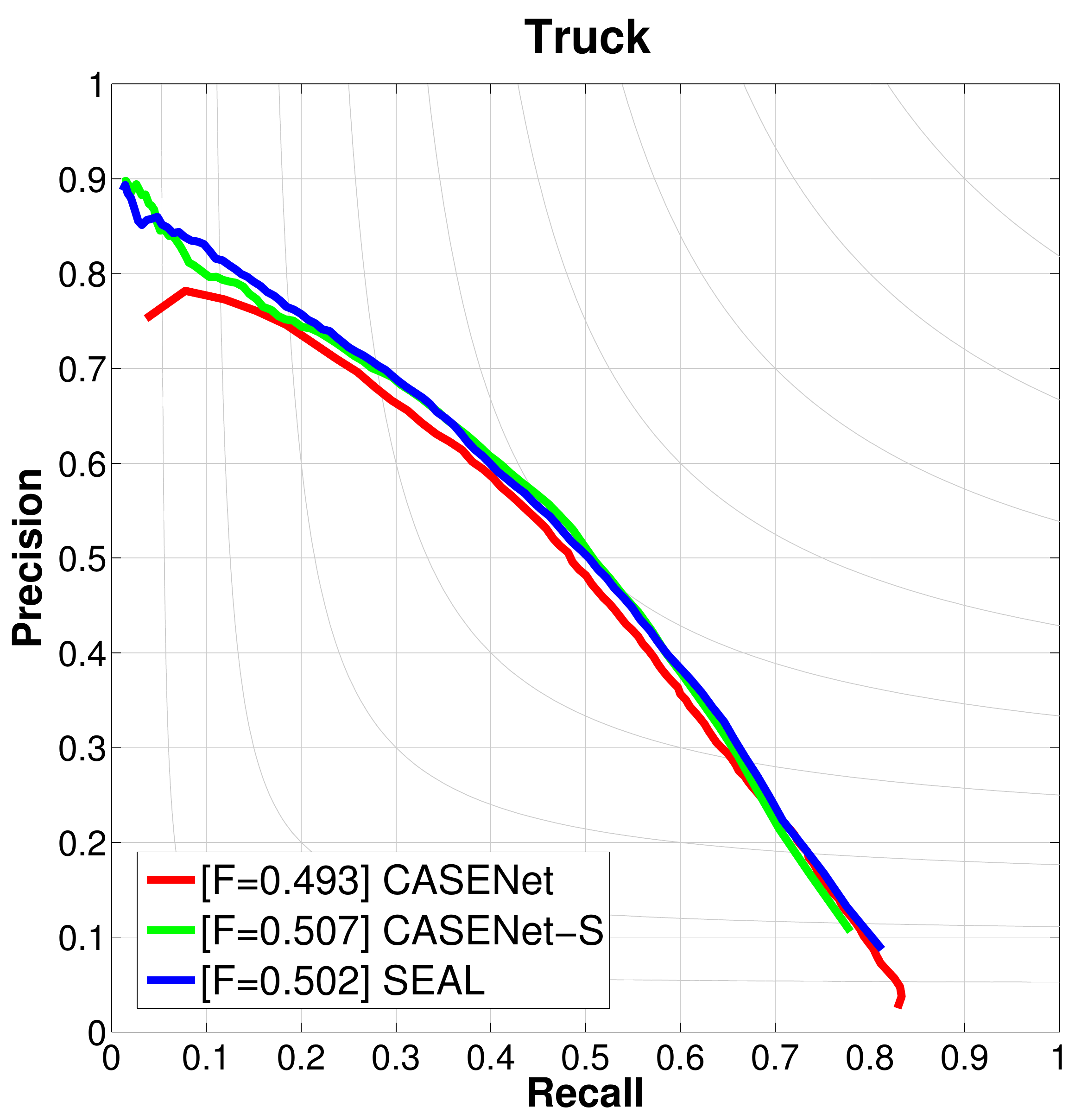}
	\includegraphics[width=.244\textwidth]{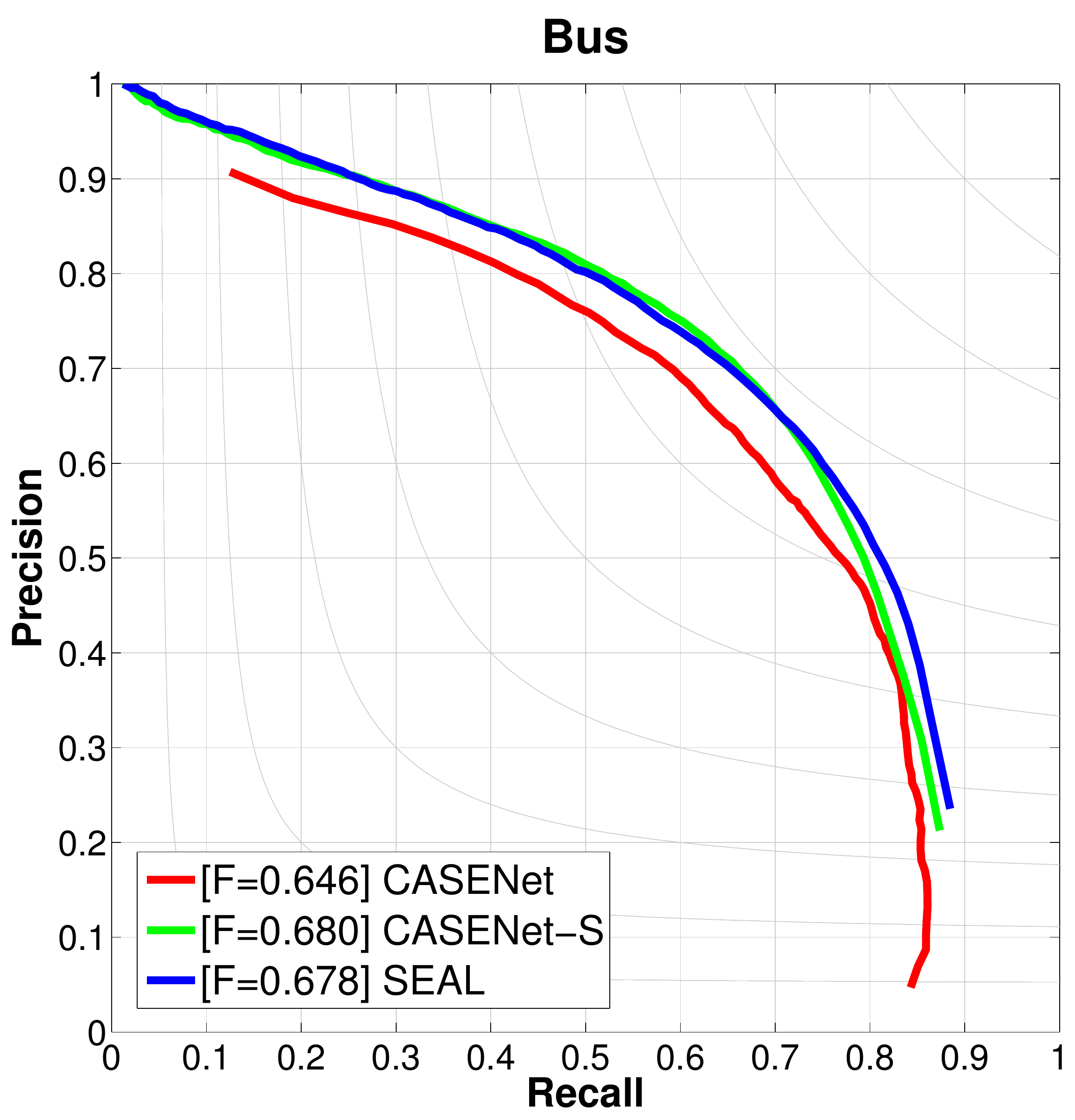}\\

	\includegraphics[width=.244\textwidth]{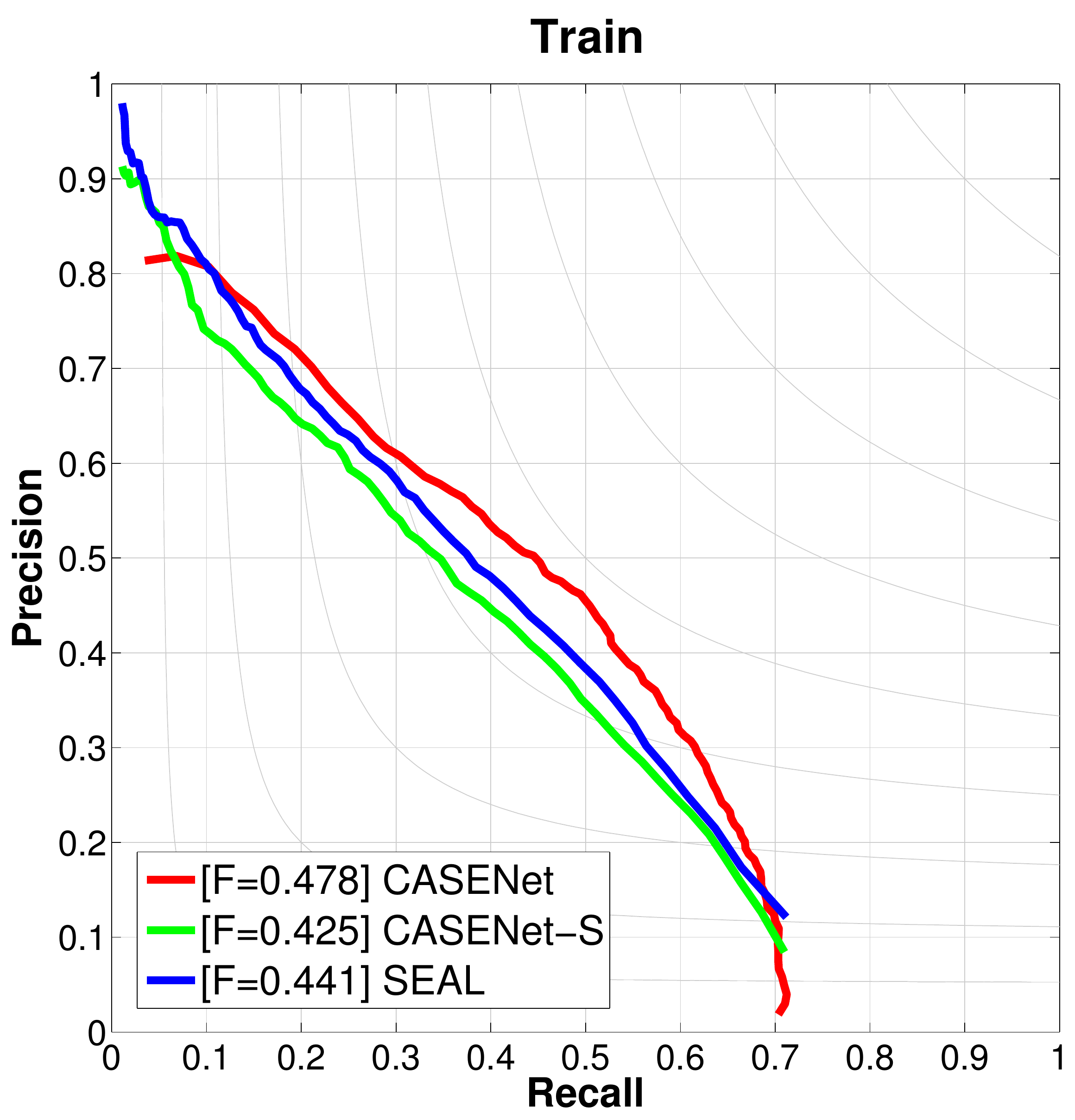}
	\includegraphics[width=.244\textwidth]{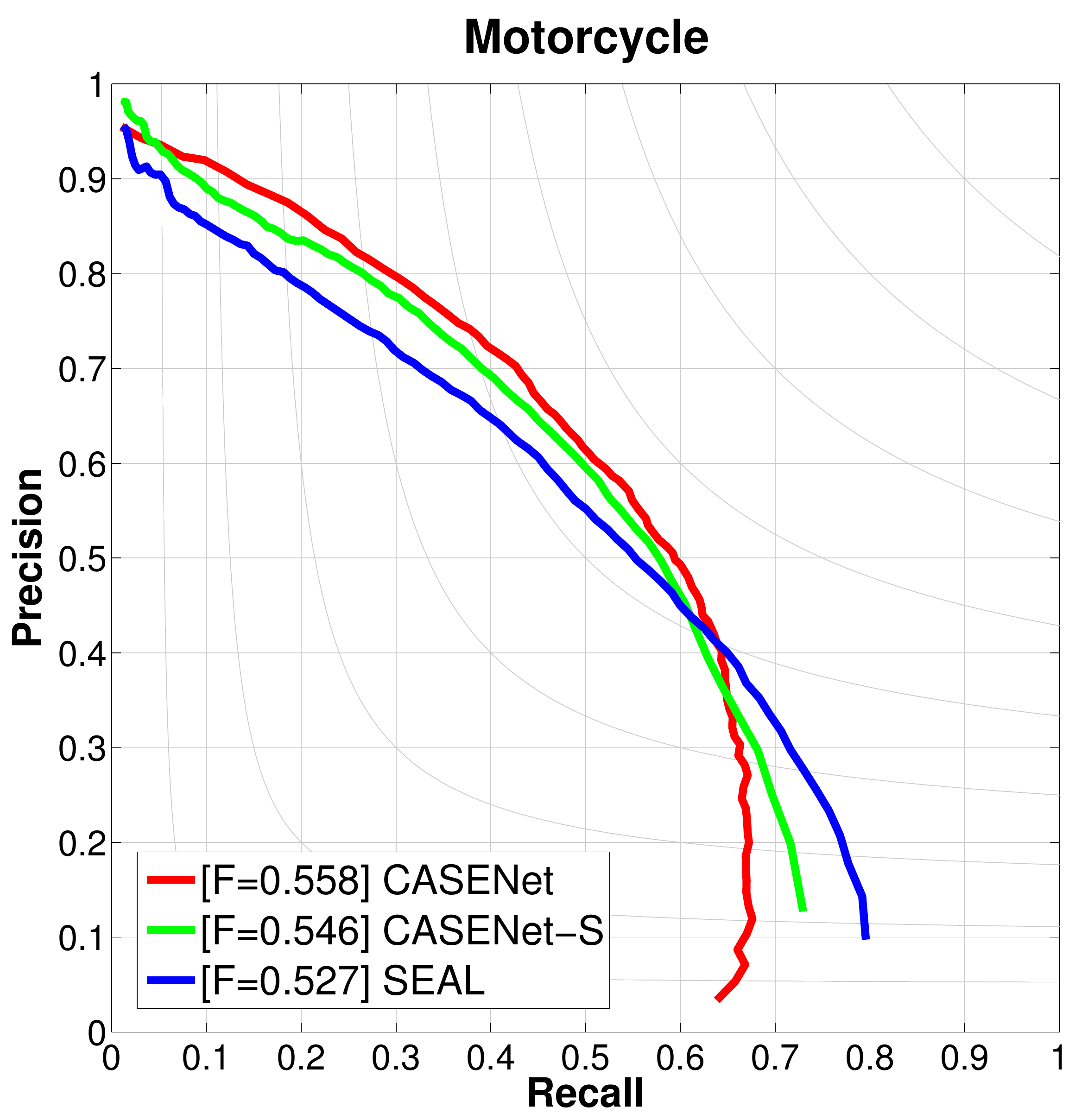}
	\includegraphics[width=.244\textwidth]{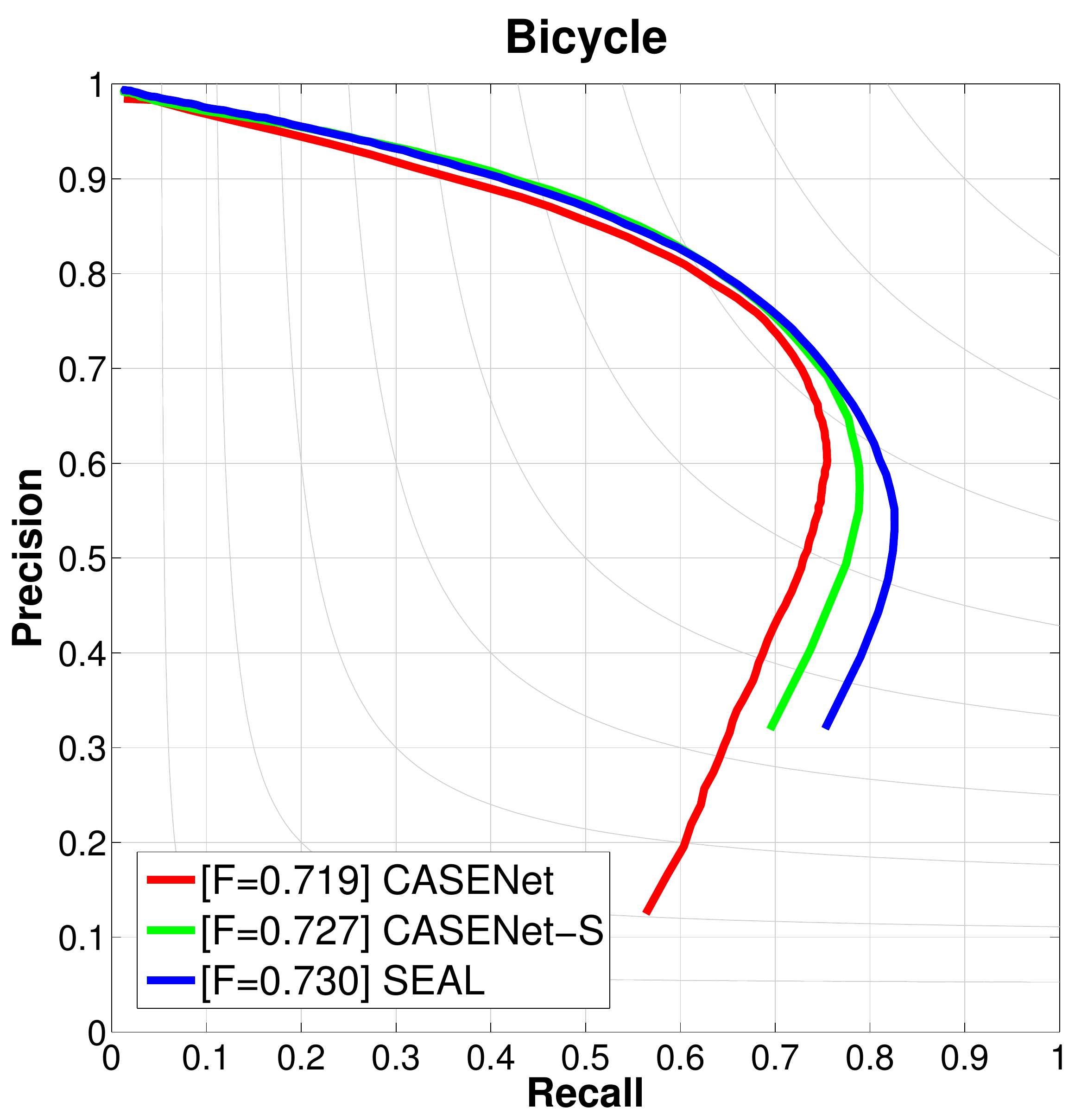}\\
	\caption{Class-wise precision-recall curves of SEAL and comparing baselines on the Cityscapes validation set under the ``Thin'' setting.}\label{pr_city_thin}
\end{figure}

\begin{figure}[tbh]
	\raggedright
	\includegraphics[width=.244\textwidth]{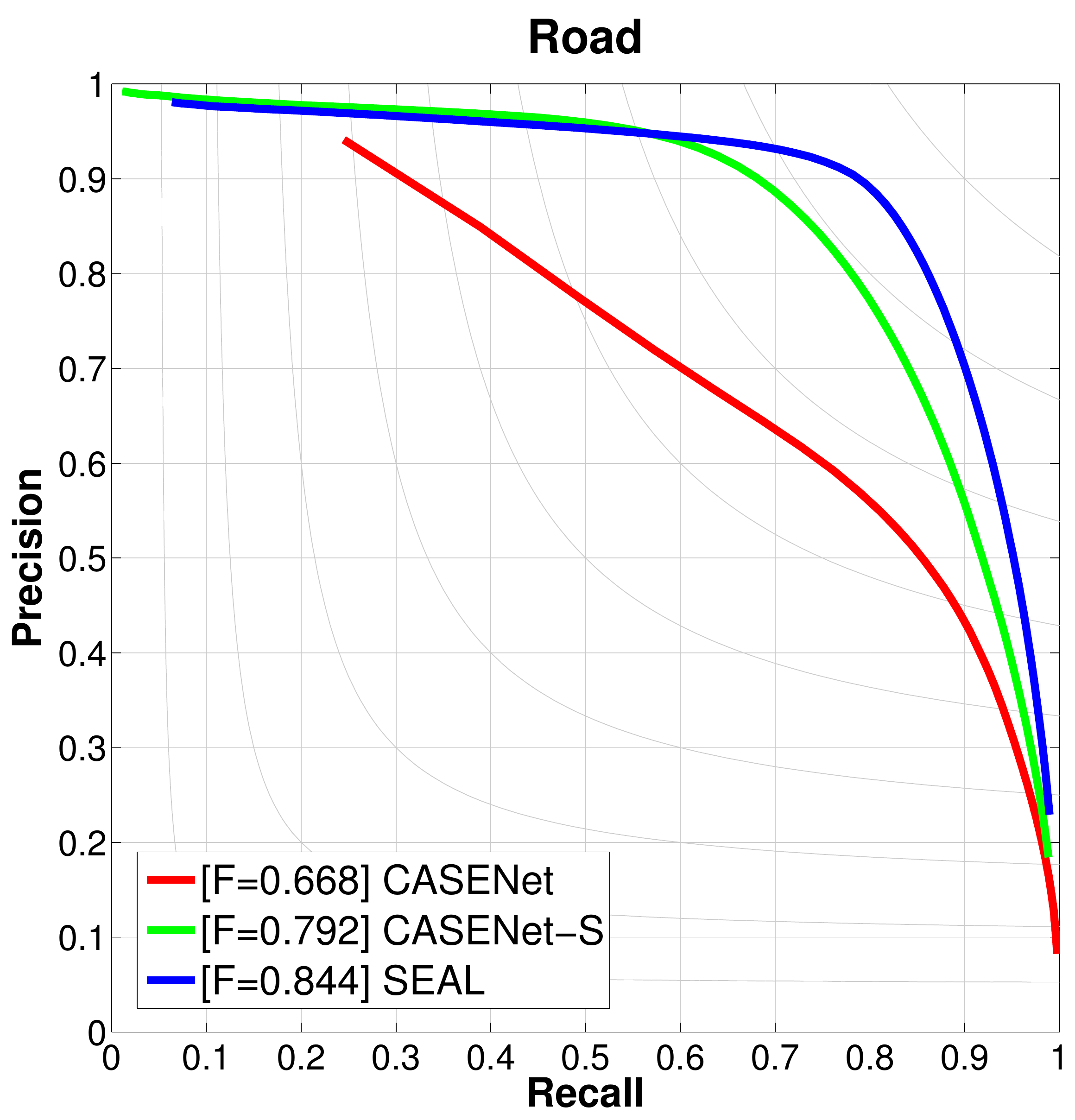}
	\includegraphics[width=.244\textwidth]{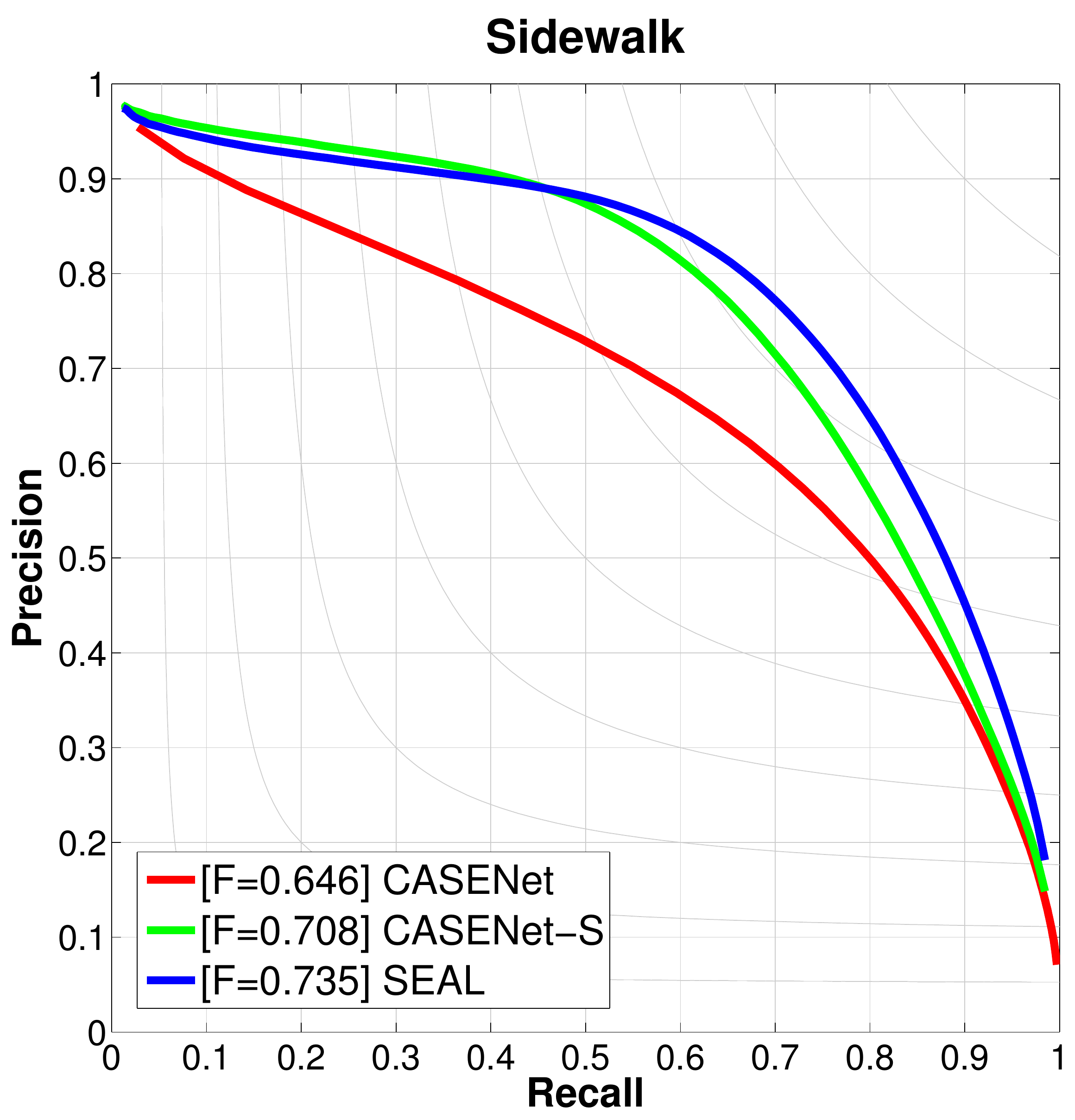}
	\includegraphics[width=.244\textwidth]{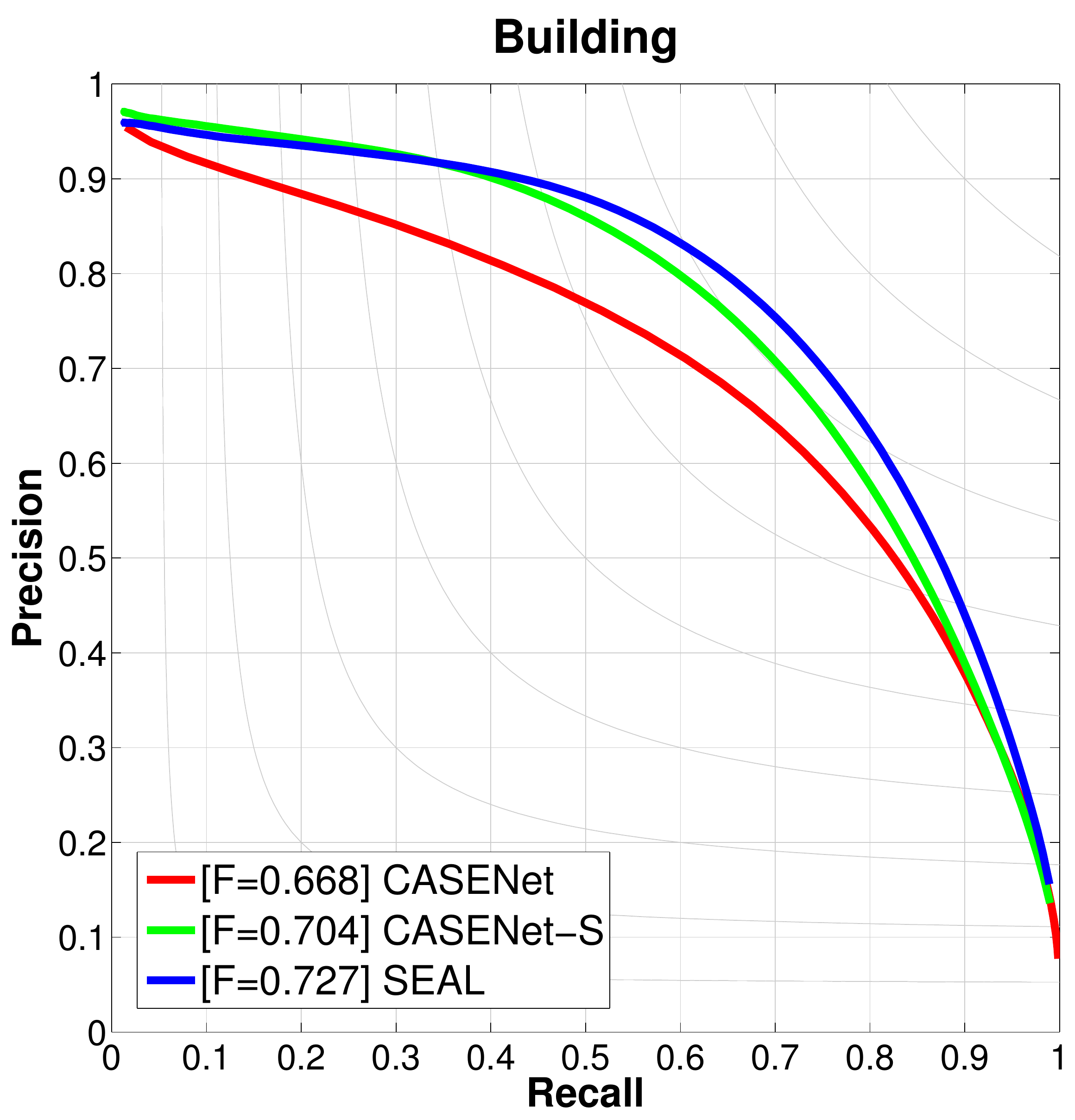}
	\includegraphics[width=.244\textwidth]{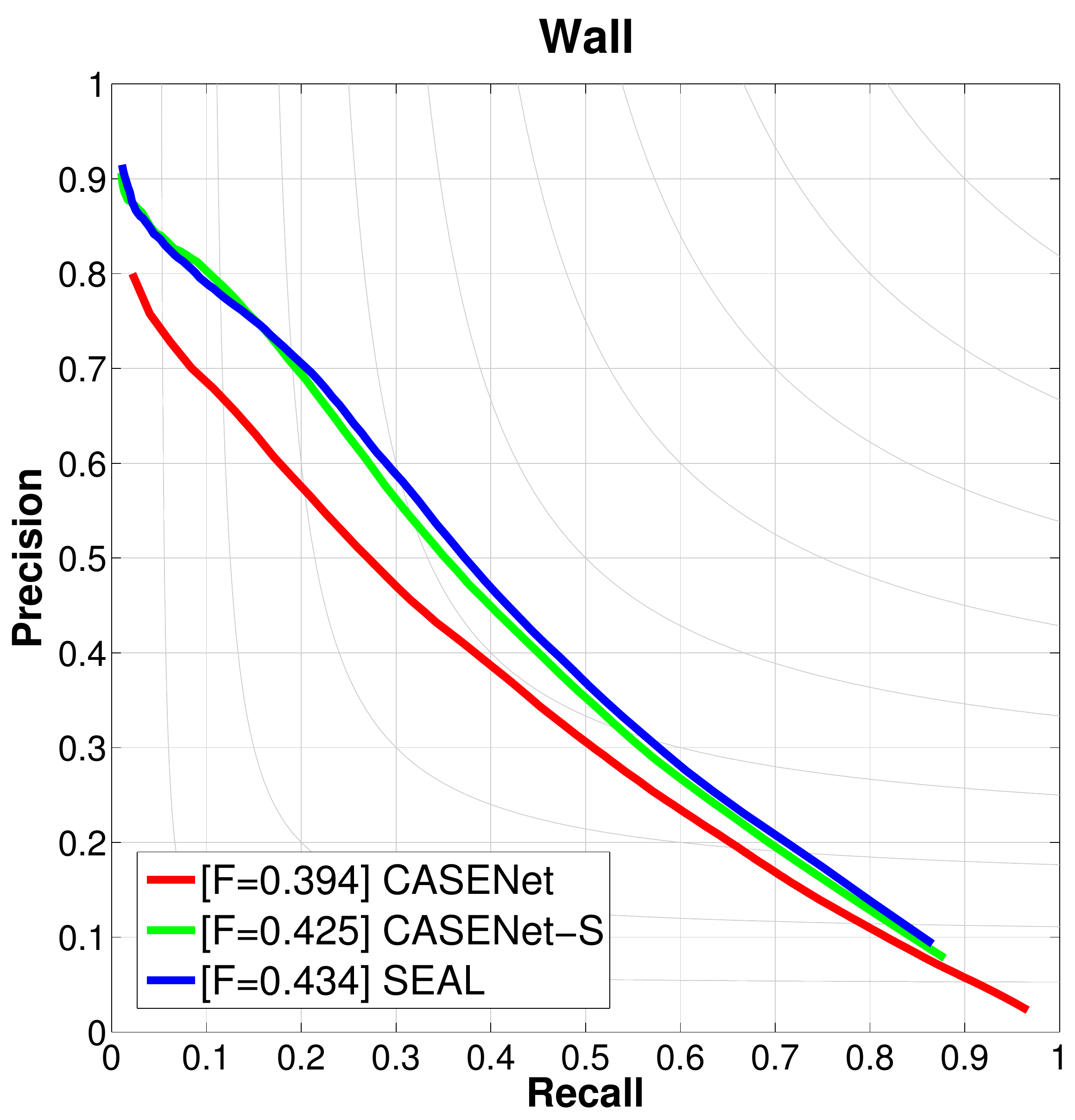}\\
	
	\includegraphics[width=.244\textwidth]{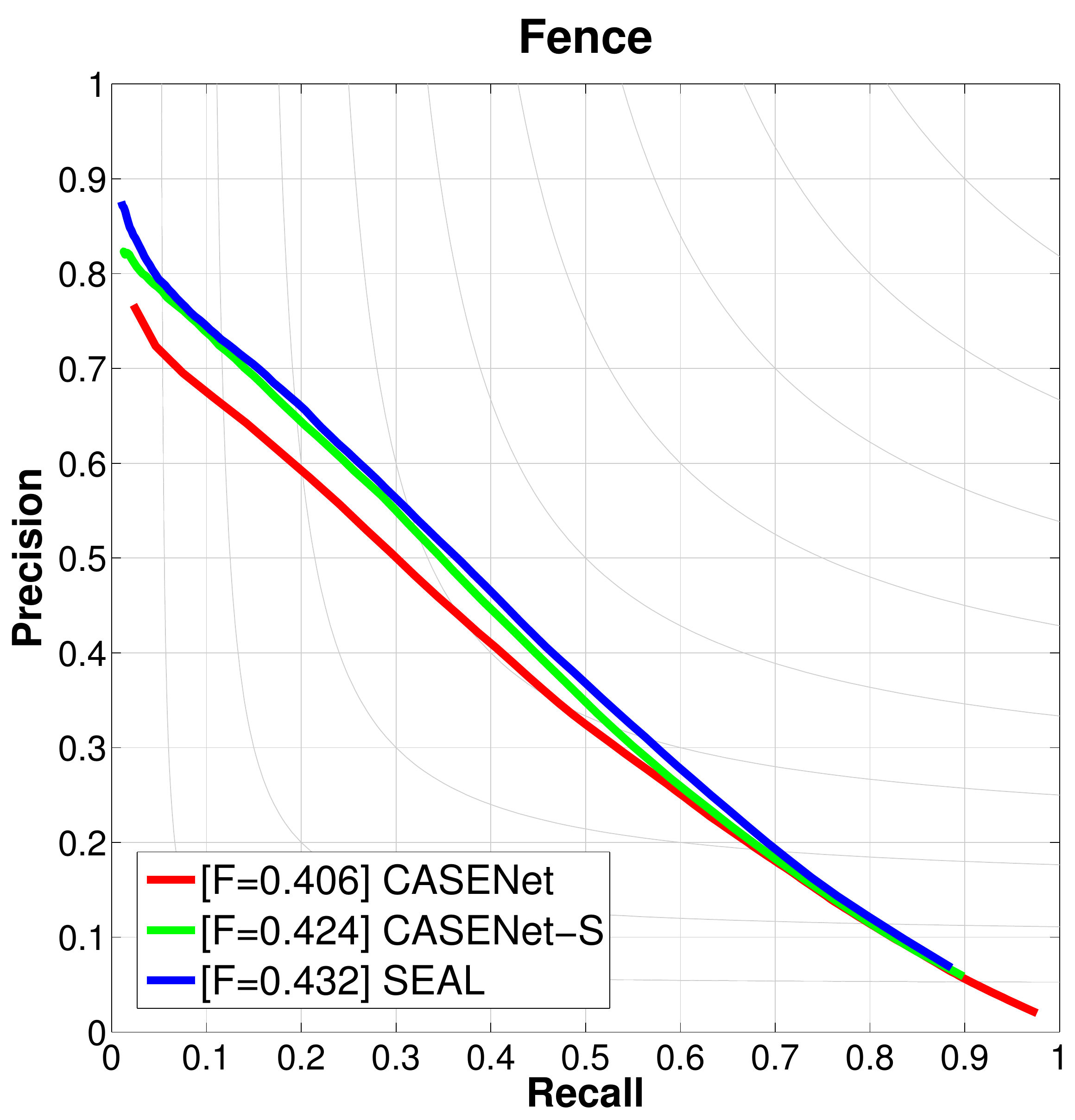}
	\includegraphics[width=.244\textwidth]{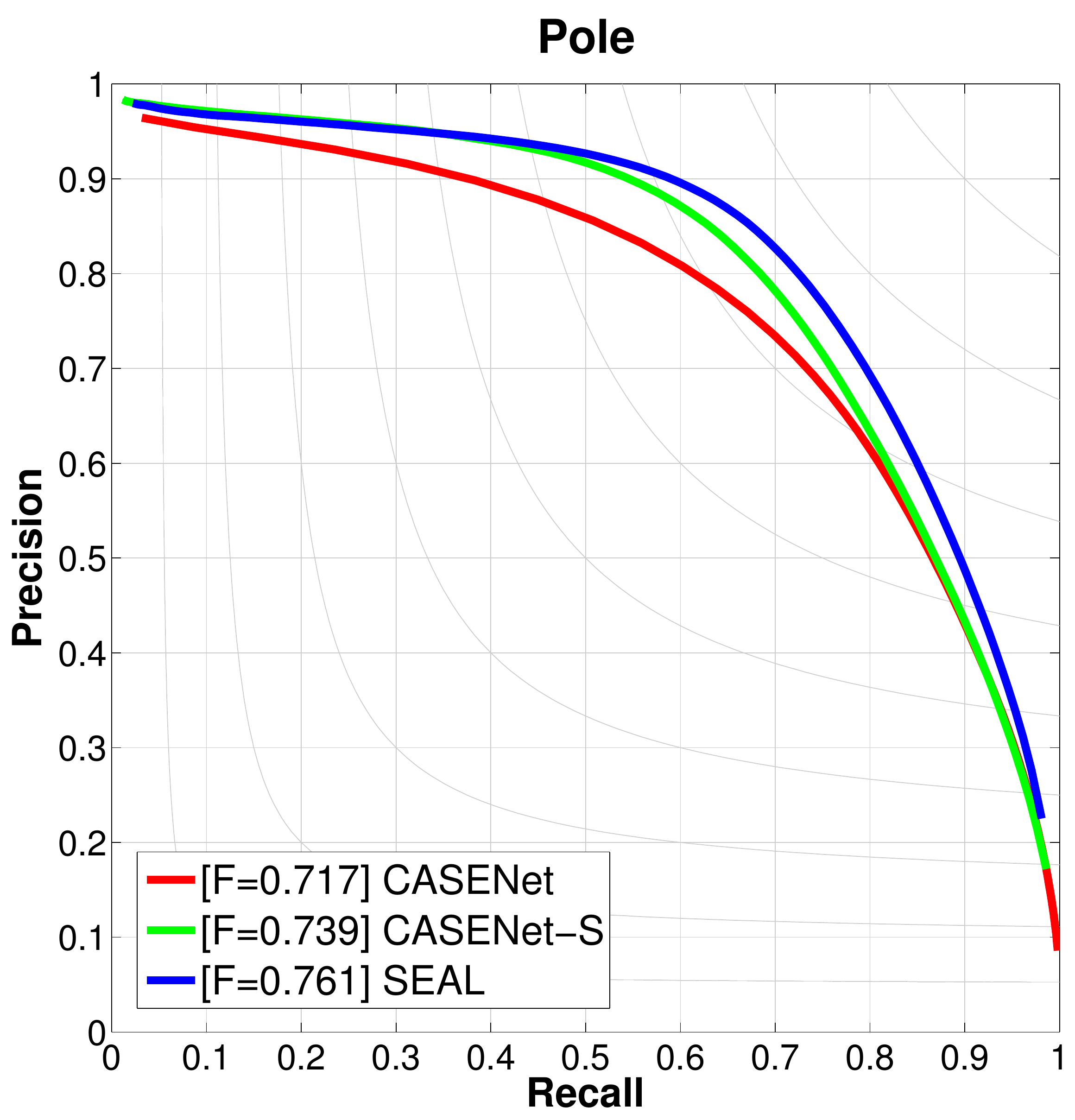}
	\includegraphics[width=.244\textwidth]{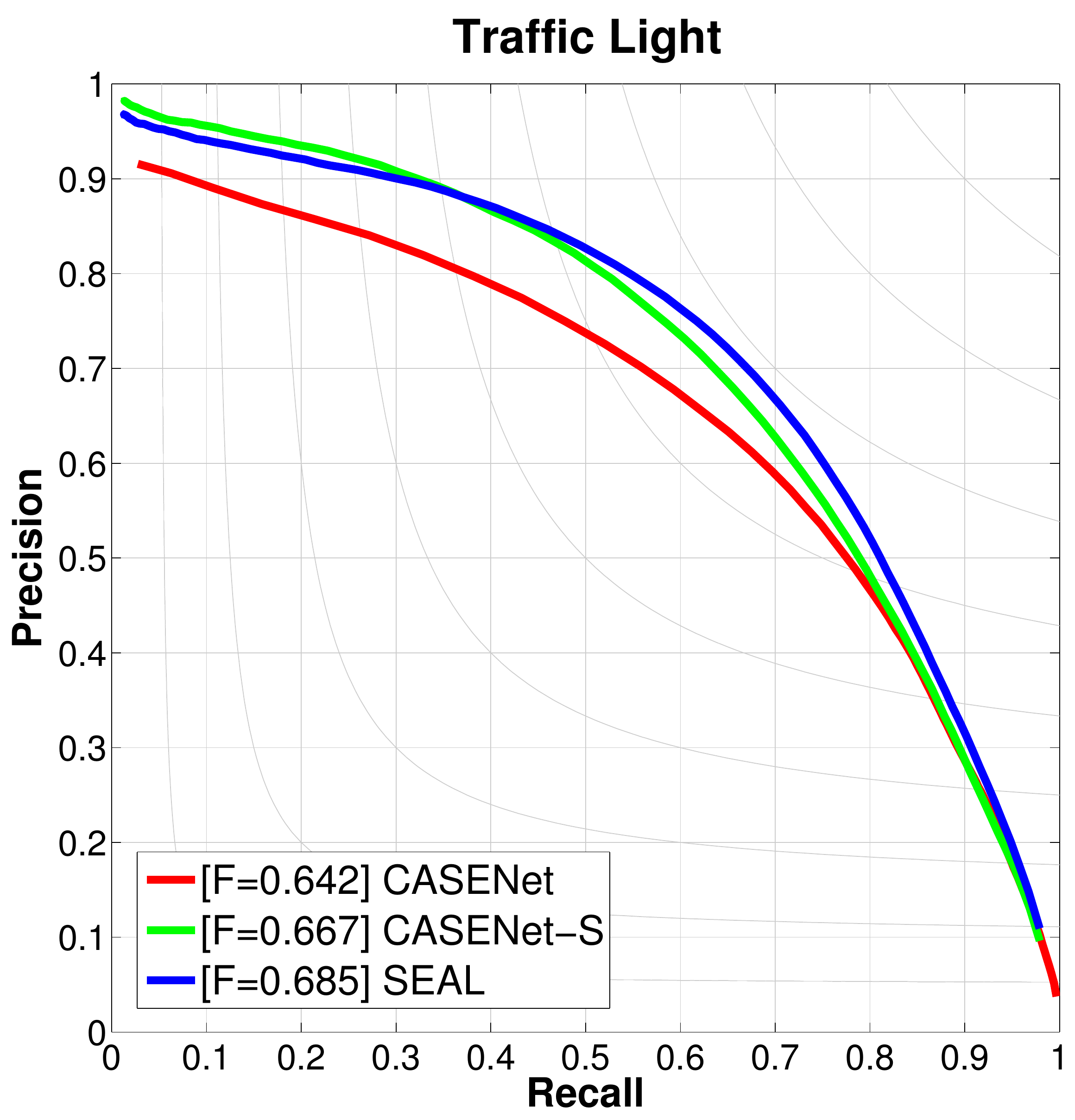}
	\includegraphics[width=.244\textwidth]{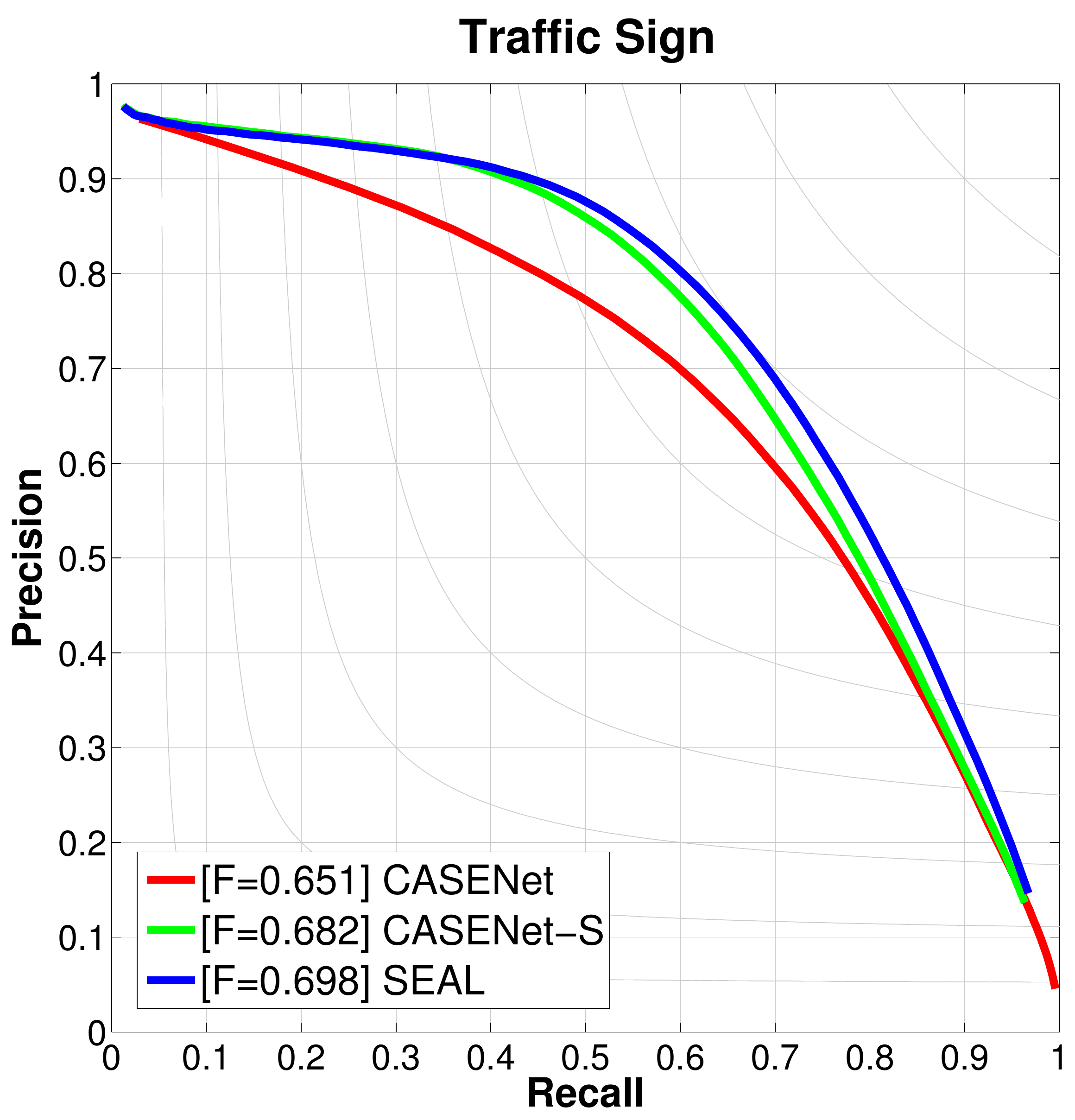}\\
	
	\includegraphics[width=.244\textwidth]{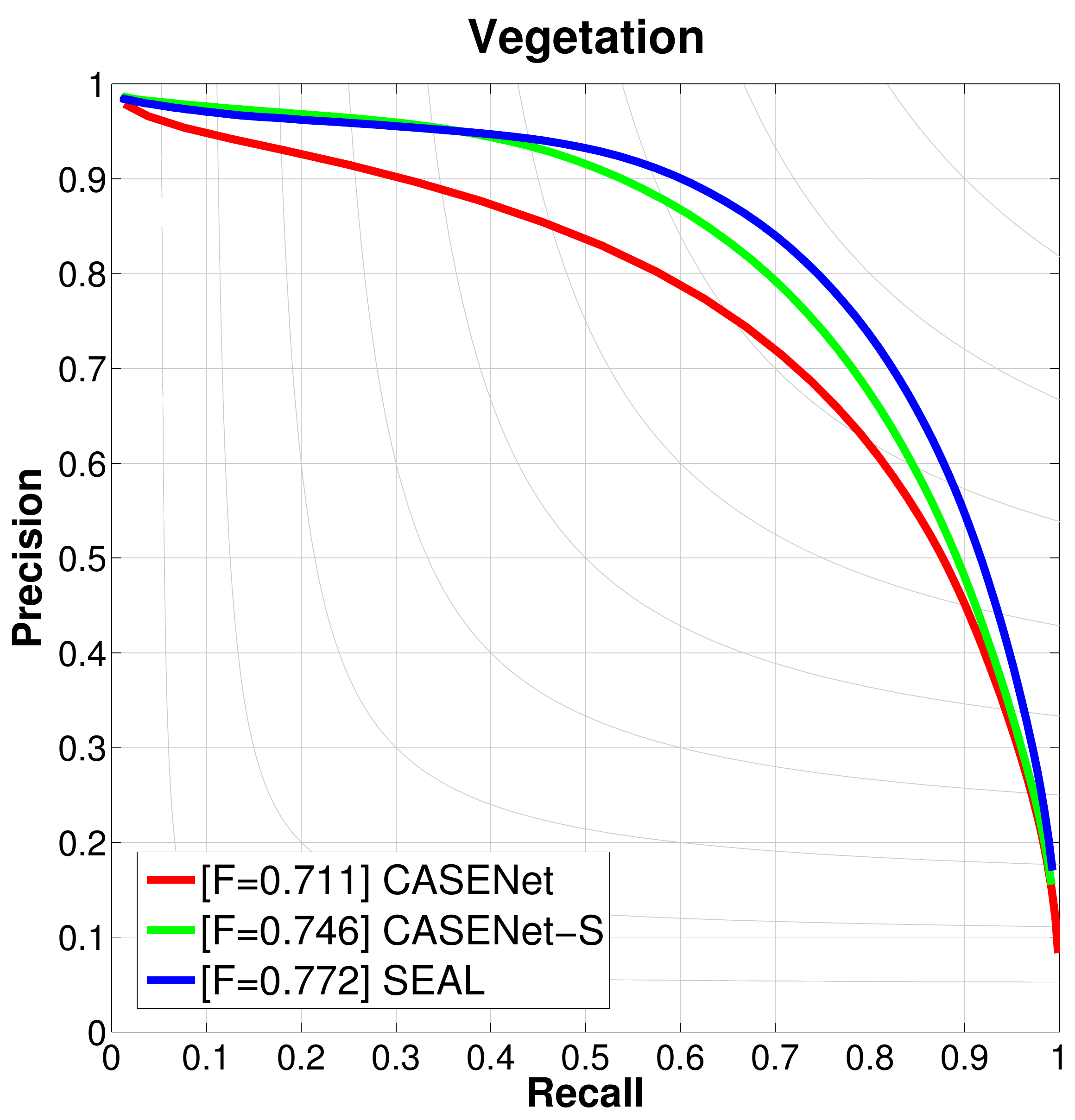}
	\includegraphics[width=.244\textwidth]{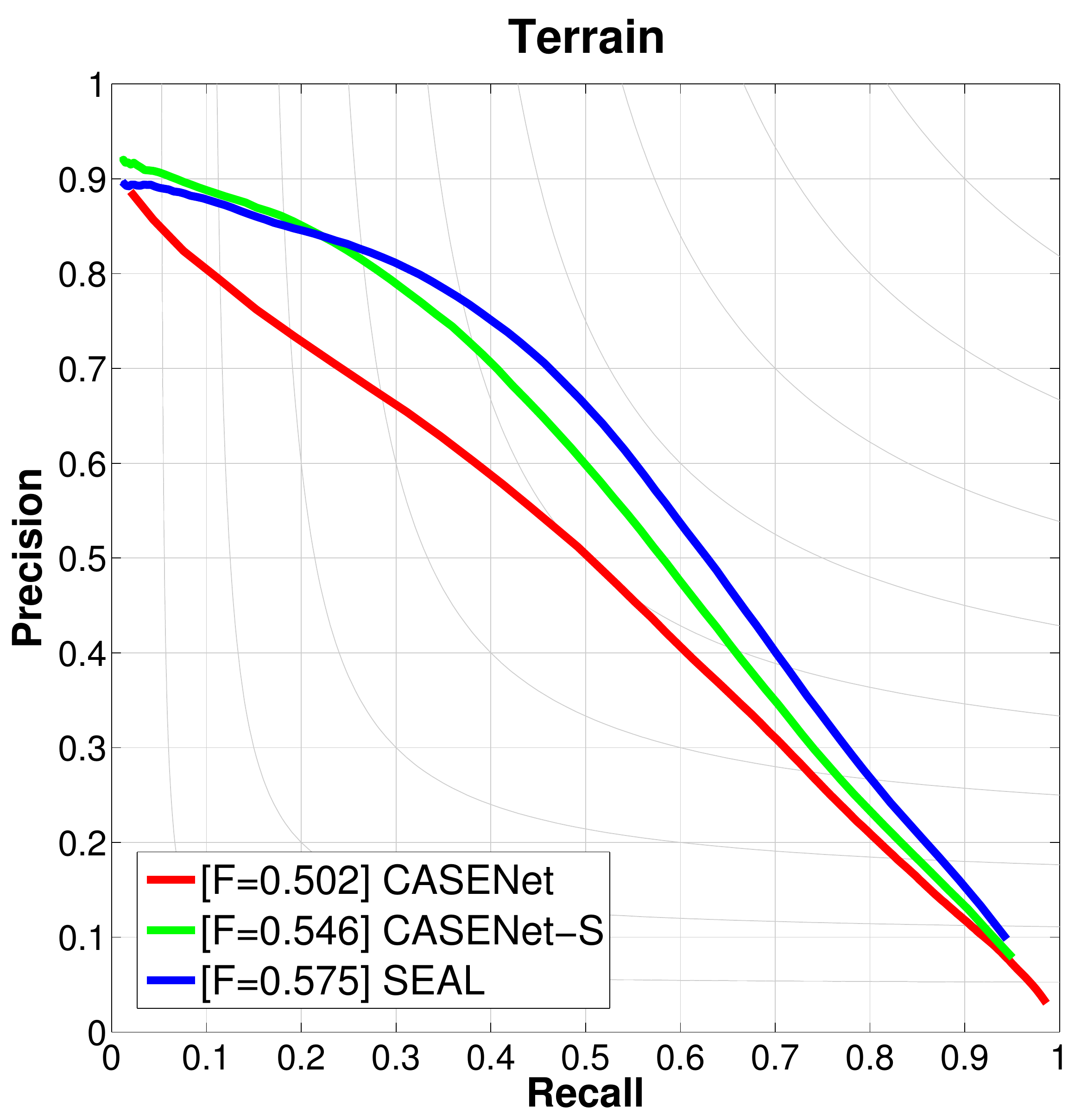}
	\includegraphics[width=.244\textwidth]{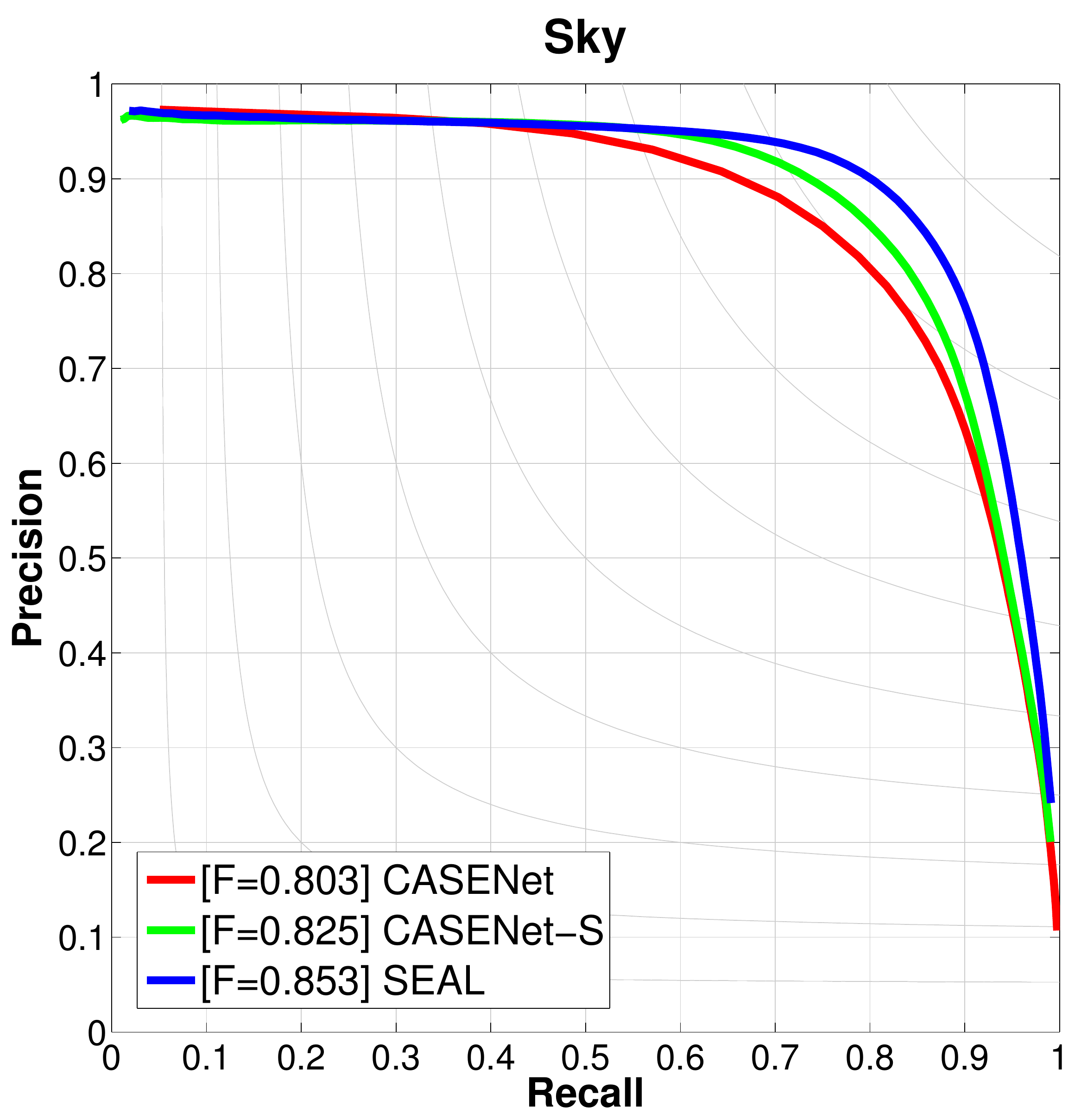}
	\includegraphics[width=.244\textwidth]{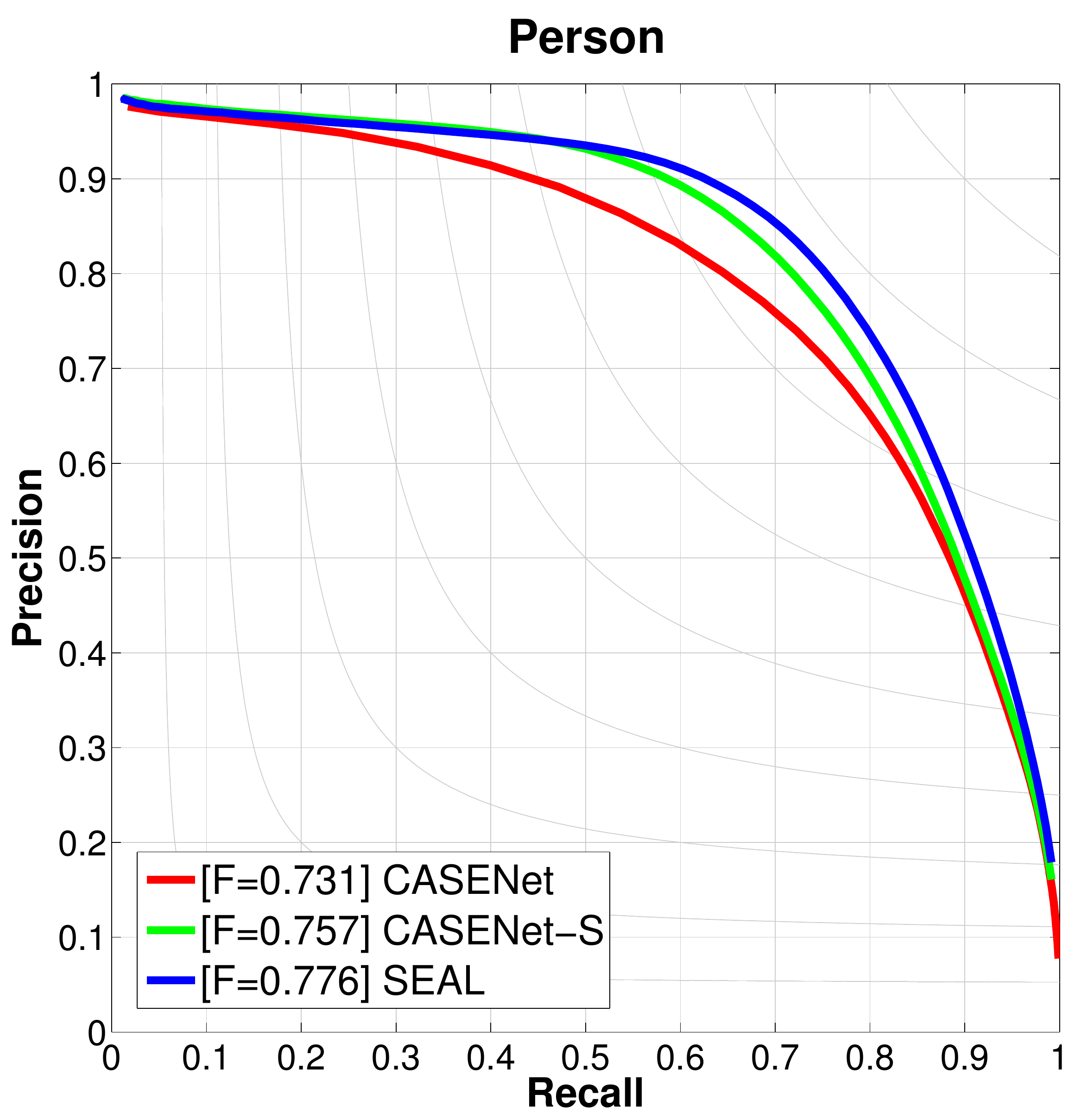}\\
	
	\includegraphics[width=.244\textwidth]{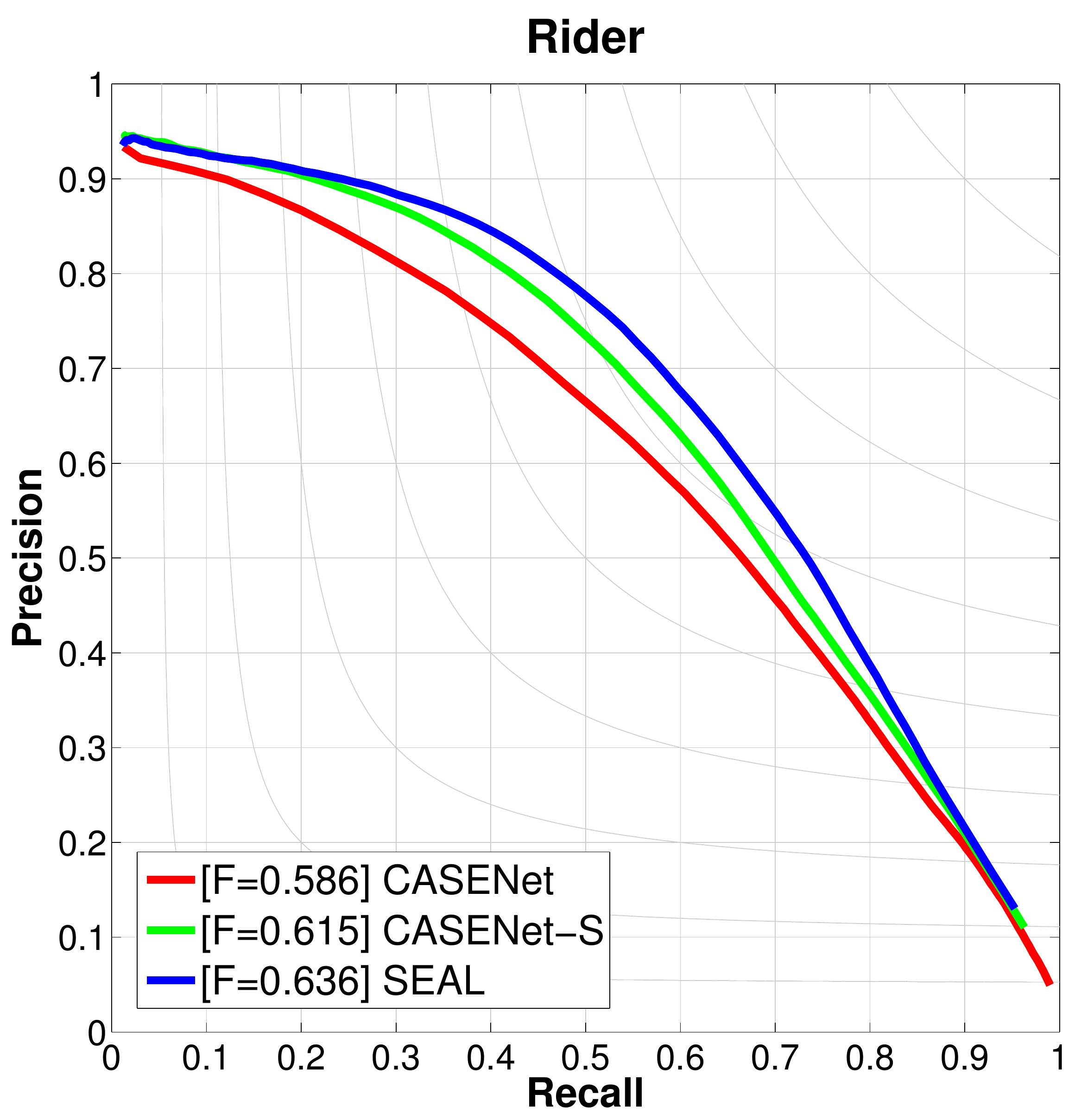}
	\includegraphics[width=.244\textwidth]{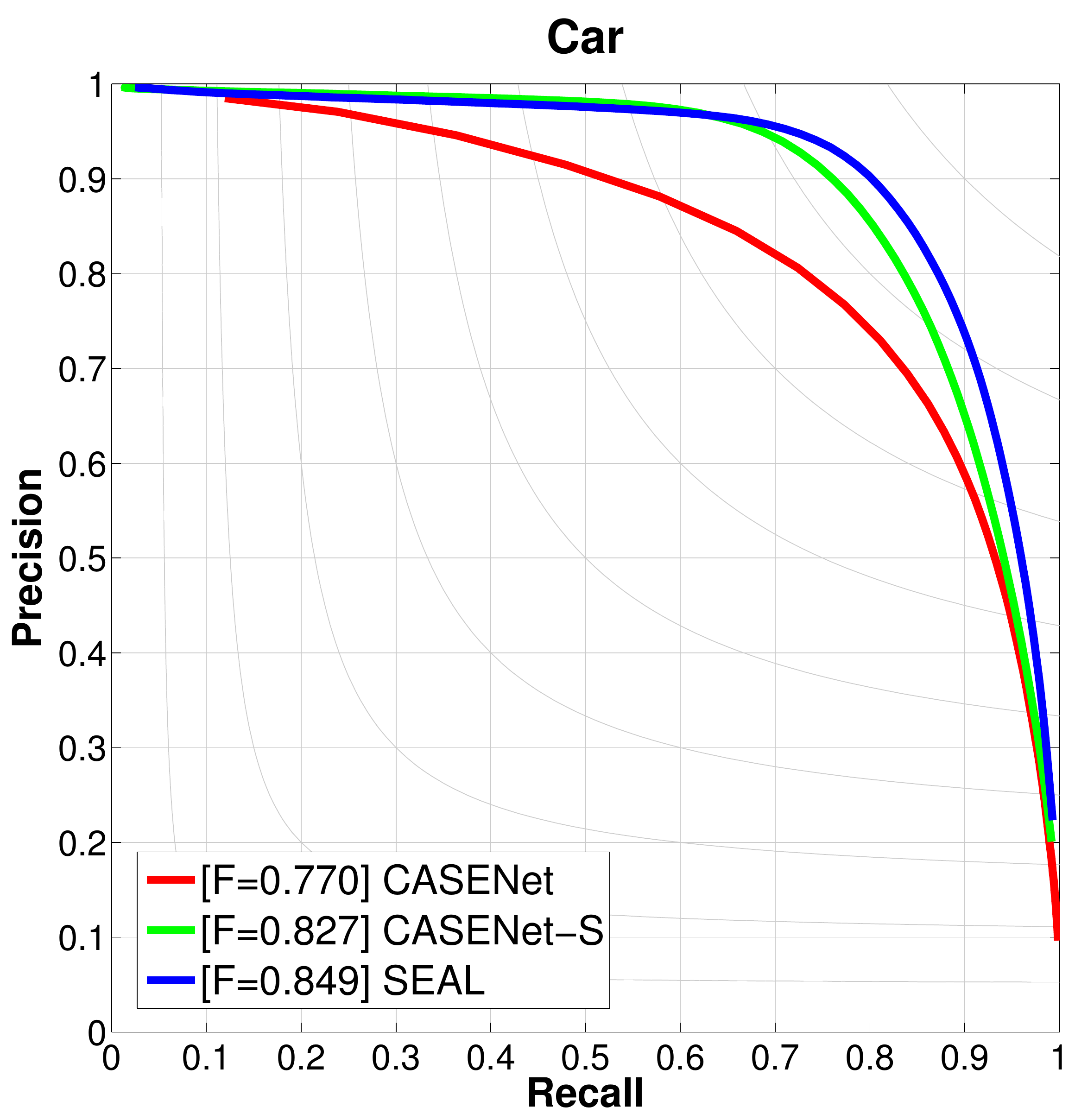}
	\includegraphics[width=.244\textwidth]{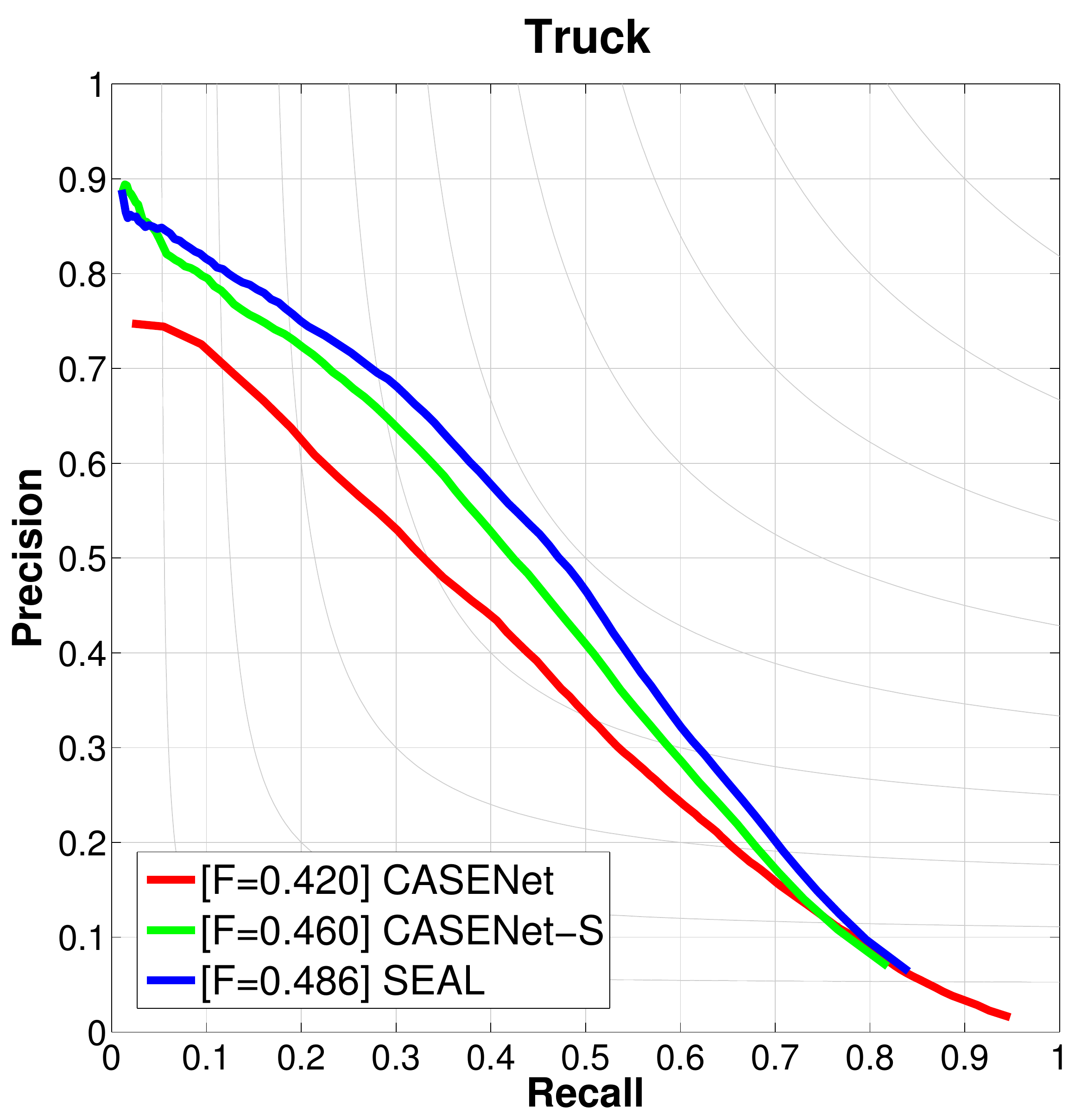}
	\includegraphics[width=.244\textwidth]{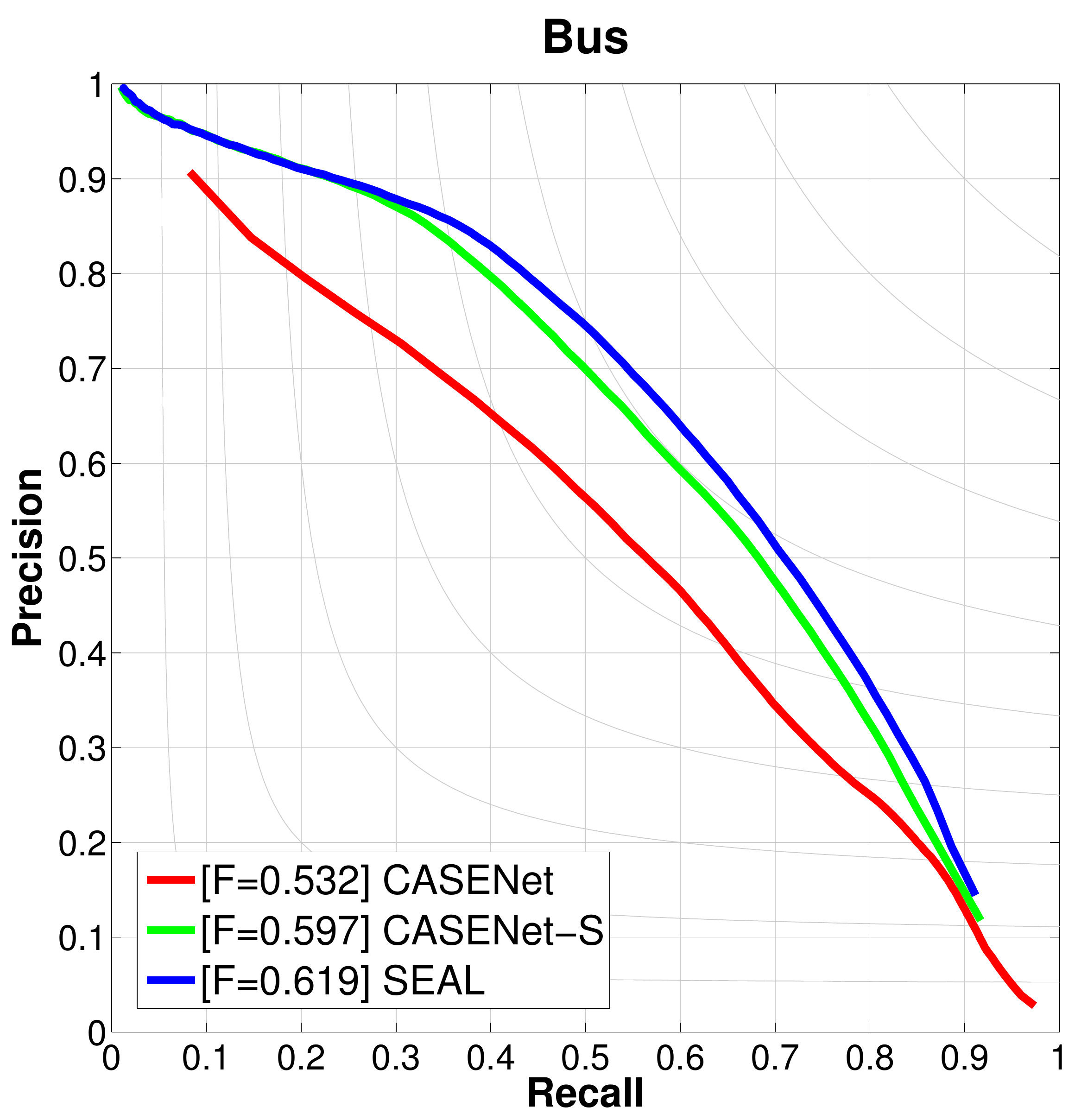}\\
	
	\includegraphics[width=.244\textwidth]{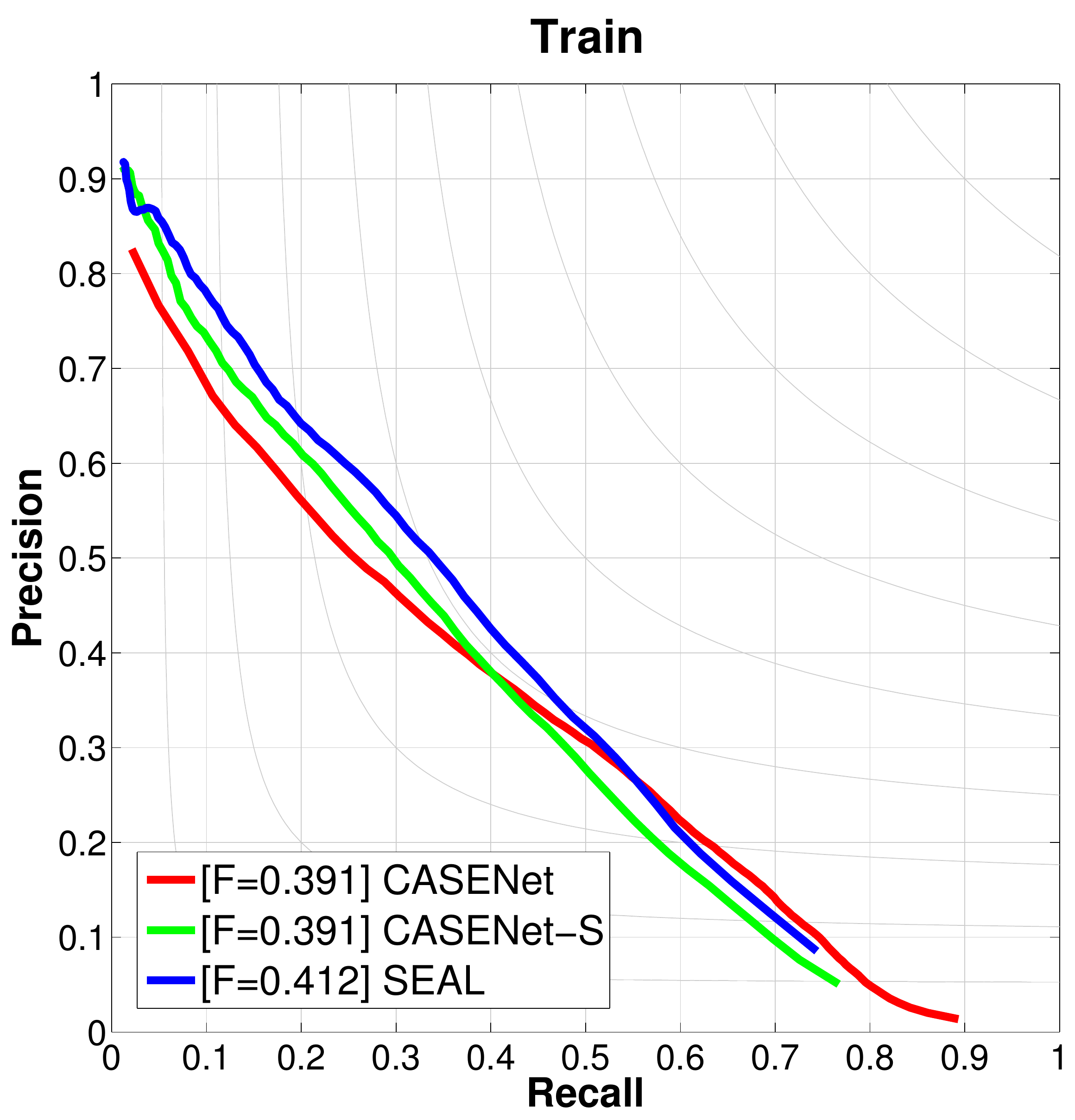}
	\includegraphics[width=.244\textwidth]{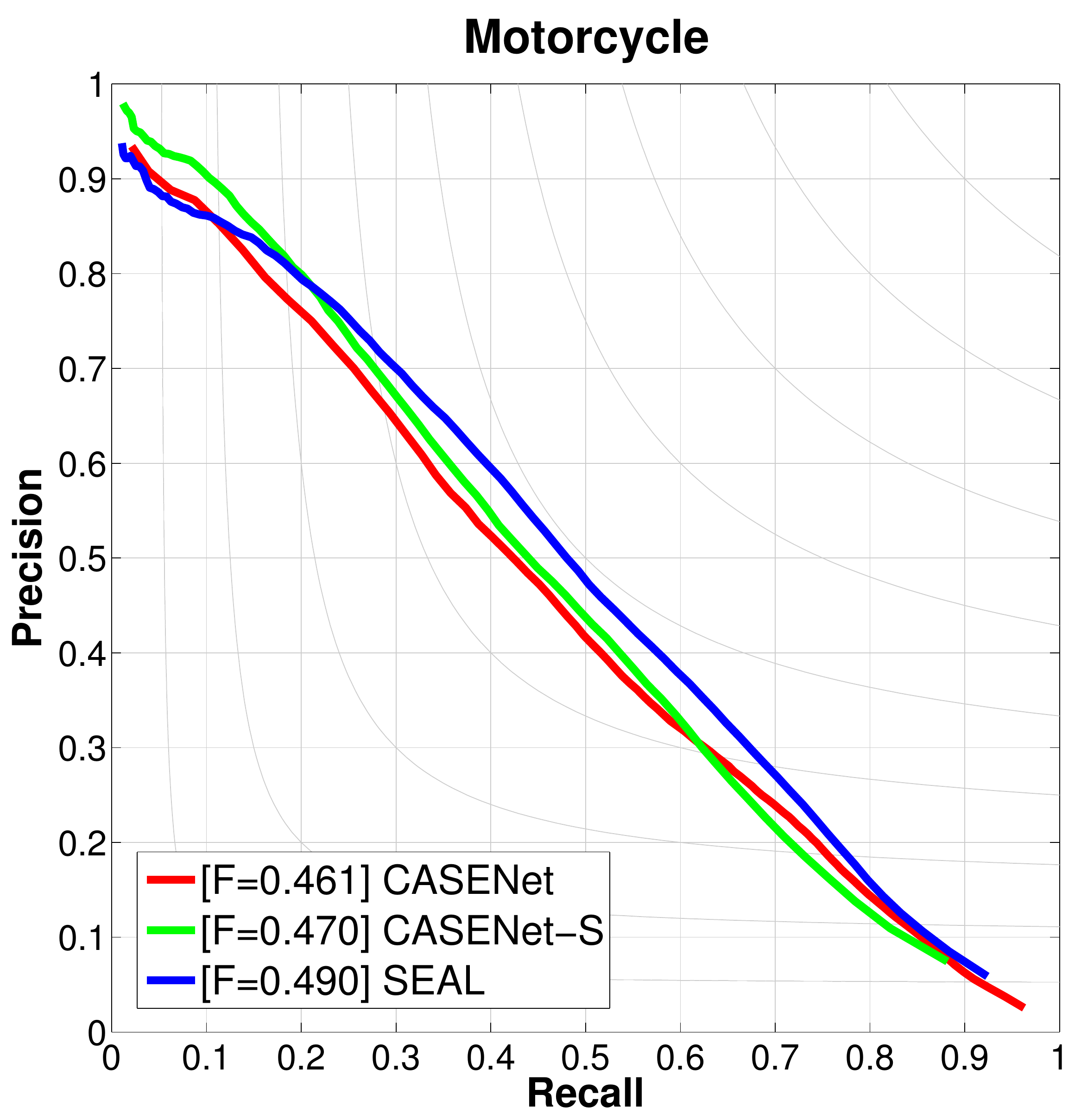}
	\includegraphics[width=.244\textwidth]{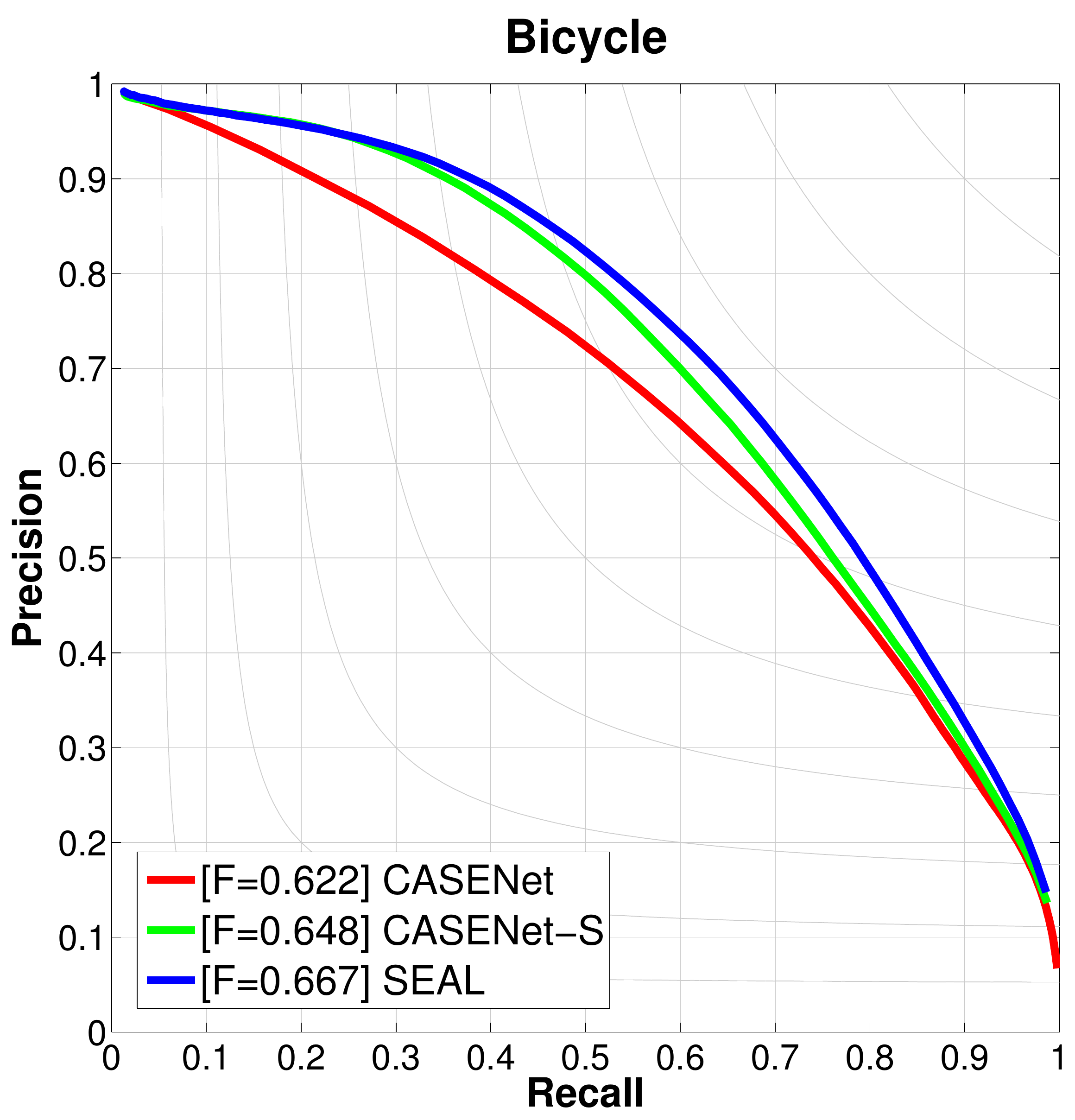}\\
	\caption{Class-wise precision-recall curves of SEAL and comparing baselines on the Cityscapes validation set under the ``Raw'' setting.}\label{pr_city_raw}
\end{figure}

In this section, we report additional results on the Cityscapes dataset. Considering the better annotation quality and the larger computation cost for edge alignment, we set $\sigma_y=3$ in the Cityscapes experiment, while keeping other SEAL parameters the same as SBD. In particular, Table~\ref{tb:cityscapes_ap} contains the average precision scores on the Cityscapes validation set under the ``Raw'' setting. In addition, the class-wise precision-recall curves under both ``Thin'' and ``Raw'' settings are illustrated in Fig.~\ref{pr_city_thin} and Fig.~\ref{pr_city_raw}. The results again indicate the better performance of SEAL over comparing baselines on Cityscapes.

\begin{table*}[t!]
\centering
\caption{AP scores on the Cityscapes validation set under the ``Raw'' setting. Results are measured by $\%$.\label{tb:cityscapes_ap}}
\resizebox{\textwidth}{!}{\begin{tabular}{l|c|c|c|c|c|c|c|c|c|c|c|c|c|c|c|c|c|c|c|c}
Method	& road & sidewalk & building & wall & fence & pole & t-light & t-sign & veg & terrain & sky & person & rider & car & truck & bus & train & motor & bike & mean \\
\hline \hline
CASENet   & 49.9 & 65.3 & 69.1 & 33.5 & 34.6 & 73.8 & 65.9 & 66.3 & 75.0 & 47.5 & 81.2 & 77.9 & 59.9 & 72.7 & 33.5 & 45.0 & 30.2 & 44.0 & 66.5 & 57.5\\
CASENet-S & \textbf{85.5} & 75.7 & 75.6 & 38.7 & 36.8 & \textbf{79.8} & 70.8 & 72.3 & 81.2 & 54.3 & 87.5 & 82.0 & 63.8 & \textbf{89.4} & 40.0 & 61.7 & 31.8 & 44.8 & 70.6 & 65.4\\
SEAL      & 82.7 & \textbf{76.9} & \textbf{77.5} & \textbf{39.6} & \textbf{38.0} & \textbf{79.8} & \textbf{71.9} & \textbf{74.0} & \textbf{83.5} & \textbf{57.0} & \textbf{88.6} & \textbf{83.5} & \textbf{65.9} & 88.9 & \textbf{43.0} & \textbf{64.0} & \textbf{34.4} & \textbf{47.1} & \textbf{72.6} & \textbf{66.8}\\
\end{tabular}}
\end{table*}

\end{document}